\documentclass{article}
 \usepackage[main, preprint]{neurips_2026}
\usepackage[utf8]{inputenc}
\usepackage[T1]{fontenc}
\usepackage{url}
\usepackage{booktabs}
\usepackage{amsfonts}
\usepackage{nicefrac}
\usepackage{microtype}
\usepackage{graphicx}
\usepackage{subcaption}
\usepackage{multirow}
\usepackage{makecell}
\usepackage{threeparttable}
\usepackage{array}
\usepackage{pifont}
\usepackage{amsmath}
\usepackage{float}
\usepackage[table]{xcolor}
\usepackage{wrapfig}
\usepackage{adjustbox}
\usepackage{placeins}
\usepackage{amssymb}
\usepackage{titlesec}
\titlespacing*{\section}{0pt}{0.6ex}{0.5ex}
\titlespacing*{\subsection}{0pt}{0.4ex}{0.3ex}
\titlespacing*{\paragraph}{0pt}{0.4ex}{0.8em}
\usepackage{hyperref} % 必须最后

\newcommand{\cmark}{\ding{51}}
\newcommand{\xmark}{\ding{55}}

\renewcommand{\arraystretch}{1.12}
\title{RA-CFGCache: From Branch-Level Criteria to Guided-Risk Control under Classifier-Free Guidance}

\author{
\textbf{Yiming Liu}$^{1,3}$ \quad
\textbf{Ben Wan}$^{2}$ \quad
\textbf{Tongxuan Liu}$^{2}$ \quad
\textbf{Ao Wang}$^{1,3}$ \quad
\textbf{Yuqi Xiong}$^{3}$
\\
\textbf{Fan Zhang}$^{1,3}$ \quad
\textbf{Hui Chen}$^{3,\dagger}$ \quad
\textbf{Guiguang Ding}$^{1,3,\dagger}$
\\
$^{1}$School of Software, Tsinghua University, Beijing, China
\\
$^{2}$JD.com, Beijing, China
\\
$^{3}$BNRist, Tsinghua University, Beijing, China
}

\begin{document}

\maketitle

\begingroup
\renewcommand{\thefootnote}{\fnsymbol{footnote}}
\footnotetext[2]{Corresponding authors.}
\endgroup

\begin{abstract}
Diffusion models enable high-quality visual generation, but iterative denoising remains computationally expensive, especially under classifier-free guidance (CFG), which requires both conditional and unconditional evaluations. Training-free caching reduces this cost by reusing previously computed features or predictions. However, branch-local reuse criteria do not explicitly account for how cache errors combine under CFG or how local perturbations affect the final output. We examine two misalignments in cache control: a \emph{branch--guided mismatch}, where guided error depends on both the magnitudes and alignment of branch errors, and a \emph{local--final mismatch}, where the downstream impact of a local error varies across timesteps.
We propose \textbf{RA-CFGCache}, a \textbf{R}isk-\textbf{A}ligned \textbf{Caching} framework under \textbf{CFG} that incorporates both factors while keeping the sampling schedule and guidance rule fixed. \emph{CFG-aware Guided-Risk Composition} combines existing branch-wise proxies using CFG coefficients and offline-calibrated cross-branch alignment. \emph{Propagation-Aware Rescaling} further weights the resulting guided-risk estimate with a timestep-dependent propagation prior calibrated from isolated reuse perturbations. An online threshold controller then determines when to jointly refresh or reuse both branches.
Experiments on FLUX.1-dev, Wan2.1-T2V-1.3B, and CogVideoX-2B demonstrate improved efficiency--fidelity trade-offs over the evaluated training-free caching baselines. Moreover, \textbf{RA-CFGCache} is compatible with diverse base proxy families, including TeaCache-, DiCache-, and MagCache-style estimators, and consistently improves fidelity at nearly unchanged latency. Code is available at \url{https://github.com/yiming-l21/RA-CFGCache.git}.
\end{abstract}

\section{Introduction}
Recent years have witnessed substantial progress in diffusion models
\citep{sohldickstein2015deepunsupervisedlearningusing,
ho2020denoisingdiffusionprobabilisticmodels,
song2020generativemodelingestimatinggradients,
dhariwal2021diffusionmodelsbeatgans,
peebles2023scalablediffusionmodelstransformers}
for visual generation. In modern conditional diffusion inference,
classifier-free guidance (CFG) \citep{ho2022classifierfreediffusionguidance}
has become a standard mechanism for improving conditional fidelity and generation quality.
However, diffusion inference remains computationally expensive because it requires repeated
network evaluations over many denoising steps, and explicit two-branch CFG further increases
the per-step cost by evaluating both conditional and unconditional predictions.

To reduce this cost, prior work has explored various diffusion acceleration strategies, including distillation~\citep{meng2023distillationguideddiffusionmodels,sauer2023adversarialdiffusiondistillation}, efficient samplers~\citep{song2022denoisingdiffusionimplicitmodels,lu2022dpmsolverfastodesolver}, pruning~\citep{zhang2024tokenpruningcachingbetter}, and feature caching~\citep{ma2023deepcacheacceleratingdiffusionmodels,Liu_2025_CVPR}. Among them, training-free caching is particularly attractive, as it improves efficiency without retraining or modifying the original model. Since model outputs at nearby timesteps
often exhibit substantial redundancy, previously computed features or predictions can be reused
when an estimated change remains sufficiently small~\citep{ma2023deepcacheacceleratingdiffusionmodels,Liu_2025_CVPR,ma2026magcache}. Under standard caching, the reuse decision is made based on a single prediction, because its cache-induced error directly enters the denoising update. Under CFG caching, existing methods often extend the same idea to the conditional and unconditional branches, using branch-wise changes as reuse signals. However, branch-local reuse criteria do not explicitly account for how the two branch errors interact after CFG composition.

% However, as illustrated in Fig.~\ref{fig:cfg_sampling}, the object that matters under CFG caching is no longer a single prediction or either branch by itself. In existing methods, reuse is often decided from the conditional and unconditional branches separately. Under CFG, however, the denoising update is driven by the guided prediction composed from both branches, so the relevant cache-induced perturbation is the guided error rather than any single branch-wise error. As a result, cache-control criteria inherited from standard caching no longer match the quantity that truly drives denoising under CFG.

\begin{figure}[t]
    \centering
    \begin{subfigure}[t]{1.0\linewidth}
        \centering
        \includegraphics[width=\linewidth]{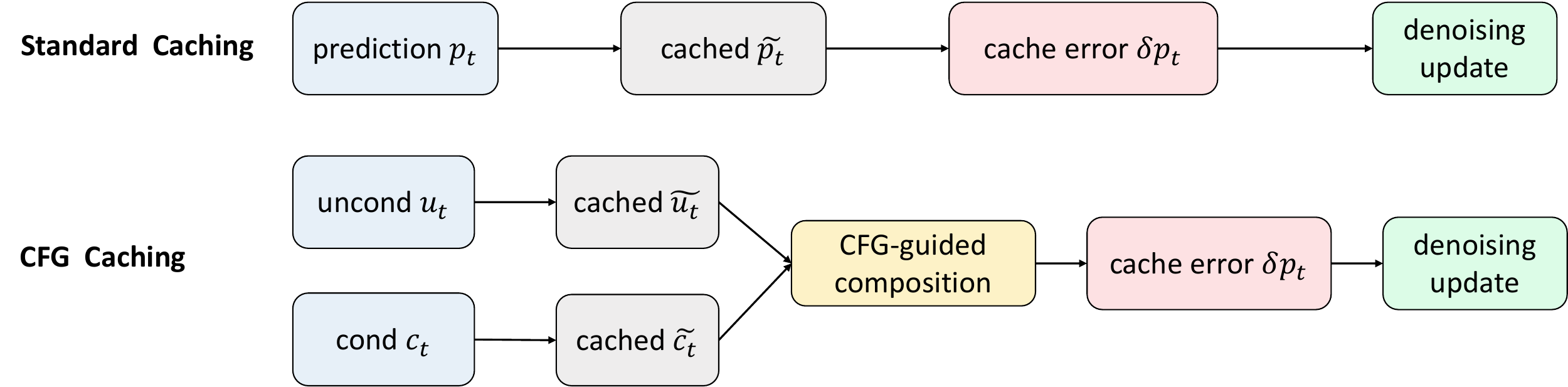}
    \end{subfigure}
    \caption{
    \textbf{Standard caching vs. CFG caching.}
    Standard caching reuses a single prediction whose cache error directly enters the denoising update, whereas CFG caching reuses conditional and unconditional branches whose errors are composed into the guided prediction.
    }
    \label{fig:cfg_sampling}
\end{figure}

\begin{figure*}[t]
    \centering

    \begin{subfigure}[t]{0.49\textwidth}
        \centering
        \includegraphics[height=0.66\textwidth, keepaspectratio]{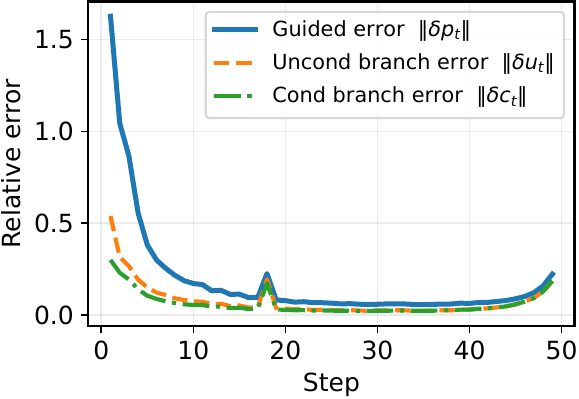}
        \caption{Branch-wise error vs. guided error.}
        \label{fig:first_mismatch}
    \end{subfigure}
    \hfill
    \begin{subfigure}[t]{0.49\textwidth}
        \centering
        \includegraphics[height=0.66\textwidth, keepaspectratio]{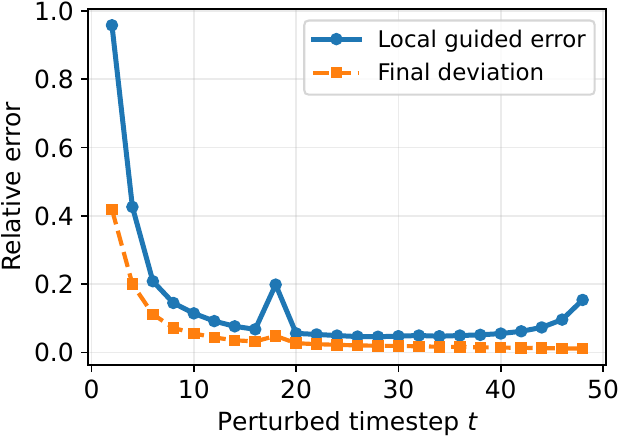}
        \caption{Local guided error vs. final deviation.}
        \label{fig:second_mismatch}
    \end{subfigure}

    \caption{
    \textbf{Two mismatches under CFG caching.}
    (a) The first mismatch: branch-wise cache errors do not directly reflect the guided error that enters the denoising update.
    (b) The second mismatch: local guided error does not directly predict the final deviation after error propagation across timesteps.
    }
    \label{fig:two_mismatches}
\end{figure*}

As illustrated in Fig.~\ref{fig:cfg_sampling}, under CFG the denoising update is determined
by the guided composition of the conditional and unconditional predictions. Consequently,
the cache-induced perturbation relevant to sampling is the error after CFG composition rather
than the error of either branch in isolation. We refer to this discrepancy as the
\emph{branch--guided mismatch}: branch-error magnitudes alone do not fully characterize the
guided error that drives the denoising update. Because the guided error depends on both
the magnitudes and relative directions of the two branch errors under the guidance scale,
their interaction affects the resulting perturbation. As shown in
Fig.~\ref{fig:two_mismatches}(a), the guided-error curve can differ substantially from the
two branch-wise error curves, especially in the early stage, showing that branch-wise error
magnitudes need not directly track the guided-error magnitude. We formally characterize this
relationship through the CFG error geometry in Section~\ref{sec:branch_guided_mismatch}.


Accounting for the \emph{branch--guided mismatch} alone is still insufficient, because the downstream effect of a local perturbation also depends on when it is introduced. Prior work has examined downstream cache effects through outcome-aware scheduling and error rectification~\citep{gao2025lemica,peng2026ertacache}. Here, we study this effect together with CFG branch-error composition. We demonstrate it through a single-step perturbation study in Fig.~\ref{fig:two_mismatches}(b): cache reuse is applied only at a specific timestep, while all other steps use full computation. We observe that similar local guided errors can lead to substantially different final deviations depending on the perturbed timestep. We refer to this discrepancy as the \emph{local--final mismatch}. This suggests that cache control should account not only for the local guided error, but also for its downstream impact on the final sample. We further quantify this effect in Section~\ref{sec:local_final_mismatch}.

Motivated by these observations, we propose \textbf{RA-CFGCache}, a \textbf{R}isk-\textbf{A}ligned \textbf{Caching} framework for \textbf{CFG}. Based on the above analysis, cache reuse under CFG should be governed by a risk estimate aligned with the cache-induced perturbation on the guided prediction, while also accounting for how this perturbation propagates to the final sample. \emph{CFG-aware Guided-Risk Composition} combines existing branch-wise proxies using the
CFG composition structure and calibrated cross-branch alignment to construct a local
guided-risk estimate, thereby addressing the \emph{branch--guided mismatch}.
\emph{Propagation-Aware Rescaling} then weights this risk estimate using a lightweight
timestep-dependent propagation prior to account for the downstream impact of local
perturbations, thereby addressing the \emph{local--final mismatch}.
The resulting risk estimate is used by a lightweight online threshold
controller to govern joint
refresh and reuse of the two branches, while keeping the sampling schedule and guidance
rule unchanged.

Our contributions are summarized as follows:
\begin{itemize}

\item \textbf{Characterizing two key misalignments under CFG caching.}
We show that cache control under CFG is affected by two distinct sources of misalignment:
the \emph{branch--guided mismatch}, where branch-wise errors do not fully characterize the
guided error that drives the denoising update, and the \emph{local--final mismatch}, where
the downstream impact of a local guided error varies across timesteps.

\item \textbf{Introducing a risk-aligned caching framework for CFG.}
We propose \textbf{RA-CFGCache}, which addresses these two misalignments through two coupled
designs: \emph{CFG-aware guided-risk composition}, which combines branch-wise proxies using
the CFG composition structure and calibrated cross-branch alignment to construct a local
guided-risk estimate, and \emph{propagation-aware rescaling}, which weights this risk estimate
using a timestep-dependent propagation prior to account for the downstream impact of local
guided errors.

\item \textbf{Demonstrating improved efficiency--fidelity trade-offs under CFG.}
Experiments and ablations on diffusion transformer pipelines show that \textbf{RA-CFGCache}
improves the efficiency--fidelity trade-off over the evaluated training-free caching baselines
under CFG. Risk diagnostics and controlled comparisons further support the proposed analysis
and control strategy.

\end{itemize}

\section{Related Work}
\label{gen_inst}
\paragraph{Training-free diffusion caching.}
Training-free diffusion caching accelerates diffusion inference by reusing intermediate
features, residuals, or model outputs across timesteps without retraining.
Existing methods broadly follow three overlapping directions.
The first follows \emph{pre-defined reuse}, where computation is reused according to fixed
layer, block, or timestep schedules, as exemplified by
DeepCache~\cite{ma2023deepcacheacceleratingdiffusionmodels},
FasterDiffusion~\cite{li2024fasterdiffusionrethinkingrole},
PAB~\cite{zhao2025realtimevideogenerationpyramid},
FORA~\cite{selvaraju2024forafastforwardcachingdiffusion}, and
$\Delta$-DiT~\cite{chen2024deltadittrainingfreeaccelerationmethod}.
The second follows \emph{adaptive reuse}, where control signals based on temporal change,
residual magnitude, token importance, sensitivity, or estimated output variation determine
when or where cached computation can be reused, as in
TeaCache~\cite{Liu_2025_CVPR},
MagCache~\cite{ma2026magcache},
ToCa~\cite{zou2025acceleratingdiffusiontransformerstokenwise},
DiCache~\cite{bu2025dicacheletdiffusionmodel}, and
SenCache~\cite{haghighi2026sencacheacceleratingdiffusionmodel}.
A third direction shifts \emph{from reuse to prediction}, estimating future features or
outputs through extrapolation, interpolation, or empirical correction, as in
TaylorSeer~\cite{Liu_2025_ICCV},
HiCache~\cite{feng2026hicachepluginscaledhermiteupgrade}, and
FoCa~\cite{zheng2025forecastcalibratefeaturecaching}.
Recent methods also consider downstream cache effects:
LeMiCa~\cite{gao2025lemica} uses final-output impact for global error-aware scheduling,
while ERTACache~\cite{peng2026ertacache} analyzes cache-error propagation and introduces
corresponding correction mechanisms.
In contrast, \textbf{RA-CFGCache} studies downstream cache effects together with CFG-specific
branch-error composition, combining cross-branch error interaction and
timestep-dependent propagation weighting in a unified criterion.

\paragraph{CFG-aware acceleration and guidance optimization.}
Classifier-free guidance (CFG)~\cite{ho2022classifierfreediffusionguidance}
improves conditional generation quality, but explicit two-branch CFG requires both
conditional and unconditional evaluations.
Existing approaches exploit \emph{cross-branch redundancy}, modify the
\emph{guidance process}, or jointly optimize \emph{guidance and caching}.
FasterCache~\cite{lv2025fastercachetrainingfreevideodiffusion} exploits redundancy
between conditional and unconditional features and introduces CFG-Cache for branch reuse.
Guidance optimization methods include per-timestep guidance scheduling such as
CFG Schedulers~\cite{wang2024analysisclassifierfreeguidanceweight},
interval-based strategies such as
Apply Guidance in Interval~\cite{kynkäänniemi2024applyingguidancelimitedinterval},
and instance-aware magnitude control such as
MAMBO-G~\cite{zhu2026mambogmagnitudeawaremitigationboosted}.
OUSAC~\cite{sun2025ousacoptimizedguidancescheduling} further jointly optimizes
CFG timesteps, guidance scales, and caching.
In contrast, \textbf{RA-CFGCache} keeps the sampling schedule and guidance rule
unchanged and controls joint branch refresh and reuse through a guided-risk
estimate that accounts for both cross-branch error interaction and
downstream propagation.

\section{Method}
\label{headings}
\subsection{Preliminaries}

We consider conditional diffusion / flow-based generative models implemented with diffusion transformers (DiTs), where inference proceeds through an iterative denoising process. At each timestep $t$, the model predicts from the current latent $x_t$ and the conditioning input, and the sampler updates the latent state using this prediction. We denote the prediction function by $f_\theta(\cdot)$, where $\theta$ denotes the fixed model parameters.

Training-free caching accelerates inference by reusing cached computation from a previous timestep instead of recomputing it at the current one. Although different methods may cache predictions, residuals, or intermediate features, their effect on the sampler can be abstracted as replacing a fresh prediction $p_t$ with the actually used prediction $\tilde p_t$. Both are defined at the same current latent $x_t$: $p_t$ is obtained by full computation, whereas $\tilde p_t$ is obtained with cache reuse. The resulting local cache-induced error is
\begin{equation}
\delta p_t = \tilde p_t - p_t.
\label{eq:generic_cache_error}
\end{equation}
Here, $p_t$ serves as an analytical reference and need not be evaluated by the online controller.

Under classifier-free guidance (CFG) \citep{ho2022classifierfreediffusionguidance}, the model evaluates both unconditional and conditional branches:
\begin{equation}
u_t = f_\theta(x_t,t,\emptyset), \qquad
c_t = f_\theta(x_t,t,y),
\label{eq:cfg_branches}
\end{equation}
where $u_t$ and $c_t$ denote the unconditional and conditional predictions, respectively. The prediction entering the sampler is
\begin{equation}
p_t^{(s)} = u_t + s(c_t-u_t),
\label{eq:cfg_guided}
\end{equation}
where $s$ is the guidance scale. This is equivalent to the original CFG formulation of \citet{ho2022classifierfreediffusionguidance} up to a reparameterization of the guidance weight. Therefore, under CFG, the denoising update is governed by the guided prediction $p_t^{(s)}$ rather than by either branch alone. Although reuse is applied at the branch level, its local effect on sampling is determined by the induced error on the guided prediction.

\subsection{Branch--Guided Mismatch}
\label{sec:branch_guided_mismatch}

Under CFG, cache reuse affects the denoising update through the guided prediction rather than through either branch alone. We formalize this by explicitly characterizing the guided error induced by branch-level reuse.

Let $u_t$ and $c_t$ denote the fresh unconditional and conditional branch predictions defined in Eq.~\eqref{eq:cfg_branches}. Under cache reuse, the actual branch predictions used in guidance composition are
\begin{equation}
\tilde{u}_t = u_t + \delta u_t, \qquad
\tilde{c}_t = c_t + \delta c_t,
\label{eq:stale_branch_preds}
\end{equation}
where $\delta u_t,\delta c_t$ are the corresponding cache-induced branch errors. The resulting guided prediction under reuse is
\begin{equation}
\tilde{p}_t^{(s)} = \tilde{u}_t + s(\tilde{c}_t-\tilde{u}_t),
\label{eq:guided_pred_reuse}
\end{equation}
so the cache-induced guided error becomes
\begin{equation}
\delta p_t
=
\tilde{p}_t^{(s)} - p_t^{(s)}
=
(1-s)\delta u_t + s\delta c_t .
\label{eq:guided_error_decomp}
\end{equation}

Let
$\rho_t = \cos(\delta u_t,\delta c_t)$
denote the cosine similarity between the two branch errors. When either error is zero, we set $\rho_t=0$ since the cross-term vanishes. Then
\begin{equation}
\|\delta p_t\|_2^2
=
(1-s)^2\|\delta u_t\|_2^2
+
s^2\|\delta c_t\|_2^2
+
2s(1-s)\|\delta u_t\|_2\|\delta c_t\|_2\rho_t .
\label{eq:guided_error_norm_rho}
\end{equation}

Equation~\eqref{eq:guided_error_norm_rho} shows that guided error depends not only on branch-error magnitudes, but also on their alignment through the cross-term. Under the practical CFG regime $s>1$, the coefficient $2s(1-s)$ is negative, so positive alignment induces partial cancellation, whereas weak or negative alignment suppresses this effect and can even lead to amplification. As shown in Fig.~\ref{fig:rho}, this alignment is timestep-dependent: the cancellation effect is weaker in earlier steps and stronger in later ones. We provide a more detailed analysis of this CFG error geometry and the origin of the alignment statistic in Appendix~\ref{app:cfg_error_geometry}. Consequently, branch-wise error no longer faithfully reflects the true local perturbation under CFG. This constitutes the \emph{branch--guided mismatch}.

\subsection{Local--Final Mismatch}
\label{sec:local_final_mismatch}
\begin{figure}[t]
    \centering
    \begin{subfigure}[t]{0.49\linewidth}
        \centering
        \includegraphics[width=\linewidth]{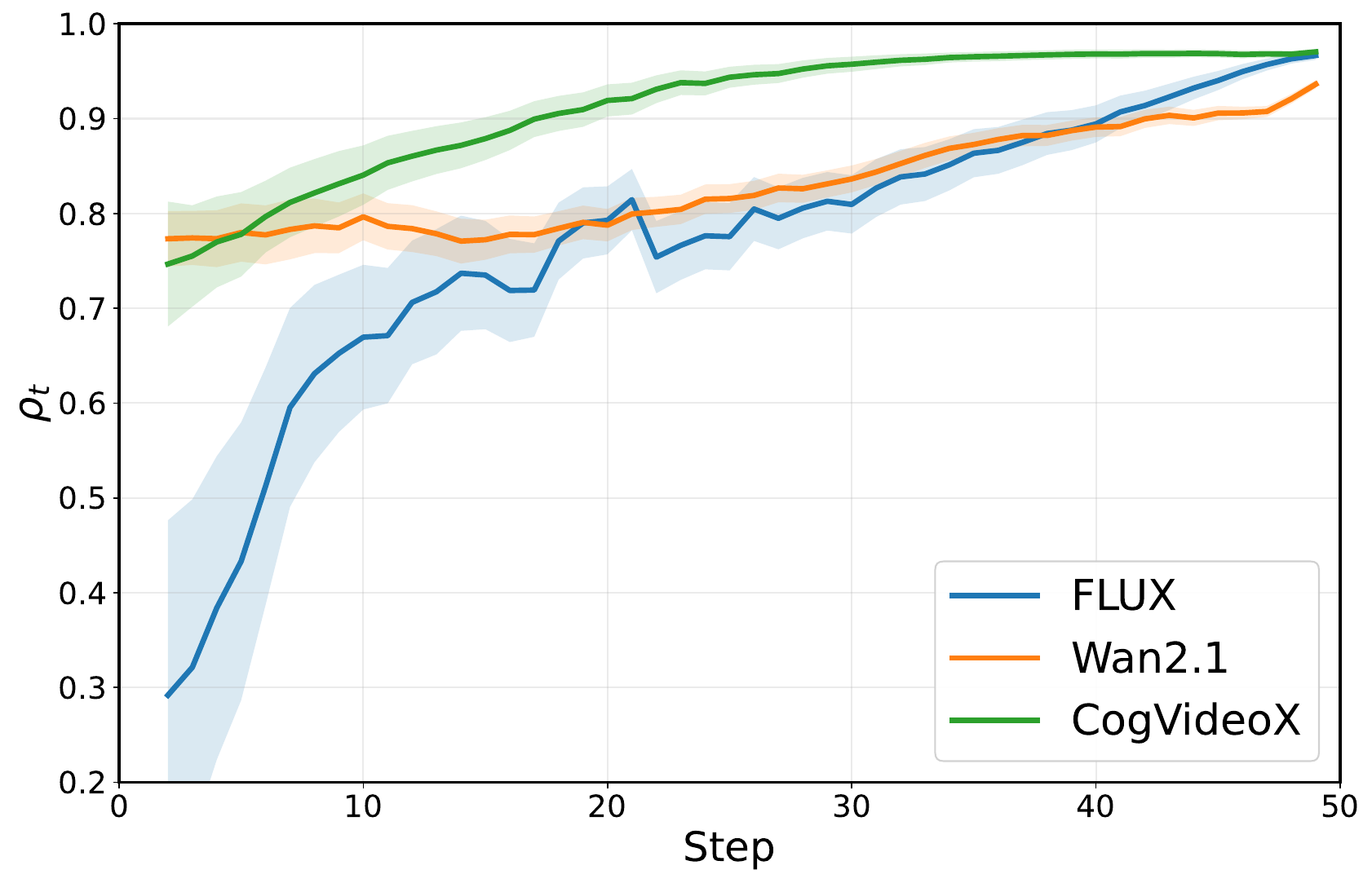}
        \caption{\textbf{Branch-error alignment}}
        \label{fig:rho}
    \end{subfigure}
    \hfill
    \begin{subfigure}[t]{0.49\linewidth}
        \centering
        \includegraphics[width=\linewidth]{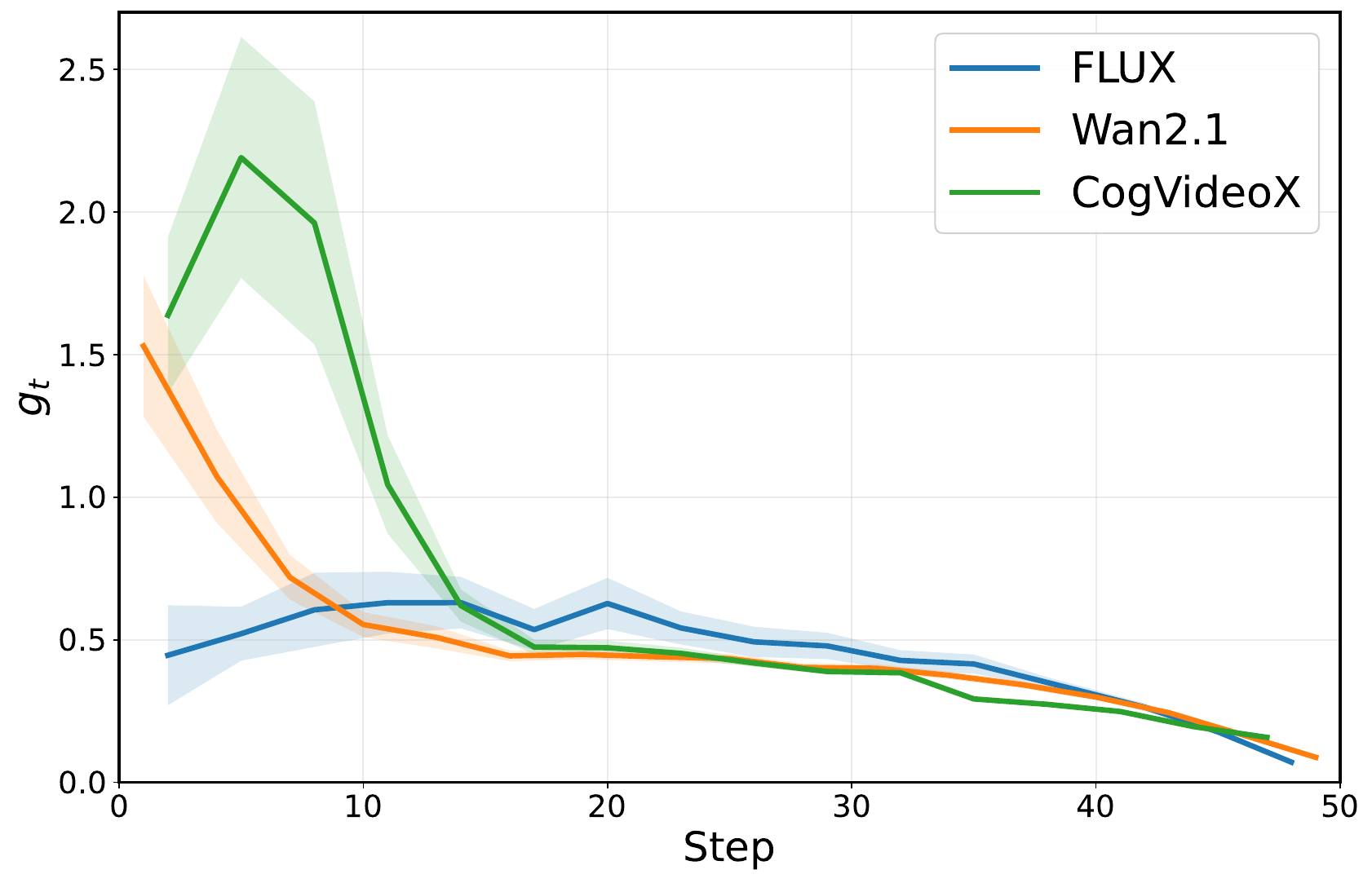}
        \caption{\textbf{Propagation gain}}
        \label{fig:prop_gain}
    \end{subfigure}
    \caption{
    \textbf{Two sources of mismatch under CFG caching.}
    (a) The alignment $\rho_t=\cos(\delta u_t,\delta c_t)$ varies across timesteps, inducing timestep-dependent cancellation or amplification in Eq.~\eqref{eq:guided_error_norm_rho}. 
    (b) The propagation gain $g_t$ varies across timesteps, showing that similar local guided errors can induce different final deviations. Solid curves show the mean over 100 prompts, and shaded bands indicate
$\pm$ one standard deviation across prompts.
    }
    \label{fig:mismatch_pair}
    \vspace{-1.0\baselineskip}
\end{figure}

Even guided error is only a local quantity at the current timestep. For cache control, what ultimately matters is its effect on the final generated sample after subsequent denoising steps.

To isolate this effect, we consider a single-step perturbation setting. The perturbed and reference trajectories are identical before timestep $t$; cache reuse is applied only at timestep $t$, while all other steps use full computation under the same initial noise and conditioning. The local guided error is defined as
\begin{equation}
e_t^{\mathrm{local}} = \|\delta p_t\|_2.
\label{eq:local_guided_error}
\end{equation}
The resulting final deviation is measured by
\begin{equation}
e_t^{\mathrm{final}} = \|x_0^{(t)} - x_0^{\mathrm{full}}\|_2,
\label{eq:final_deviation}
\end{equation}
where $x_0^{(t)}$ and $x_0^{\mathrm{full}}$ denote the final latent states
of the perturbed and full-computation trajectories, respectively. For $e_t^{\mathrm{local}}>0$, we then define the propagation gain as
\begin{equation}
g_t=\frac{e_t^{\mathrm{final}}}{e_t^{\mathrm{local}}}.
\label{eq:prop_gain}
\end{equation}
We provide additional details of propagation analysis in Appendix~\ref{app:local_final_analysis}.

Figure~\ref{fig:mismatch_pair}(b) shows that the propagation gain varies systematically across timesteps: similar local guided errors can induce different final deviations depending on when they occur. We refer to this as the \emph{local--final mismatch}, motivating a timestep-dependent propagation prior for cache control.

\subsection{RA-CFGCache}
\label{sec:ra_cfgcache}

\begin{figure*}[t]
    \centering
    \includegraphics[width=\textwidth]{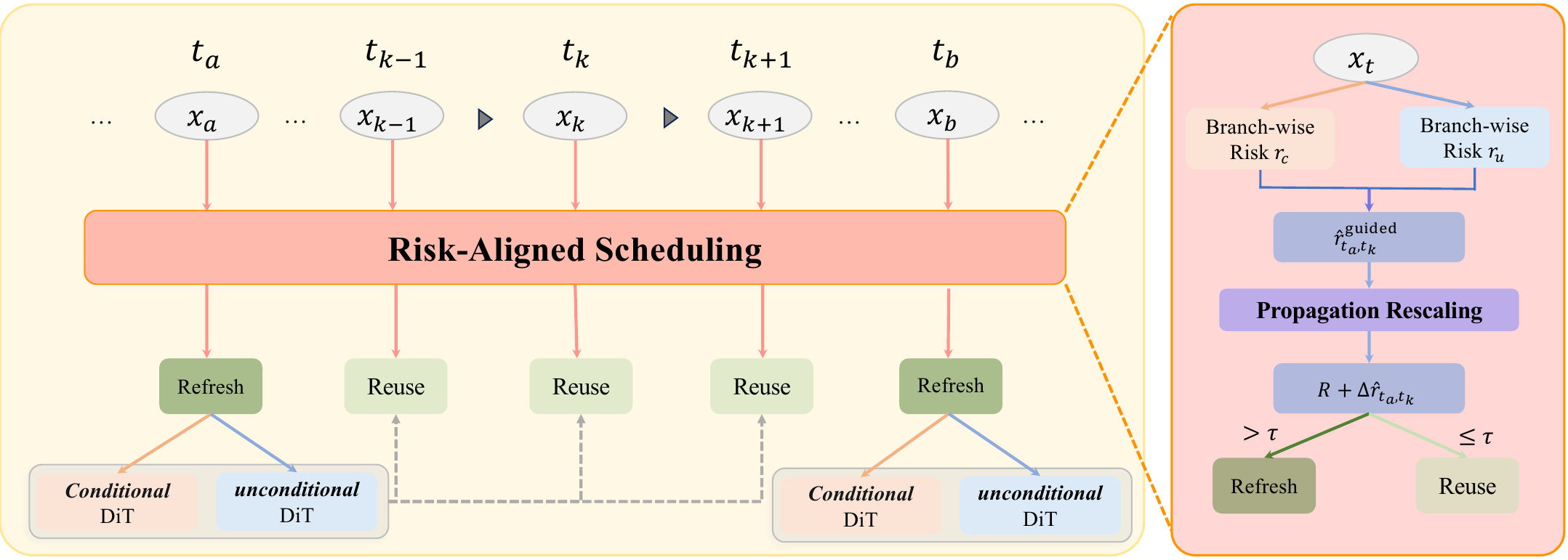}
    \caption{
\textbf{Overview of RA-CFGCache.}
Under classifier-free guidance (CFG), cache control should be aligned with the guided prediction rather than either branch alone. \textbf{RA-CFGCache} first combines branch-wise proxy estimates into a guided-risk estimate, then rescales it using a timestep-dependent propagation prior, and finally uses the resulting risk estimate for online refresh decisions. 
}
\vspace{-1em}
    \label{fig:main_overview}
\end{figure*}

\textbf{RA-CFGCache} realizes cache control under CFG as \emph{guided-risk control}, as illustrated in Fig.~\ref{fig:main_overview}. It contains two components: \emph{CFG-aware guided-risk composition}, which constructs a CFG-informed guided-risk estimate from branch-wise proxies, and \emph{propagation-aware rescaling}, which weights this risk estimate using a calibrated timestep-dependent propagation prior.

\paragraph{CFG-aware Guided-Risk Composition.}
Since true branch errors are unavailable online, we start from nonnegative branch-wise proxy estimates provided by a base caching method, such as TeaCache~\citep{Liu_2025_CVPR}, MagCache~\citep{ma2026magcache}, or DiCache~\citep{bu2025dicacheletdiffusionmodel}. Let $\hat r_u(t_a,t_b)$ and $\hat r_c(t_a,t_b)$ denote the proxy estimates for the current candidate reuse from anchor timestep $t_a$ to current timestep $t_b$; accumulation is handled separately by the controller below. The exact proxy definitions used in our experiments are provided in Appendix~\ref{app:proxy_details}. Motivated by the exact CFG error decomposition in Eq.~\eqref{eq:guided_error_norm_rho}, we combine them using the same quadratic form:
\begin{equation}
\bigl(\hat r_{t_a,t_b}^{\mathrm{guided}}\bigr)^2
=
(1-s)^2 \hat r_u^2(t_a,t_b)
+
s^2 \hat r_c^2(t_a,t_b)
+
2s(1-s)\,\bar\rho_{t_a,t_b}\,
\hat r_u(t_a,t_b)\hat r_c(t_a,t_b),
\label{eq:online_guided_risk}
\end{equation}
and take the nonnegative square root to obtain
$\hat r_{t_a,t_b}^{\mathrm{guided}}$.
Here, $\bar\rho_{t_a,t_b}$ is an offline estimate of cross-branch alignment. On a held-out calibration split, we compute the cosine similarity between conditional and unconditional temporal differences of the full-computation residual outputs for each pair $(t_a,t_b)$ and average it across samples. We refer to $\hat r_{t_a,t_b}^{\mathrm{guided}}$ as a proxy-based guided-risk estimate used for cache control, rather than an exact estimate of the true guided-error magnitude.

\paragraph{Propagation-aware Rescaling.}
To account for non-uniform error propagation, we further rescale the guided-risk estimate using a timestep-dependent propagation prior. The single-step perturbation study reveals systematic variation across timesteps, which we summarize with a lightweight power-law surrogate:
\begin{equation}
\hat g_k
=
\lambda\Bigl(\frac{T-k}{T}\Bigr)^\alpha+\beta,
\label{eq:gain_fit}
\end{equation}
where $k$ is the increasing sampler-step index, $T$ is the total number of steps, and $\lambda,\alpha,\beta$ are fitted from the calibration measurements. The power-law form is empirical; fitting details are provided in Appendix~\ref{app:propagation_fit}. Here, $a$ and $b$ denote the sampler-step indices corresponding to timesteps $t_a$ and $t_b$. The propagation-aware risk contribution is
\begin{equation}
\Delta \hat r_{t_a,t_b}
=
\max\left(0,\hat g_b\,\hat r_{t_a,t_b}^{\mathrm{guided}}\right).
\label{eq:online_risk}
\end{equation}
This quantity incorporates both CFG error composition and the timestep-dependent downstream effect of local perturbations.

\paragraph{Online Control Policy.}
We maintain an accumulated-risk score $R$ within each reuse interval. At the beginning of an interval, the anchor step is set to $t_a$ and $R$ is initialized to zero. At the current step $t_b$, the controller computes $\Delta \hat r_{t_a,t_b}$ in Eq.~\eqref{eq:online_risk}. A refresh is triggered when
\begin{equation}
R+\Delta \hat r_{t_a,t_b} > \tau,
\label{eq:refresh_rule}
\end{equation}
where $\tau$ controls the reuse--refresh trade-off. Otherwise, reuse continues and
\begin{equation}
R \leftarrow R+\Delta \hat r_{t_a,t_b}.
\end{equation}
After a refresh, the anchor is reset to the current step and $R$ is reinitialized to zero. The accumulated-risk score is an interval-level control quantity rather than an exact estimate or formal bound on the final output deviation.

\paragraph{Reuse Action.}
We adopt joint reuse of both CFG branches as the default action, allowing the controller to account for their combined proxy risk. When branch errors are positively aligned and $s>1$, their contributions can partially cancel under CFG composition. Joint reuse also skips recomputation of the cached components in both branches. Under parallel branch execution, one-sided reuse may provide limited latency savings because the guided prediction must still wait for the recomputed branch. Detailed comparisons are provided in Appendix~\ref{app:reuse_action} and Appendix~\ref{app:parallel_consistency}.

\section{Experiments}
\label{others}
\subsection{Experimental Setup}
\label{sec:exp_setup}

\begin{figure*}[t]
    \centering
    \hspace{-0.8cm}
    \includegraphics[height=3.5cm,keepaspectratio]{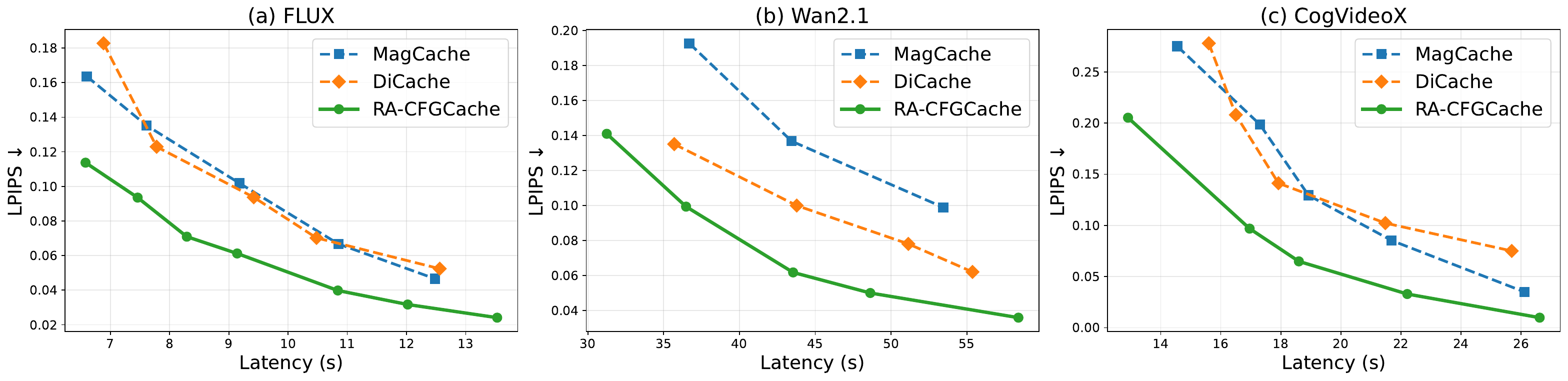}
    \caption{\textbf{Trade-off curves.}
    \textbf{RA-CFGCache} achieves lower LPIPS at comparable latency across models.}
\vspace{-1em}
    \label{fig:pareto_curves}
    
\end{figure*}
\begin{table*}[t]
\centering
\footnotesize
\begin{threeparttable}
\caption{\textbf{Main results on text-to-image and text-to-video generation.}
Comparison of \textbf{RA-CFGCache} with recent diffusion acceleration baselines.}
\label{tab:main_results}
\setlength{\tabcolsep}{3.5pt}
\renewcommand{\arraystretch}{1.12}
\begin{tabular}{llccccc}
\toprule
\textbf{Model}
& \textbf{Method}
& \textbf{LPIPS} $\downarrow$
& \textbf{SSIM} $\uparrow$
& \textbf{PSNR} $\uparrow$
& \textbf{Speedup} $\uparrow$
& \textbf{Latency (s)} $\downarrow$ \\
\midrule

\multirow{10}{*}{\textbf{FLUX.1-dev}}
& Vanilla ($T=50$)                              & 0.0000 & 1.0000 & $\infty$ & 1.00$\times$ & 21.02 \\
& Vanilla ($T=25$)                              & 0.4556 & 0.6207 & 13.09 & 1.96$\times$ & 10.71 \\
& TeaCache ($\tau$=0.4)                         & 0.4471 & 0.6174 & 13.80 & 2.53$\times$ & 8.31 \\
& TaylorSeer ($\mathcal{N}$=3,$\mathcal{O}$=1)  & 0.4379 & 0.6398 & 13.75 & 2.37$\times$ & 8.86 \\
& DiCache ($m=2,\tau$=0.4)                      & 0.1431 & 0.8229 & 22.13 & 2.94$\times$ & 7.15 \\
& MagCache ($\tau$=0.24)                        & 0.1636 & 0.8177 & 21.46 & 3.18$\times$ & 6.60 \\
& HiCache ($\mathcal{N}$=3,$\mathcal{O}$=1)     & 0.3937 & 0.6646 & 14.72 & 2.35$\times$ & 8.94 \\
& FasterCache ($I_{\Delta}=8, I_{\mathrm{attn}}=8$) & 0.2158 & 0.7593 & 20.88 & 2.09$\times$ & 10.06 \\
& \cellcolor{gray!10}\textbf{RA-CFGCache-Slow ($\tau$=0.18)}
& \cellcolor{gray!10}\textbf{0.1137}
& \cellcolor{gray!10}\textbf{0.8680}
& \cellcolor{gray!10}\textbf{24.02}
& \cellcolor{gray!10}3.19$\times$
& \cellcolor{gray!10}6.58 \\

& \cellcolor{gray!10}\textbf{RA-CFGCache-Fast ($\tau$=0.3)}
& \cellcolor{gray!10}0.1484
& \cellcolor{gray!10}0.8340
& \cellcolor{gray!10}22.75
& \cellcolor{gray!10}\textbf{3.95$\times$}
& \cellcolor{gray!10}\textbf{5.32} \\
\midrule

\multirow{10}{*}{\textbf{Wan2.1-T2V-1.3B}}
& Vanilla ($T=50$)                              & 0.0000 & 1.0000 & $\infty$ & 1.00$\times$ & 93.32 \\
& Vanilla ($T=25$)                              & 0.4371 & 0.5658 & 15.87 & 1.93$\times$ & 48.41 \\
& TeaCache ($\tau$=0.15)                        & 0.1661 & 0.7955 & 23.08 & 2.44$\times$ & 38.24 \\
& TaylorSeer ($\mathcal{N}$=3,$\mathcal{O}$=1)  & 0.3632 & 0.6187 & 16.96 & 2.13$\times$ & 43.79 \\
& DiCache ($m=2,\tau$=0.4)                      & 0.1350 & 0.8361 & 24.90 & 2.61$\times$ & 35.70 \\
& MagCache ($\tau$=0.1)                         & 0.1926 & 0.7696 & 22.16 & 2.54$\times$ & 36.69 \\
& HiCache ($\mathcal{N}$=3,$\mathcal{O}$=1)     & 0.2916 & 0.6711 & 18.79 & 2.14$\times$ & 43.49 \\
& FasterCache ($I_{\Delta}=8, I_{\mathrm{attn}}=8$) & 0.3483 & 0.6532 & 20.37 & 2.41$\times$ & 38.71 \\
& \cellcolor{gray!10}\textbf{RA-CFGCache-Slow ($\tau$=0.4)}
& \cellcolor{gray!10}\textbf{0.0993}
& \cellcolor{gray!10}\textbf{0.8778}
& \cellcolor{gray!10}\textbf{28.03}
& \cellcolor{gray!10}2.56$\times$
& \cellcolor{gray!10}36.48 \\
& \cellcolor{gray!10}\textbf{RA-CFGCache-Fast ($\tau$=0.6)}
& \cellcolor{gray!10}0.1409
& \cellcolor{gray!10}0.8386
& \cellcolor{gray!10}26.12
& \cellcolor{gray!10}\textbf{2.99$\times$}
& \cellcolor{gray!10}\textbf{31.25} \\
\midrule

\multirow{8}{*}{\textbf{CogVideoX-2B}}
& Vanilla ($T=50$)                              & 0.0000 & 1.0000 & $\infty$ & 1.00$\times$ & 40.48 \\
& Vanilla ($T=25$)                              & 0.4751 & 0.5445 & 13.61 & 1.86$\times$ & 21.79 \\
& TeaCache ($\tau$=0.2)                         & 0.2040 & 0.7538 & 21.48 & 2.42$\times$ & 16.72 \\
& DiCache ($m=2,\tau$=0.4)                      & 0.2080 & 0.7710 & 22.35 & 2.44$\times$ & 16.57 \\
& MagCache ($\tau$=0.08)                        & 0.2386 & 0.7230 & 20.33 & 2.34$\times$ & 17.30 \\
& FasterCache ($I_{\Delta}=6, I_{\mathrm{attn}}=6$)                         & 0.1157 & 0.8573 & 25.48 & 2.03$\times$ & 19.94 \\
& \cellcolor{gray!10}\textbf{RA-CFGCache-Slow ($\tau$=0.6)}
& \cellcolor{gray!10}\textbf{0.0968}
& \cellcolor{gray!10}\textbf{0.8746}
& \cellcolor{gray!10}\textbf{26.93}
& \cellcolor{gray!10}2.39$\times$
& \cellcolor{gray!10}16.96 \\
& \cellcolor{gray!10}\textbf{RA-CFGCache-Fast ($\tau$=1.4)}
& \cellcolor{gray!10}0.2052
& \cellcolor{gray!10}0.7614
& \cellcolor{gray!10}21.89
& \cellcolor{gray!10}\textbf{3.13$\times$}
& \cellcolor{gray!10}\textbf{12.91} \\
\bottomrule
\end{tabular}
\end{threeparttable}
\vspace{-2em}
\end{table*}

\paragraph{Models and baselines.}
We evaluate our method on diffusion transformer pipelines under classifier-free guidance, including FLUX.1-dev~\citep{flux2024} for text-to-image generation, Wan2.1-T2V-1.3B~\citep{wan2025wanopenadvancedlargescale} and CogVideoX-2B~\citep{yang2025cogvideoxtexttovideodiffusionmodels} for text-to-video generation. We compare against representative training-free acceleration baselines, including TeaCache~\citep{Liu_2025_CVPR}, TaylorSeer~\citep{Liu_2025_ICCV}, DiCache~\citep{bu2025dicacheletdiffusionmodel}, MagCache~\citep{ma2026magcache}, HiCache~\citep{feng2026hicachepluginscaledhermiteupgrade}, and the CFG-related FasterCache~\citep{lv2025fastercachetrainingfreevideodiffusion}. All methods use the same models, prompts, denoising schedules, and hardware unless otherwise specified.

\paragraph{Main evaluation settings.}
Unless otherwise specified, all main experiments use $50$ denoising steps.
For FLUX.1-dev, we generate $1024\times1024$ images with true CFG scale and guidance embedding both set to $3.5$.
For Wan2.1-T2V-1.3B, we generate $832\times480$ videos with $81$ frames, $16$ FPS and guidance scale $5.0$.
For CogVideoX-2B, we generate $720\times480$ videos with $49$ frames, $8$ FPS and guidance scale $6.0$.

\paragraph{Data and metrics.}
For text-to-image evaluation, we use 200 DrawBench prompts~\citep{saharia2022photorealistictexttoimagediffusionmodels}. For text-to-video evaluation, we use 100 prompts sampled from the \texttt{all\_dimension.txt} prompt set of VBench~\citep{huang2023vbenchcomprehensivebenchmarksuite}. Unless otherwise specified, all accelerated outputs are compared against the corresponding vanilla outputs under the same prompt and random seed. We report end-to-end latency and speedup as efficiency metrics, and LPIPS~\citep{zhang2018unreasonableeffectivenessdeepfeatures}, SSIM~\citep{1284395}, and PSNR against vanilla outputs as fidelity metrics.

\begin{figure*}[t]
\centering
\setlength{\tabcolsep}{0pt}
\renewcommand{\arraystretch}{0.95}
\scriptsize

\begin{tabular}{@{}c@{}c@{}c@{}c@{}c@{}}
Original & TeaCache (2.8$\times$) & MagCache (2.76$\times$) & DiCache (2.71$\times$) & Ours (2.83$\times$) \\
\includegraphics[width=0.19\textwidth]{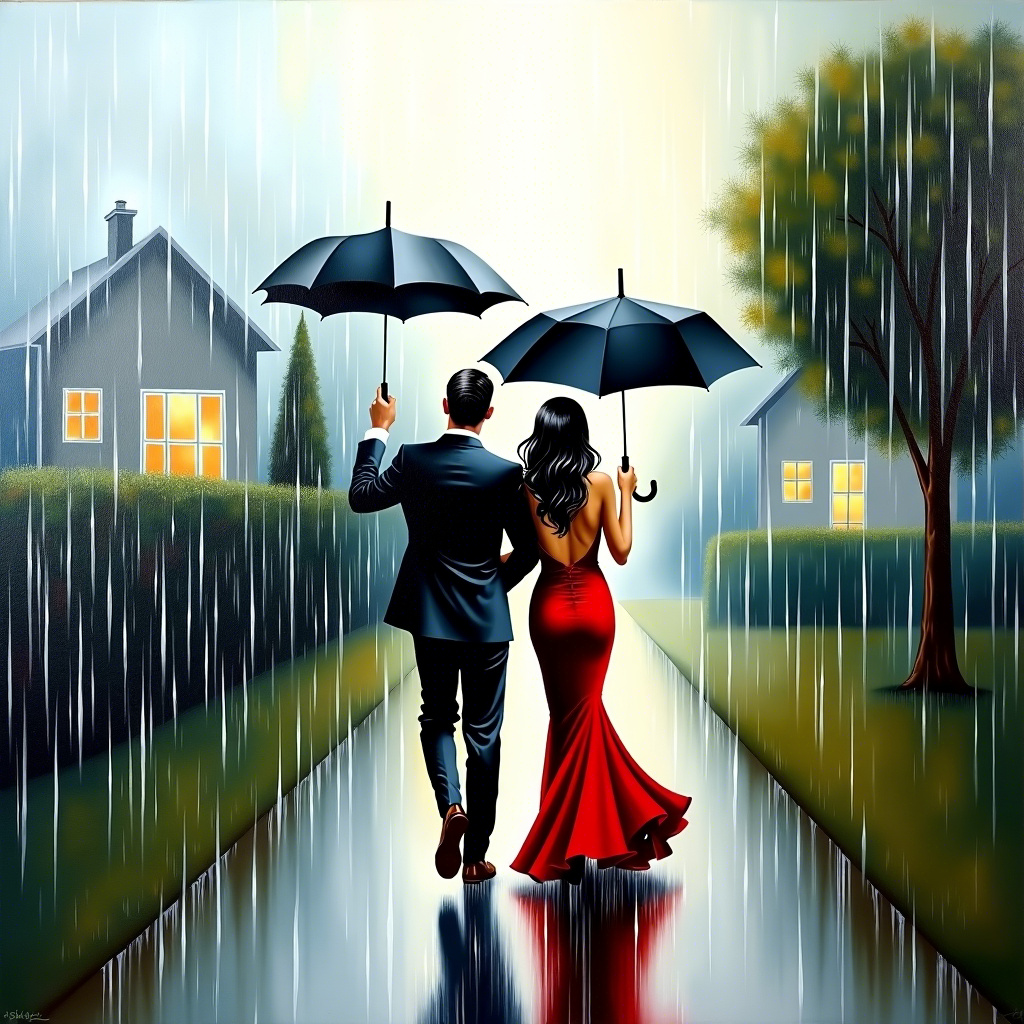}
& \includegraphics[width=0.19\textwidth]{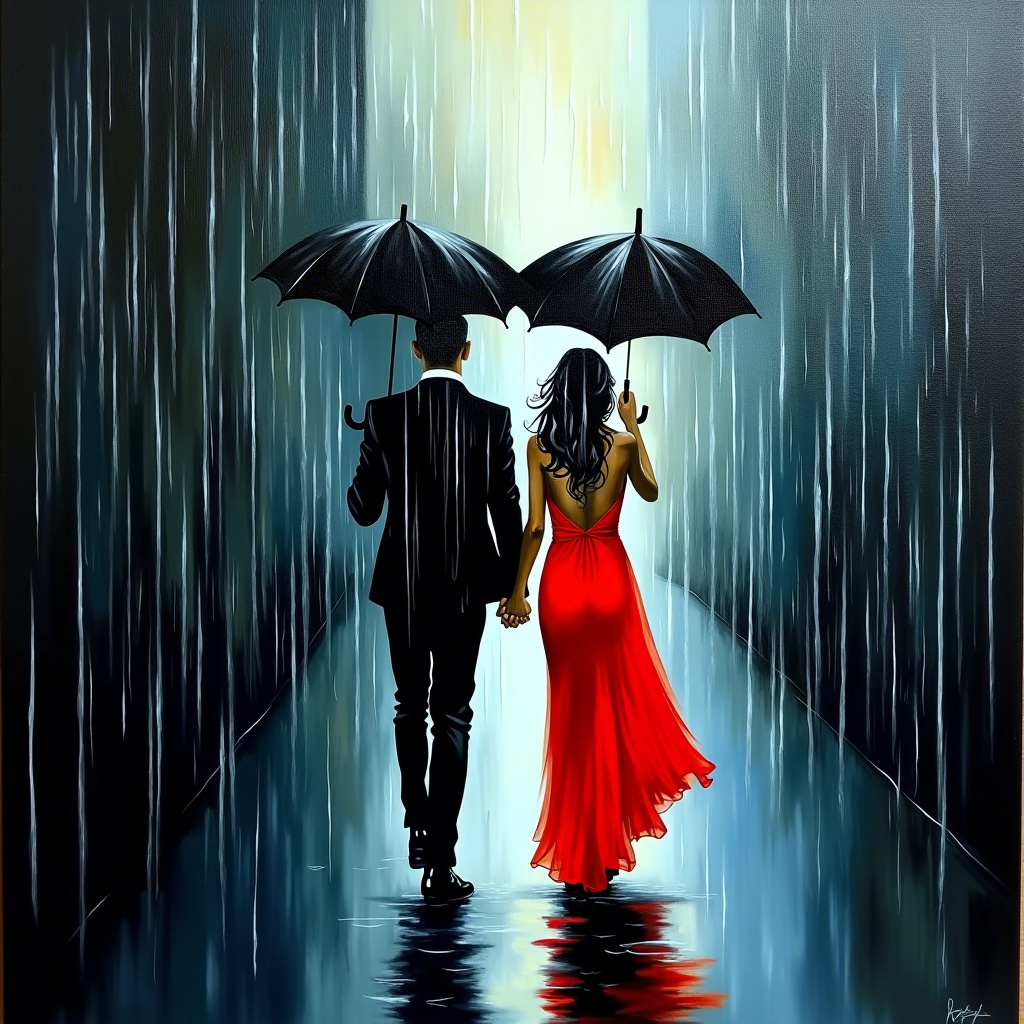}
& \includegraphics[width=0.19\textwidth]{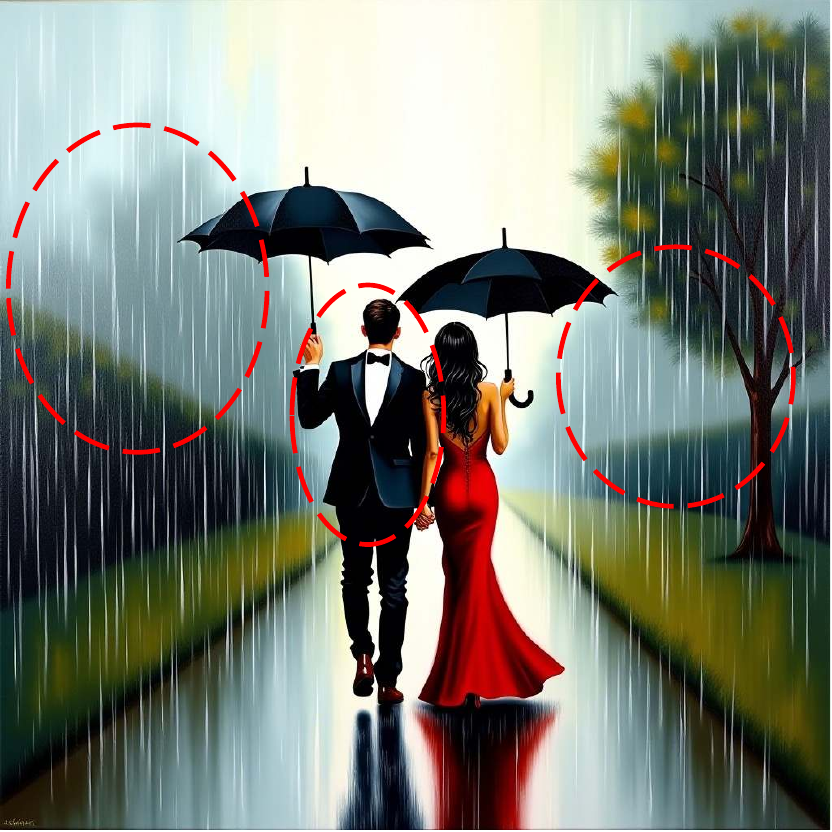}
& \includegraphics[width=0.19\textwidth]{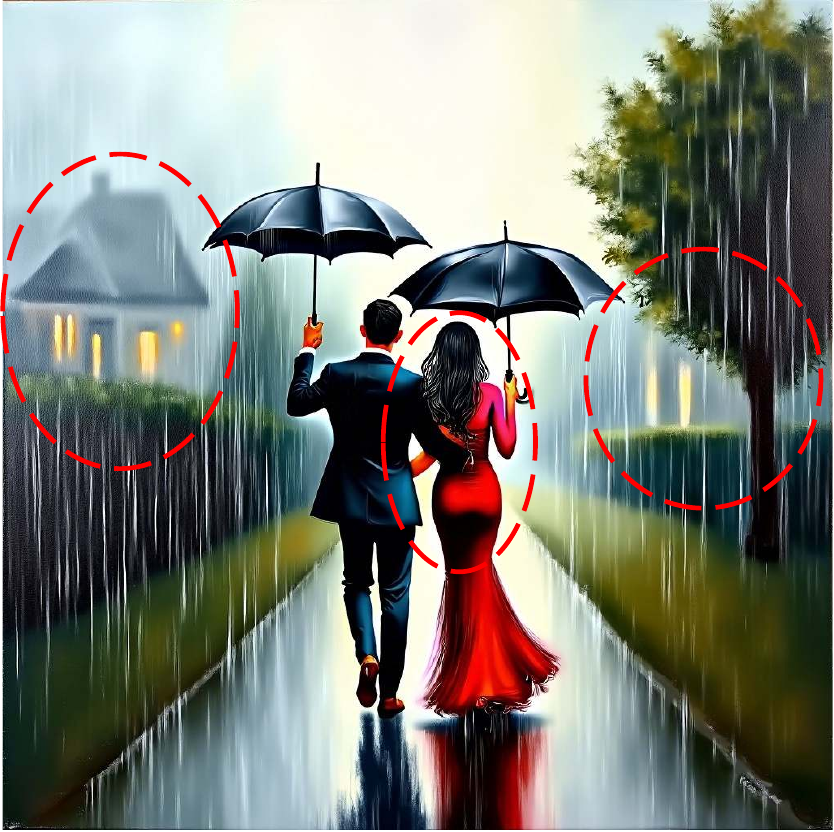}
& \includegraphics[width=0.19\textwidth]{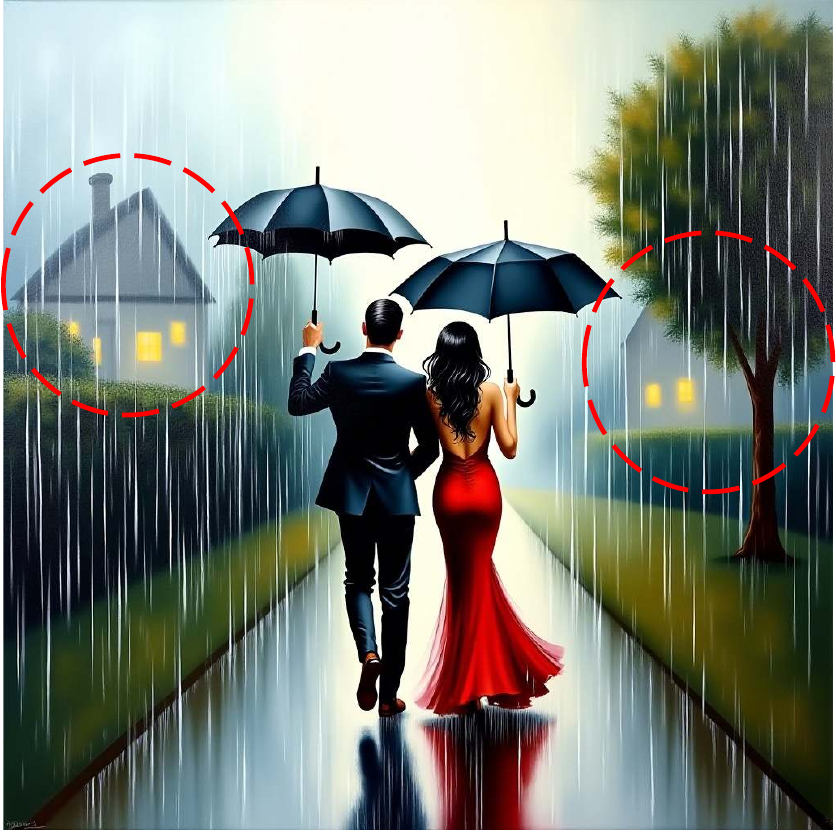}
\\[-0.15em]

\includegraphics[width=0.19\textwidth]{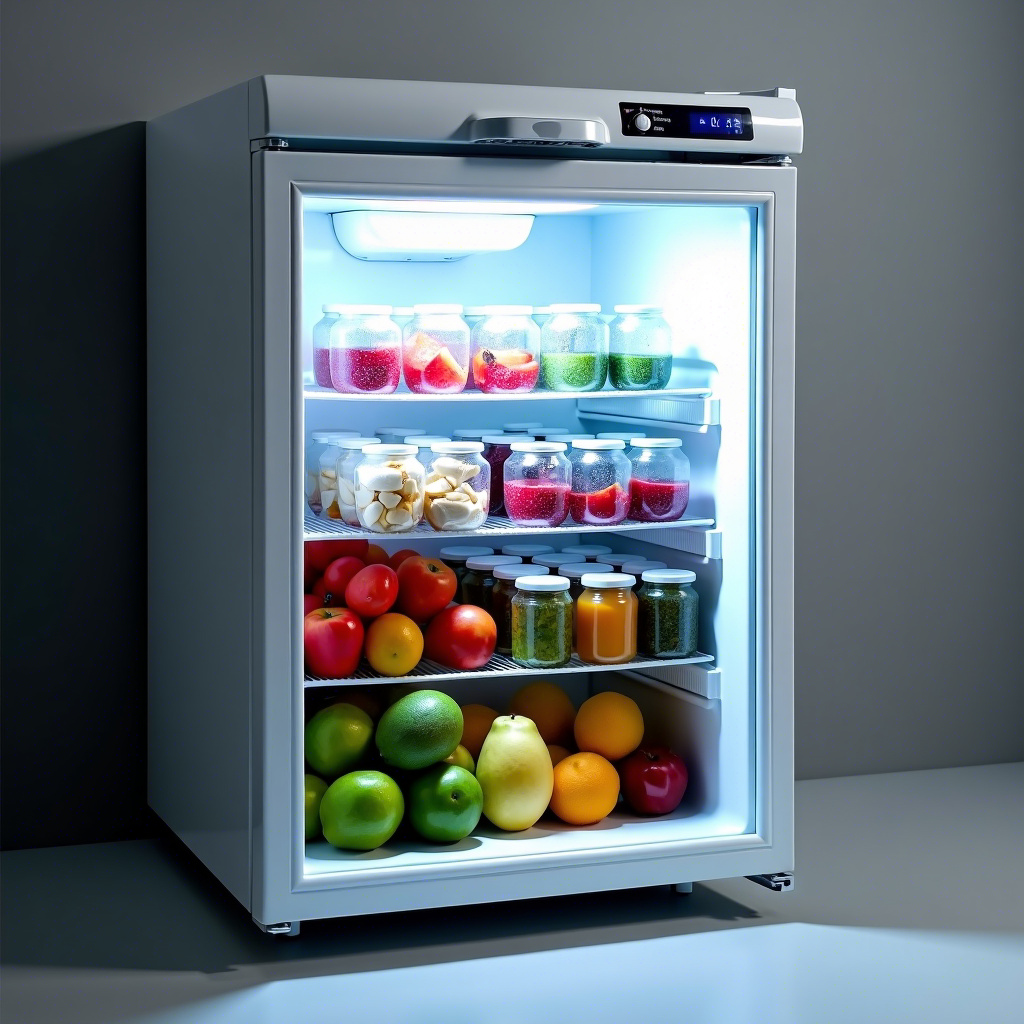}
& \includegraphics[width=0.19\textwidth]{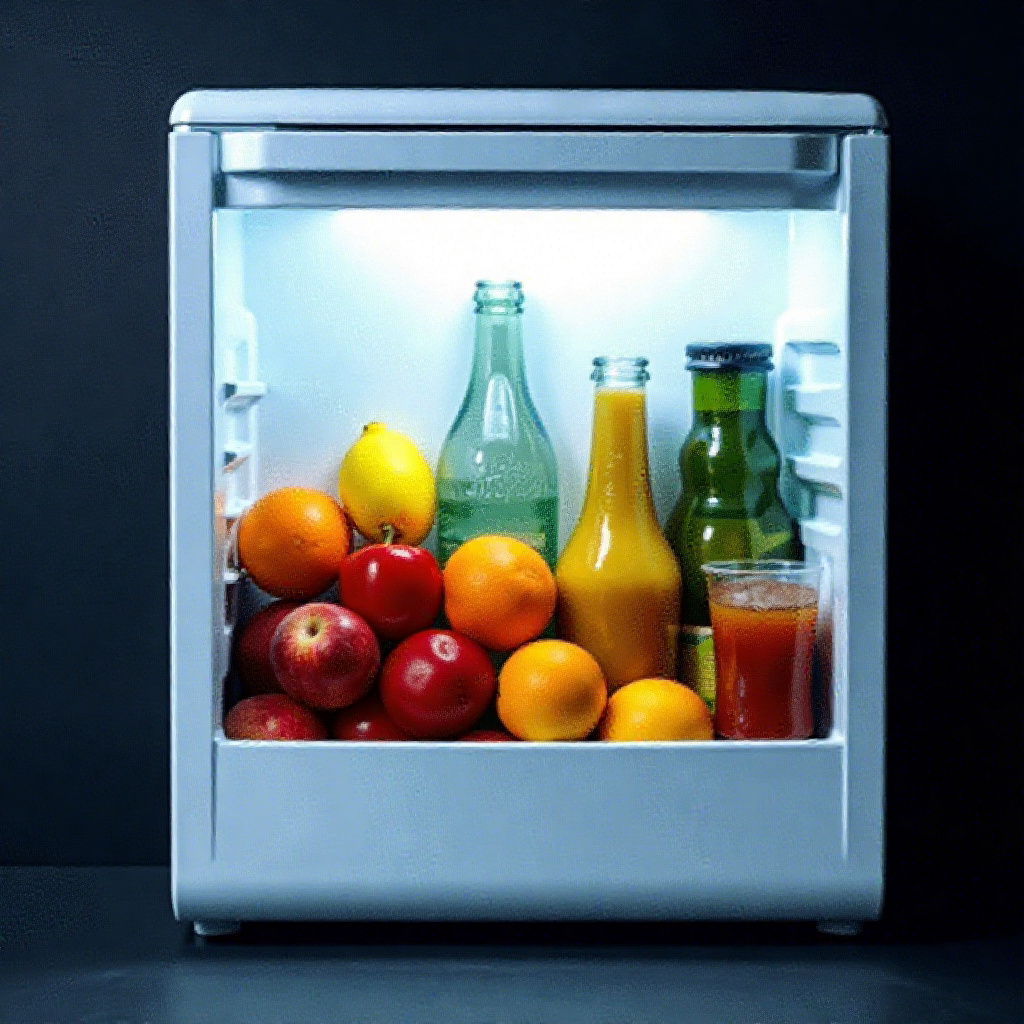}
& \includegraphics[width=0.19\textwidth]{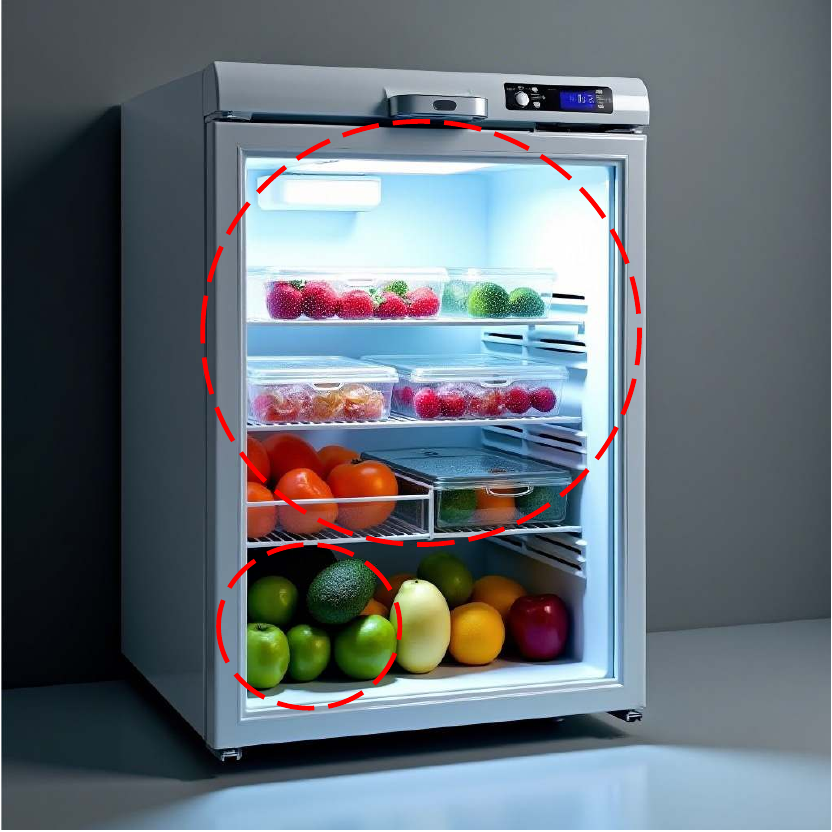}
& \includegraphics[width=0.19\textwidth]{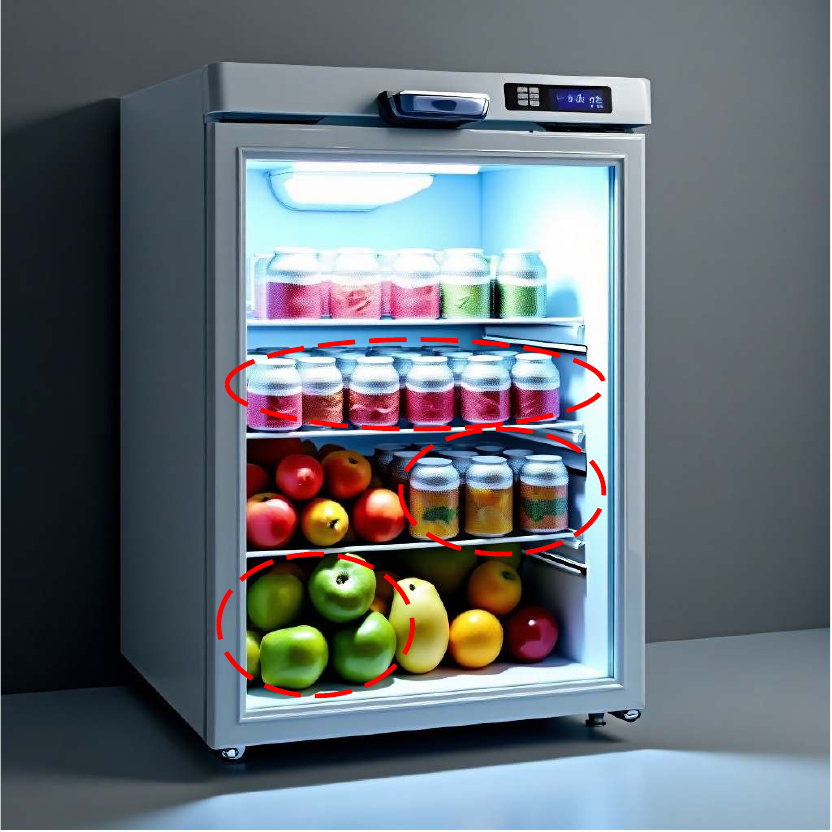}
& \includegraphics[width=0.19\textwidth]{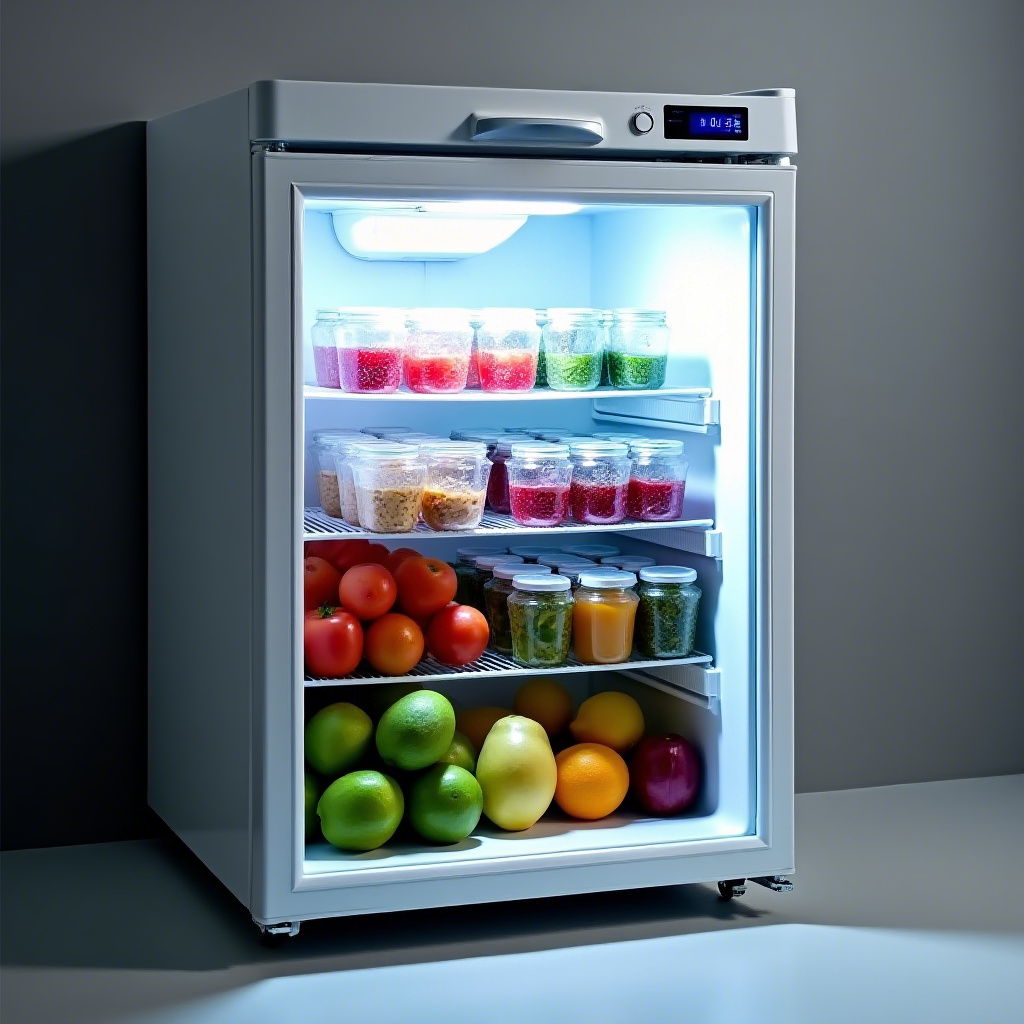}
\\[-0.15em]

\includegraphics[width=0.19\textwidth]{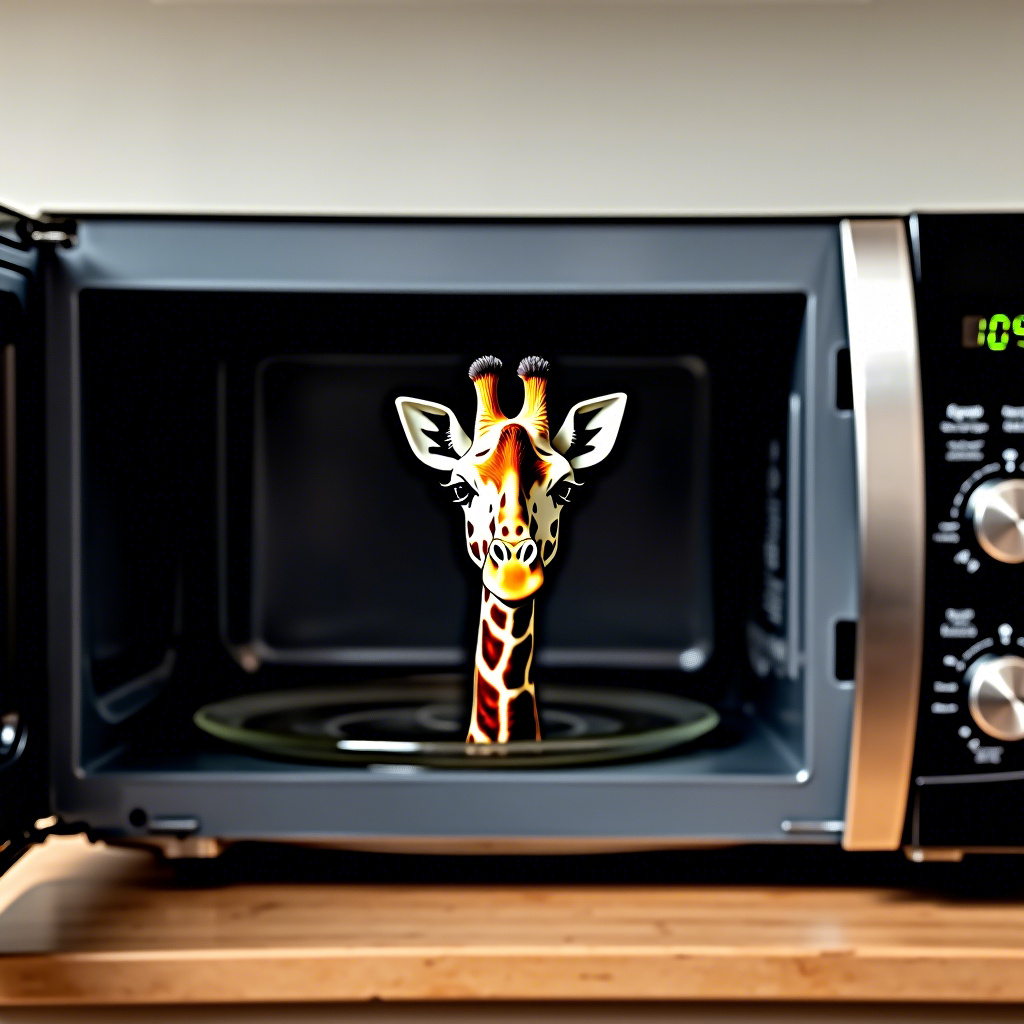}
& \includegraphics[width=0.19\textwidth]{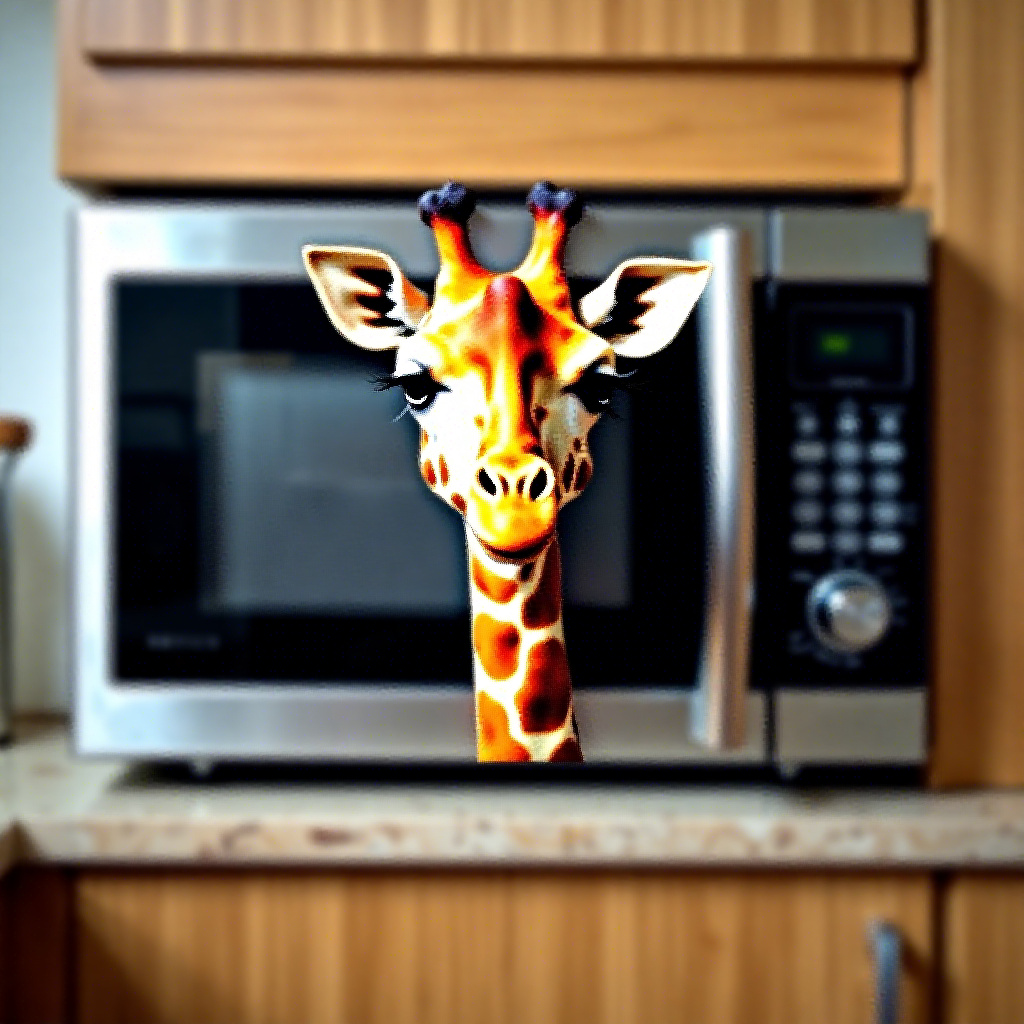}
& \includegraphics[width=0.19\textwidth]{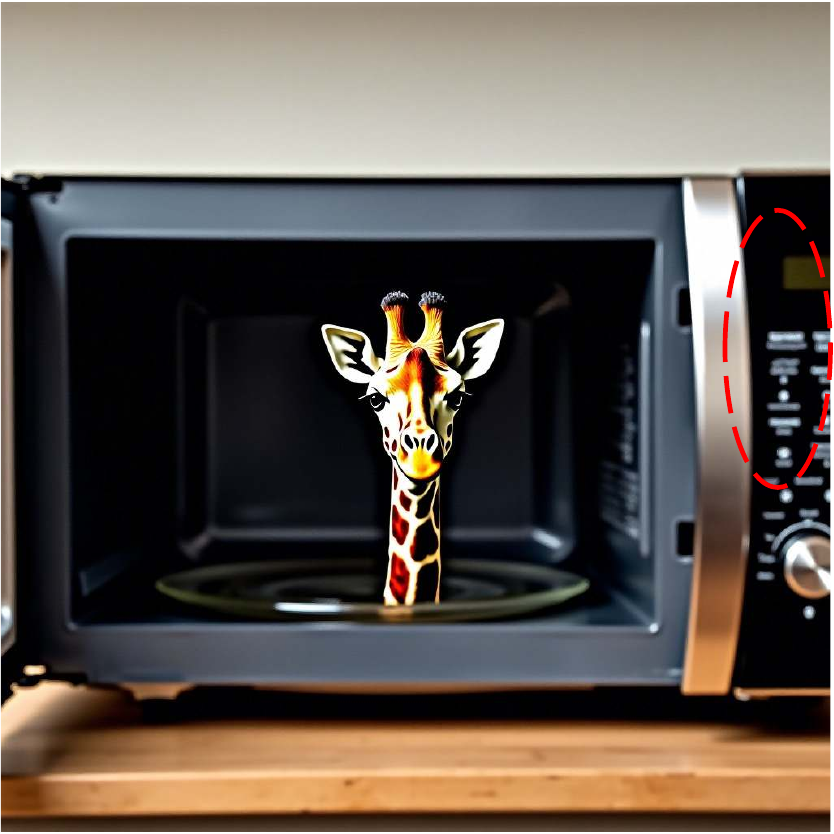}
& \includegraphics[width=0.19\textwidth]{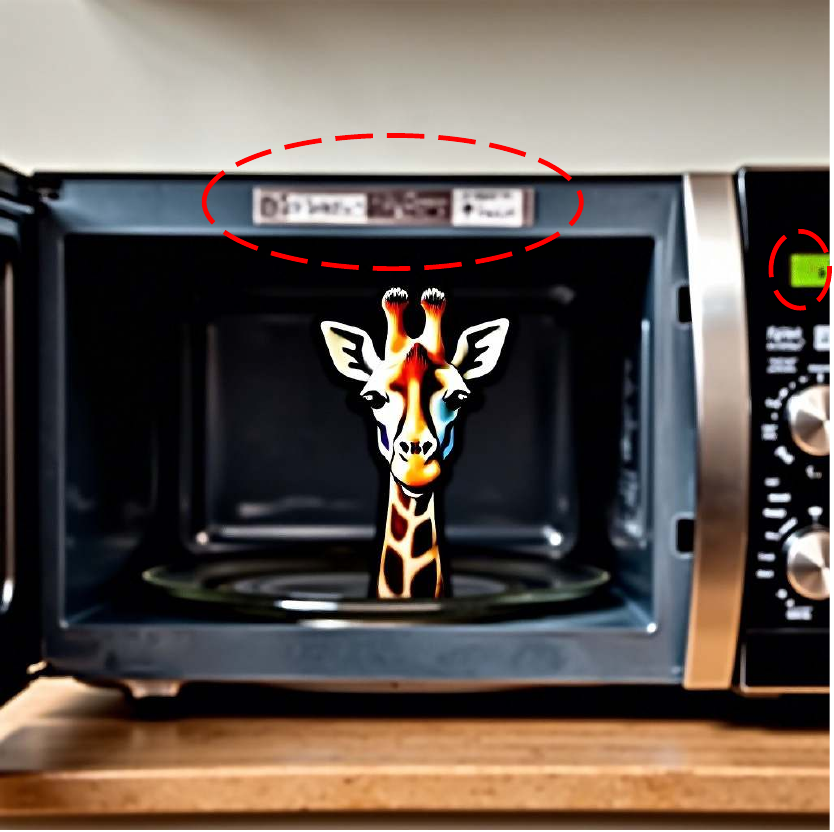}
& \includegraphics[width=0.19\textwidth]{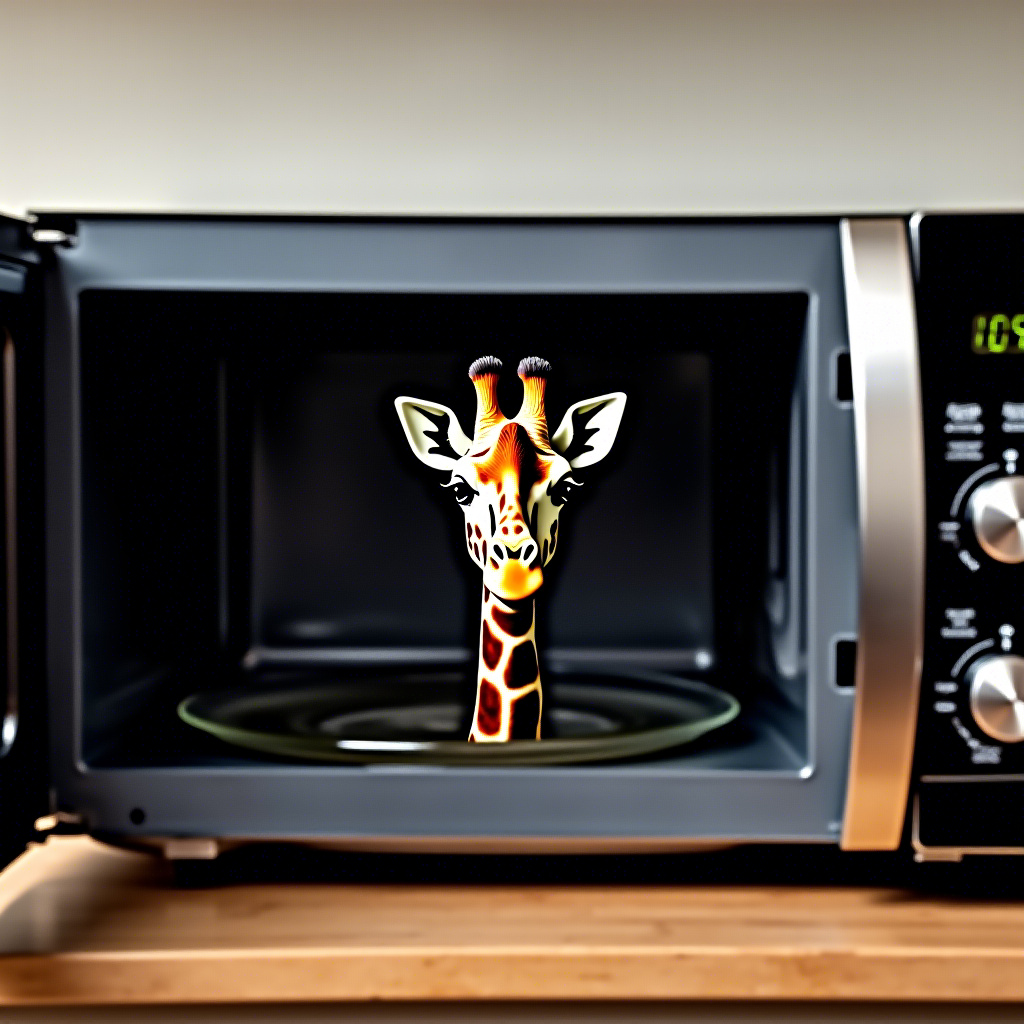}
\\
\end{tabular}

\caption{
\textbf{Qualitative comparison on FLUX.1-dev.}
We show three representative text-to-image examples under the same prompts and random seeds.
Compared with TeaCache, MagCache, and DiCache, \textbf{RA-CFGCache} better preserves local textures, object structures, and semantic details under comparable acceleration settings.
}
\vspace{-1em}
\label{fig:qualitative_compare}
\end{figure*}
\begin{table*}[t]
\centering
\hspace*{-0.02\textwidth}%
\begin{minipage}[t]{0.48\textwidth}
\vspace{0pt}
\centering
\scriptsize
\caption{\textbf{Contribution of each component.} Each component improves the trade-off, and combining both gives the best result.}
\label{tab:ablation_components}
\setlength{\tabcolsep}{1.6pt}
\renewcommand{\arraystretch}{1}
\begin{tabular}{cccccc}
\toprule
\textbf{CFG-aware} & \textbf{Propagation} & \textbf{Speedup}$\uparrow$ & \textbf{LPIPS}$\downarrow$ & \textbf{SSIM}$\uparrow$ & \textbf{PSNR}$\uparrow$ \\
\textbf{Composition} & \textbf{Rescaling} &  &  &  &  \\
\midrule
\xmark & \xmark & 3.42$\times$ & 0.1732 & 0.8076 & 21.12 \\
\cmark & \xmark & 3.41$\times$ & 0.1391 & 0.8462 & 22.82 \\
\xmark & \cmark & 3.41$\times$ & 0.1423 & 0.8354 & 22.47 \\
\rowcolor{gray!10}
\cmark & \cmark & \textbf{3.43}$\times$ & \textbf{0.1262} & \textbf{0.8565} & \textbf{23.45} \\
\bottomrule
\end{tabular}
\end{minipage}
\hspace{0.005\textwidth}
\begin{minipage}[t]{0.46\textwidth}
\vspace{0pt}
\centering
\scriptsize
\caption{\textbf{Effect of the reused quantity.} Joint branch reuse gives the best overall result under a matched refresh budget.}
\label{tab:ablation_reuse_action}
\setlength{\tabcolsep}{2.2pt}
\renewcommand{\arraystretch}{1}
\begin{tabular}{lcccc}
\toprule
\textbf{Reused quantity} & \textbf{Speedup}$\uparrow$ & \textbf{LPIPS}$\downarrow$ & \textbf{SSIM}$\uparrow$ & \textbf{PSNR}$\uparrow$ \\
\midrule
Always refresh           & 1.00$\times$ & 0.0000 & 1.0000 & $\infty$ \\
Reuse unconditional only & 1.55$\times$ & 0.2943 & 0.7281 & 19.80 \\
Reuse conditional only   & 1.47$\times$ & 0.2317 & 0.7630 & 18.62 \\
Reuse $\Delta$ only      & 1.55$\times$ & 0.1939 & 0.7940 & 19.91 \\
\rowcolor{gray!10}
Reuse both branches      & \textbf{3.38$\times$} & \textbf{0.1729} & \textbf{0.8112} & \textbf{21.17} \\
\bottomrule
\end{tabular}
\end{minipage}
\vspace{-2em}
\end{table*}

\paragraph{Implementation details.}
All experiments are implemented in PyTorch on NVIDIA H200 GPUs, primarily using bfloat16 with selected operations in float32. Each sample uses one GPU, with the two CFG branches evaluated sequentially in the main experiments. \textbf{RA-CFGCache} uses the threshold-based scheduler in Section~\ref{sec:ra_cfgcache} with joint branch reuse by default. Unless otherwise specified, it uses a MagCache-style proxy for FLUX.1-dev and a TeaCache-style proxy for Wan2.1-T2V-1.3B and CogVideoX-2B.
We report \textbf{RA-CFGCache-Fast} and \textbf{RA-CFGCache-Slow} using different risk thresholds $\tau$.
For each model, offline calibration estimates the pairwise alignment statistic $\bar{\rho}_{t_a,t_b}$ and timestep-dependent rescaling coefficients, which are fixed during evaluation. Reported latency includes online cache-control overhead but excludes offline calibration. Additional details are provided in Appendix~\ref{app:implementation_details}.

\subsection{Main Results}
\label{sec:main_results}

\paragraph{Quantitative Comparison.}
Table~\ref{tab:main_results} reports the main quantitative results under CFG. Across the evaluated text-to-image and text-to-video pipelines, \textbf{RA-CFGCache} achieves favorable efficiency--fidelity trade-offs relative to the evaluated baselines. At comparable latency, it improves fidelity over strong caching baselines, while its fast variants provide additional acceleration with a corresponding fidelity trade-off.

\paragraph{Trade-off curves.}
Figure~\ref{fig:pareto_curves} compares latency--LPIPS trade-offs across FLUX.1-dev, Wan2.1-T2V-1.3B, and CogVideoX-2B.
Across the evaluated operating range, \textbf{RA-CFGCache} achieves lower LPIPS at comparable latency than MagCache and DiCache.
The improvement across multiple operating points shows that the gain is not limited to a single tuned threshold.

\textbf{Qualitative comparison.}
Figure~\ref{fig:qualitative_compare} compares \textbf{RA-CFGCache} with TeaCache, MagCache, and DiCache on FLUX.1-dev at comparable speedups.
The baselines introduce visible scene changes or local structural artifacts in the highlighted regions, while \textbf{RA-CFGCache} remains closer to the original samples and better preserves textures, structures, and semantic details.
Additional qualitative results on Wan2.1 and CogVideoX are provided in Appendix~\ref{app:comparison}.

\begin{table*}[t]
\centering

% =========================================================
% Left: CFG-scale robustness
% =========================================================
\begin{minipage}[t]{0.50\textwidth}
\centering
\scriptsize
\caption{\textbf{Robustness across CFG scales.}
\textbf{RA-CFGCache} maintains a strong efficiency--fidelity trade-off
across different guidance scales.}
\label{tab:ablation_cfgscale}

\setlength{\tabcolsep}{3.8pt}
\renewcommand{\arraystretch}{1.11}

\begin{tabular}{llcccc}
\toprule
\textbf{CFG}
& \textbf{Method}
& \textbf{Speedup}$\uparrow$
& \textbf{LPIPS}$\downarrow$
& \textbf{SSIM}$\uparrow$
& \textbf{PSNR}$\uparrow$ \\
\midrule

\multirow{4}{*}{1.5}
& DiCache      & 3.11$\times$ & 0.2042 & 0.8114 & 22.55 \\
& MagCache     & 3.18$\times$ & 0.1682 & 0.8268 & 22.42 \\
& FasterCache  & 2.09$\times$ & 0.2002 & 0.7814 & 21.58 \\
& \cellcolor{gray!10}\textbf{Ours}
& \cellcolor{gray!10}\textbf{3.38}$\times$
& \cellcolor{gray!10}\textbf{0.1430}
& \cellcolor{gray!10}\textbf{0.8553}
& \cellcolor{gray!10}\textbf{24.08} \\

\midrule

\multirow{4}{*}{3.5}
& DiCache      & 2.94$\times$ & 0.1431 & 0.8229 & 22.13 \\
& MagCache     & 3.18$\times$ & 0.1636 & 0.8177 & 21.46 \\
& FasterCache  & 2.09$\times$ & 0.2158 & 0.7593 & 20.88 \\
& \cellcolor{gray!10}\textbf{Ours}
& \cellcolor{gray!10}\textbf{3.43}$\times$
& \cellcolor{gray!10}\textbf{0.1262}
& \cellcolor{gray!10}\textbf{0.8565}
& \cellcolor{gray!10}\textbf{23.45} \\

\midrule

\multirow{4}{*}{5.5}
& DiCache      & 2.91$\times$ & 0.1378 & 0.8460 & 24.08 \\
& MagCache     & 3.18$\times$ & 0.1558 & 0.8423 & 23.13 \\
& FasterCache  & 2.01$\times$ & 0.2710 & 0.6969 & 19.11 \\
& \cellcolor{gray!10}\textbf{Ours}
& \cellcolor{gray!10}\textbf{3.35}$\times$
& \cellcolor{gray!10}\textbf{0.1153}
& \cellcolor{gray!10}\textbf{0.8779}
& \cellcolor{gray!10}\textbf{25.30} \\

\midrule

\multirow{4}{*}{7.5}
& DiCache      & 2.88$\times$ & 0.1163 & 0.8692 & 25.65 \\
& MagCache     & 3.18$\times$ & 0.1295 & 0.8673 & 24.81 \\
& FasterCache  & 2.00$\times$ & 0.2784 & 0.6695 & 19.34 \\
& \cellcolor{gray!10}\textbf{Ours}
& \cellcolor{gray!10}\textbf{3.37}$\times$
& \cellcolor{gray!10}\textbf{0.1031}
& \cellcolor{gray!10}\textbf{0.8867}
& \cellcolor{gray!10}\textbf{26.41} \\

\bottomrule
\end{tabular}
\end{minipage}
\hfill
% =========================================================
% Right: mechanism-validation tables
% =========================================================
\begin{minipage}[t]{0.48\textwidth}
\centering

% ---------------- Guided-risk ----------------
\scriptsize
\caption{\textbf{Validation of Guided-Risk Composition.}
The full formulation improves guided-error alignment and high-risk
reuse detection under an identical refresh budget.}
\label{tab:guided_risk_main}

\setlength{\tabcolsep}{3.0pt}
\renewcommand{\arraystretch}{1.04}

\begin{tabular}{lcccc}
\toprule
\textbf{Variant}
& \textbf{Pearson}$\uparrow$
& \textbf{AUPRC}$\uparrow$
& \textbf{P95}$\downarrow$
& \textbf{P99}$\downarrow$ \\
\midrule

No-cross
& 0.6082
& 0.5671
& 0.3247
& 0.5536 \\

\rowcolor{gray!10}
\textbf{Full Eq.~\ref{eq:online_guided_risk}}
& 0.7575
& 0.6799
& 0.2809
& 0.4635 \\

\midrule

Proxy mag. + true align.
& 0.8325
& 0.7601
& 0.2662
& 0.4145 \\

True mag. + est. align.
& 0.9358
& 0.9178
& 0.1976
& 0.2840 \\

Oracle geometry
& 1.0000
& 1.0000
& 0.1744
& 0.1874 \\

\bottomrule
\end{tabular}

\vspace{1.3em}

% ---------------- Propagation ----------------
\scriptsize
\caption{\textbf{Validation of Propagation-Aware Rescaling.}
Propagation weighting improves configuration-level prediction
of mean final LPIPS under repeated reuse.}
\label{tab:propagation_main}

\setlength{\tabcolsep}{5.0pt}
\renewcommand{\arraystretch}{1.04}

\begin{tabular}{lccc}
\toprule
\textbf{Predictor}
& \textbf{Spearman}$\uparrow$
& LOO $\boldsymbol{R^2}\uparrow$
& LOO \textbf{MAE}$\downarrow$ \\
\midrule

Unweighted
& 0.6571
& $-0.1773$
& 0.0248 \\

\textbf{Prop.-weighted}
& \textbf{0.9429}
& \textbf{0.5274}
& \textbf{0.0157} \\

\bottomrule
\end{tabular}

\end{minipage}

\vspace{-1em}
\end{table*}
\subsection{Ablation Studies}
\label{sec:ablation}

Unless otherwise specified, all ablations are conducted on FLUX.1-dev under the main experimental setting. We examine the contributions of the proposed components and reuse policy.

\paragraph{Contribution of each component.}

We ablate the two components of \textbf{RA-CFGCache}: CFG-aware guided-risk composition and propagation-aware rescaling. As shown in Table~\ref{tab:ablation_components}, each component improves fidelity at similar latency, and their combination performs best. These results support their complementary benefits in the evaluated setting.

\paragraph{Effect of the reused quantity.}
We compare different reuse targets under the same refresh count. As shown in Table~\ref{tab:ablation_reuse_action}, joint branch reuse achieves the best fidelity among the evaluated actions and larger latency savings by skipping recomputation in both branches. A separate parallel-execution study in Appendix~\ref{app:parallel_consistency}
compares independent branch-wise and joint controllers at matched
branch-evaluation counts, showing lower latency for joint control.
Appendix~\ref{app:reuse_action} further analyzes reuse-object selection
through CFG error geometry and computational savings.

\paragraph{Effect of CFG scale.}
Table~\ref{tab:ablation_cfgscale} shows that \textbf{RA-CFGCache} achieves a strong efficiency--fidelity trade-off across different guidance scales.
Across all tested CFG scales, it remains among the best methods in trade-off, while maintaining stable speedup and strong fidelity relative to competing baselines.
This indicates that risk-aligned cache control remains effective across the evaluated guidance strengths.

\paragraph{Validation of risk composition and propagation rescaling.}
Table~5 evaluates guided-risk composition on 37,400 reuse events
from 50 held-out FLUX.1-dev prompts disjoint from offline calibration.
Pearson correlation measures agreement with true guided-error magnitude,
while AUPRC measures detection of the top 20\% highest-error events.
Under the same 30\% refresh budget, P95 and P99 report true-error
percentiles among accepted reuse events.
The full composition improves correlation and high-risk discrimination
while reducing these tail errors.
Oracle-assisted variants substitute true branch-error magnitudes
and/or alignment for diagnosis only and are unavailable online.

Table~6 evaluates propagation weighting on six configurations:
three thresholds $\tau\in\{0.12,0.18,0.30\}$, each with and without
rescaling, using 50 held-out prompts per configuration.
For each trajectory, we sum the unweighted guided-risk contributions
or their propagation-weighted counterparts over all accepted reuse steps,
including across refresh intervals.
We then average each score and final LPIPS over prompts.
Spearman correlation is computed across the six configuration means.
For $R^2$ and MAE, we use leave-one-configuration-out affine regression:
each configuration's mean LPIPS is predicted from its mean risk
using a model fitted on the other five configurations.
Propagation weighting increases Spearman correlation from 0.6571
to 0.9429 and improves held-out $R^2$ and MAE, supporting its
predictive value at the configuration level.

\begin{table*}[t]
\centering

% ==================== Left: rho robustness ====================
\begin{minipage}[t]{0.49\textwidth}
\centering
\scriptsize

\caption{\textbf{Robustness of calibration on $\bar{\rho}_{t_a,t_b}$.}
Downstream fidelity and speedup remain relatively stable across the tested calibration-set sizes and prompt subsets.}
\label{tab:robust_rho}

\setlength{\tabcolsep}{4.3pt}
\renewcommand{\arraystretch}{1.04}

\begin{tabular}{lccccc}
\toprule
\textbf{Strategy}
& \textbf{\#}
& \textbf{LPIPS}$\downarrow$
& \textbf{SSIM}$\uparrow$
& \textbf{PSNR}$\uparrow$
& \textbf{Speedup}$\uparrow$ \\
\midrule

Random
& 1
& 0.1345
& 0.8472
& 23.01
& 3.47$\times$ \\

Random
& 10
& 0.1301
& 0.8507
& 23.28
& 3.46$\times$ \\

Outlier
& 10
& 0.1326
& 0.8495
& 23.19
& 3.45$\times$ \\

Random
& 50
& 0.1308
& 0.8500
& 23.26
& 3.46$\times$ \\

Random
& 100
& \textbf{0.1297}
& \textbf{0.8513}
& \textbf{23.29}
& 3.46$\times$ \\

\bottomrule
\end{tabular}

\end{minipage}
\hfill
% ==================== Right: gt robustness ====================
\begin{minipage}[t]{0.49\textwidth}
\centering
\scriptsize

\caption{\textbf{Robustness of calibration on $g_t$.}
Downstream fidelity and speedup remain relatively stable across the tested calibration-set sizes and prompt subsets.}
\label{tab:robust_gt}

\setlength{\tabcolsep}{4.3pt}
\renewcommand{\arraystretch}{1.04}

\begin{tabular}{lccccc}
\toprule
\textbf{Strategy}
& \textbf{\#}
& \textbf{LPIPS}$\downarrow$
& \textbf{SSIM}$\uparrow$
& \textbf{PSNR}$\uparrow$
& \textbf{Speedup}$\uparrow$ \\
\midrule

Random
& 1
& 0.1426
& 0.8380
& 22.94
& 3.46$\times$ \\

Random
& 10
& 0.1329
& 0.8497
& 23.21
& 3.46$\times$ \\

Outlier
& 10
& 0.1448
& 0.8379
& 22.80
& 3.46$\times$ \\

Random
& 50
& 0.1297
& 0.8513
& 23.29
& 3.46$\times$ \\

Random
& 100
& \textbf{0.1262}
& \textbf{0.8579}
& \textbf{23.80}
& 3.46$\times$ \\

\bottomrule
\end{tabular}

\end{minipage}

\vspace{-0.8em}
\end{table*}
\paragraph{Validation of alignment and propagation calibration.}
We further evaluate the robustness of the calibrated quantities by varying only the calibration prompt subset while keeping the evaluation protocol and controller unchanged. As shown in Tables~\ref{tab:robust_rho} and~\ref{tab:robust_gt}, downstream fidelity and speedup remain relatively stable across the tested random and outlier calibration subsets and calibration-set sizes. These results indicate robustness to the tested calibration-set variations, rather than invariance of the calibrated alignment or propagation curves themselves.

\section{Conclusion}
We revisit training-free diffusion caching under CFG as guided-error control. 
\textbf{RA-CFGCache} aligns reuse decisions with the guided prediction error rather than branch-local criteria, and further rescales this risk by its timestep-dependent propagation to the final sample. 
By combining CFG-aware guided-risk composition, propagation-aware rescaling, and threshold-based scheduling, \textbf{RA-CFGCache} consistently improves the efficiency--fidelity trade-off over the evaluated training-free caching baselines on text-to-image and text-to-video diffusion transformers.

\newpage
\section*{Acknowledgments and Disclosure of Funding}

This work was supported by the National Natural Science Foundation of China
(Grant Nos.~62525103 and 62271281) and the Beijing Natural Science Foundation
(Grant No.~L247026). The authors declare no competing financial interests.

\bibliographystyle{unsrtnat}

%%%%%%%%%%%%%%%%%%%%%%%%%%%%%%%%%%%%%%%%%%%%%%%%%%%%%%%%%%%
\newpage
\appendix
\section*{Appendix}
\section{Limitations}
\label{app:limitations}
While \textbf{RA-CFGCache} shows consistent gains under the evaluated
classifier-free guidance settings, the current formulation still has several
methodological limitations.

First, the propagation gain is an empirical timestep-dependent surrogate
motivated by first-order propagation analysis, rather than a complete
model of how local guided error affects the final output. It is estimated
from isolated single-step perturbations and therefore does not explicitly
model higher-order interactions among errors introduced by consecutive
reuse decisions.

Second, although the cumulative scheduler goes beyond purely local
thresholding, it remains an online approximation to the underlying
sequential control problem and does not explicitly optimize long-horizon
interactions among reuse decisions or provide a formal optimality guarantee.

Third, the alignment and propagation statistics rely on offline calibration
for a given inference configuration, and their transfer across substantially
different models, samplers, timestep schedules, or guidance settings may
require validation or recalibration. Improving such transferability while
retaining lightweight online control is an important direction for future work.
\section{Implementation Details}
\label{app:implementation_details}
\subsection{Details of the Compared Methods}

\textbf{TeaCache}~\citep{Liu_2025_CVPR} is a training-free,
architecture-agnostic caching method that exploits the correlation between
timestep embedding changes and model output differences across adjacent steps.
It activates caching through an accumulated error-based thresholding strategy.
We adapt its official implementation to our evaluation repository and use
model-specific representative thresholds as reported in
Table~\ref{tab:main_results}: $\tau=0.4$ for FLUX.1-dev,
$\tau=0.15$ for Wan2.1-T2V-1.3B, and $\tau=0.2$ for CogVideoX-2B.

\textbf{TaylorSeer}~\citep{Liu_2025_ICCV} is a training-free acceleration
method based on the ``cache-then-forecast'' paradigm. Instead of directly
reusing cached features, it predicts future features using Taylor-series-based
extrapolation. We adapt the released implementation and use the reported
configuration with activation/cache interval $N=3$ and Taylor expansion order $O=1$.

\textbf{DiCache}~\citep{bu2025dicacheletdiffusionmodel} is a training-free
caching method that reuses diffusion outputs through a lightweight temporal
reuse policy. We adapt its official implementation and use probe depth $m=2$
and threshold $\tau=0.4$ across the reported models.

\textbf{MagCache}~\citep{ma2026magcache} controls acceleration through a
maximum skip length and an accumulated error threshold. We adapt its official
implementation, use the default fast-mode maximum skip length $K=4$, and use
model-specific thresholds as reported in Table~\ref{tab:main_results}:
$\tau=0.24$ for FLUX.1-dev, $\tau=0.1$ for Wan2.1-T2V-1.3B, and
$\tau=0.08$ for CogVideoX-2B. The magnitude-ratio lists are obtained from
the official repository or the corresponding official calibration procedure
in \texttt{diffusers}.

\textbf{HiCache}~\citep{feng2026hicachepluginscaledhermiteupgrade} is a
higher-order caching method that improves temporal reuse using a scaled
Hermite-style update. We adapt the released implementation and use cache
interval $\mathcal{N}=3$ and update order $\mathcal{O}=1$.

\textbf{FasterCache}~\citep{lv2025fastercachetrainingfreevideodiffusion}
is a training-free CFG acceleration method that exploits redundancy between
the conditional and unconditional branches and reuses self-attention
computations. We use the released implementation for CogVideoX and
backend-specific adapters for FLUX and Wan, without changing the model,
sampling steps, resolution, or guidance. For FLUX and Wan, we set
\(I_{\Delta}=8\) and \(I_{\mathrm{attn}}=8\), use an extrapolation coefficient
of \(0.3\) and \(t_0=0.5\), and disable frequency gains. For CogVideoX, we
retain the released frequency compensation, set
\(I_{\Delta}=I_{\mathrm{attn}}=6\), and keep the extrapolation coefficient at
\(0.3\). Configurations are selected using calibration prompts only and frozen
for evaluation; the FLUX configuration is transferred to Wan without further
tuning.

We additionally include a reduced-step vanilla baseline,
\textbf{Vanilla} $(T=25)$, which uses the same model and sampling
configuration as the full vanilla baseline except for halving the denoising
steps from $T=50$ to $T=25$. For all baselines, we use the same model
checkpoints, prompts, random seeds, and hardware as in the main experiments.
The 50-step schedule is retained except for the explicitly identified
reduced-step baseline. For methods with multiple thresholds or operating
modes, operating points are selected to target the reported speedup ranges;
the exact configurations are listed in Table~\ref{tab:main_results}.

\subsection{Branch-wise Proxy Definitions Used by RA-CFGCache}
\label{app:proxy_details}
\textbf{RA-CFGCache} does not introduce a new branch-local error estimator.
Instead, it takes the control signal of a base caching method as the
branch-wise proxy in Eq.~\ref{eq:online_guided_risk}. For each branch
$j\in\{u,c\}$, we denote this signal by
$\hat r_j(t_a,t_b)$. The proxy is evaluated independently for the two CFG
branches before guided-risk composition. Importantly, the proxy families
used in our experiments differ in whether they depend on the current
sample: Tea-style and DiCache-style proxies use online feature signals,
whereas the MagCache-style proxy is derived from fixed offline-calibrated
magnitude-ratio statistics.

\paragraph{MagCache-style proxy.}
For the FLUX.1-dev main experiments, we use the fixed magnitude-ratio
statistics of MagCache~\citep{ma2026magcache}. Let $\mu_j(k)$ denote the
pre-calibrated magnitude ratio associated with branch $j$ at execution
step $k$. For a candidate reuse interval from anchor $a$ to current step
$b$, we form the cumulative ratio
\begin{equation}
M_j(a,b)=\prod_{k=a+1}^{b}\mu_j(k),
\end{equation}
and use the corresponding deviation
\begin{equation}
\hat r_j(t_a,t_b)=\left|1-M_j(a,b)\right|
\end{equation}
as the branch-wise proxy supplied to the CFG-aware composition.
The magnitude-ratio sequence is fixed before evaluation and no
sample-specific feature measurement is used to construct this proxy.
Thus, under a fixed model, timestep schedule, calibration table, and
threshold, the MagCache-style proxy should not be interpreted as an
instance-adaptive estimator.

\paragraph{TeaCache-style proxy.}
Following the core design of TeaCache~\citep{Liu_2025_CVPR}, we construct
a sample-dependent control signal from the relative change of the
timestep-modulated feature. For each CFG branch $j$, let $m_j(k)$ denote
the current modulated feature and $m_j^{\mathrm{ref}}$ the stored reference
feature maintained by the corresponding TeaCache-compatible implementation.
We compute
\begin{equation}
d_j(k)=
\frac{
\operatorname{mean}\!\left|m_j(k)-m_j^{\mathrm{ref}}\right|
}{
\operatorname{mean}\!\left|m_j^{\mathrm{ref}}\right|+10^{-6}
}.
\end{equation}
If the reference mean absolute magnitude is zero, the raw change signal
$d_j(k)$ is set to zero before rescaling. We then apply the same
model-specific polynomial $p_{\mathrm{model}}$ as the corresponding
TeaCache-compatible baseline and define
\begin{equation}
\hat r_j(k)=\max\!\left\{p_{\mathrm{model}}(d_j(k)),\,0\right\},
\end{equation}
with no upper clipping. Thus, the proxy preserves TeaCache's basic
relative-change and polynomial-rescaling construction while serving as
the branch-wise online signal for \textbf{RA-CFGCache}. The resulting proxy is
sample-dependent; \textbf{RA-CFGCache} replaces the original TeaCache decision
rule with the guided-risk composition and cumulative controller described
in Section~\ref{sec:ra_cfgcache}.

\paragraph{DiCache-style proxy.}
For the DiCache-style compatibility experiments, we use the online
shallow-probe signal of DiCache~\citep{bu2025dicacheletdiffusionmodel}.
For each branch $j$, the first $m$ Transformer blocks are evaluated and
the relative $\ell_1$ change of the resulting shallow feature is used as
the branch-wise proxy:
\begin{equation}
\hat r_j(k)=
\frac{\|z^{(m)}_j(k)-z^{(m)}_j(k-1)\|_1}
     {\|z^{(m)}_j(k-1)\|_1+\epsilon},
\end{equation}
where $z^{(m)}_j(k)$ denotes the feature after the probe depth $m$.
We use $m=2$ in the reported experiments. Because the probe feature is
computed from the current branch trajectory, this proxy is
sample-dependent. The probe computation is included in the reported
end-to-end latency.

\paragraph{Normalization and accumulation.}
No additional sample-wise normalization is introduced by \textbf{RA-CFGCache}
beyond the normalization or calibration already used by the corresponding
base proxy. The two branch scores are composed through
Eq.~\ref{eq:online_guided_risk}; propagation rescaling is then applied
as in Eq.~\ref{eq:online_risk}. The interval-level accumulation
variable $R$ in Eq.~\ref{eq:refresh_rule} belongs to \textbf{RA-CFGCache} and is
separate from the definition of the branch-wise proxy itself. 

For the Tea-style and DiCache-style estimators, the base proxy is an
adjacent-step online change signal rather than a direct estimate of the
anchor-to-target prediction error. \textbf{RA-CFGCache} uses this signal as the
current branch-wise contribution within the active reuse interval, while
the cross-branch alignment statistic is indexed by the current cache
anchor and target step.

\subsection{Calibration Details}

The offline calibration procedure estimates two quantities used by
\textbf{RA-CFGCache}: the anchor--target alignment statistic
$\bar{\rho}_{t_a,t_b}$ and the timestep-dependent propagation gain $g_t$.
The resulting calibration quantities are fixed during evaluation.

\noindent\textbf{Calibration of $\bar{\rho}$.}
We first run full, non-accelerated inference and record the hidden-space
residual cached by the corresponding implementation at each timestep.
Let $h^c_{t_k}$ and $h^u_{t_k}$ denote the conditional and unconditional
residual tensors at sampler step $k$, whose diffusion time is $t_k$.
These hidden-space quantities are distinct from the final branch predictions
$c_{t_k}$ and $u_{t_k}$ used in the CFG error formulation.

For FLUX.1-dev, the cached residual is the change in image-token hidden states
across the complete double-stream and single-stream Transformer stack:
the hidden states immediately before \texttt{final\_layer} minus the image
hidden states after \texttt{img\_in} and before the block stack. Only the
image-stream residual is cached. For Wan2.1-T2V-1.3B, it is the change in
video hidden states across the complete Transformer stack, before
\texttt{norm\_out}, modulation, and \texttt{proj\_out}. For CogVideoX-2B,
the corresponding video-token and text-token residuals across the full
Transformer stack are cached separately before the final normalization and
projection layers.

For anchor step $a$ and target step $b>a$, we construct
\begin{equation}
d_c(t_a,t_b)
=
\operatorname{vec}\!\left(h^c_{t_a}-h^c_{t_b}\right),
\qquad
d_u(t_a,t_b)
=
\operatorname{vec}\!\left(h^u_{t_a}-h^u_{t_b}\right),
\end{equation}
where $\operatorname{vec}(\cdot)$ flattens the tensor into a vector.
We average their cosine similarity over calibration samples to obtain
$\bar{\rho}_{t_a,t_b}$. Indexed by execution order, $\bar{\rho}$ is an
upper-triangular anchor--target matrix. It describes cross-branch alignment
of hidden-residual drift and serves as a calibrated proxy for the
prediction-error alignment appearing in the exact CFG error identity.
It is not itself the actual prediction-error alignment for every online
reuse event.

\noindent\textbf{Calibration of propagation-aware rescaling.}
Propagation calibration uses an isolated one-step prediction perturbation.
At a selected execution step, both the conditional and unconditional
branch predictions used by the sampler are replaced by their corresponding
predictions from the preceding execution step. The reused predictions are
then combined using the same CFG rule as in normal inference. All other
steps use fresh computation on their current latent states, so each perturbed
trajectory contains only one injected reuse event.

The perturbed and reference trajectories use the same prompt, initial noise,
and sampling randomness. We measure the discrepancy between the perturbed
and fresh guided predictions at the injected step and the resulting
deviation between their final latents before VAE decoding. For each
calibration sample, we form the ratio between final and local discrepancies,
and then average these per-sample ratios at each probed step. Samples with
invalid or vanishing local discrepancy are excluded from the ratio average.

We probe one eligible step every $\Delta_g=3$ execution steps, excluding
the first denoising step because no preceding prediction is available.
The resulting timestep-dependent mean gains are fitted using the power-law
surrogate in Eq.~\ref{eq:gain_fit} by least squares. Fitted gains are
lower-bounded at zero when used for risk rescaling.

\noindent\textbf{Offline calibration cost and online overhead.}
Let $C$ denote the number of calibration prompts, $T$ the number of denoising
steps, $F_m$ the cost of one full two-branch denoising evaluation, and $D_m$
the size of a saved hidden residual tensor. Alignment calibration consists
of full-trajectory collection followed by pairwise reduction, with complexity
\[
\mathcal{O}(CTF_m)+\mathcal{O}(CT^2D_m).
\]
The quadratic term operates only on stored residual tensors and does not
require $\mathcal{O}(T^2)$ diffusion-model evaluations. Propagation
calibration probes one perturbation anchor every $\Delta_g$ steps, giving
an approximate rollout cost
\[
\mathcal{O}(CT^2F_m/\Delta_g).
\]

For a 100-prompt FLUX.1-dev calibration configuration with $T=50$,
alignment and propagation calibration require approximately $0.60$ and
$8.3$ GPU-hours, respectively, for $8.9$ GPU-hours in total. Raw calibration
traces temporarily occupy approximately 11 GB at $1024\times1024$ resolution
and 42 GB at $2048\times2048$, whereas the final calibrated statistics occupy
only tens of kilobytes. Calibration is entirely offline.

At test time, the \textbf{RA-CFGCache} control layer adds only constant-time lookup
and scalar composition/rescaling beyond the computations required by the
base proxy, and introduces no additional diffusion-model forward pass.
This online overhead is included in the reported end-to-end latency.

\noindent\textbf{Calibration reuse and transfer.}
Once calibrated, $\bar{\rho}_{t_a,t_b}$ and $g_t$ are fixed and reused
across evaluation prompts, random seeds, and operating thresholds $\tau$.
The alignment statistic is calibrated separately for the evaluated CFG
scales, while the same propagation prior is reused across those scale
settings. Changes to the model or sampler/noise schedule require
recalibration; changes to the timestep grid or prompt domain should be
validated on the new setting and recalibrated when necessary.

Separate calibration-size ablations show stable performance when reducing
alignment calibration to 50 prompts or propagation calibration to 10 prompts
in the evaluated setting. These individual ablations do not establish the
performance of a jointly reduced 50/10-prompt configuration.

\section{Mechanism Analysis}
\label{app:analysis}

\subsection{CFG Error Geometry under Classifier-Free Guidance}
\label{app:cfg_error_geometry}

\begin{figure*}[t]
    \centering
    % 放大子图宽度，同时保留间距
    \begin{subfigure}[t]{0.33\textwidth}
        \centering
        % 宽度改为 \textwidth，利用子图的全部空间
        \includegraphics[width=\textwidth]{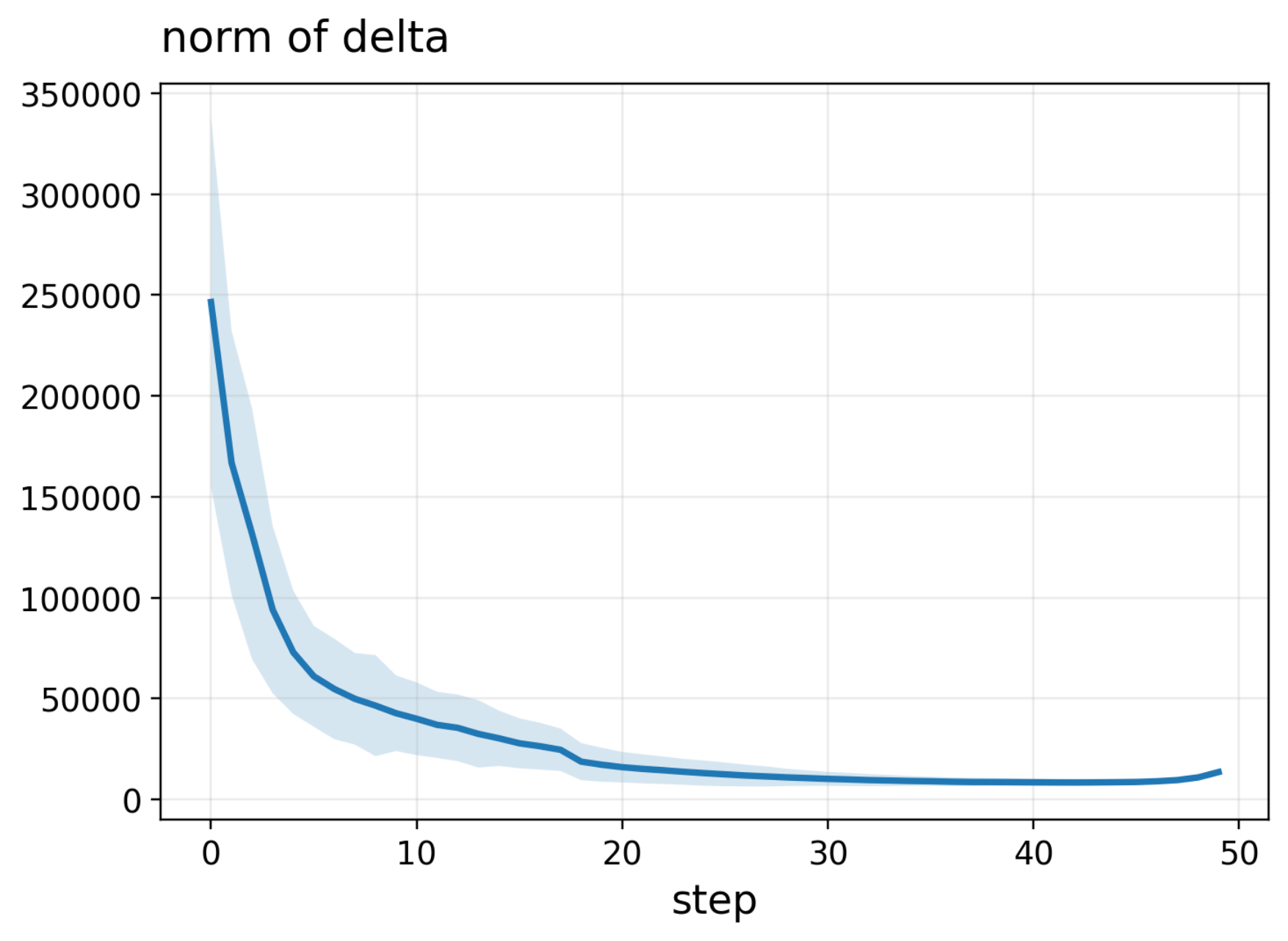}
        \caption{$\|\Delta_t\|_1$}
        \label{fig:delta_norm}
    \end{subfigure}\hfill
    \begin{subfigure}[t]{0.33\textwidth}
        \centering
        \includegraphics[width=\textwidth]{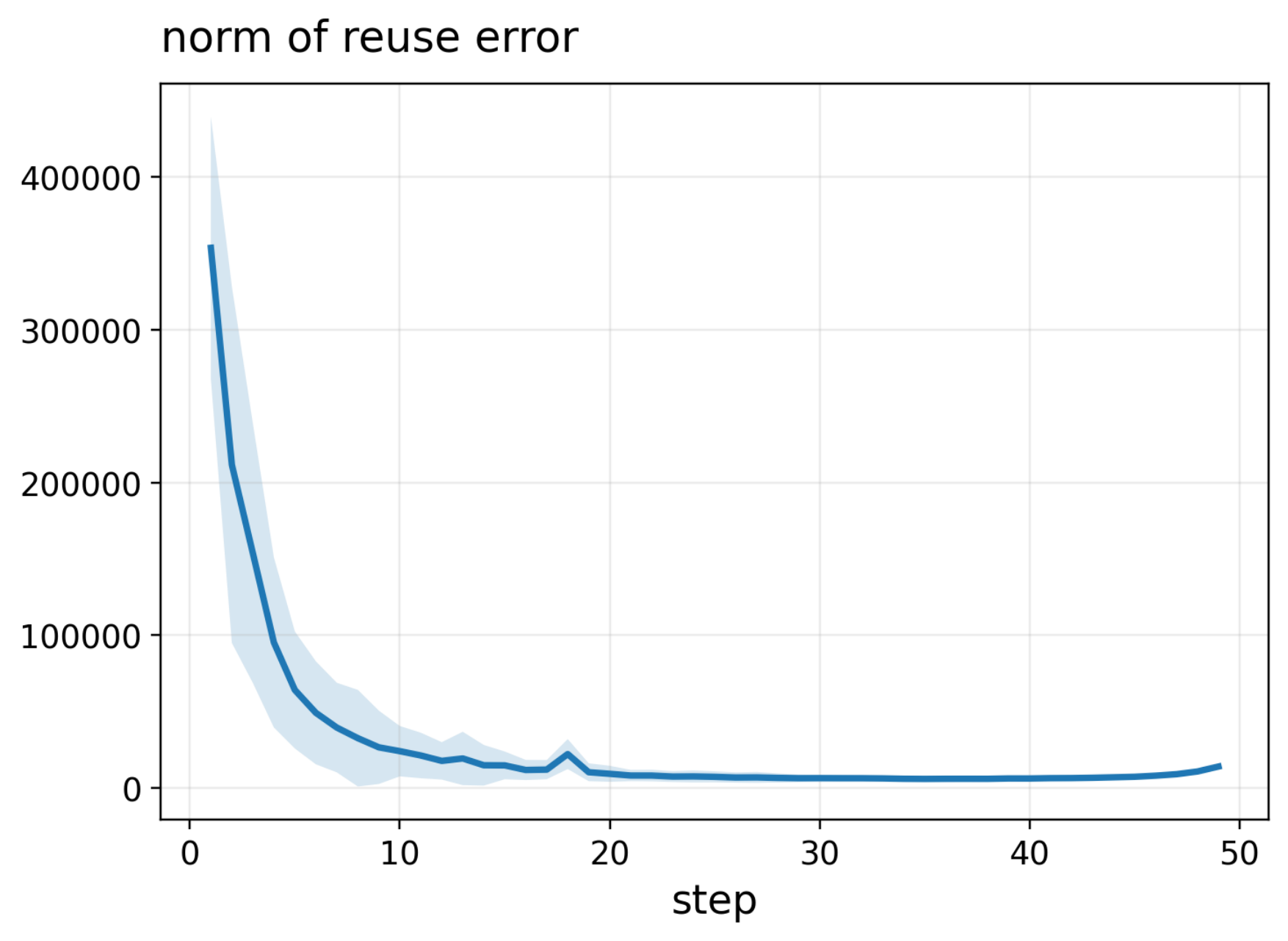}
        \caption{$\|\Delta_t-\Delta_{t-1}\|_1$}
        \label{fig:delta_diff}
    \end{subfigure}\hfill
    \begin{subfigure}[t]{0.33\textwidth}
        \centering
        \includegraphics[width=\textwidth]{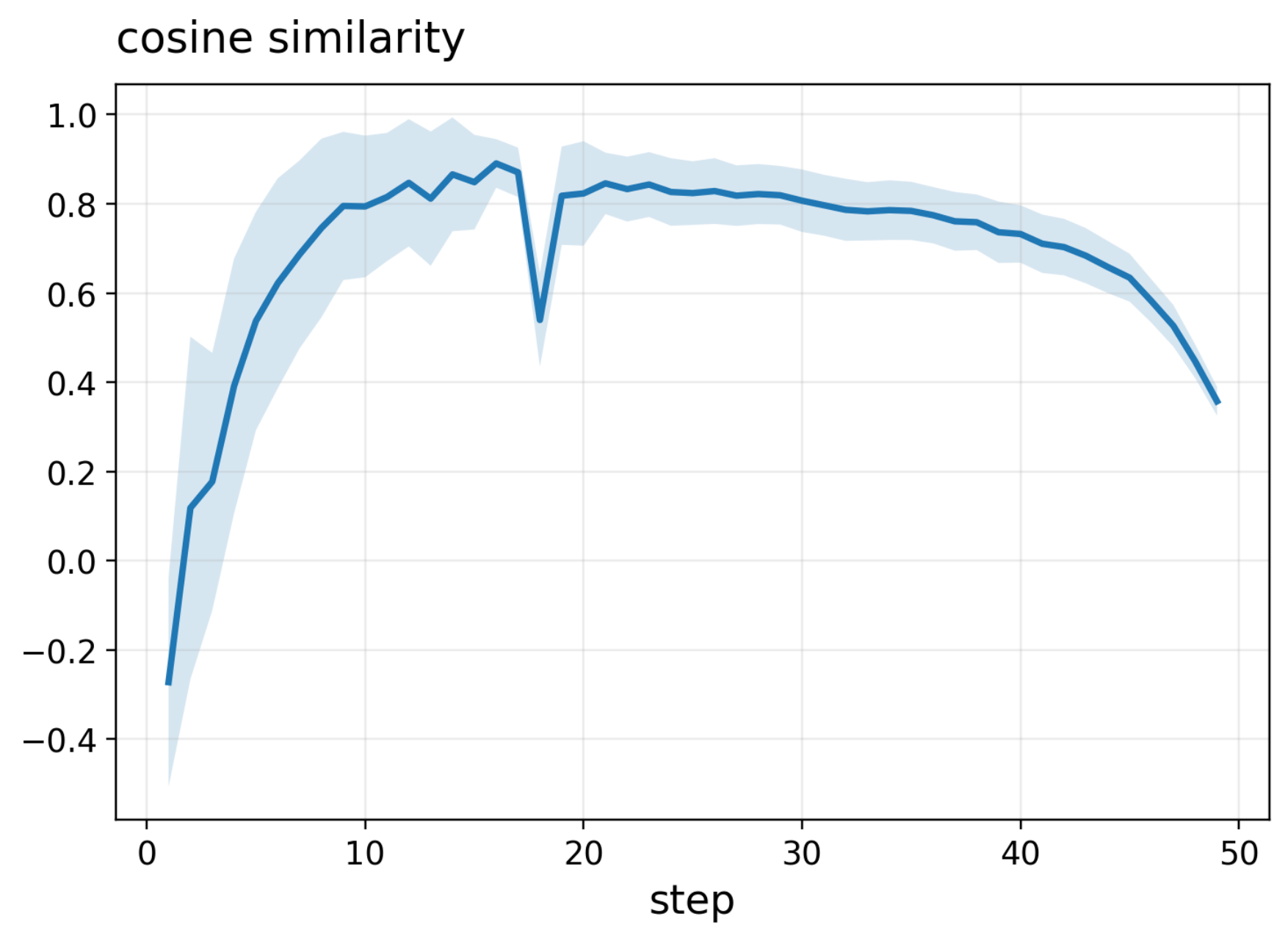}
        \caption{$\cos(\Delta_t,\Delta_{t-1})$}
        \label{fig:delta_cos}
    \end{subfigure}
    \caption{\textbf{Temporal dynamics of the conditional correction signal under CFG.}
From left to right: the norm of $\Delta_t$, the adjacent-step variation $\|\Delta_t-\Delta_{t-1}\|_1$, and the cosine similarity $\cos(\Delta_t,\Delta_{t-1})$ between consecutive correction vectors.}
    \vspace{-2em}\label{fig:delta_dynamics}
\end{figure*}

\noindent\textbf{Branch-Error Alignment and Conditional-Correction Drift.}

We examine how the evolution of the conditional correction relates to
branch-error alignment. Let
\begin{equation}
\Delta_t=c_t-u_t,\qquad
\tilde\Delta_t=\tilde c_t-\tilde u_t.
\end{equation}
Here $\Delta_t$ is the difference between the fresh conditional and
unconditional final branch predictions. For arbitrary branch approximations,
the following identity holds:
\begin{equation}
\delta c_t
=
\delta u_t+(\tilde\Delta_t-\Delta_t).
\end{equation}
Thus, for nonzero branch errors,
\begin{equation}
\rho_t
=
\cos\!\left(
\delta u_t,\,
\delta u_t+(\tilde\Delta_t-\Delta_t)
\right).
\end{equation}

For direct prediction reuse from anchor $t_a$,
$\tilde\Delta_t=\Delta_{t_a}$. In particular, adjacent-step prediction
reuse gives the correction drift
$\Delta_{t_{k-1}}-\Delta_{t_k}$ at execution step $k$.
For intermediate-feature or residual caching, this temporal-drift
interpretation serves as a proxy rather than an exact identity for the
induced prediction error.

Alignment depends on both the magnitude and direction of the correction
error relative to $\delta u_t$. If the correction error is small relative
to a nonzero unconditional error, the two branch errors remain closely
aligned. A larger correction error may preserve or weaken alignment depending
on its direction. Neither the correction norm alone nor its unnormalized
temporal variation determines $\rho_t$.

Figure~\ref{fig:delta_dynamics} examines the correction norm,
adjacent-step variation, and adjacent-step cosine similarity using the
final branch predictions $\Delta_t=c_t-u_t$. In the displayed setting,
the correction norm decreases over sampling, while its temporal variation
is large early, smaller in the middle, and rises again near the end.
These observations describe changing correction dynamics and are consistent
with timestep-dependent branch alignment. They do not, by themselves,
establish that correction-norm decay causes increased alignment; the
relative drift magnitudes and directions also matter.

\vspace{0.5em}
\noindent\textbf{Risk Mismatch and Refresh Redistribution under CFG.}

\begin{figure*}[t]
    \centering
    \begin{subfigure}[t]{0.48\textwidth}
        \centering
        \includegraphics[width=\textwidth]{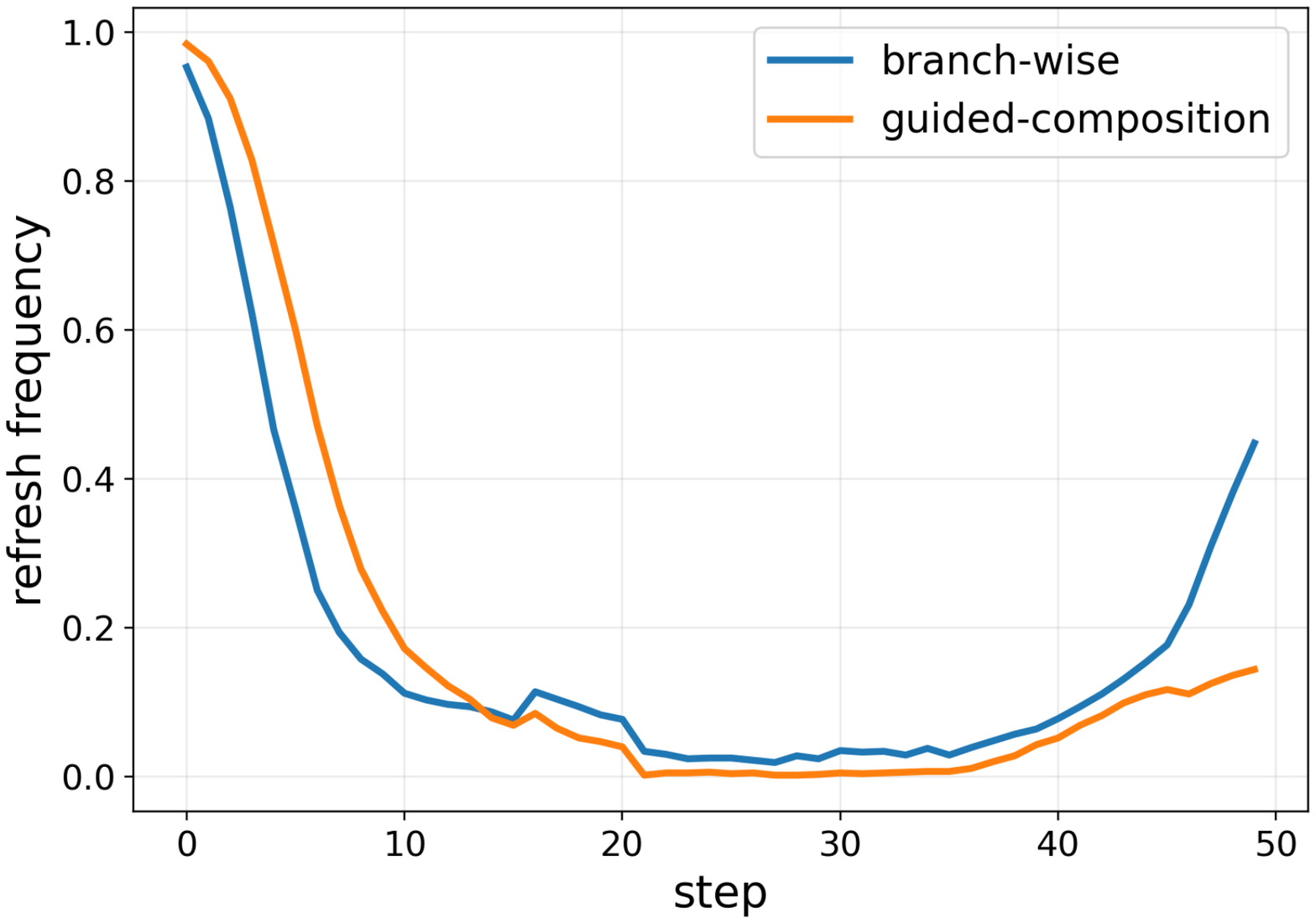}
    \end{subfigure}\hfill
    \begin{subfigure}[t]{0.48\textwidth}
        \centering
        \includegraphics[width=\textwidth]{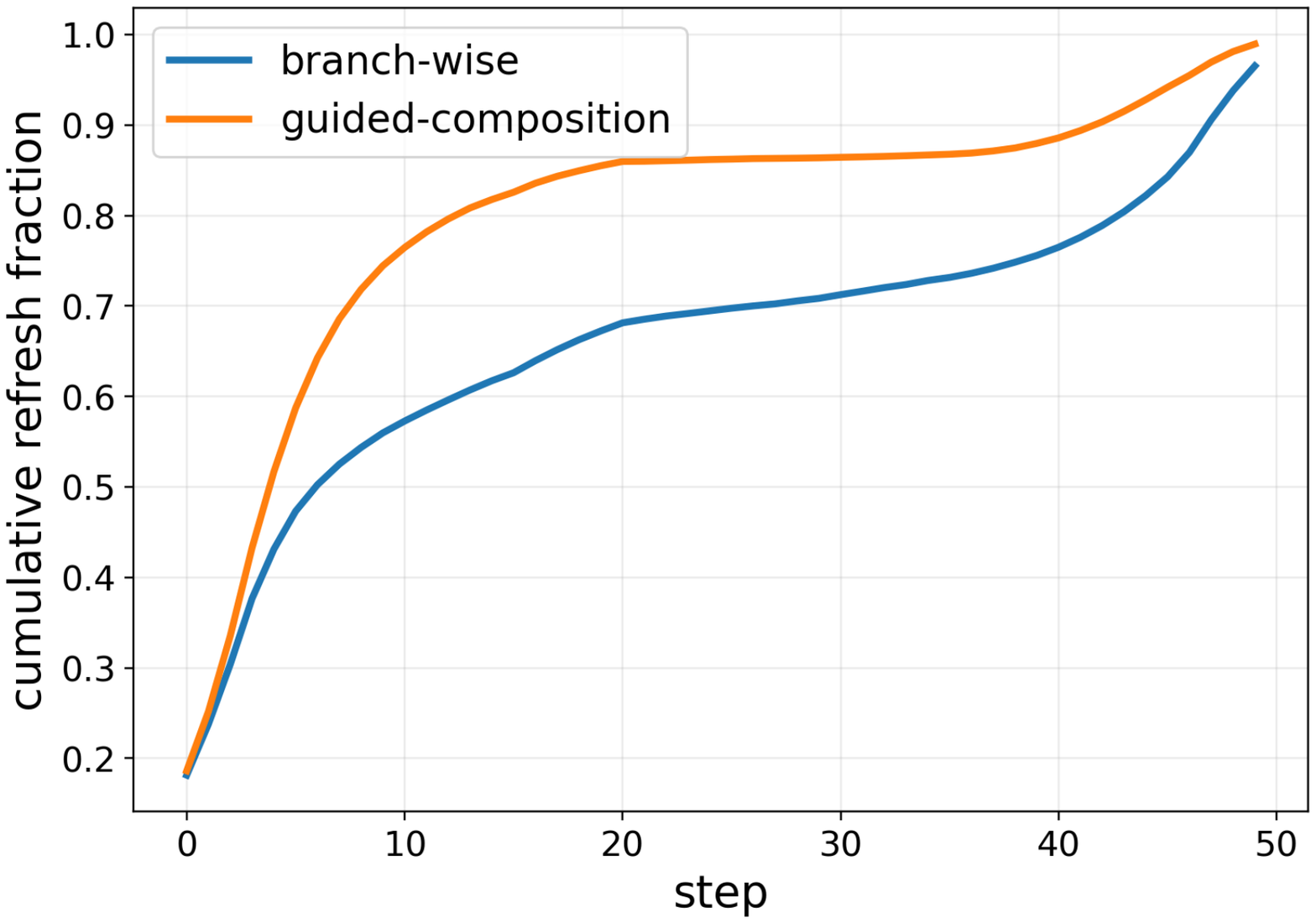}
    \end{subfigure}
    \caption{\textbf{Refresh redistribution under matched budget.}
    Left: per-step refresh frequency of the branch-wise and guided-composition controllers. Right: cumulative refresh fraction over timesteps. Guided-composition shifts refresh decisions toward earlier timesteps, showing that preserving CFG error geometry changes the actual scheduling behavior rather than only the local risk value.}
    \vspace{-1.5em}\label{fig:refresh_redistribution}
\end{figure*}

The calibrated alignment term changes the temporal profile of the proxy
score. We examine whether this change also affects refresh allocation
through an offline replay on precomputed branch traces. The replay isolates
scheduling behavior and does not measure end-to-end generation quality.

We compare two cumulative-threshold controllers using the same base proxy,
reuse action, and replay protocol: a no-cross-term controller and a
guided-composition controller that includes the calibrated cross-term.
Their thresholds are tuned separately to match the average refresh count.

Figure~\ref{fig:refresh_redistribution} shows that guided composition shifts
refreshes toward earlier timesteps at the matched average count. The mean
refresh step decreases from 16.87 to 10.03, while the fraction of refreshes
assigned to the first 20\% of steps increases from 56.1\% to 75.0\%.
These results show that the calibrated cross-term changes the allocation
of refresh decisions in the replay setting. Whether this redistribution
improves output fidelity is evaluated separately through controlled
quality comparisons.

\subsection{Propagation-aware Rescaling under Diffusion Dynamics}
\label{app:local_final_analysis}

The guided error describes a local prediction perturbation, whereas its
effect on the final latent also depends on the sampler and the remaining
denoising dynamics. We use a first-order analysis to motivate the empirical
propagation prior.

\paragraph{First-order propagation analysis.}
To distinguish execution order from diffusion time, let $z_k$ denote the
latent entering sampler step $k$, with corresponding diffusion time $t_k$.
Write one complete update as
\[
z_{k+1}
=
F_k\!\left(z_k,p_{t_k}(z_k)\right).
\]
Let $e_k$ be a guided-prediction perturbation introduced only at step $k$.
Evaluating derivatives along the unperturbed trajectory gives
\[
\delta z_{k+1}
\approx
B_ke_k,
\qquad
B_k
:=
\frac{\partial F_k}{\partial p},
\]
and define the effective Jacobian of a subsequent complete step as
\[
A_j
:=
\frac{d}{dz_j}
F_j\!\left(z_j,p_{t_j}(z_j)\right)
=
\frac{\partial F_j}{\partial z_j}
+
\frac{\partial F_j}{\partial p}
\frac{\partial p_{t_j}}{\partial z_j}.
\]
The final perturbation is therefore approximated by the ordered product
\[
\delta z_{T+1}^{(k)}
\approx
J_ke_k,
\qquad
J_k
:=
A_TA_{T-1}\cdots A_{k+1}B_k,
\]
where the product preceding $B_k$ is the identity for the final update.
For multistep samplers, the state can be augmented to include the history
required by the update. For stochastic samplers, the comparison conditions
on shared sampling randomness.

For a nonzero perturbation, the directional amplification factor is
\[
g_{t_k}(e_k)
:=
\frac{\|\delta z_{T+1}^{(k)}\|_2}{\|e_k\|_2}
\approx
\frac{\|J_ke_k\|_2}{\|e_k\|_2}.
\]
This factor depends on both the trajectory and the perturbation direction.
We summarize this behavior using a timestep-dependent calibration average,
\[
\bar g_{t_k}
:=
\mathbb{E}_{\mathrm{cal}}
\left[
\frac{\|\delta z_{T+1}^{(k)}\|_2}{\|e_k\|_2}
\right].
\]
In practice, we do not explicitly compute the Jacobian chain. Instead,
the calibration uses complete perturbed rollouts and measures the downstream
effect of an isolated one-step perturbation. The fitted $\hat g_{t_k}$ is
therefore an empirical timestep-dependent surrogate motivated by the
first-order analysis rather than a direct Jacobian estimate.

\paragraph{From single-step sensitivity to cache control.}
Under repeated reuse, earlier errors change the states at which later model
evaluations occur, and perturbations may interact. The single-step analysis
therefore does not provide an exact decomposition of final error under
arbitrary multi-step caching.

We use the fitted gain as a lightweight timestep-dependent prior. Given the
guided-risk contribution $\hat r^{\mathrm{guided}}_{t_a,t_b}$ for a
candidate reuse decision from anchor $t_a$ to target $t_b$, we define
\[
\Delta\hat r_{t_a,t_b}
=
\max\left(
0,
\hat g_{t_b}\hat r^{\mathrm{guided}}_{t_a,t_b}
\right).
\]
When reuse is accepted, the accumulated control score is updated as
\[
R
\leftarrow
R+\Delta\hat r_{t_a,t_b}.
\]
Thus, propagation weighting is applied before accumulation. The resulting
$R$ is an online control statistic, not a bound on the final output
deviation.

\subsection{Reuse Object Selection: Why Joint Branch Reuse Is a Strong Default}
\label{app:reuse_action}

\begin{figure*}[t]
    \centering
    \begin{subfigure}[t]{0.5\textwidth}
        \centering
        \includegraphics[width=\textwidth]{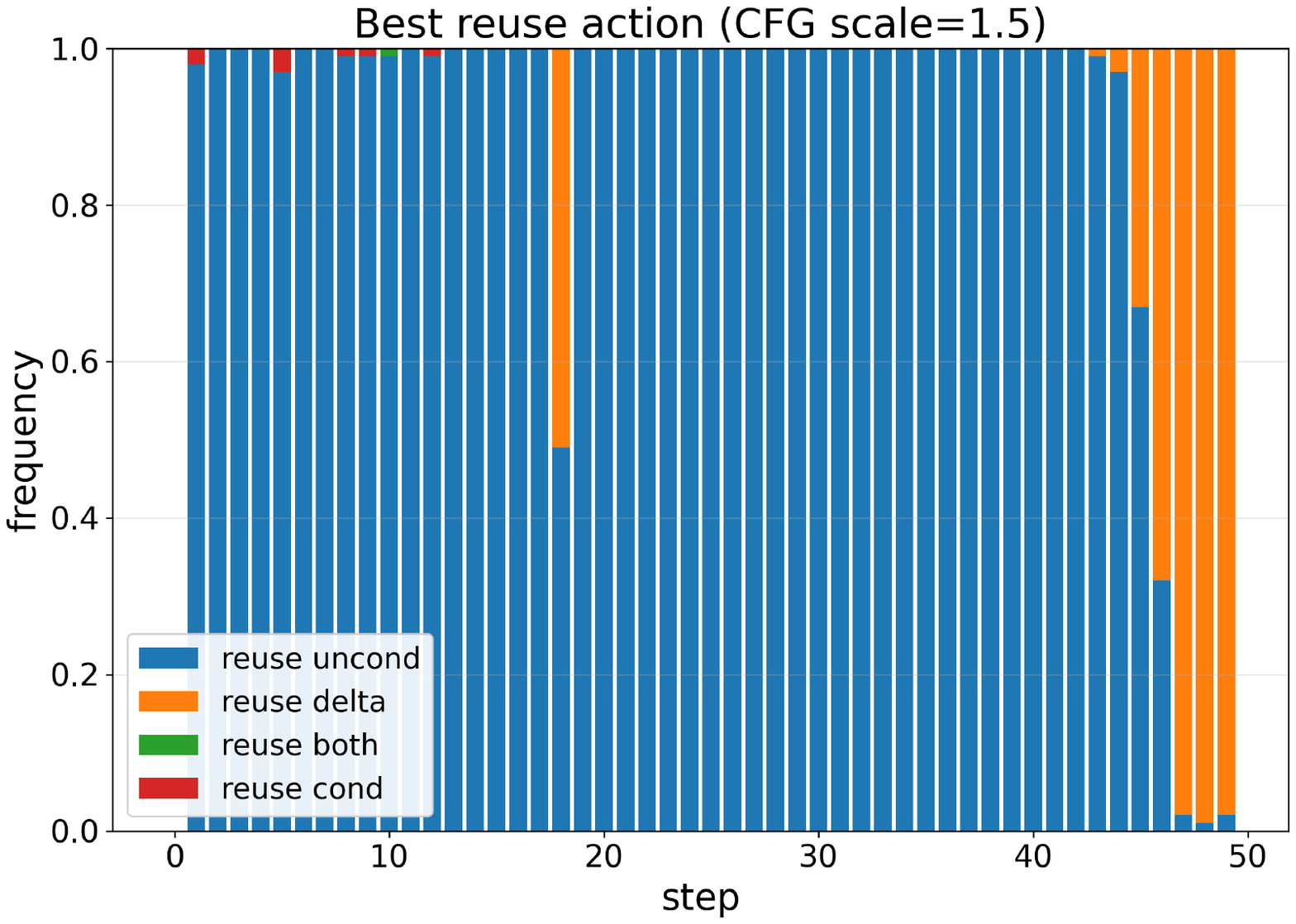}
    \end{subfigure}\hfill
    \begin{subfigure}[t]{0.5\textwidth}
        \centering
        \includegraphics[width=\textwidth]{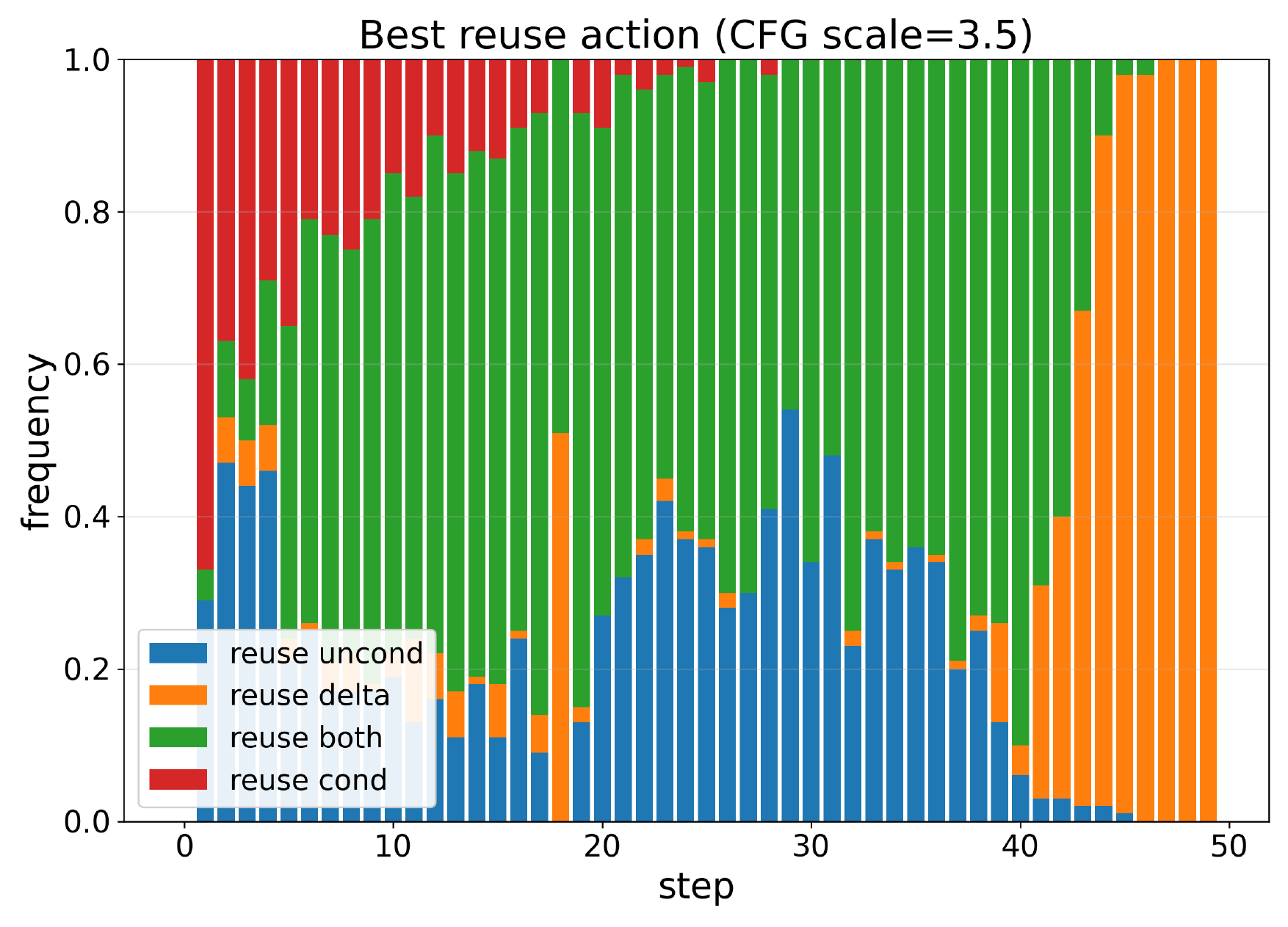}
    \end{subfigure}

    \vspace*{-2pt} 

    \begin{subfigure}[t]{0.5\textwidth}
        \centering
        \includegraphics[width=\textwidth]{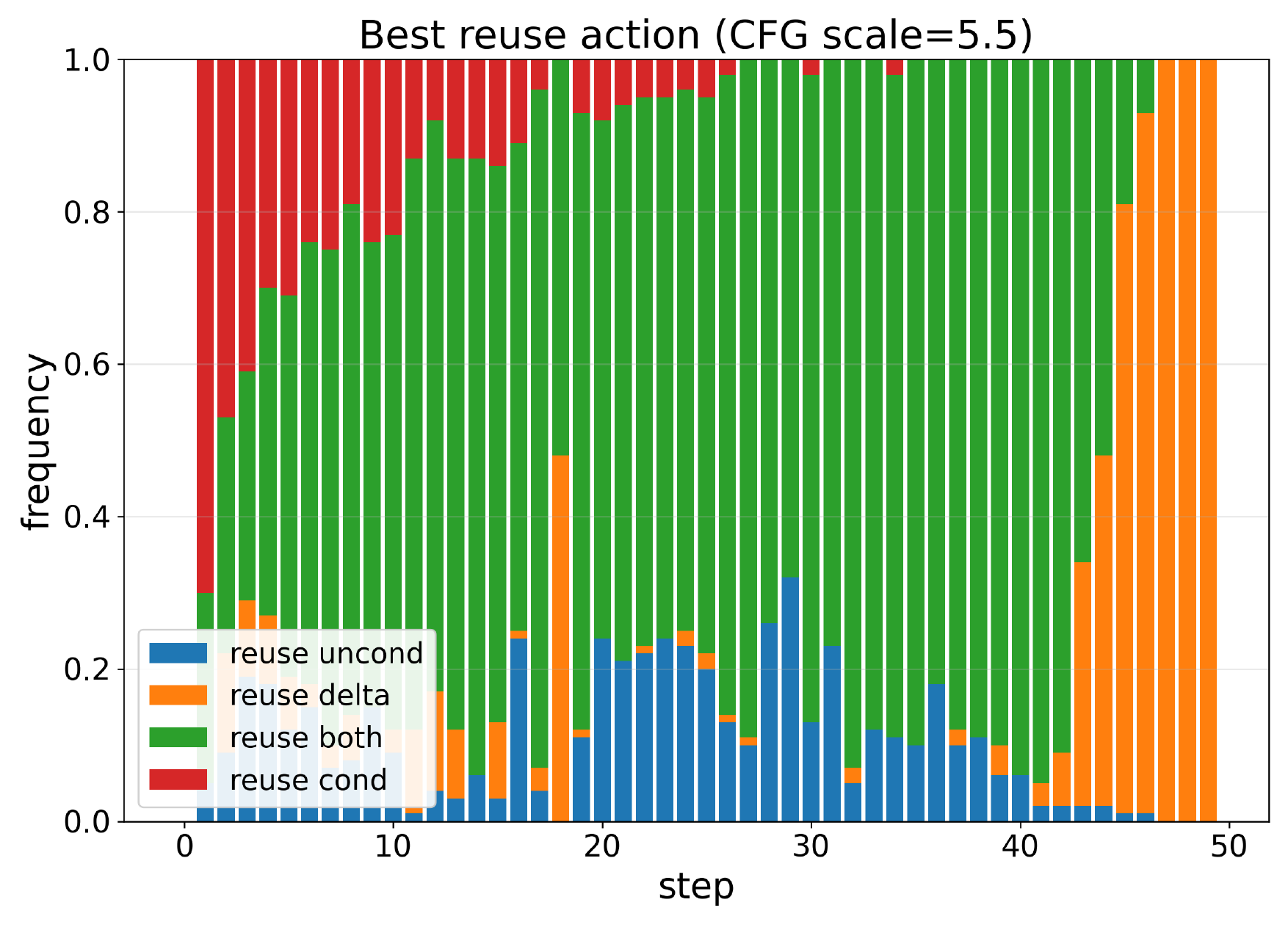}
    \end{subfigure}\hfill
    \begin{subfigure}[t]{0.5\textwidth}
        \centering
        \includegraphics[width=\textwidth]{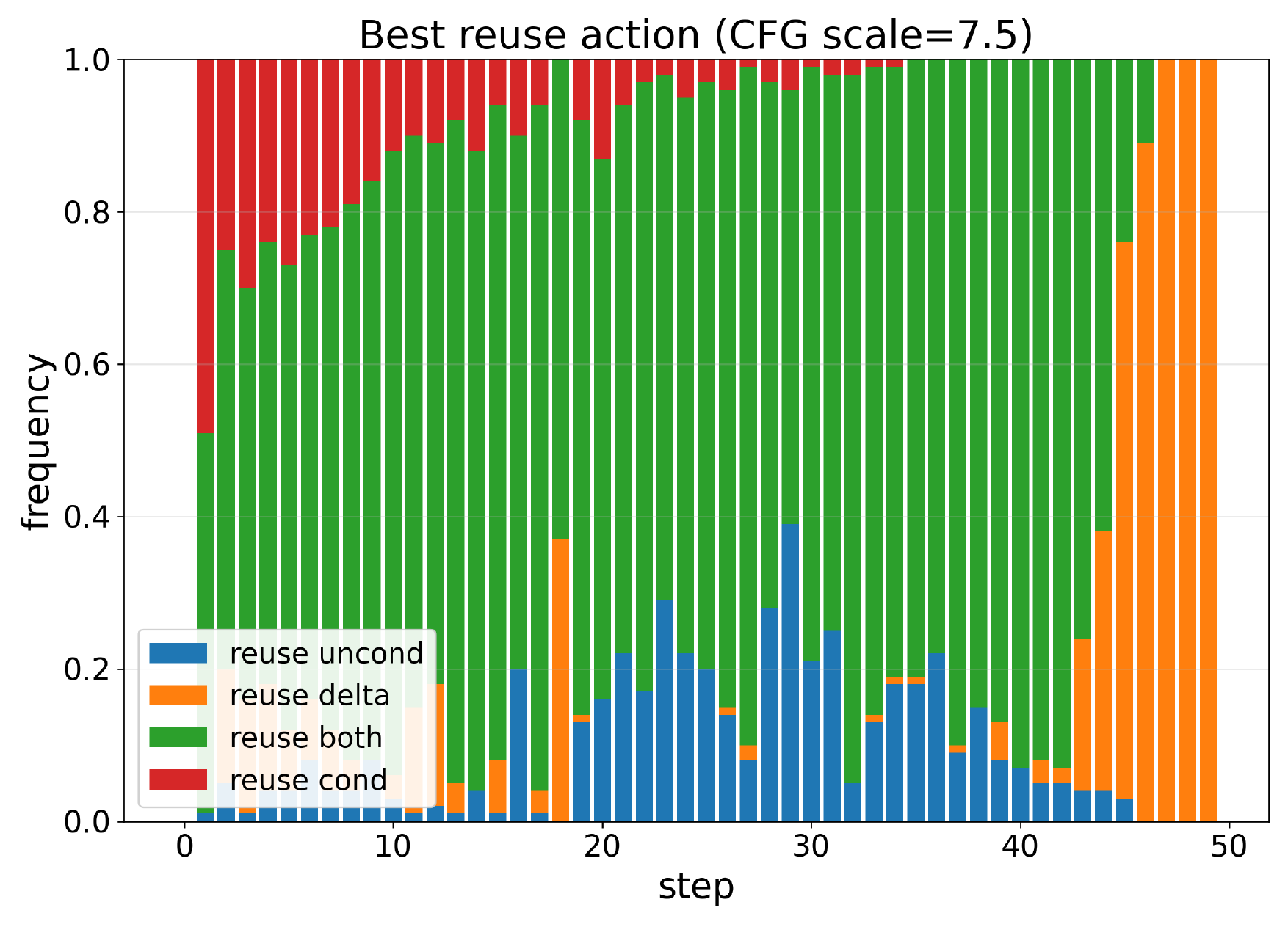}
    \end{subfigure}

    \caption{\textbf{Best reuse action frequency across timesteps under different CFG scales}
    (top-left: 1.5, top-right: 3.5, bottom-left: 5.5, bottom-right: 7.5). As CFG scale increases, reuse-both dominates a larger portion of timesteps, while reuse-\(\Delta\) becomes favorable mainly in the very late stage. Reuse-cond is rarely optimal throughout.}
    
    \label{fig:best_action_frequency_multi_cfg}
    \vspace*{-12pt} 
\end{figure*}

Beyond refresh scheduling, CFG caching introduces a second decision dimension:
what should be reused between two refresh steps. Under true CFG,
\[
p_t
=
u_t+s\Delta_t,
\qquad
\Delta_t:=c_t-u_t,
\]
or equivalently,
\[
p_t
=
(1-s)u_t+sc_t.
\]

For direct prediction reuse from anchor $t_a$, different actions induce the
following exact guided errors:
\[
\delta p_t^{(u)}
=
(1-s)(u_{t_a}-u_t),
\]
\[
\delta p_t^{(c)}
=
s(c_{t_a}-c_t),
\]
\[
\delta p_t^{(\Delta)}
=
s(\Delta_{t_a}-\Delta_t),
\]
and
\[
\delta p_t^{(\mathrm{both})}
=
(1-s)(u_{t_a}-u_t)
+
s(c_{t_a}-c_t).
\]
For intermediate-feature or residual caching, these equations describe the
corresponding idealized prediction-reuse actions rather than the exact
internal computation.

In the direct-prediction analysis, reuse-$\Delta$ recomputes the unconditional
prediction and reuses the cached CFG delta, saving one branch evaluation.
Reuse-uncond and reuse-cond also save one branch evaluation, whereas
reuse-both saves two. For intermediate caching, the exact saving depends
on the cached components and any computations retained for proxy evaluation.

We first examine object selection using a timestep-wise offline action oracle
based on true guided error. For each CFG scale, we enumerate candidate reuse
actions at every timestep and record, over 100 prompts, how often each action
attains the smallest guided action error relative to full computation.
As shown in Fig.~\ref{fig:best_action_frequency_multi_cfg}, the preferred
reuse object is phase-dependent. At low CFG scale, reuse-uncond remains
competitive over a larger portion of the trajectory. As CFG scale increases,
reuse-both is selected more frequently, while reuse-$\Delta$ becomes
competitive mainly in the late stage. Reuse-cond is rarely preferred.

The late-stage competitiveness of reuse-$\Delta$ is consistent with the
CFG geometry. The CFG-delta drift is
\[
\Delta_{t_a}-\Delta_t
=
(c_{t_a}-c_t)-(u_{t_a}-u_t).
\]
When the conditional and unconditional branch drifts are both highly aligned
and similar in magnitude, this difference can become small. In that regime,
directly reusing the CFG delta can induce a relatively small raw guided error.

A default online reuse action should also account for computational saving.
We therefore additionally compare actions by guided error per saved branch
evaluation. Under this diagnostic ratio, reuse-both performs best overall
across the evaluated CFG scales in our offline analysis. This ratio is an
auxiliary diagnostic and does not establish an optimal online policy.

Taken together, object choice under CFG is phase-dependent, while joint
branch reuse provides a strong practical default in the evaluated settings.
It is frequently preferred at moderate-to-large CFG scales and provides
favorable error per saved branch evaluation. We therefore use joint branch
reuse as the default action in the main controller.

\section{Extended Experimental Results}
\label{app:exp}

\subsection{Comparison with CFG-aware Guidance and Acceleration Methods}
\label{app:cfg_aware_comparison}

We further compare \textbf{RA-CFGCache} with related CFG-aware guidance
and acceleration methods, including MAMBO-G~\cite{zhu2026mambogmagnitudeawaremitigationboosted},
CFG Scheduler~\cite{wang2024analysisclassifierfreeguidanceweight},
and OUSAC~\cite{sun2025ousacoptimizedguidancescheduling}, under the same
FLUX.1-dev evaluation protocol.

Importantly, \textbf{RA-CFGCache does not tune the CFG scale or modify the
guidance schedule}. It preserves the original 50-step sampler and fixed CFG
rule throughout inference and only controls whether intermediate computation
is refreshed or reused. Its objective is therefore to accelerate a fixed
vanilla CFG sampler while preserving its outputs as closely as possible,
rather than to optimize absolute generation quality or preference-oriented
metrics.

In contrast, MAMBO-G dynamically adjusts guidance magnitudes, CFG Scheduler
varies guidance weights across timesteps, and OUSAC jointly optimizes sparse
guidance scheduling and caching. These approaches may therefore alter the
target generation trajectory itself, whereas \textbf{RA-CFGCache} treats the vanilla
CFG sampler as fixed.

\begin{table*}[t]
\centering
\caption{\textbf{Comparison with CFG-aware guidance and acceleration
methods on FLUX.1-dev.}
\textbf{RA-CFGCache} preserves the original sampler and fixed CFG rule.
MAMBO-G, CFG Scheduler, and OUSAC modify the guidance policy,
while FasterCache exploits CFG branch redundancy.
LPIPS, SSIM, and PSNR measure sampler fidelity, with CLIPScore,
ImageReward, and HPSv2 reported as complementary quality metrics.}
\label{tab:cfg_aware_comparison}

\footnotesize
\setlength{\tabcolsep}{4.2pt}
\renewcommand{\arraystretch}{1.08}

\begin{tabular*}{0.98\textwidth}{
@{\extracolsep{\fill}}
lccccccc
@{}
}
\toprule
\textbf{Method}
& \textbf{Speedup}$\uparrow$
& \textbf{LPIPS}$\downarrow$
& \textbf{SSIM}$\uparrow$
& \textbf{PSNR}$\uparrow$
& \textbf{CLIP}$\uparrow$
& \textbf{ImageReward}$\uparrow$
& \textbf{HPSv2}$\uparrow$ \\
\midrule

Vanilla ($T=50$)
& 1.00$\times$
& 0.0000
& 1.0000
& $\infty$
& 27.9063
& 0.8003
& 0.2776 \\

Vanilla ($T=25$)
& 1.96$\times$
& 0.4556
& 0.6207
& 13.09
& 27.0429
& 0.2629
& 0.2364 \\

FasterCache
& 2.09$\times$
& 0.2158
& 0.7593
& 20.88
& 27.4779
& 0.6241
& 0.2643 \\

MAMBO-G (T=25)
& 1.98$\times$
& 0.5616
& 0.5192
& 11.64
& \textbf{28.4720}
& 0.9800
& 0.2930 \\

CFG Scheduler
& 1.00$\times$
& 0.5327
& 0.5503
& 12.26
& 28.1390
& \textbf{1.0160}
& \textbf{0.3000} \\

OUSAC
& 3.44$\times$
& 0.5530
& 0.5092
& 11.74
& 27.8770
& 0.9300
& 0.2880 \\

\textbf{RA-CFGCache-Slow}
& 3.19$\times$
& \textbf{0.1137}
& \textbf{0.8680}
& \textbf{24.02}
& 27.8843
& 0.7672
& 0.2748 \\

\textbf{RA-CFGCache-Fast}
& \textbf{3.95$\times$}
& 0.1484
& 0.8340
& 22.75
& 27.9071
& 0.7403
& 0.2694 \\

\bottomrule
\end{tabular*}
\end{table*}

At the closest high-speed operating point, \textbf{RA-CFGCache-Fast} achieves a
$3.95\times$ speedup compared with $3.44\times$ for OUSAC, while producing
outputs substantially closer to the vanilla outputs. Specifically, LPIPS
decreases from 0.5530 to 0.1484, SSIM increases from 0.5092 to 0.8340,
and PSNR increases from 11.74 to 22.75. The corresponding CLIPScore remains
comparable (27.9071 vs.\ 27.8770). \textbf{RA-CFGCache-Slow} provides a more
fidelity-oriented operating point, achieving LPIPS 0.1137, SSIM 0.8680,
and PSNR 24.02 at a $3.19\times$ speedup.

MAMBO-G, CFG Scheduler, and OUSAC obtain higher ImageReward or HPSv2 scores
in several settings. This distinction is consistent with their different
objectives: these methods explicitly modify the guidance policy, whereas
\textbf{RA-CFGCache} does not optimize CFG parameters or preference-oriented
generation quality. Our claim is therefore a stronger
\emph{speed--sampler-fidelity trade-off}, rather than universal superiority
in absolute generation quality.

The two directions are therefore complementary: guidance-policy optimization
changes the target trajectory to improve generation characteristics, whereas
\textbf{RA-CFGCache} accelerates a fixed guidance policy and aims to reproduce its
outputs as faithfully as possible.

\subsection{Generality of Propagation-aware Rescaling}

\begin{table}[t]
\centering
\small
\begin{threeparttable}
\caption{\textbf{Generality of propagation-aware rescaling in the standard single-prediction setting.}
We apply propagation-aware rescaling to several representative non-CFG caching baselines and observe consistent fidelity improvements across the evaluated proxy families.}
\label{tab:ablation_propagation_universal}

\renewcommand{\arraystretch}{1.12}

\begin{tabular*}{\linewidth}{
@{\extracolsep{\fill}}
llcccc
@{}
}
\toprule
\textbf{Base method}
& \textbf{Variant}
& \textbf{Speedup}$\uparrow$
& \textbf{LPIPS}$\downarrow$
& \textbf{SSIM}$\uparrow$
& \textbf{PSNR}$\uparrow$ \\
\midrule

TeaCache
& Original
& 3.27$\times$
& 0.3617
& 0.7112
& 17.05 \\

TeaCache
& + Propagation Rescaling
& \textbf{3.32$\times$}
& \textbf{0.3207}
& \textbf{0.7305}
& \textbf{18.02} \\

\midrule

MagCache
& Original
& 3.11$\times$
& 0.1964
& 0.8169
& 22.05 \\

MagCache
& + Propagation Rescaling
& \textbf{3.12$\times$}
& \textbf{0.1822}
& \textbf{0.8266}
& \textbf{22.57} \\

\midrule

DiCache
& Original
& \textbf{3.20$\times$}
& 0.2999
& 0.7744
& 21.58 \\

DiCache
& + Propagation Rescaling
& 3.19$\times$
& \textbf{0.2393}
& \textbf{0.7966}
& \textbf{22.38} \\

\bottomrule
\end{tabular*}
\end{threeparttable}
\vspace{-1em}
\end{table}

Although propagation-aware rescaling is introduced together with our CFG-aware guided-risk formulation, its role is more general: it corrects local reuse scores by accounting for the non-uniform downstream impact of errors across timesteps. We therefore evaluate it in a standard single-prediction caching setting, where reuse decisions are made on a single model prediction rather than CFG-composed branch outputs.

Specifically, we apply the same propagation-aware rescaling to representative proxy baselines, including TeaCache-style, MagCache-style, and DiCache-style controllers. 
We keep their original proxy signals, thresholds, and reuse policies unchanged, and modify only the local reuse score by incorporating the timestep-dependent propagation coefficient.

Table~\ref{tab:ablation_propagation_universal} shows that propagation-aware rescaling consistently improves the final efficiency--fidelity trade-off across all tested baselines. In particular, it reduces LPIPS and improves SSIM and PSNR while maintaining nearly identical speedup.

These results indicate that propagation-aware rescaling is not merely an implementation detail tied to \textbf{RA-CFGCache}, but a broadly applicable correction principle: local reuse error should be adjusted according to its timestep-dependent downstream propagation before being used for decision-making. This supports our formulation that effective cache control requires not only accurate local error estimation, but also a proper modeling of how such errors propagate to the final output.

\subsection{Compatibility with Different Base Proxy Estimators}

\begin{table*}[t]
\centering
\small
\begin{threeparttable}
\caption{\textbf{Compatibility with different base proxy estimators.}
We apply \textbf{RA-CFGCache} on top of different branch-wise proxy families and compare it against their original branch-wise accumulation controllers.
Across all proxy types, \textbf{RA-CFGCache} consistently improves fidelity at nearly unchanged latency, showing that its gain is complementary to the underlying proxy design rather than tied to a specific estimator.}
\label{tab:ablation_proxy_compatibility}
\setlength{\tabcolsep}{4.2pt}
\renewcommand{\arraystretch}{1.12}
\begin{tabular}{llccccc}
\toprule
\textbf{Base proxy family} & \textbf{Controller} & \textbf{Latency$\downarrow$} & \textbf{Speedup$\uparrow$} & \textbf{LPIPS$\downarrow$} & \textbf{SSIM$\uparrow$} & \textbf{PSNR$\uparrow$} \\
\midrule
\multirow{2}{*}{Tea-style proxy}
& Branch-wise accumulation & 6.16 s & 3.41$\times$ & 0.1852 & 0.7977 & 20.60 \\
& + RA-CFGCache wrapper       & \textbf{6.16 s} & \textbf{3.41$\times$} & \textbf{0.1492} & \textbf{0.8297} & \textbf{22.39} \\
\midrule
\multirow{2}{*}{Mag-style proxy}
& Branch-wise accumulation & 6.18 s & 3.40$\times$ & 0.1683 & 0.8144 & 21.35 \\
& + RA-CFGCache wrapper       & \textbf{6.17 s} & \textbf{3.41$\times$} & \textbf{0.1232} & \textbf{0.8590} & \textbf{23.53} \\
\midrule
\multirow{2}{*}{DiCache-style proxy}
& Branch-wise accumulation & 6.13 s & 3.43$\times$ & 0.2177 & 0.7706 & 19.48 \\
& + RA-CFGCache wrapper       & \textbf{6.13 s} & \textbf{3.43$\times$} & \textbf{0.1453} & \textbf{0.8364} & \textbf{22.57} \\
\bottomrule
\end{tabular}
\end{threeparttable}
\vspace{-2em}
\end{table*}

\textbf{RA-CFGCache} is designed as a control framework that operates on top of branch-wise proxy signals, rather than relying on a specific proxy design. To verify this plug-in property, we evaluate our method with several representative proxy families, including Tea-style, Mag-style, and DiCache-style estimators.

For each proxy family, we compare its original branch-wise accumulation controller with the corresponding \textbf{RA-CFGCache} version, which replaces the original accumulation rule with CFG-aware guided-risk composition and propagation-aware rescaling, while keeping the underlying proxy signals unchanged.

As shown in Table~\ref{tab:ablation_proxy_compatibility}, \textbf{RA-CFGCache} consistently improves fidelity across all proxy families at nearly identical latency. For example, under the Tea-style proxy, LPIPS is significantly reduced with improved SSIM and PSNR, and similar improvements are observed for Mag-style and DiCache-style proxies.

These results demonstrate that the gain of \textbf{RA-CFGCache} does not come from designing a stronger proxy signal, but from correcting how proxy signals are composed and accumulated under CFG. This confirms that \textbf{RA-CFGCache} can be viewed as a general wrapper that upgrades existing proxy-based caching methods into a risk-aligned control paradigm.

\subsection{Additional Calibration Visualizations}

\begin{figure*}[t]
    \centering
    \begin{subfigure}[t]{0.48\textwidth}
        \centering
        \includegraphics[height=4.8cm]{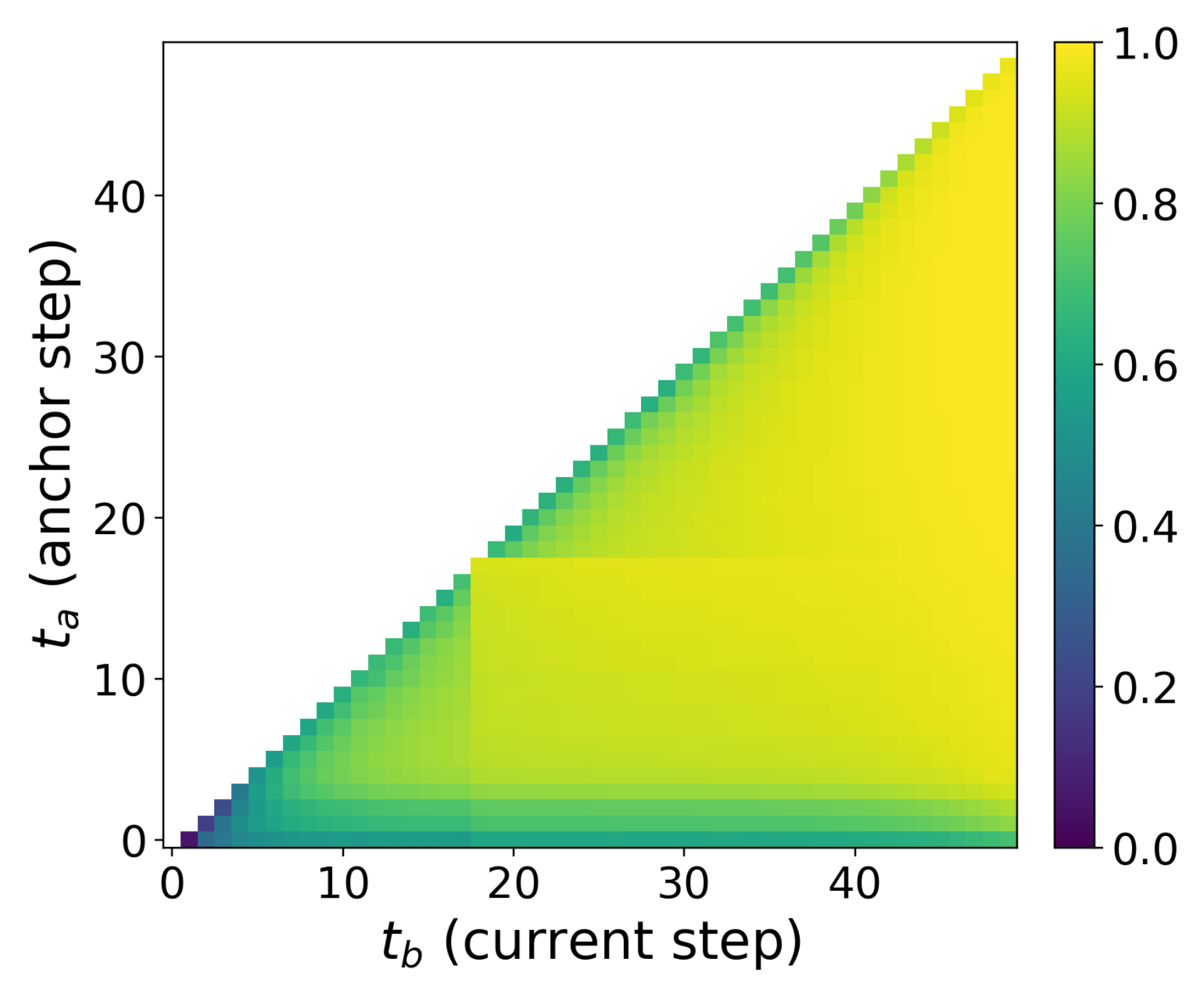}
        \caption{Pairwise alignment heatmap $\bar{\rho}_{t_a,t_b}$.}
    \end{subfigure}
    \hfill
    \begin{subfigure}[t]{0.48\textwidth}
        \centering
        \includegraphics[height=4.8cm]{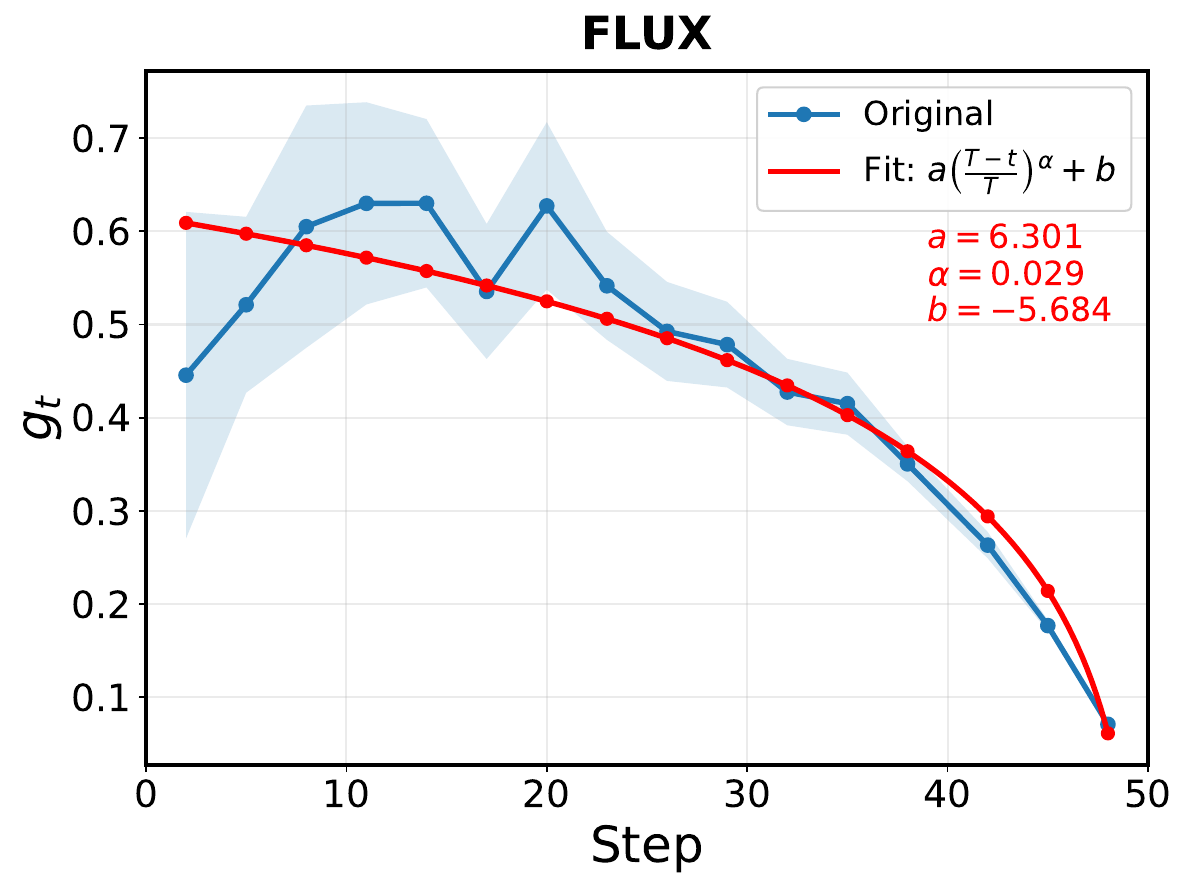}
        \caption{Power-law fit of the propagation gain curve $g_t$ on FLUX.}
    \end{subfigure}

    \vspace{0.5em}

    \begin{subfigure}[t]{0.48\textwidth}
        \centering
        \includegraphics[height=4.8cm]{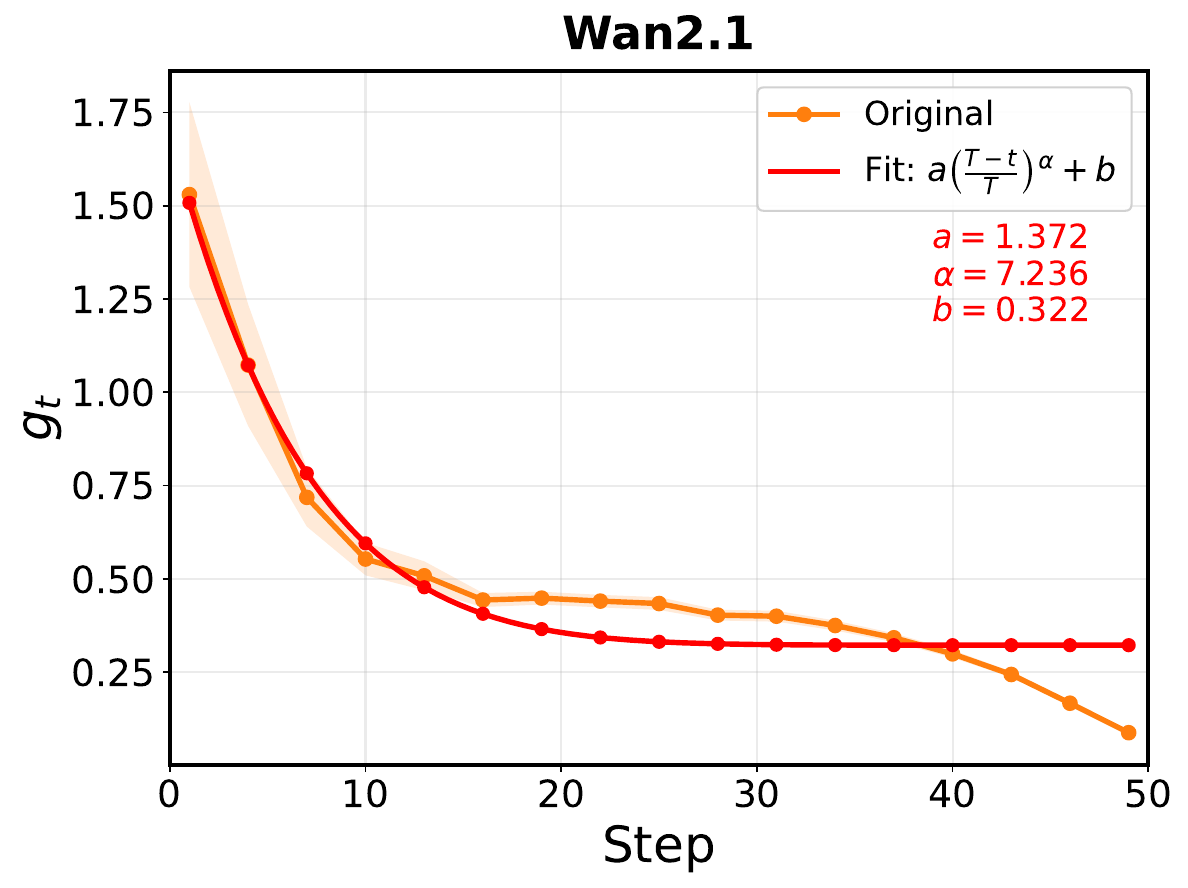}
        \caption{Power-law fit of the propagation gain curve $g_t$ on Wan2.1-T2V-1.3B.}
    \end{subfigure}
    \hfill
    \begin{subfigure}[t]{0.48\textwidth}
        \centering
        \includegraphics[height=4.8cm]{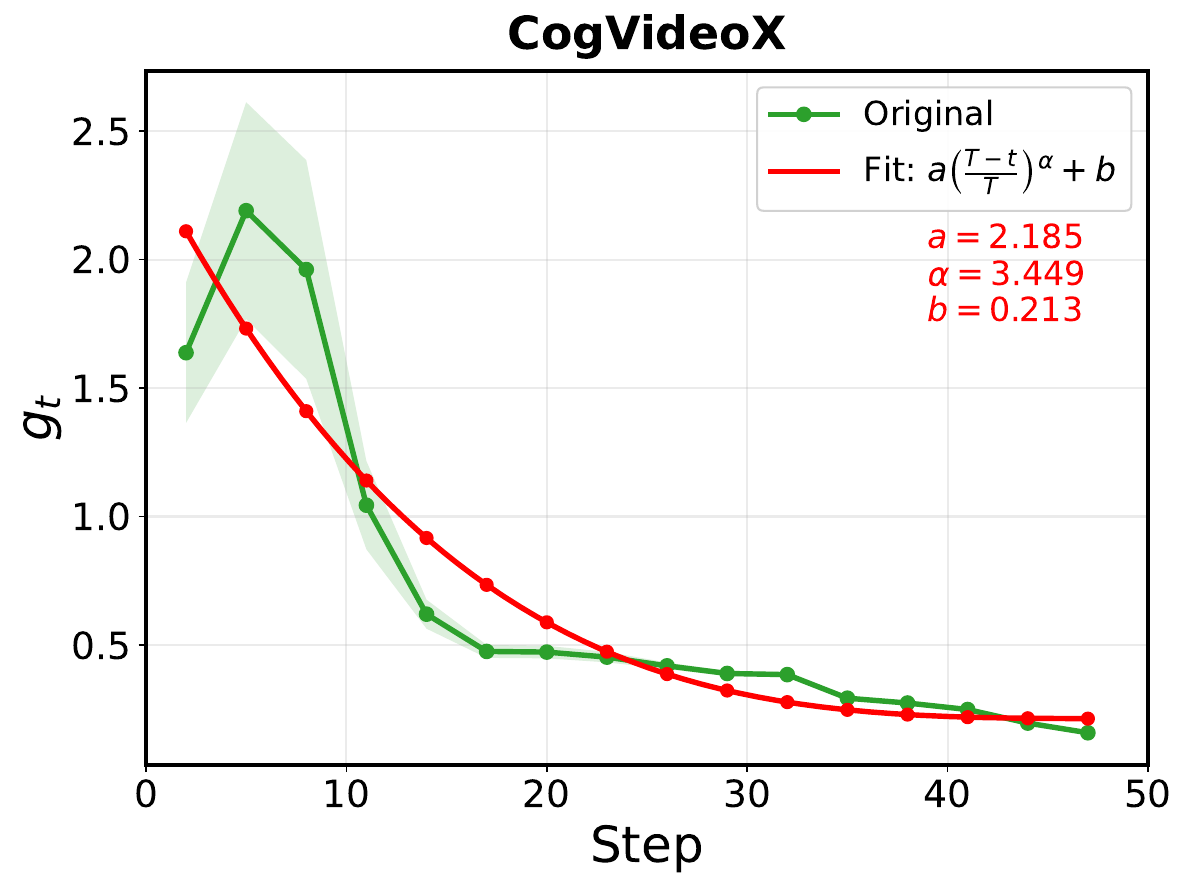}
        \caption{Power-law fit of the propagation gain curve $g_t$ on CogVideoX.}
    \end{subfigure}

    \caption{\textbf{Additional calibration visualizations.}
    Top-left: pairwise branch-error alignment heatmap.
    The other three panels show model-specific power-law fits of the propagation gain curve $g_t$.}
    \label{fig:calibration_visualizations}
    \vspace{-1em}
\end{figure*}

We visualize the two calibrated quantities used by our method:
the anchor--target alignment statistic $\bar{\rho}_{t_a,t_b}$ and the
timestep-dependent propagation gain $g_t$.

Figure~\ref{fig:calibration_visualizations}(a) shows the alignment heatmap
between anchor step $t_a$ and current step $t_b$. The heatmap exhibits clear
structure rather than an unstructured noisy pattern. Alignment is weaker
and more variable in some early regions and stronger over broad later
regions. This motivates retaining anchor--target dependence in the calibrated
alignment statistic, while the heatmap alone does not establish that a full
pairwise matrix is necessary relative to simpler parameterizations.

Figure~\ref{fig:calibration_visualizations}(b)--(d) show the measured
propagation-gain curve and its fitted power-law surrogate. The measured
gain is non-uniform across timesteps: perturbations introduced at different
steps lead to substantially different final deviations. The fitted curve
captures the dominant trend and provides a compact representation for
online control. Its usefulness for cache control is evaluated separately
through the surrogate ablation.

Together, these visualizations illustrate the two calibration objects
used by \textbf{RA-CFGCache}: an anchor--target calibrated alignment statistic
for guided-risk composition and timestep-dependent propagation weighting for
control-score accumulation.

\subsection{Cross-model Alignment and Propagation Patterns}

\begin{figure*}[t]
    \centering

    \begin{subfigure}[t]{0.32\textwidth}
        \centering
        \includegraphics[width=\textwidth]{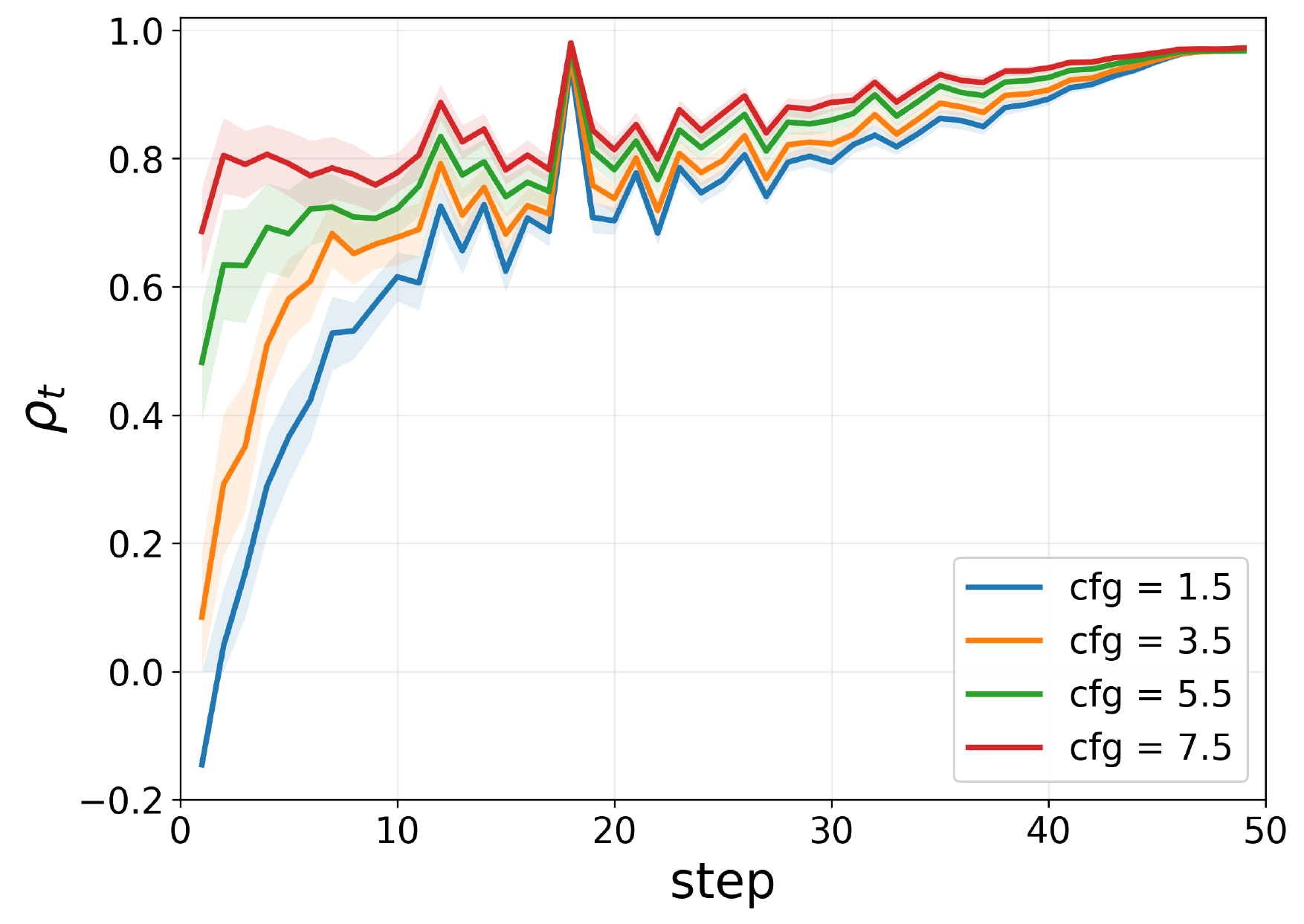}
        \caption{FLUX}
        \label{fig:rho_flux_cfg}
    \end{subfigure}
    \hfill
    \begin{subfigure}[t]{0.32\textwidth}
        \centering
        \includegraphics[width=\textwidth]{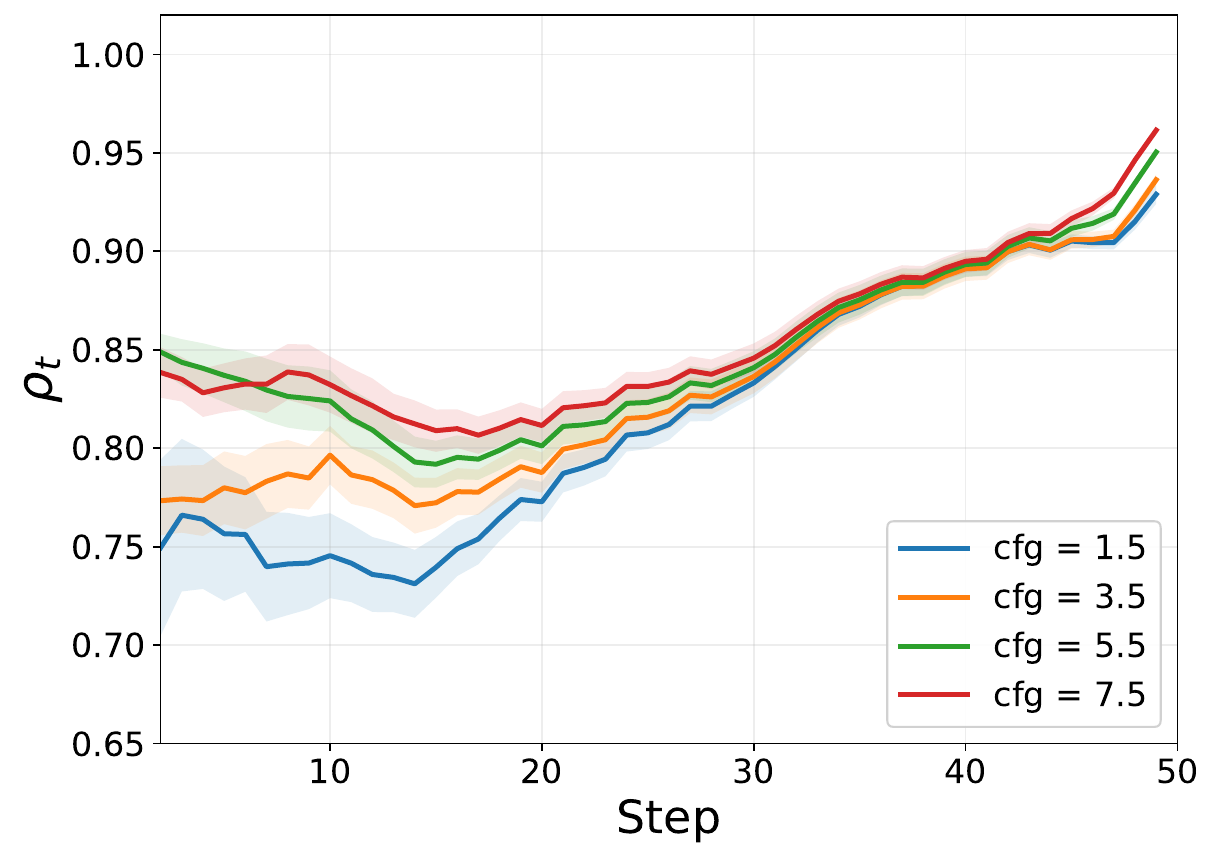}
        \caption{Wan2.1}
        \label{fig:rho_wan_cfg}
    \end{subfigure}
    \hfill
    \begin{subfigure}[t]{0.32\textwidth}
        \centering
        \includegraphics[width=\textwidth]{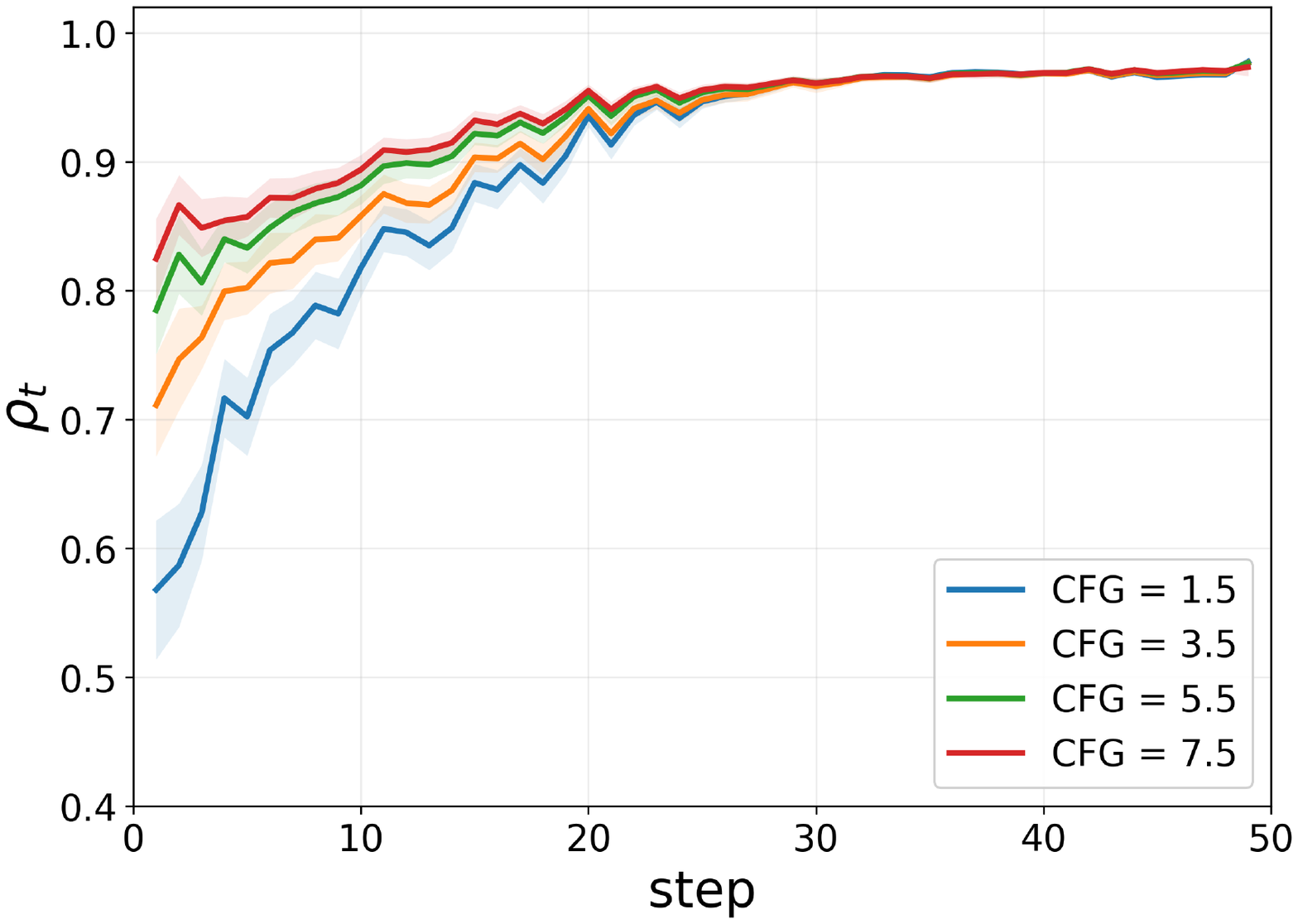}
        \caption{CogVideoX}
        \label{fig:rho_cog_cfg}
    \end{subfigure}

    \caption{\textbf{Alignment curves across CFG scales for different diffusion models.}
    Across models, $\rho_t$ generally remains lower or less stable in the early-to-middle stage and tends toward a higher-alignment regime in later timesteps, although detailed local fluctuations are model-dependent.}
    \label{fig:rho_curves_multi_model_cfg}
    \vspace{-1em}
\end{figure*}
\begin{figure*}[t]
    \centering

    \begin{subfigure}[t]{0.32\textwidth}
        \centering
        \includegraphics[width=\textwidth]{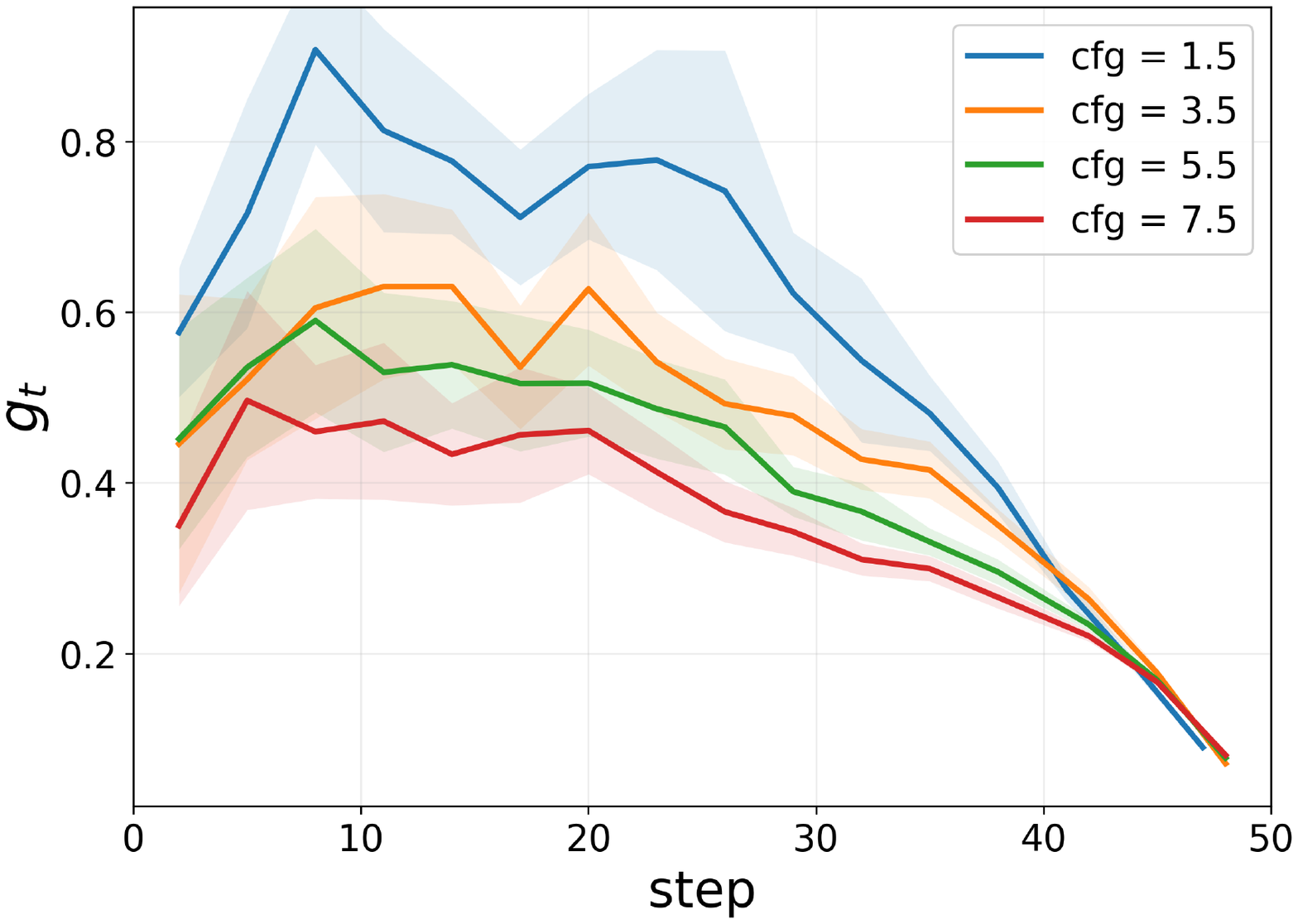}
        \caption{FLUX}
        \label{fig:gt_flux_cfg}
    \end{subfigure}
    \hfill
    \begin{subfigure}[t]{0.32\textwidth}
        \centering
        \includegraphics[width=\textwidth]{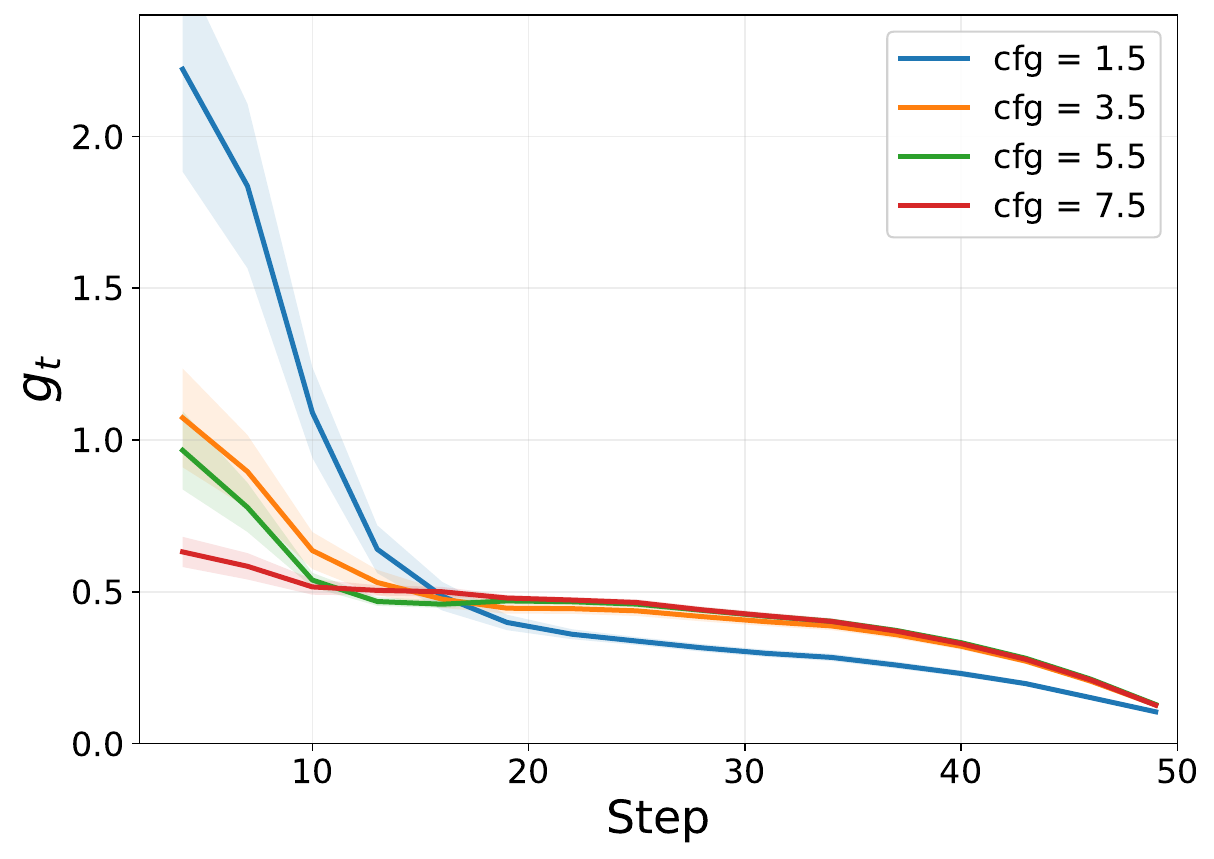}
        \caption{Wan2.1}
        \label{fig:gt_wan_cfg}
    \end{subfigure}
    \hfill
    \begin{subfigure}[t]{0.32\textwidth}
        \centering
        \includegraphics[width=\textwidth]{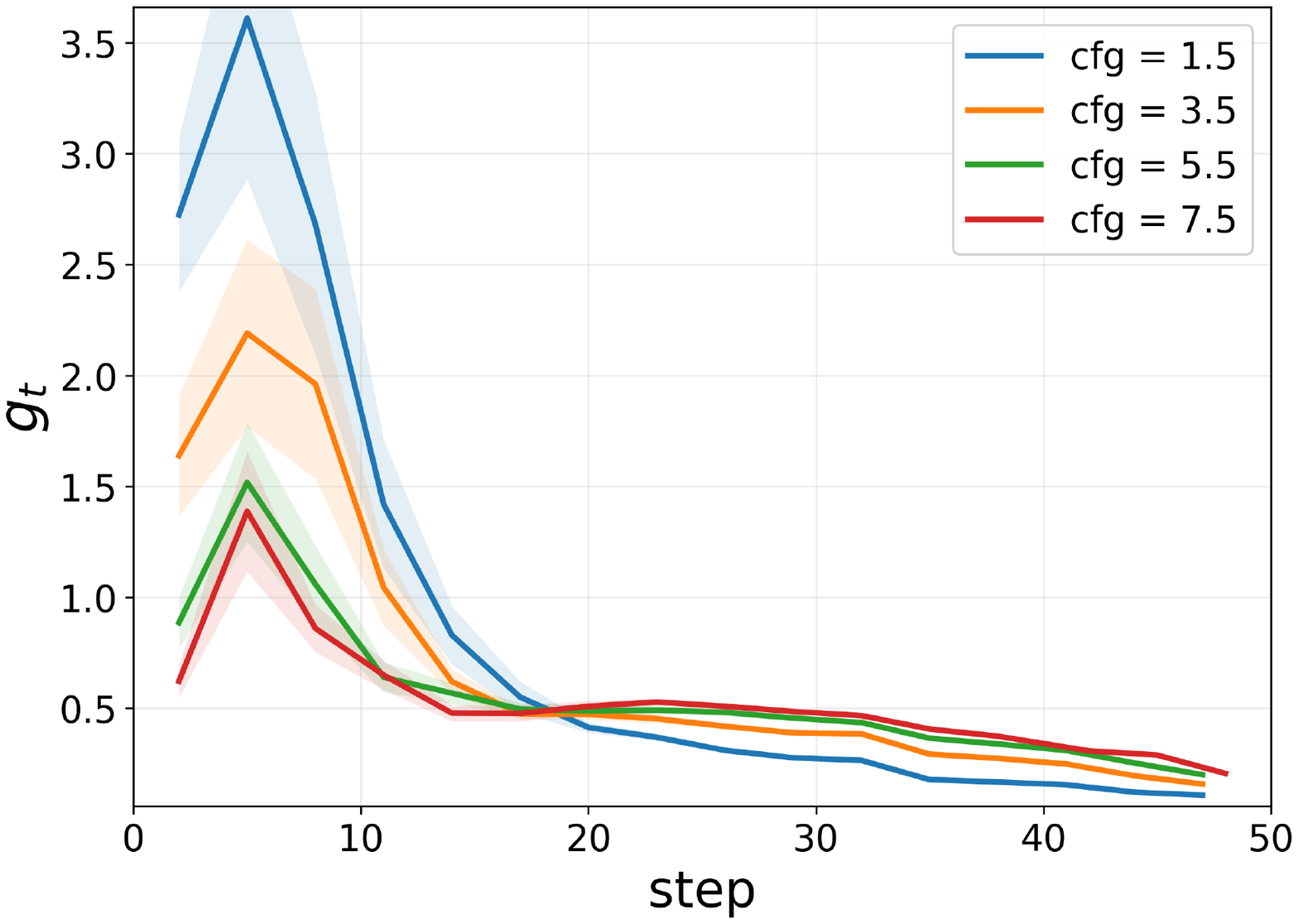}
        \caption{CogVideoX}
        \label{fig:gt_cog_cfg}
    \end{subfigure}

    \caption{
    \textbf{Propagation gain curves $g_t$ across CFG scales for different diffusion models.}
    The propagation gain is strongly timestep-dependent, with larger values in the early or early-middle stages and much smaller values in later steps. Across CFG scales, the curves show model-dependent magnitude differences and tend to become closer in later steps, indicating that local guided errors have substantially different final impacts depending on their injection timestep.
    }
    \label{fig:gt_curves_multi_model_cfg}
\end{figure*}

To further examine the empirical patterns relevant to our calibration
procedure, we visualize alignment and propagation statistics across
diffusion models and CFG settings.

Figure~\ref{fig:rho_curves_multi_model_cfg} shows the timestep-wise alignment curves
across multiple CFG scales for the evaluated models. Despite differences in
architecture and absolute magnitude, the curves exhibit structured
timestep-dependent variation, with many settings moving toward a stronger
alignment regime later in sampling.

Figure~\ref{fig:gt_curves_multi_model_cfg} presents the corresponding
propagation-gain curves. Across the evaluated models, perturbations introduced
at different timesteps have markedly different downstream effects, with the
dominant trend generally decreasing toward later sampling stages.

These results show related empirical patterns across the evaluated models
and CFG configurations. They are consistent with our use of model-specific
alignment calibration together with a shared low-dimensional functional
family for the propagation prior. However, similar qualitative patterns do
not imply direct transfer of calibrated parameters across models, samplers,
or guidance configurations.

\subsection{Why a Power-law Surrogate for Propagation Gain}
\label{app:propagation_fit}

After measuring the propagation gain through offline perturbation calibration,
we require a practical surrogate for online scheduling. Our goal is not to
reproduce every local fluctuation of the measured curve, but to obtain a
smooth control signal that captures the dominant timestep dependence.

Using the increasing execution-step index $k$, we model the gain as
\[
\hat g_k
=
\lambda\Bigl(\frac{T-k}{T}\Bigr)^\alpha+\beta,
\]
where $t_k$ denotes the corresponding diffusion time and
$\lambda,\alpha,\beta$ are fitted from the offline perturbation measurements.

\begin{table}[t]
\centering
\small
\caption{
\textbf{Effect of propagation-gain surrogate.}
We compare compact surrogate forms for propagation-aware rescaling.
The power-law surrogate provides the best fidelity among the tested smooth parameterizations, achieving the lowest LPIPS and highest SSIM with competitive PSNR.
}
\label{tab:ablation_target_approx}
\setlength{\tabcolsep}{4.0pt}
\renewcommand{\arraystretch}{1.05}
\begin{tabular}{lcccccc}
\toprule
\textbf{Surrogate} 
& \textbf{$R^2 \uparrow$} 
& \textbf{NMAE $\downarrow$} 
& \textbf{Speedup $\uparrow$}
& \textbf{LPIPS $\downarrow$} 
& \textbf{SSIM $\uparrow$} 
& \textbf{PSNR $\uparrow$} \\
\midrule
No prop. rescaling          & --              & --              & 3.74$\times$ & 0.1699 & 0.8162 & 21.41 \\
Fixed linear decay          & -0.0024         & 0.3174          & 3.75$\times$ & 0.1469 & 0.8297 & 22.74 \\
Fitted linear decay         & 0.7164          & 0.1521          & 3.76$\times$ & 0.1448 & 0.8379 & 22.80 \\
\rowcolor{gray!10}
Power-law decay             & \textbf{0.8201} & \textbf{0.1010} & \textbf{3.77$\times$} & \textbf{0.1406} & \textbf{0.8398} & \textbf{22.93} \\
\bottomrule
\end{tabular}
\vspace{-1em}
\end{table}

The power-law form has three practical advantages. First, it provides a
smooth approximation to the dominant propagation trend and avoids reacting
to local calibration noise. Second, $(T-k)/T$ gives a dimensionless
parameterization of sampling progress, although this normalization alone
does not guarantee transfer across step counts or noise schedules.
Third, the surrogate requires only three scalar parameters and therefore
adds negligible online cost.

Figure~\ref{fig:gt_curves_multi_model_cfg} shows that the measured gains
contain model-dependent transients and local fluctuations but exhibit a
clear coarse timestep dependence. Table~\ref{tab:ablation_target_approx}
compares several propagation-gain surrogates. Removing propagation rescaling
gives the weakest final fidelity in the reported comparison, supporting the
benefit of timestep-dependent downstream weighting. Compared with the fixed
and fitted linear alternatives, the power-law surrogate achieves the best
LPIPS and SSIM with competitive PSNR at comparable speedup. The $R^2$ and
NMAE columns characterize approximation quality, whereas LPIPS, SSIM, and
PSNR evaluate the downstream effect of the resulting controller.

The power-law form should therefore be understood as a practical scheduling
prior rather than a strict law of diffusion error propagation.
\subsection{Parallel Consistency under CFG Parallelism}
\label{app:parallel_consistency}

We separately evaluate a parallel CFG implementation, distinct from the
single-GPU sequential branch execution used in the main experiments.
Each sample uses two GPUs, with the conditional and unconditional branches
placed on separate GPUs. Their outputs are synchronized before CFG
composition, and the associated synchronization overhead is included in the
reported end-to-end latency.

\begin{table}[t]
\centering
\small
\setlength{\tabcolsep}{3.2pt}
\caption{\textbf{Effect of caching strategies under CFG parallelism.}
Controlled FLUX.1-dev study at 1024$\times$1024 resolution, CFG scale 3.5, 50 denoising steps, comparing independent branch-wise caching and joint guided caching under a matched branch-evaluation budget.}
\label{tab:parallel_consistency}
\begin{tabular}{lcccc}
\toprule
\textbf{Controller} 
& \textbf{Reuse both (\%)} $\uparrow$ 
& \textbf{One-sided reuse (\%)} $\downarrow$ 
& Full-step refresh (\%)
& \textbf{Latency (s)} $\downarrow$ \\
\midrule
Branch-wise caching & 35.1 & 53.8 & 11.1 & 6.81 \\
Ours (reuse both)  & \textbf{62.0} & \textbf{0.0} & 38.0 & \textbf{4.21} \\
\bottomrule
\end{tabular}
\vspace{-2em}
\end{table}

We compare joint branch decisions against independent TeaCache-style
controllers applied separately to the conditional and unconditional branches.
The thresholds are tuned to match the aggregate number of branch evaluations.

As shown in Table~\ref{tab:parallel_consistency}, independent branch-wise
control uses one-sided reuse at 53.8\% of steps and joint reuse at 35.1\%.
One-sided reuse reduces branch computation, but may provide limited
wall-clock benefit when the recomputed branch remains on the critical path.
Joint control avoids these one-sided decisions and skips recomputation of
cached components in both branches at each accepted joint reuse step.

At the matched branch-evaluation count, independent branch-wise control
requires 6.81\,s, whereas joint control requires 4.21\,s. These results
support joint reuse as an effective execution choice in the evaluated
parallel implementation. They do not imply that one-sided reuse provides
no computational saving or that the same latency advantage necessarily
holds across different hardware configurations or execution backends.
\section{Broader impacts}
\label{app:impacts}
\textbf{RA-CFGCache} improves the efficiency of CFG-based diffusion inference, which can lower deployment cost and reduce energy consumption for visual generation systems.
The method does not introduce new generative capabilities, datasets, or pretrained models, and we do not identify negative societal impacts unique to the proposed caching algorithm.
As with other efficiency improvements for generative models, practical deployment should still follow the safety and content-moderation policies of the underlying models.

\newpage
\section{Additional Qualitative Results}
\label{app:comparison}
\begin{figure}[H]
\centering
\setlength{\tabcolsep}{0pt}
\renewcommand{\arraystretch}{0.92}
\scriptsize

\resizebox{0.92\textwidth}{!}{%
\begin{tabular}{@{}c@{}c@{}c@{}c@{}c@{}}
Original & TeaCache (2.8$\times$) & MagCache (2.76$\times$) & DiCache (2.71$\times$) & Ours (2.83$\times$) \\
\includegraphics[width=0.19\textwidth]{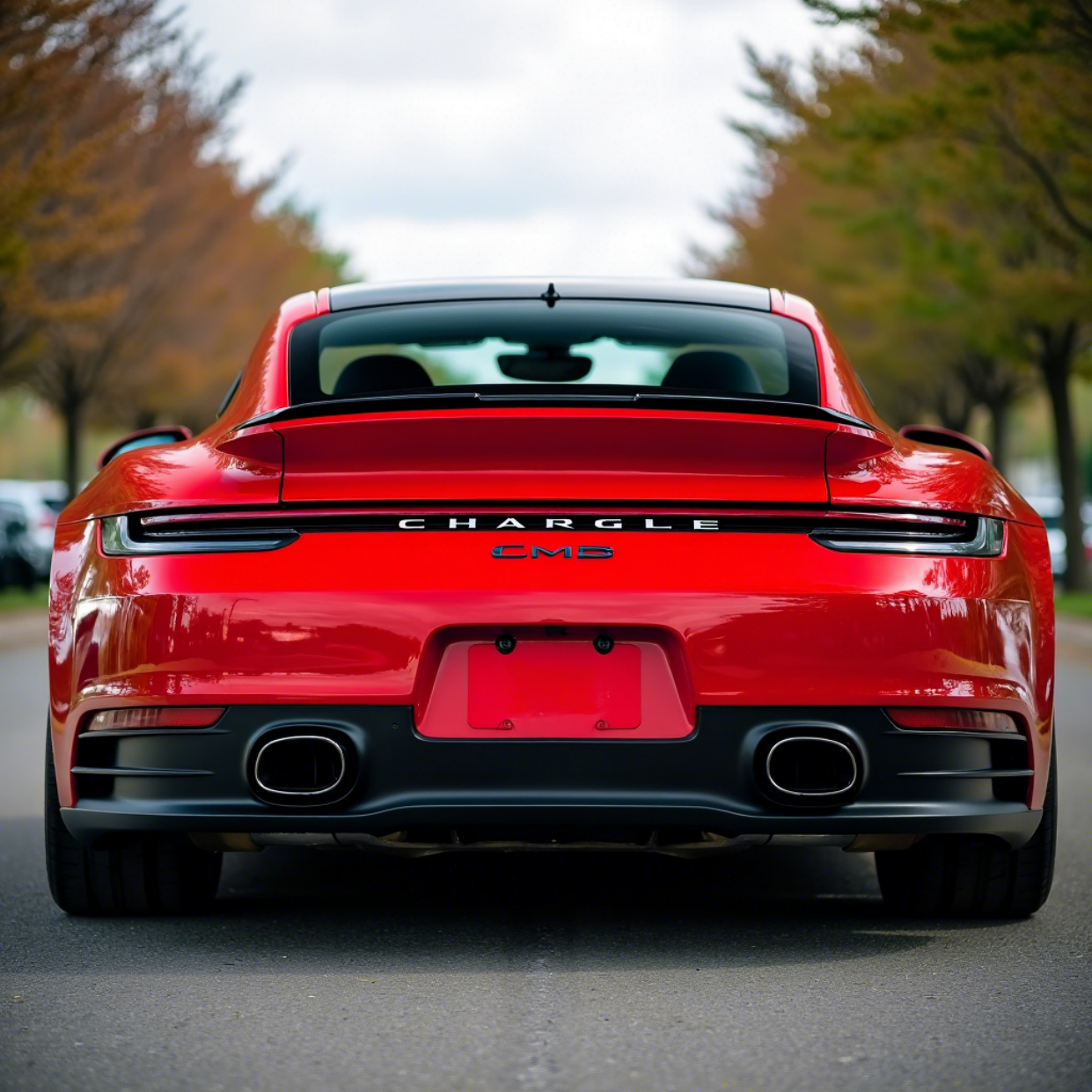}
& \includegraphics[width=0.19\textwidth]{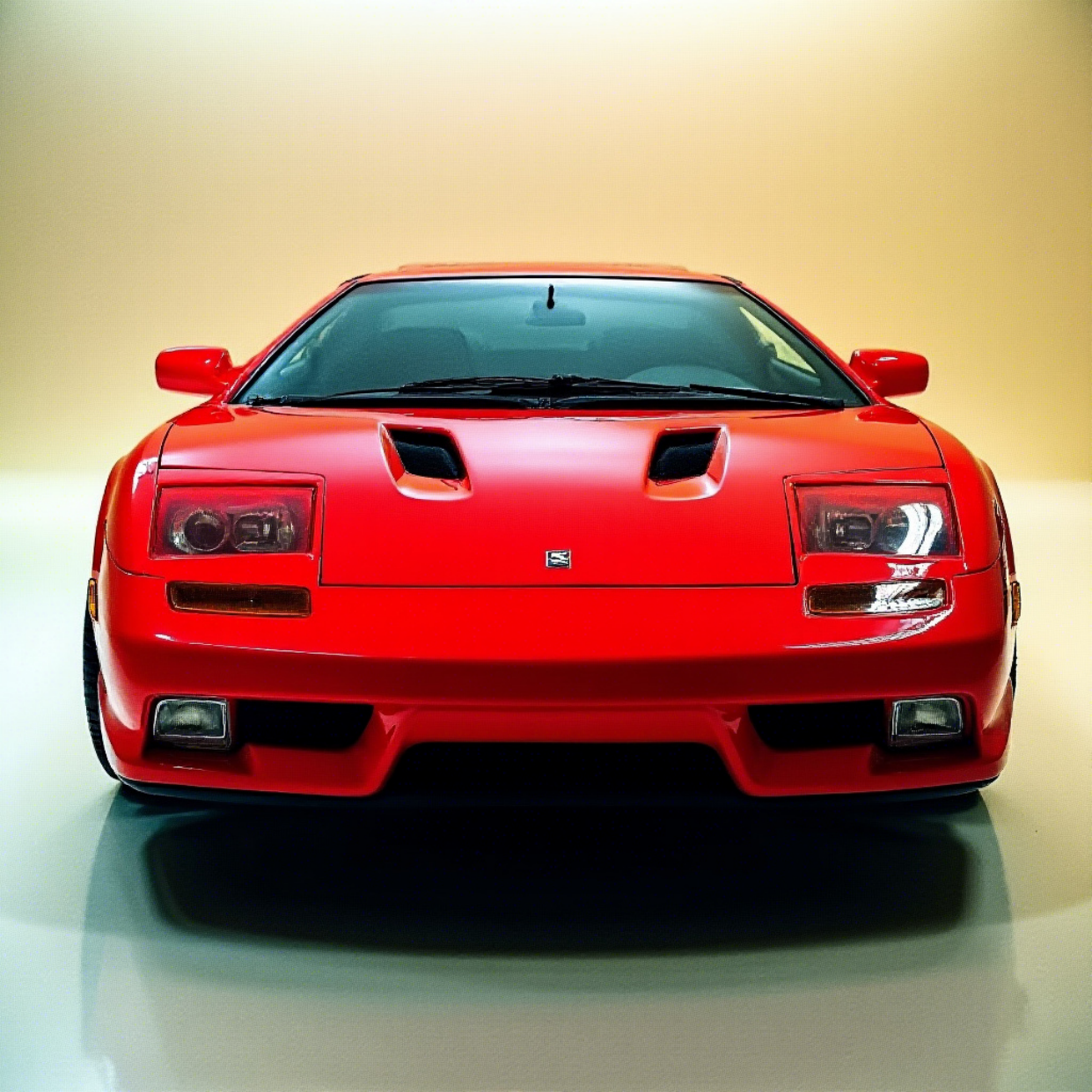}
& \includegraphics[width=0.19\textwidth]{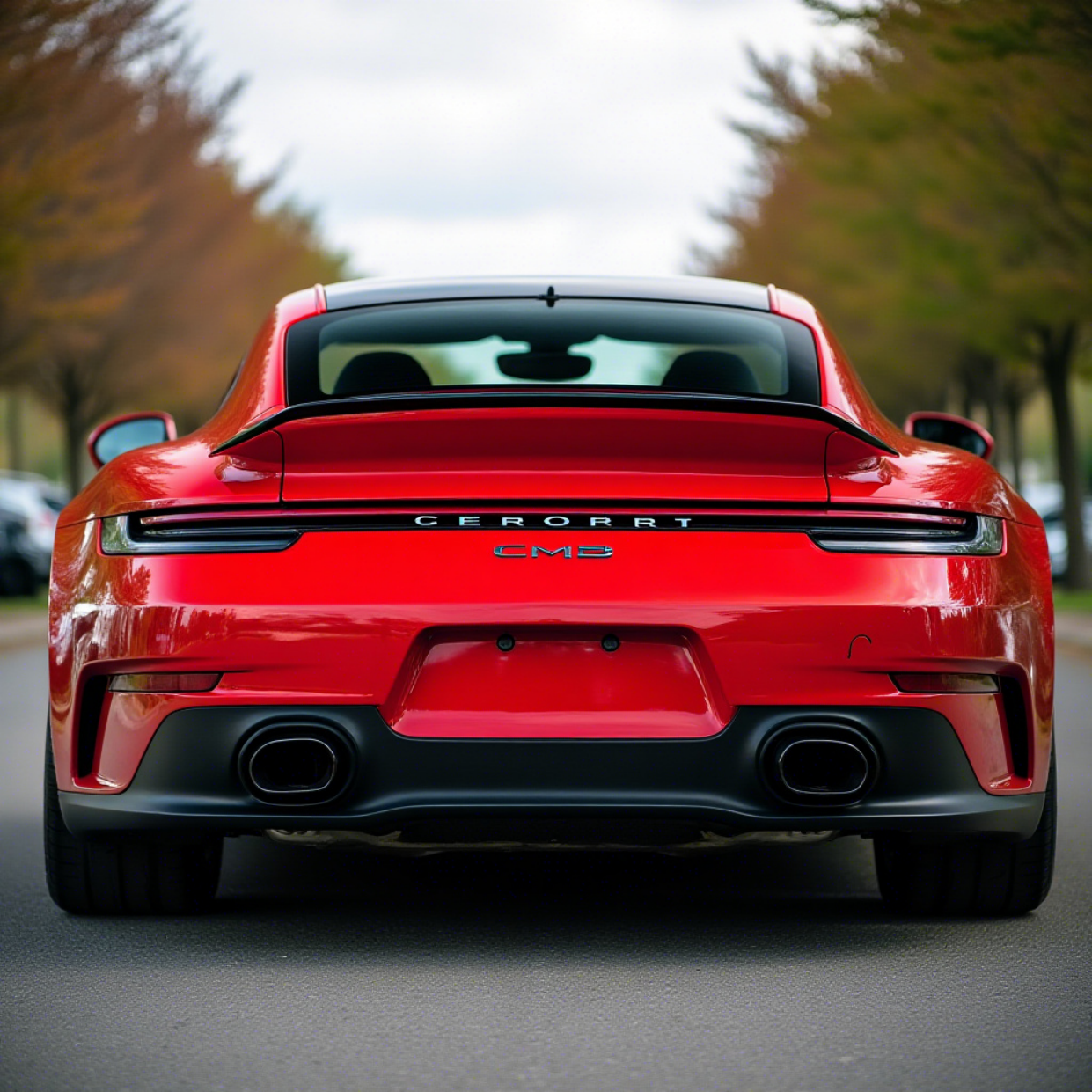}
& \includegraphics[width=0.19\textwidth]{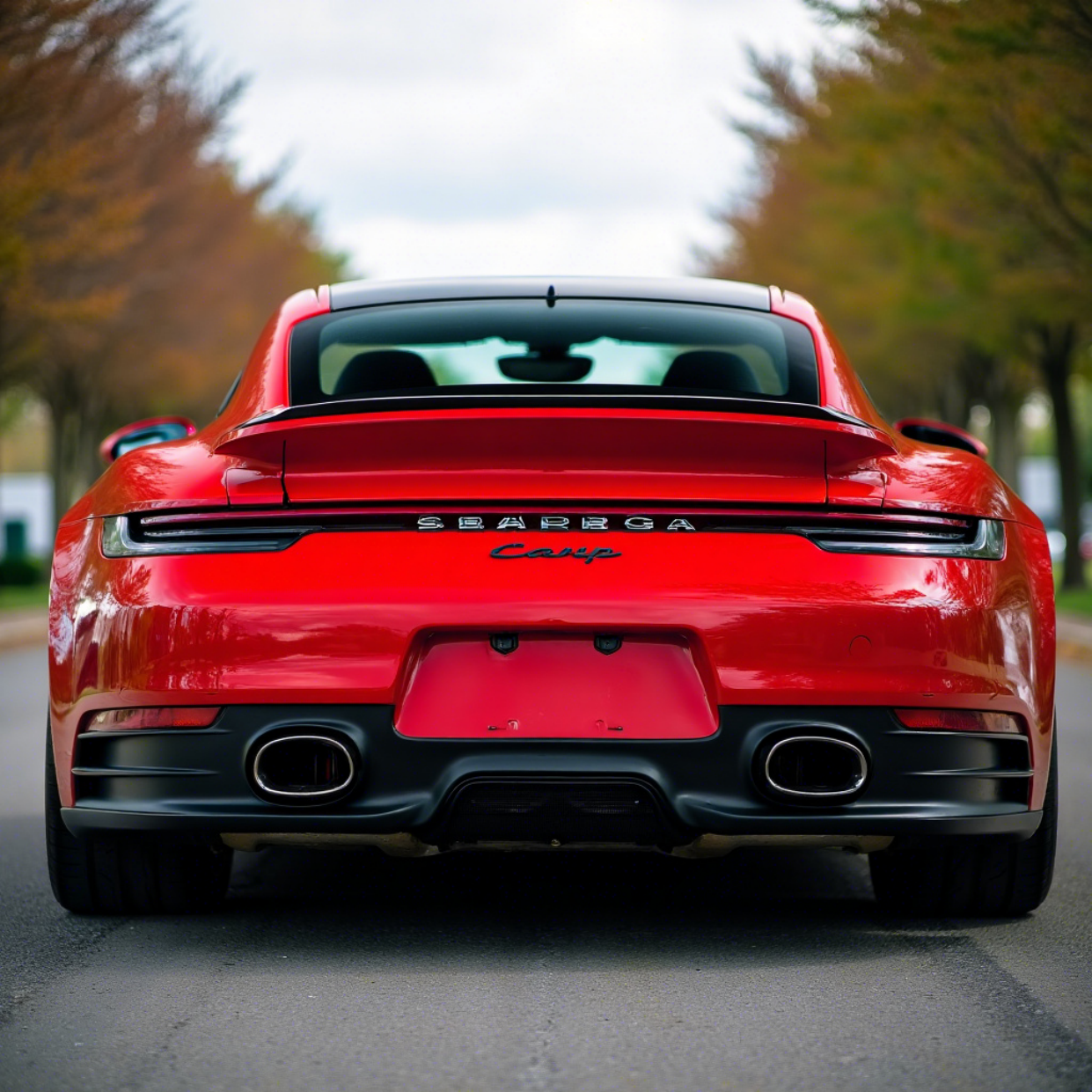}
& \includegraphics[width=0.19\textwidth]{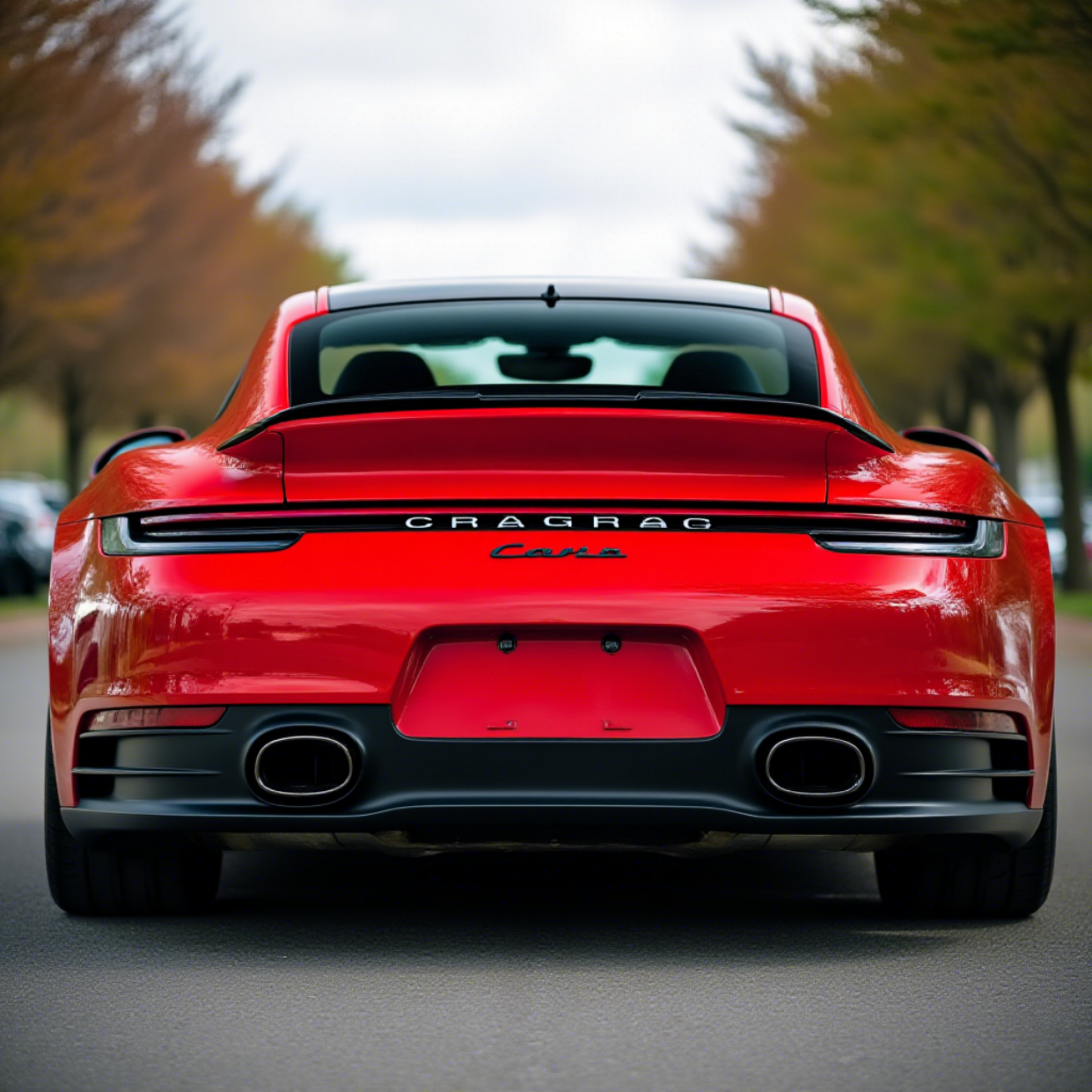}
\\[-0.25em]

\includegraphics[width=0.19\textwidth]{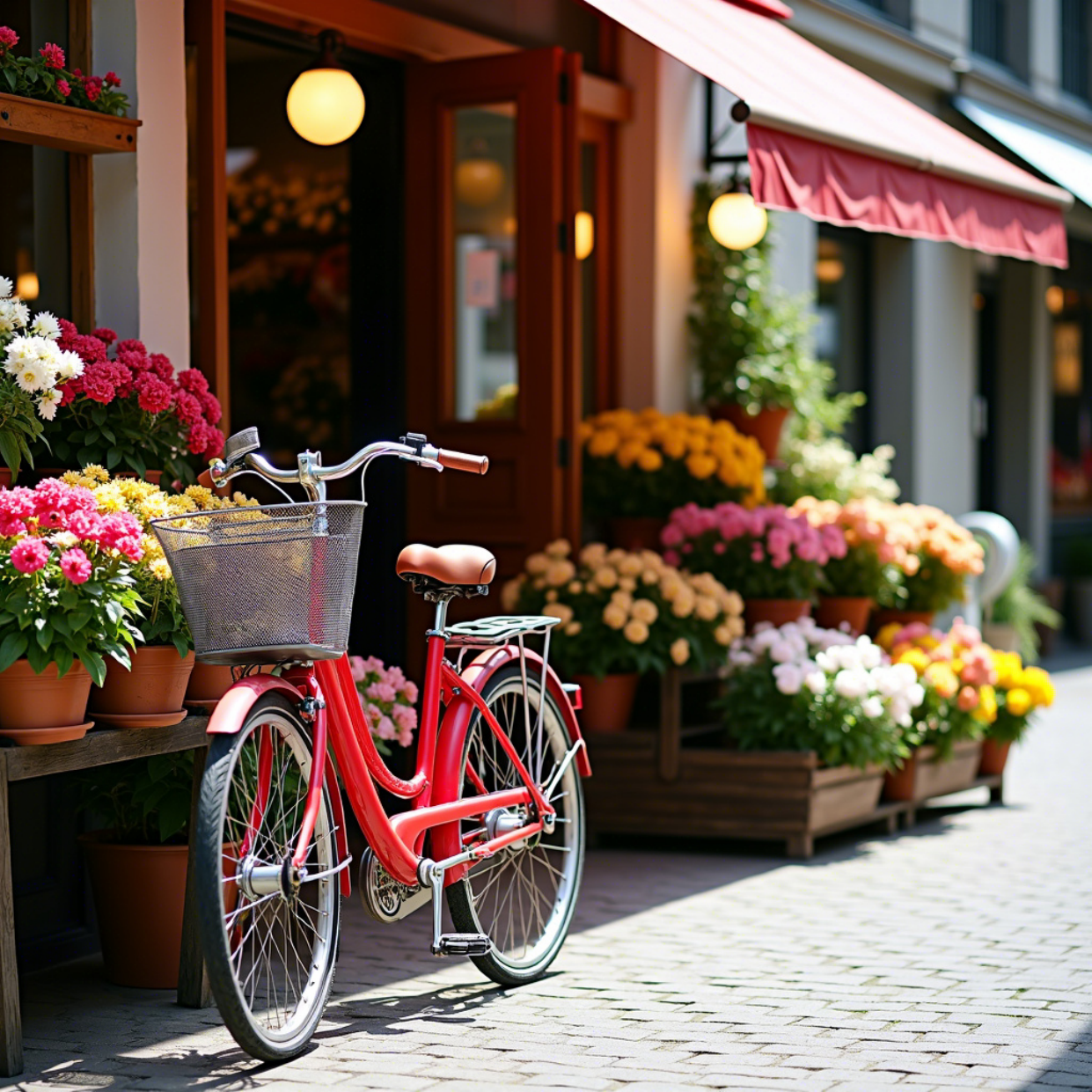}
& \includegraphics[width=0.19\textwidth]{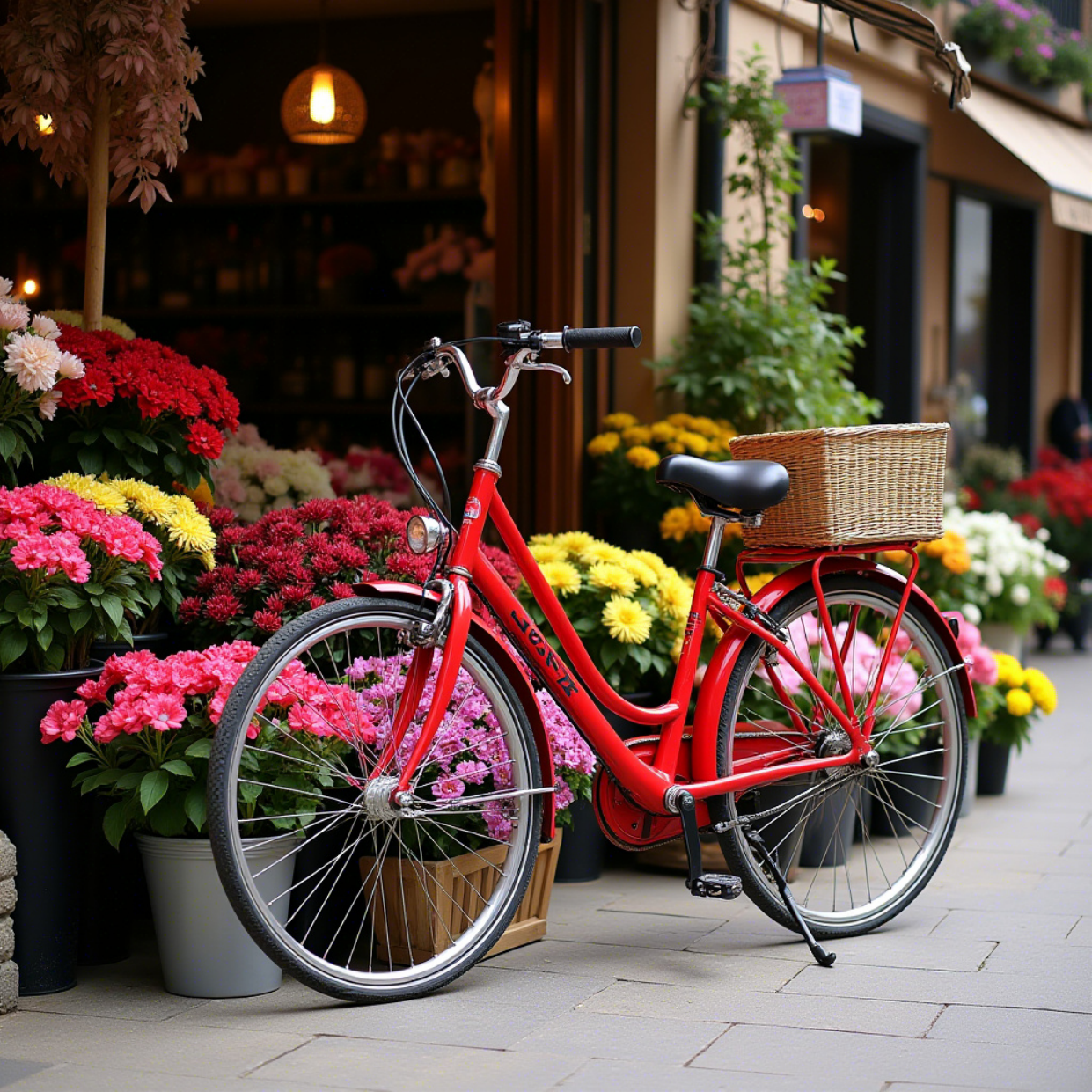}
& \includegraphics[width=0.19\textwidth]{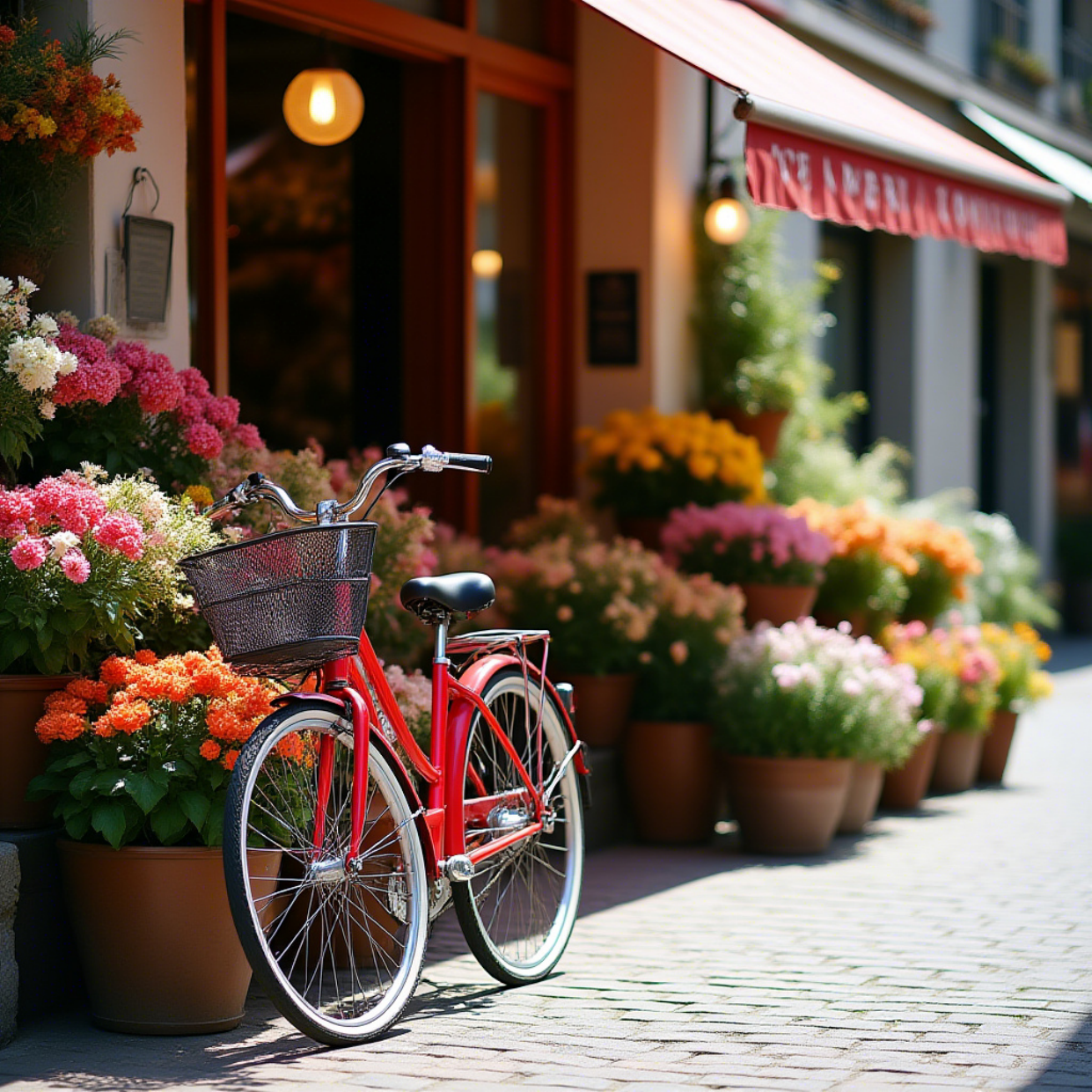}
& \includegraphics[width=0.19\textwidth]{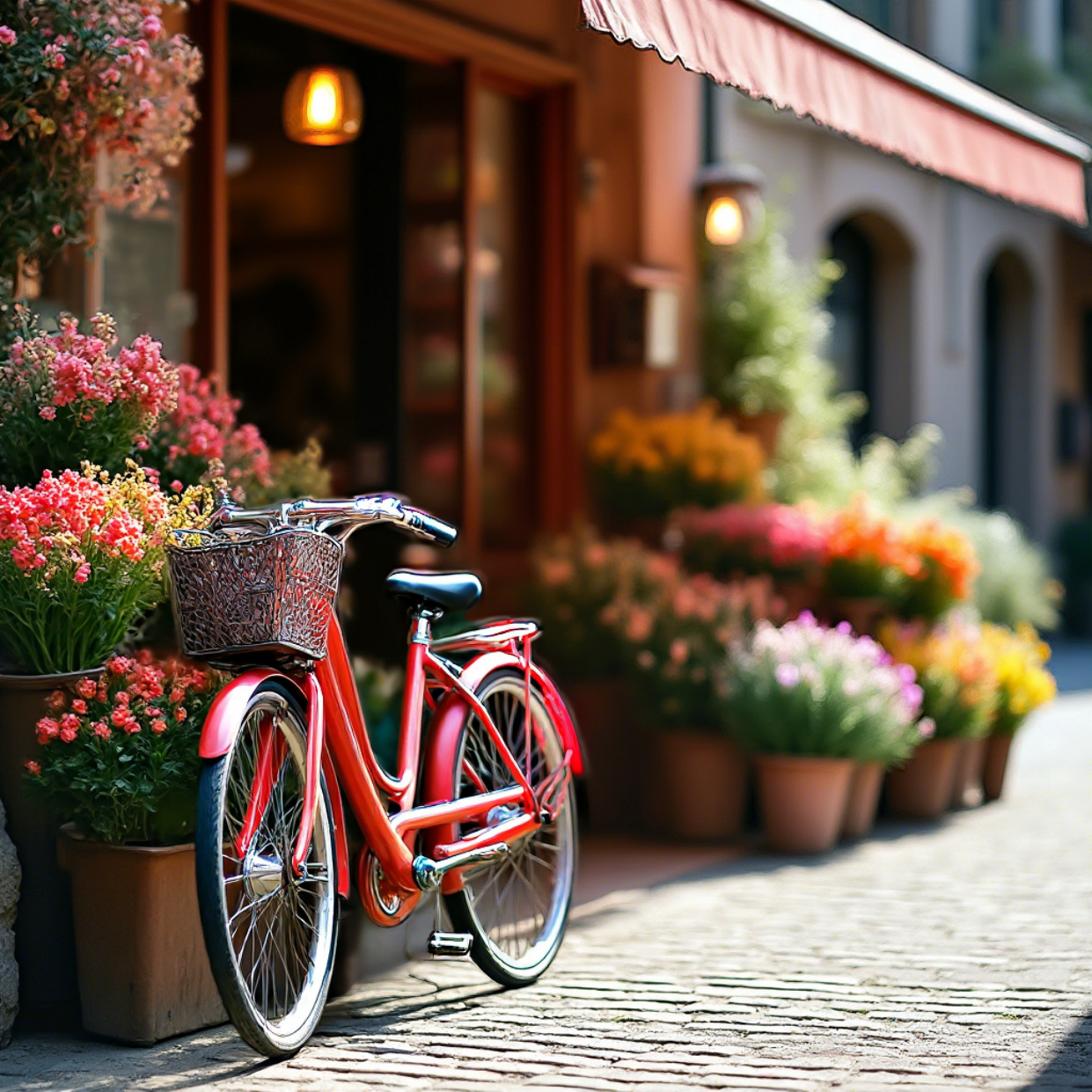}
& \includegraphics[width=0.19\textwidth]{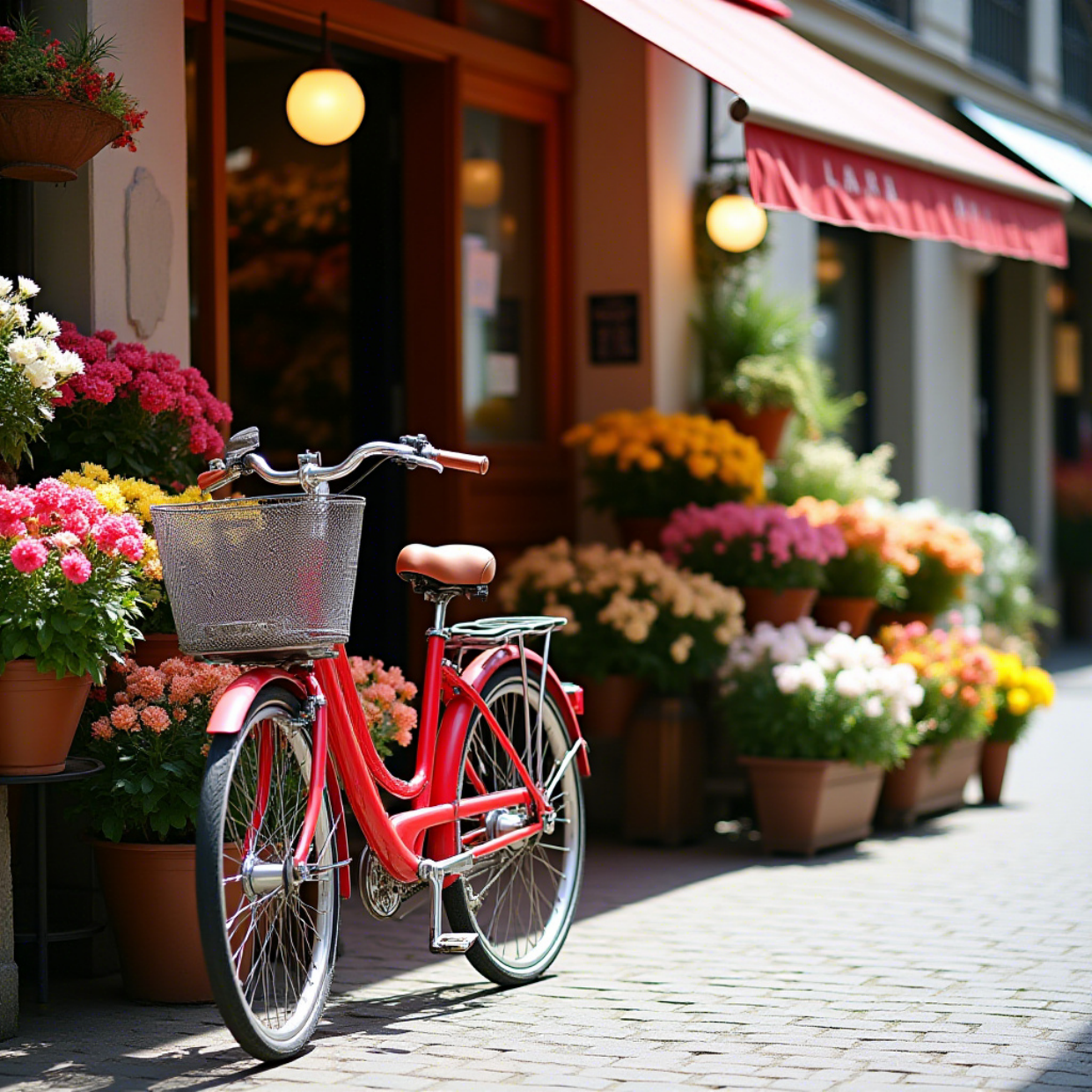}
\\[-0.25em]
\includegraphics[width=0.19\textwidth]{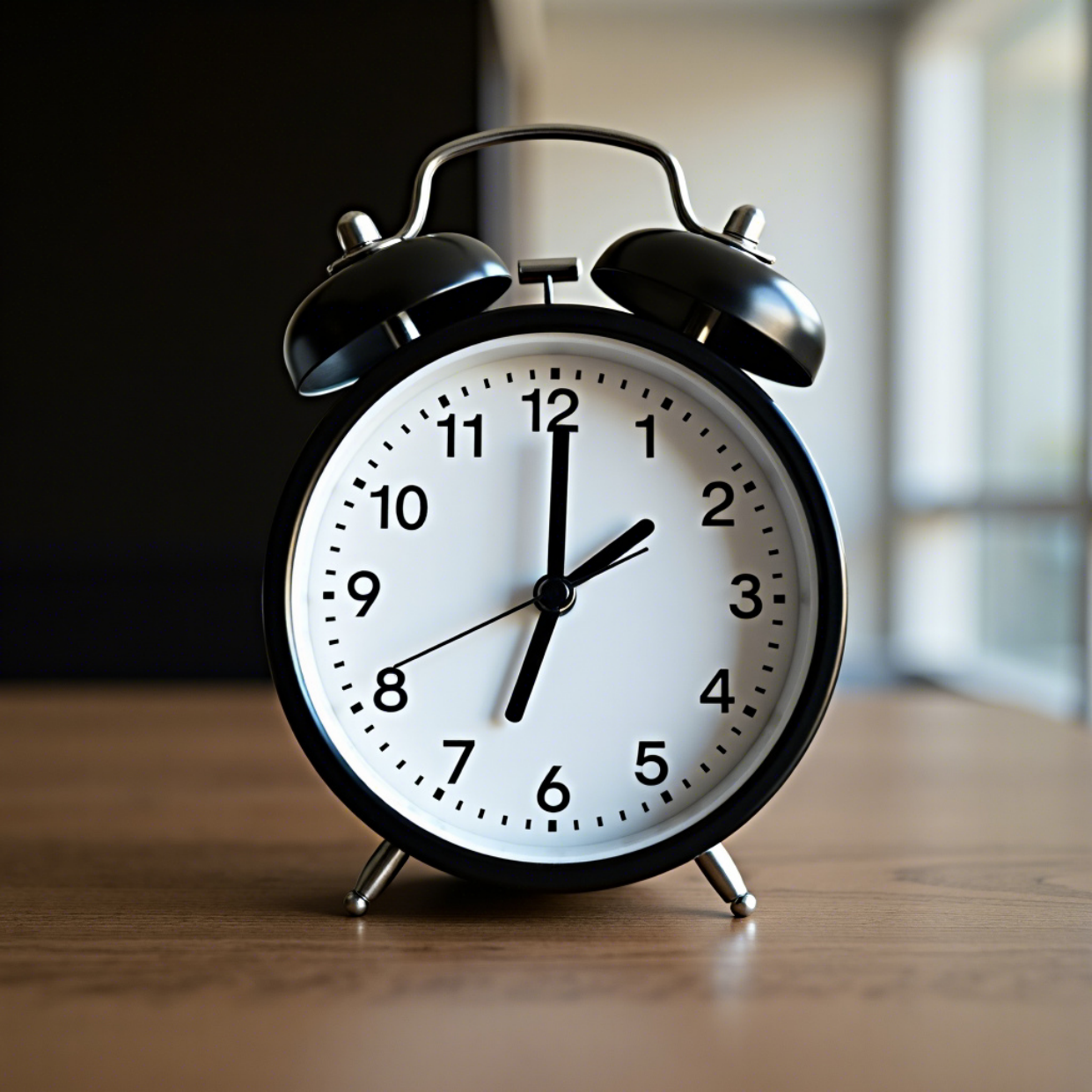}
& \includegraphics[width=0.19\textwidth]{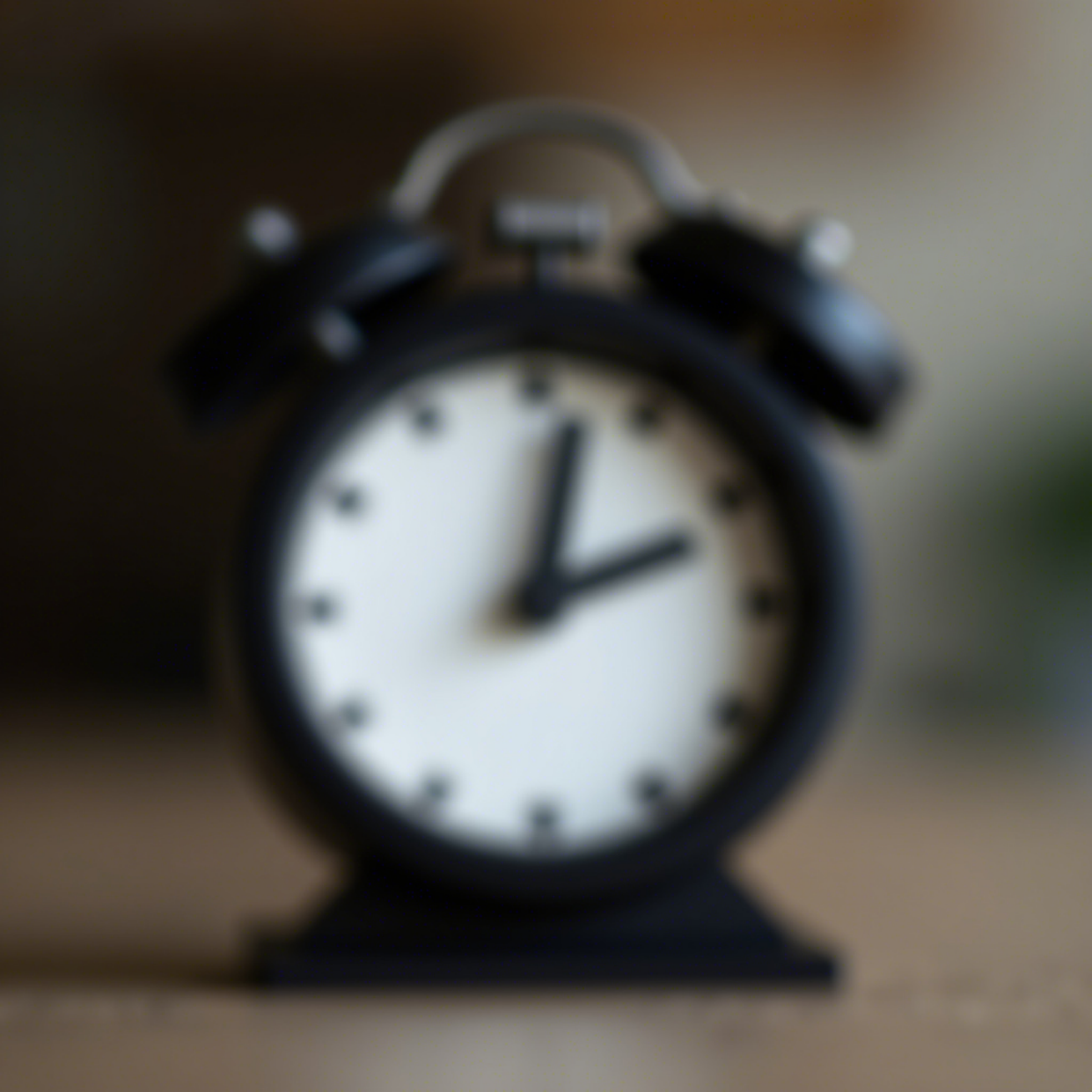}
& \includegraphics[width=0.19\textwidth]{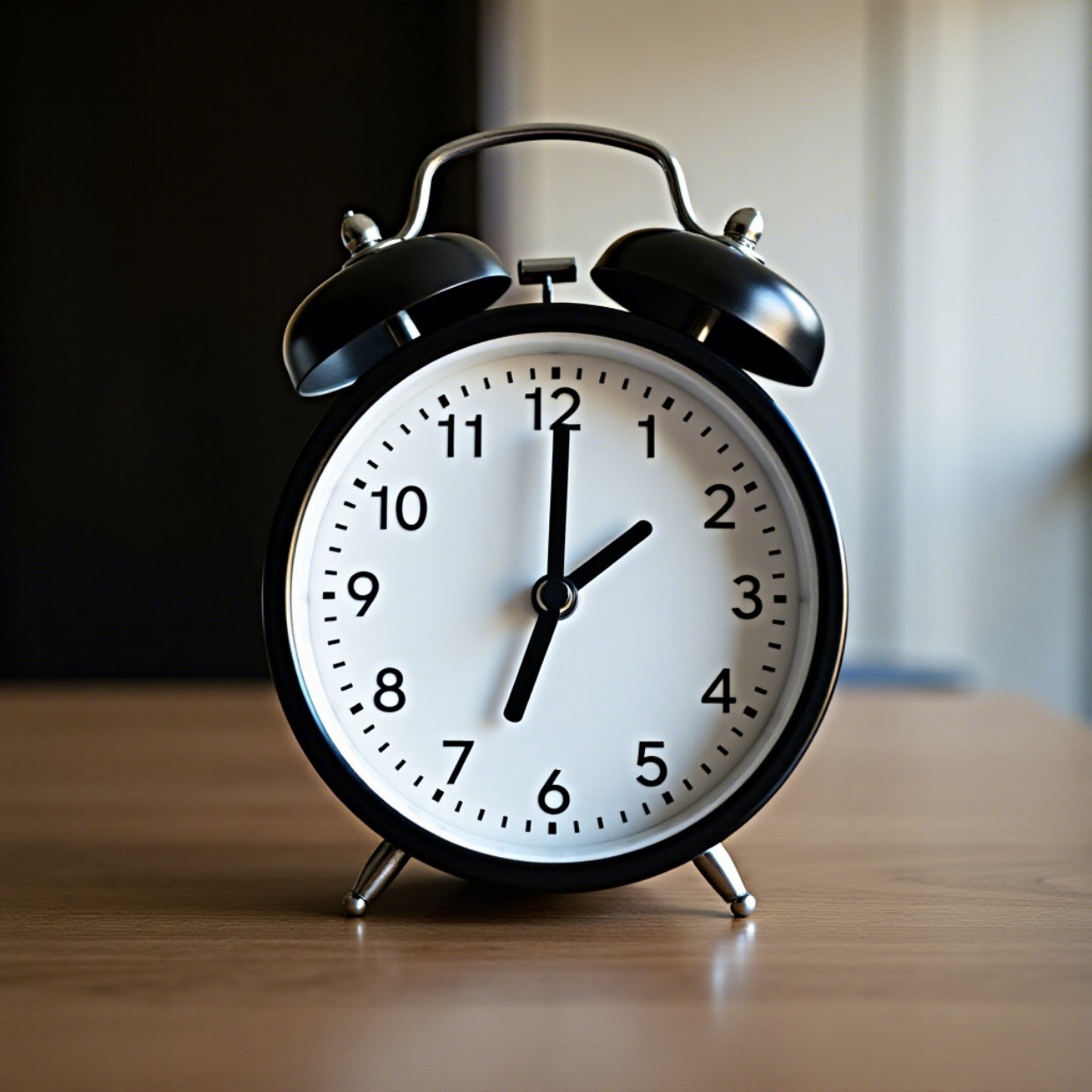}
& \includegraphics[width=0.19\textwidth]{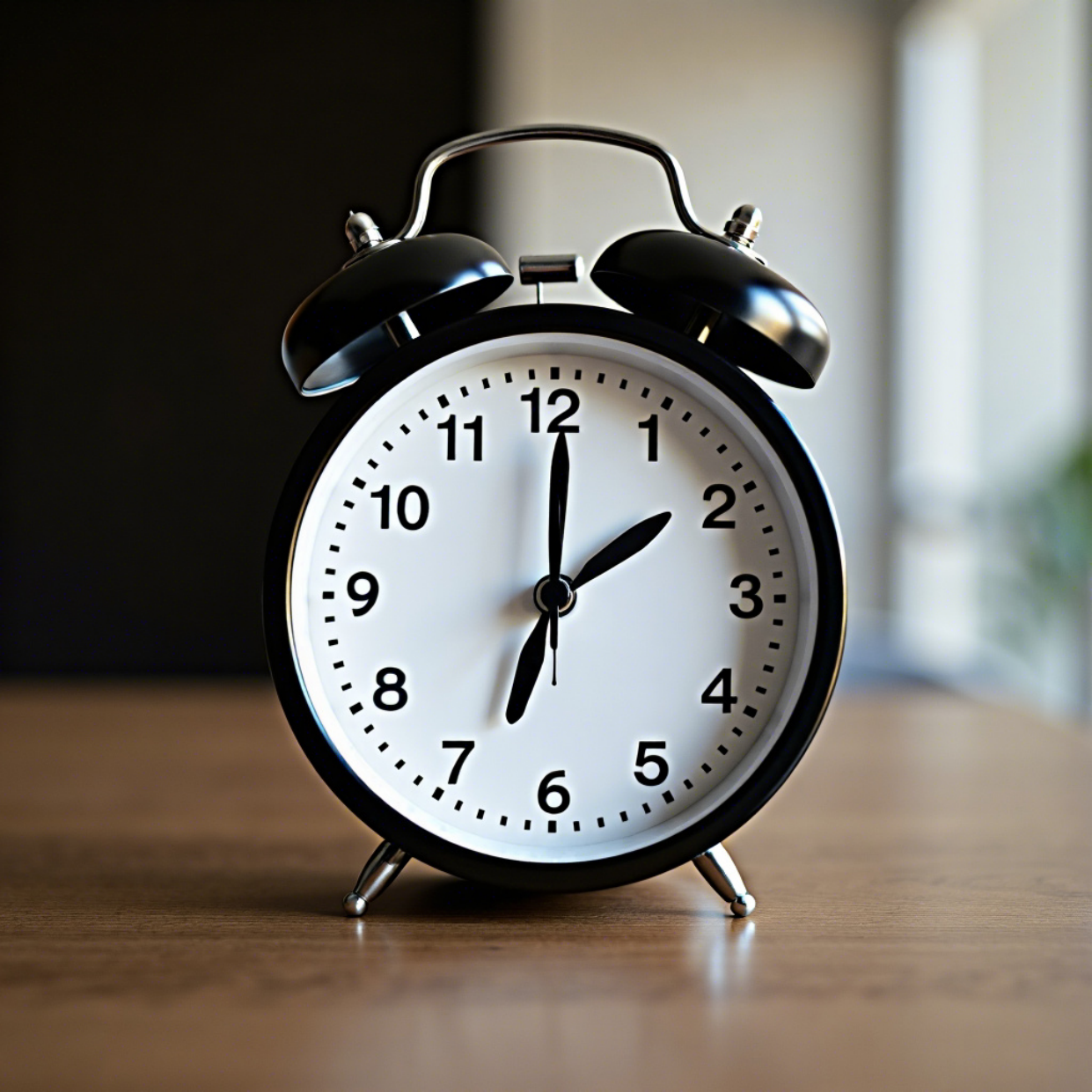}
& \includegraphics[width=0.19\textwidth]{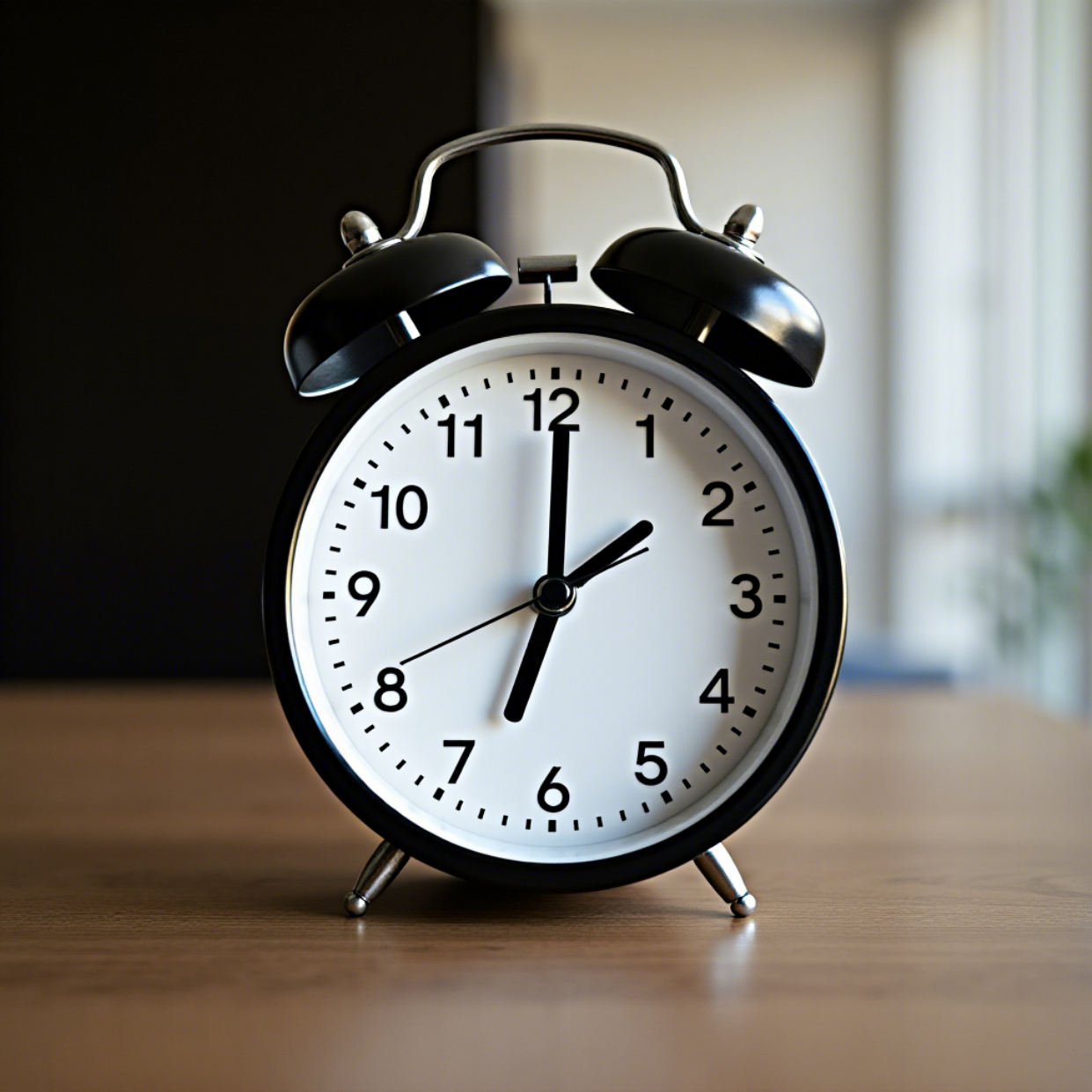}
\\[-0.25em]

\includegraphics[width=0.19\textwidth]{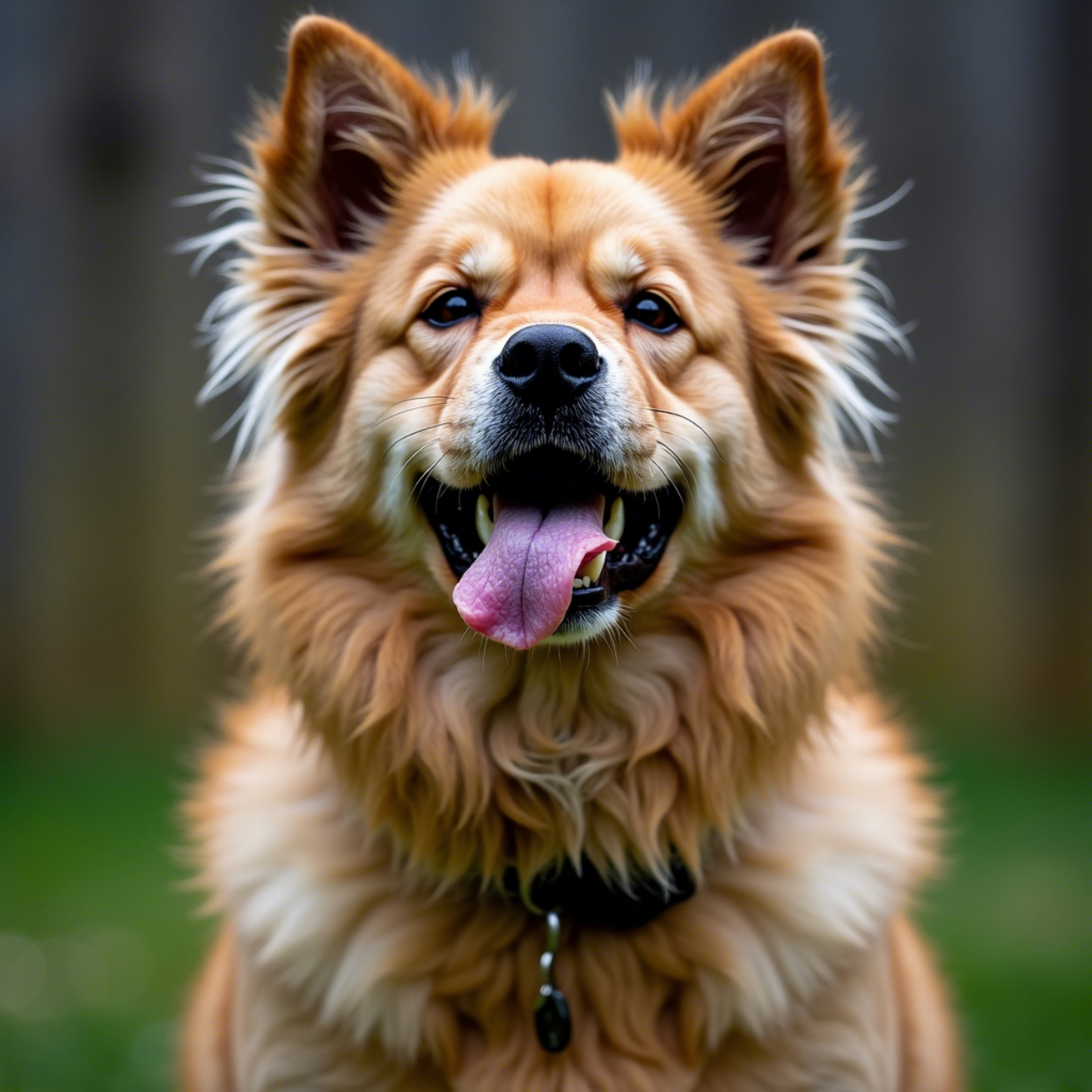}
& \includegraphics[width=0.19\textwidth]{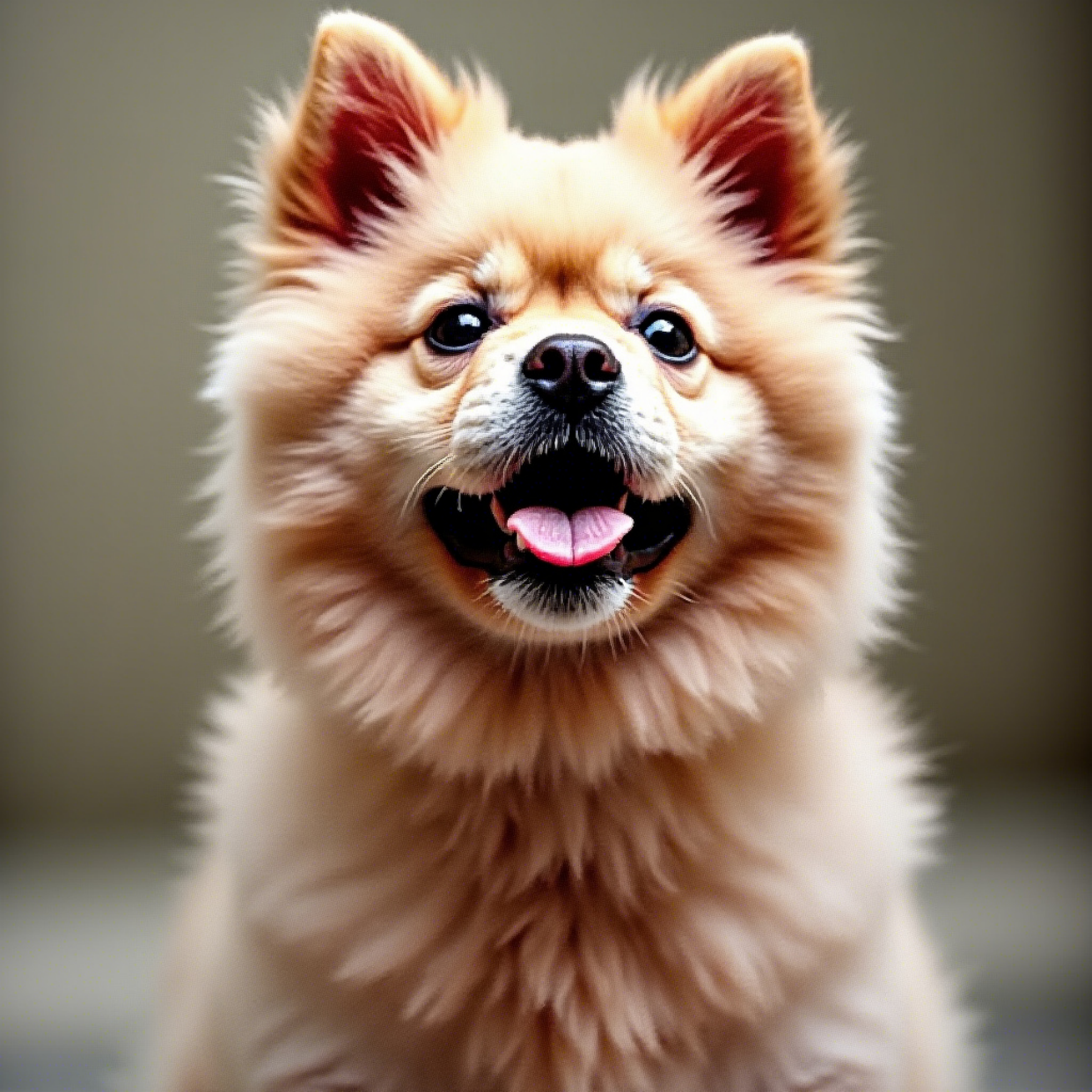}
& \includegraphics[width=0.19\textwidth]{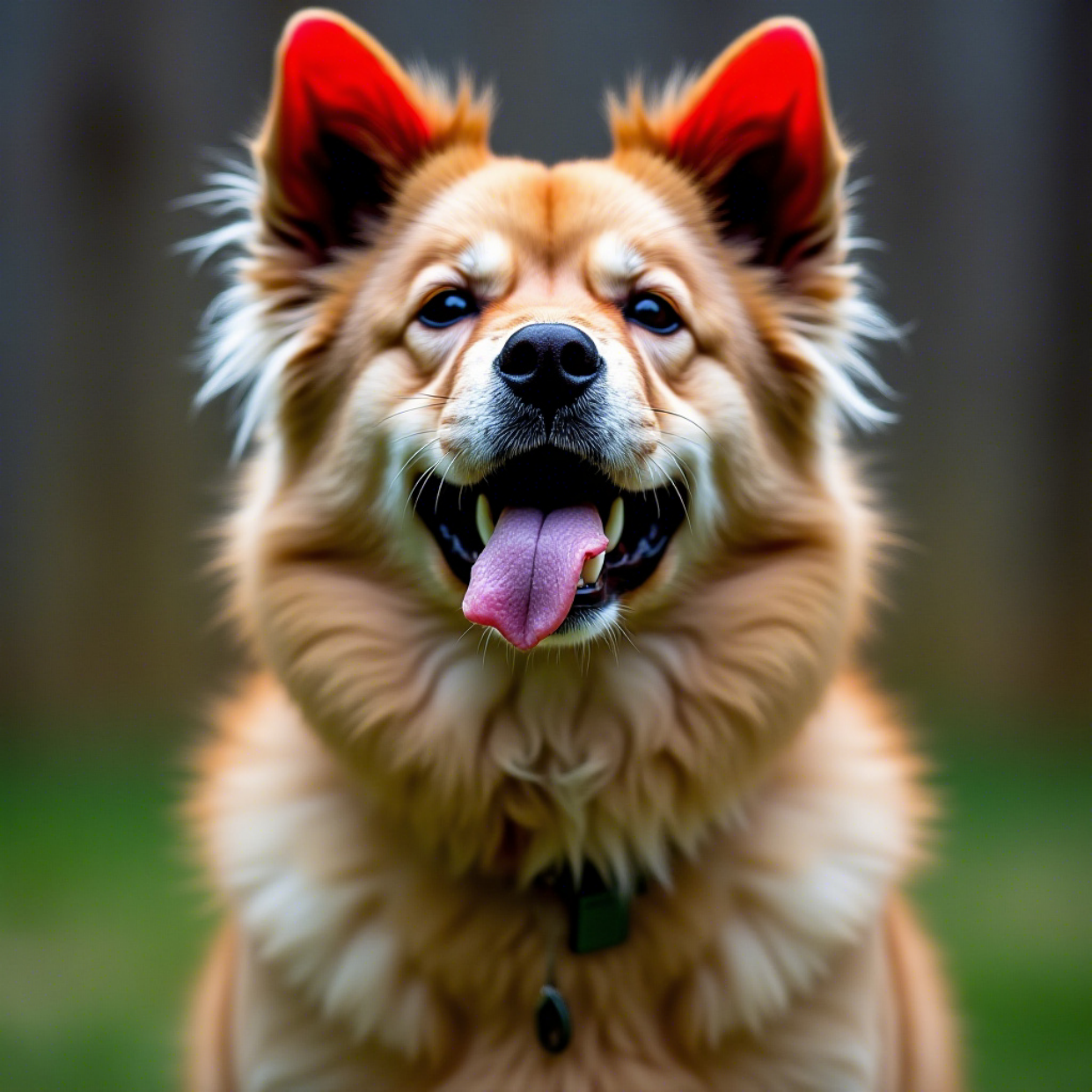}
& \includegraphics[width=0.19\textwidth]{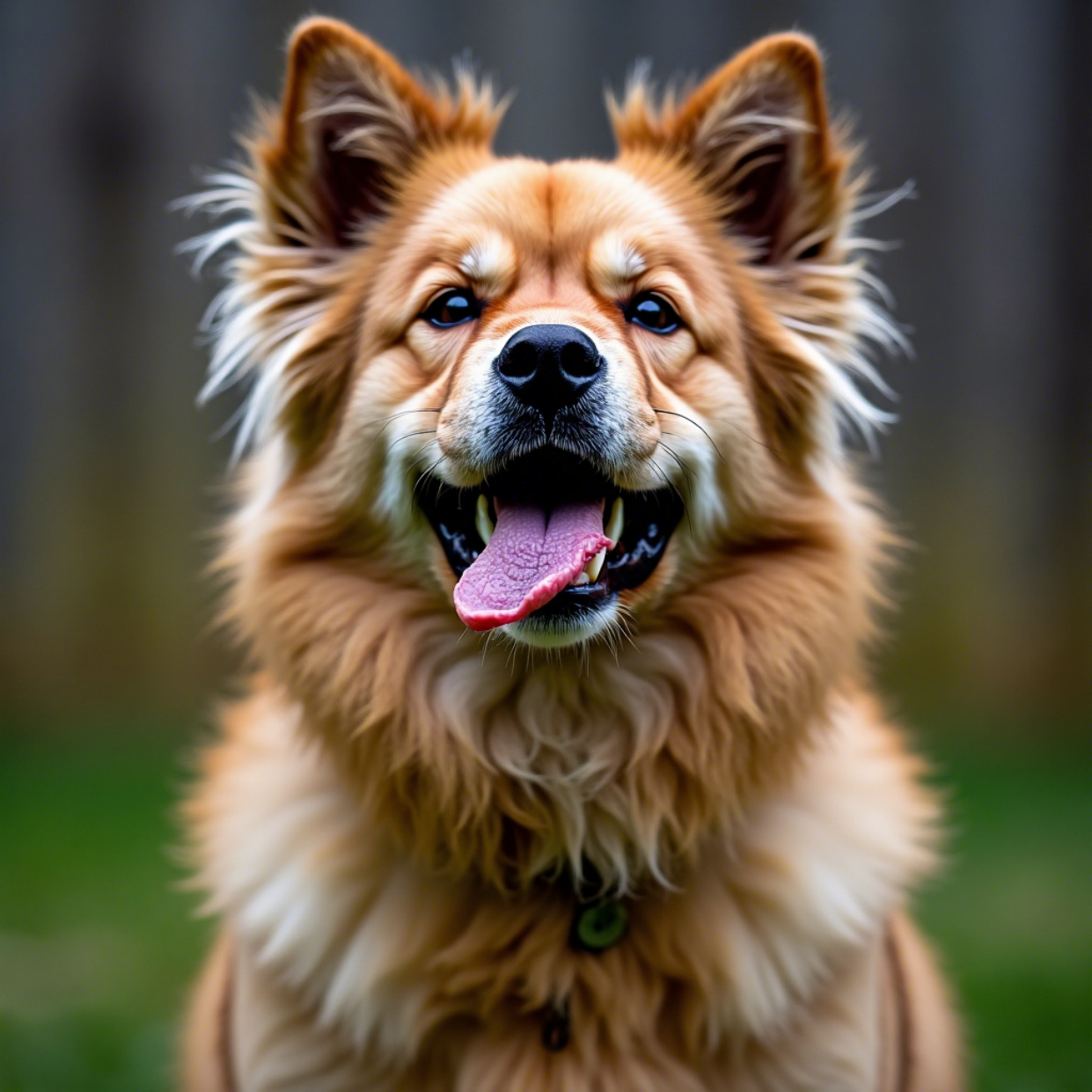}
& \includegraphics[width=0.19\textwidth]{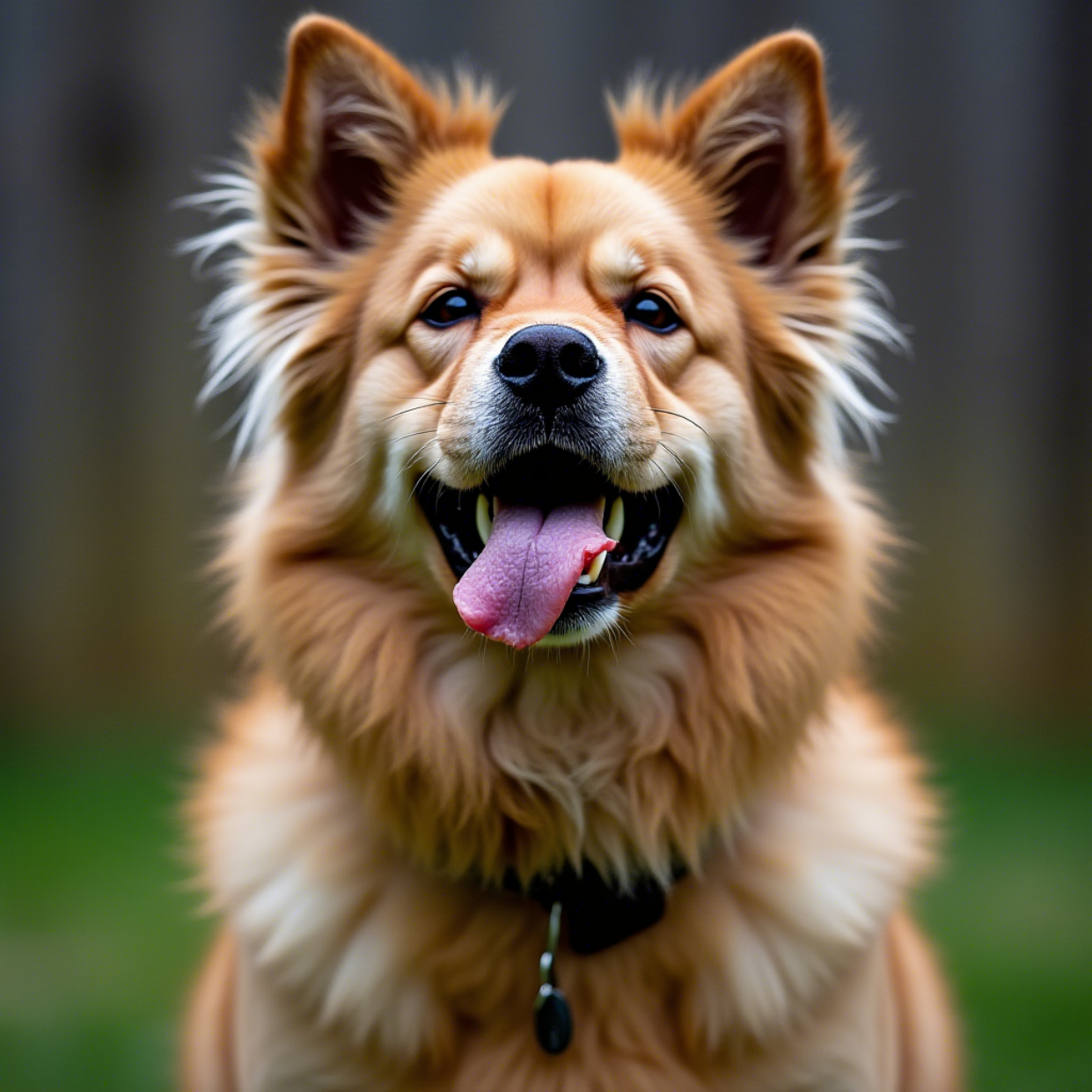}
\\[-0.25em]

\includegraphics[width=0.19\textwidth]{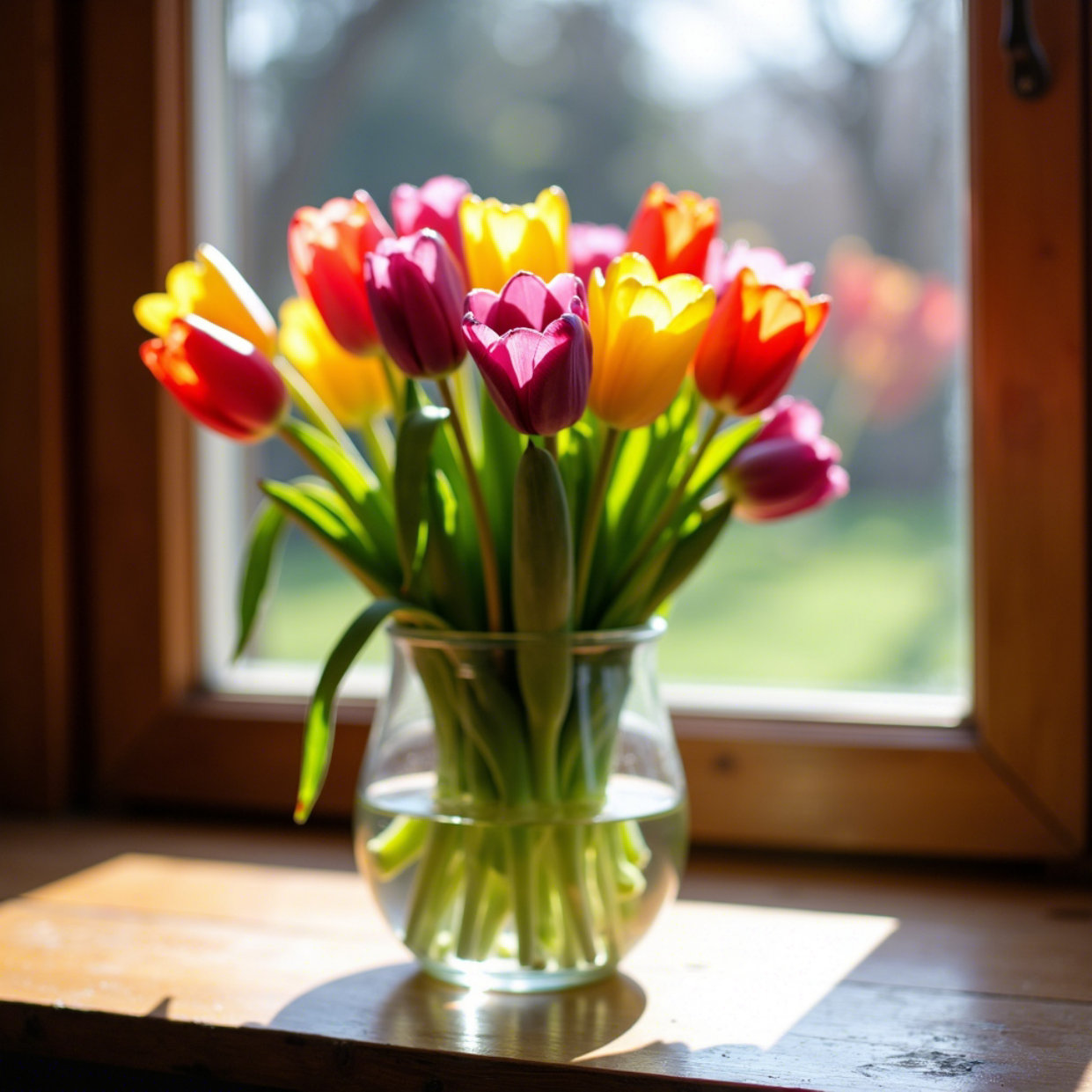}
& \includegraphics[width=0.19\textwidth]{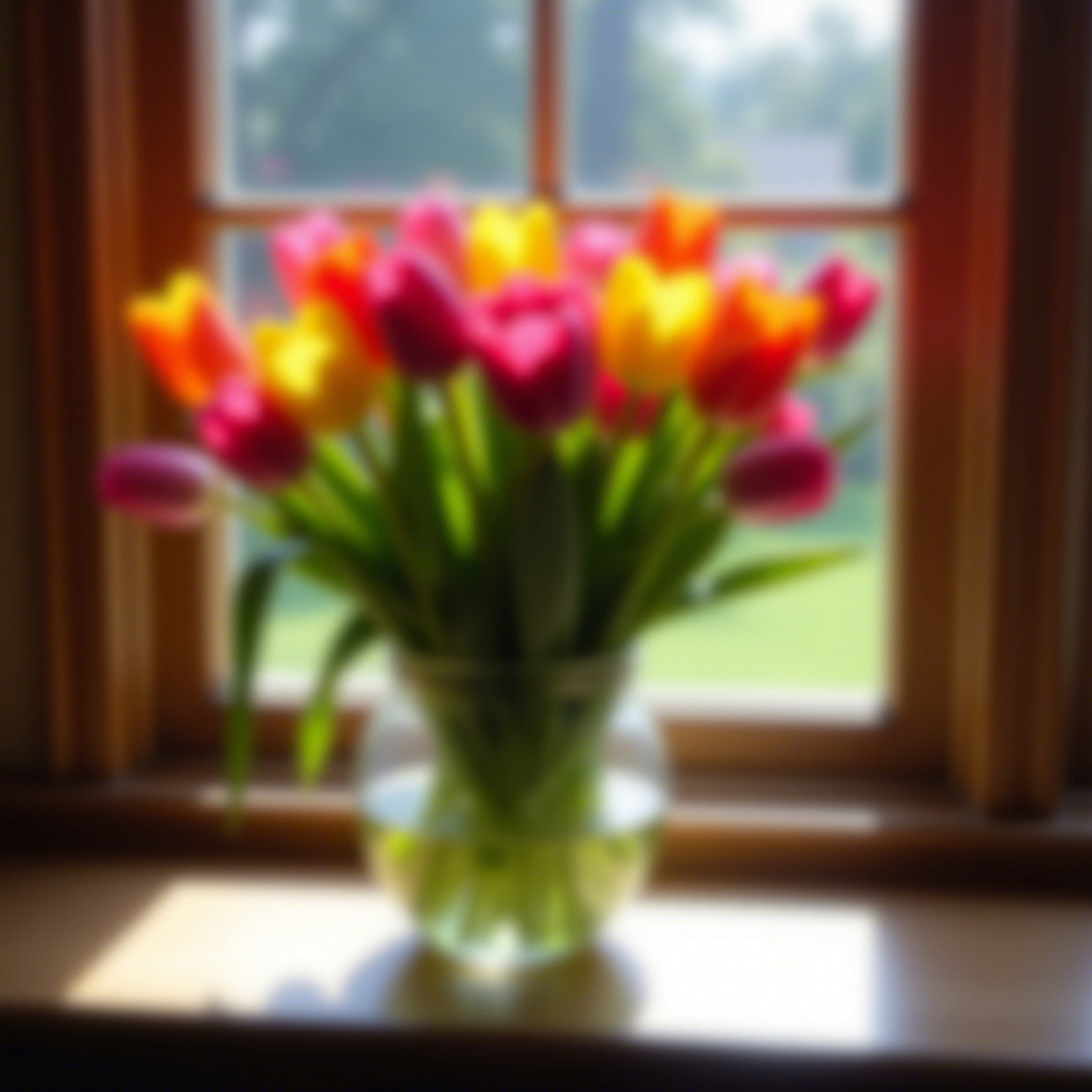}
& \includegraphics[width=0.19\textwidth]{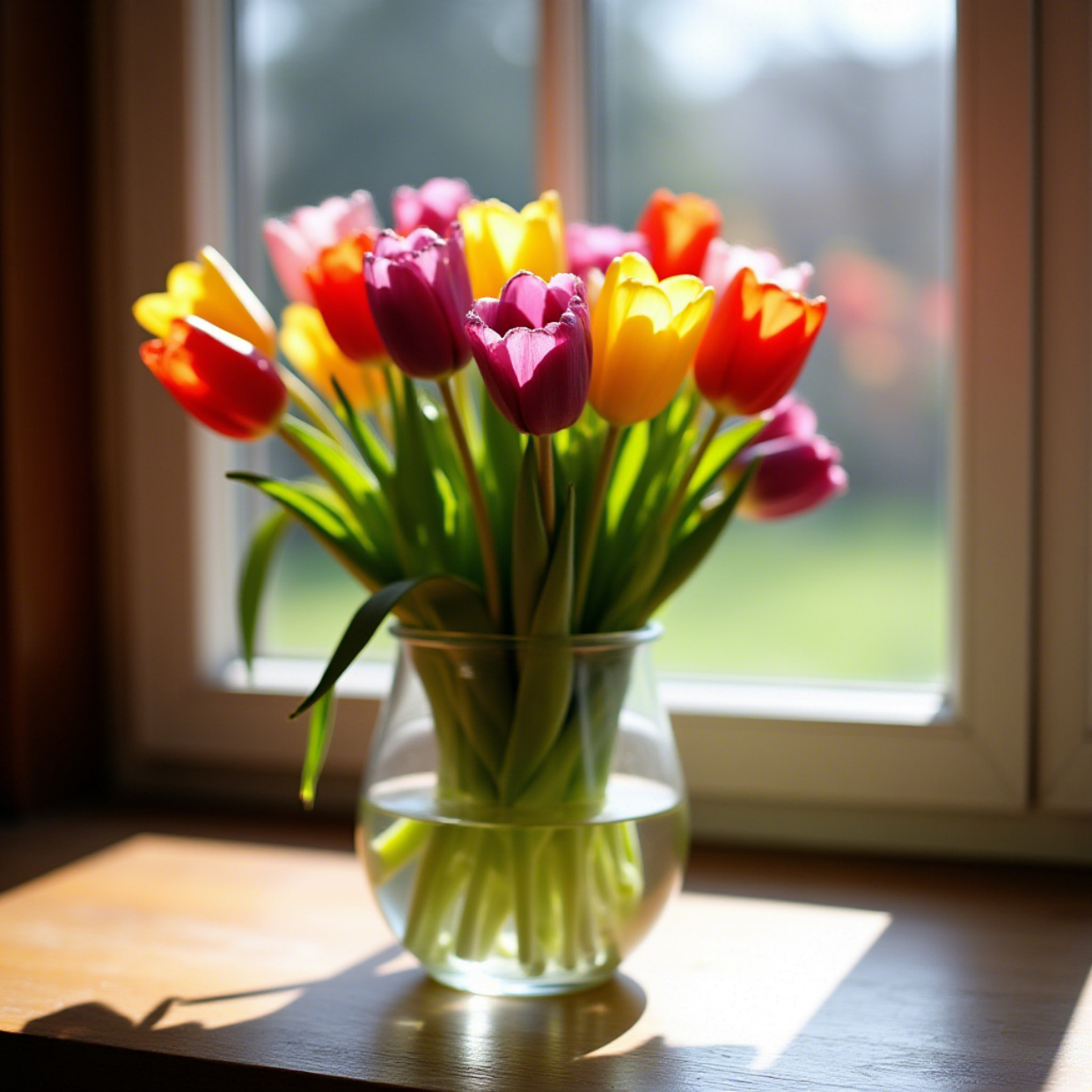}
& \includegraphics[width=0.19\textwidth]{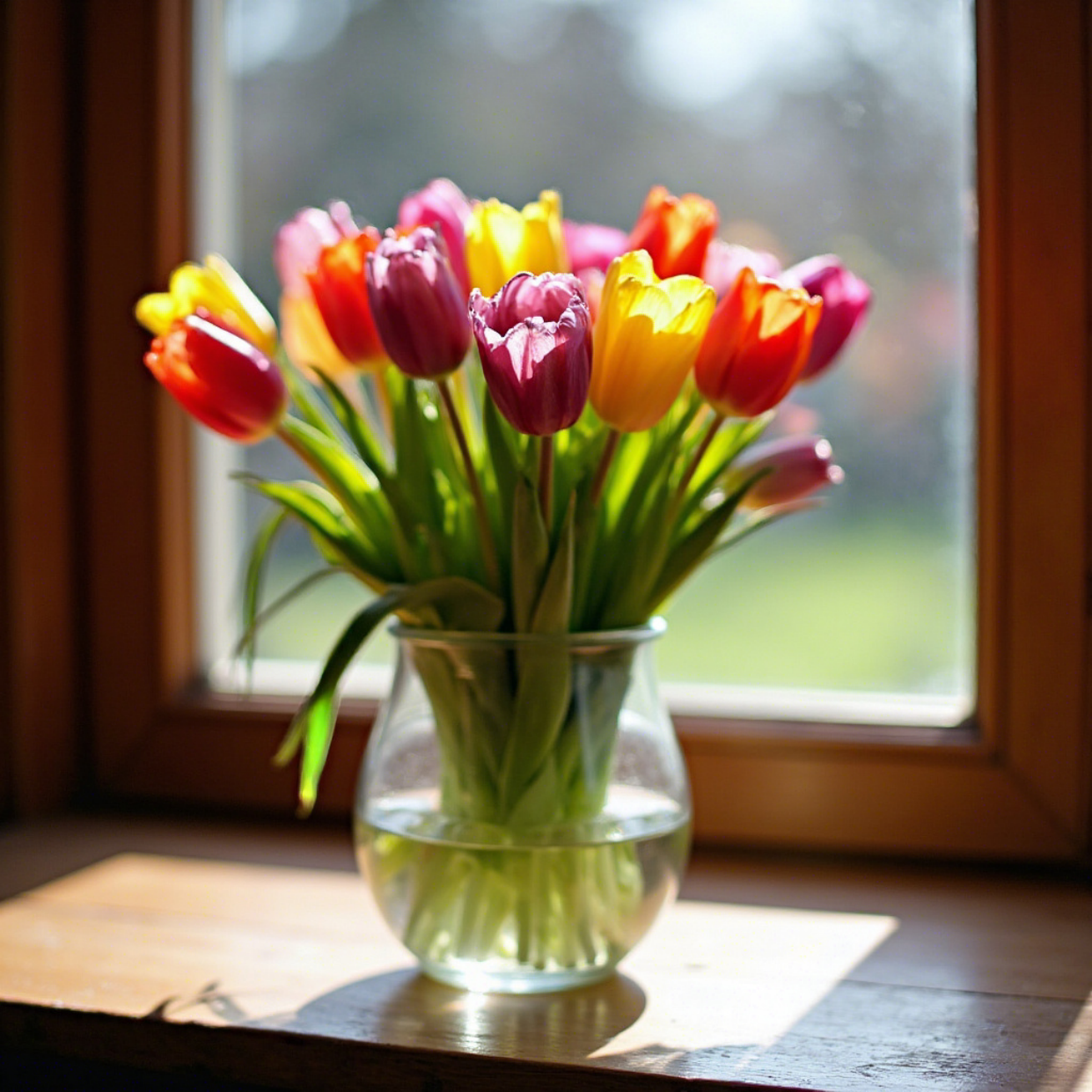}
& \includegraphics[width=0.19\textwidth]{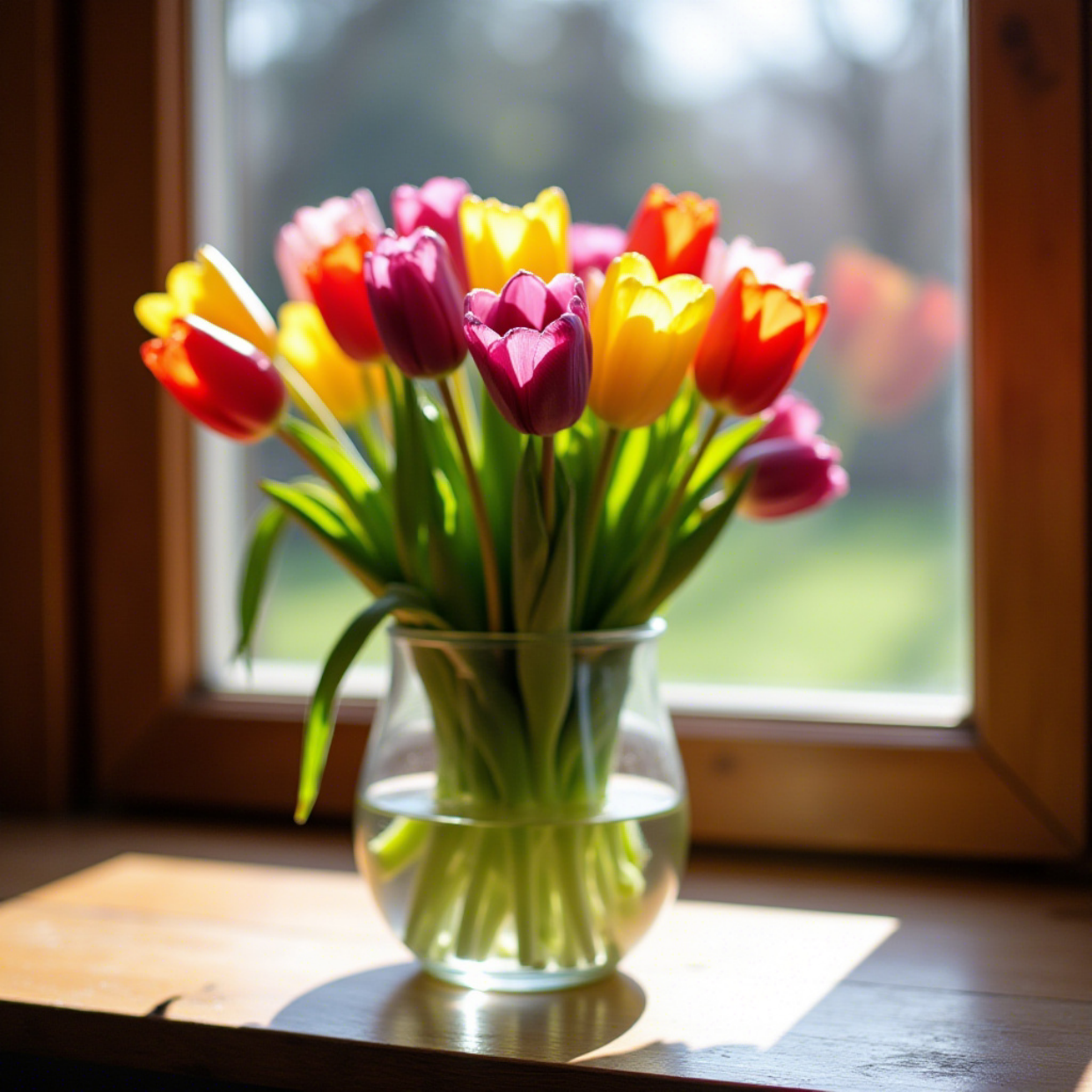}
\\[-0.25em]

\includegraphics[width=0.19\textwidth]{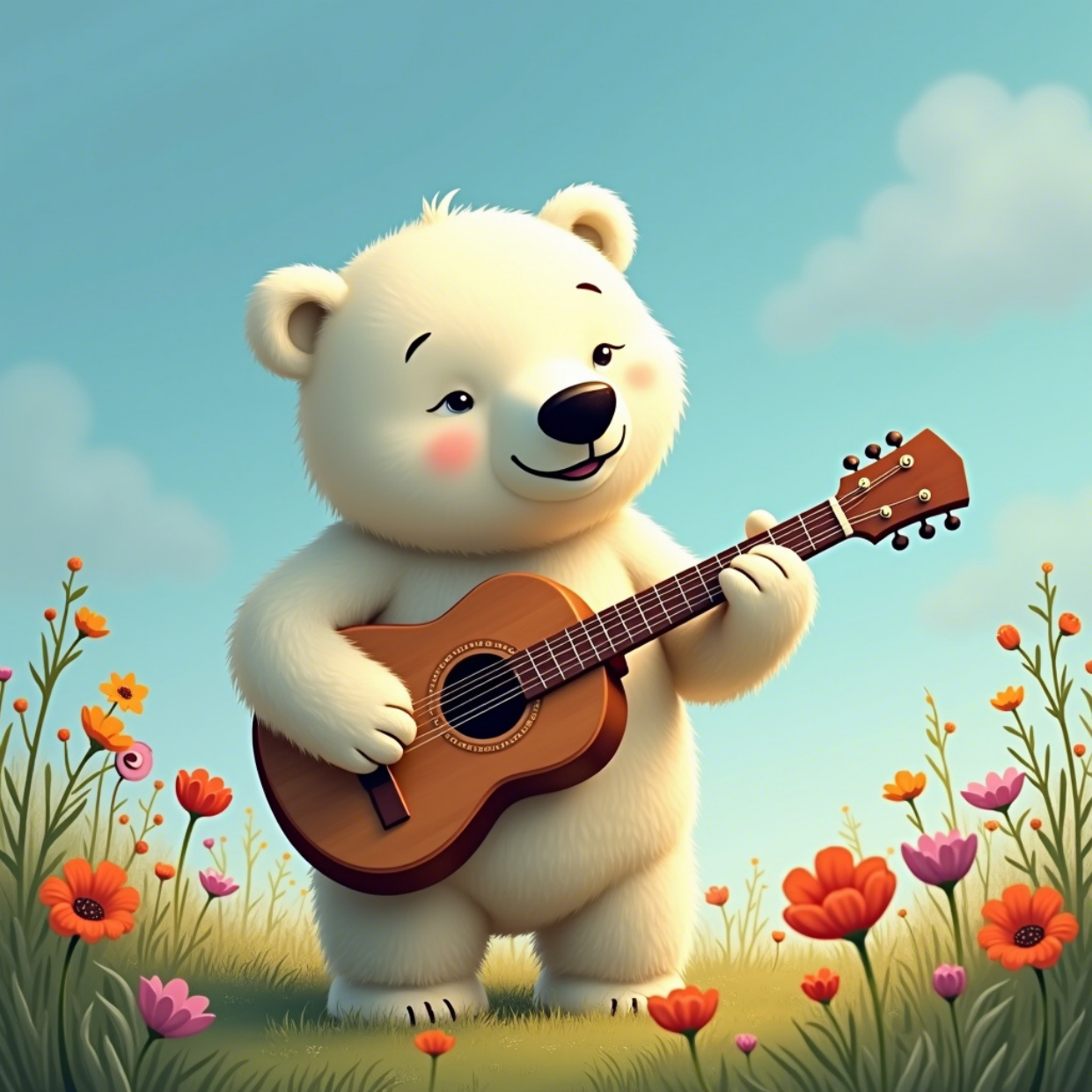}
& \includegraphics[width=0.19\textwidth]{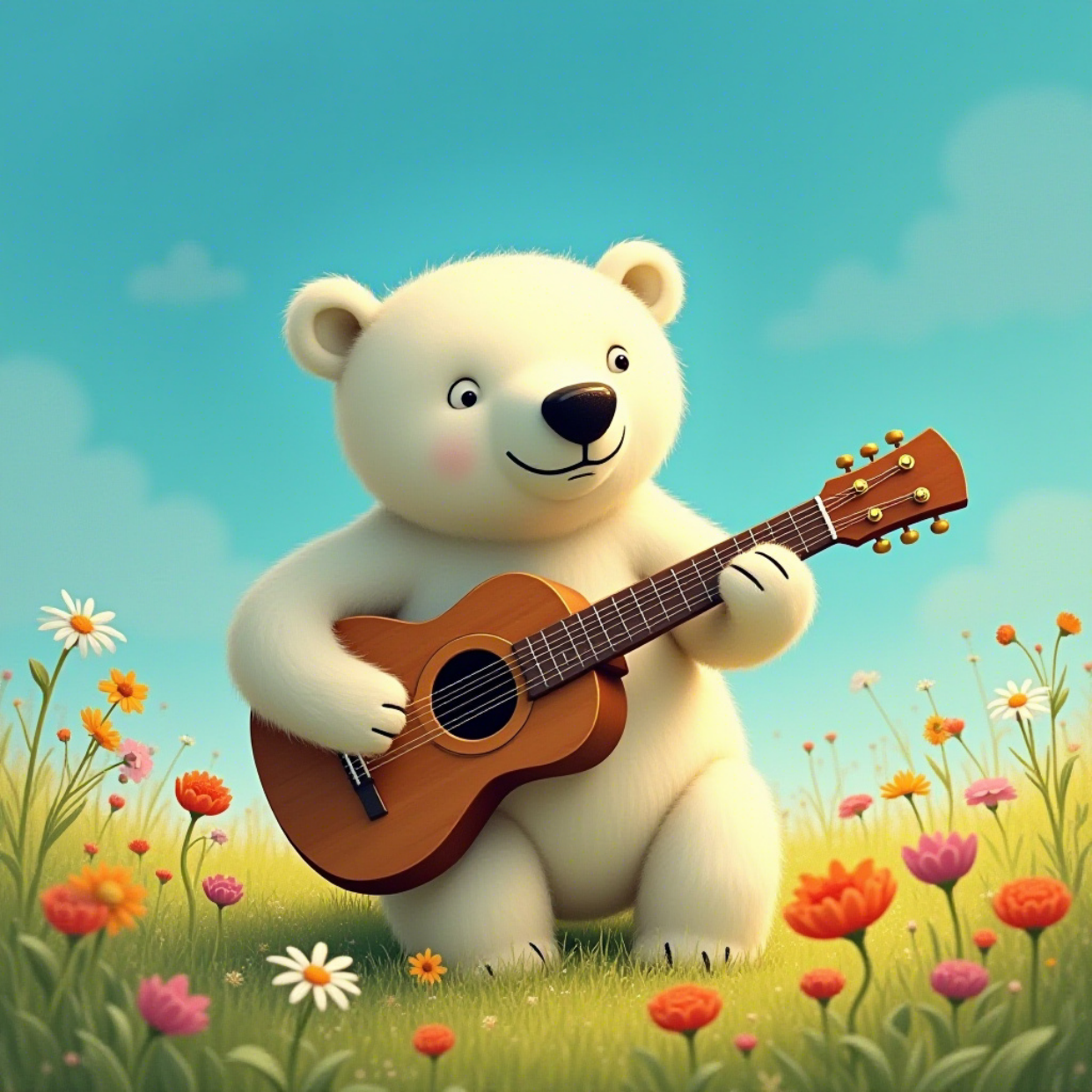}
& \includegraphics[width=0.19\textwidth]{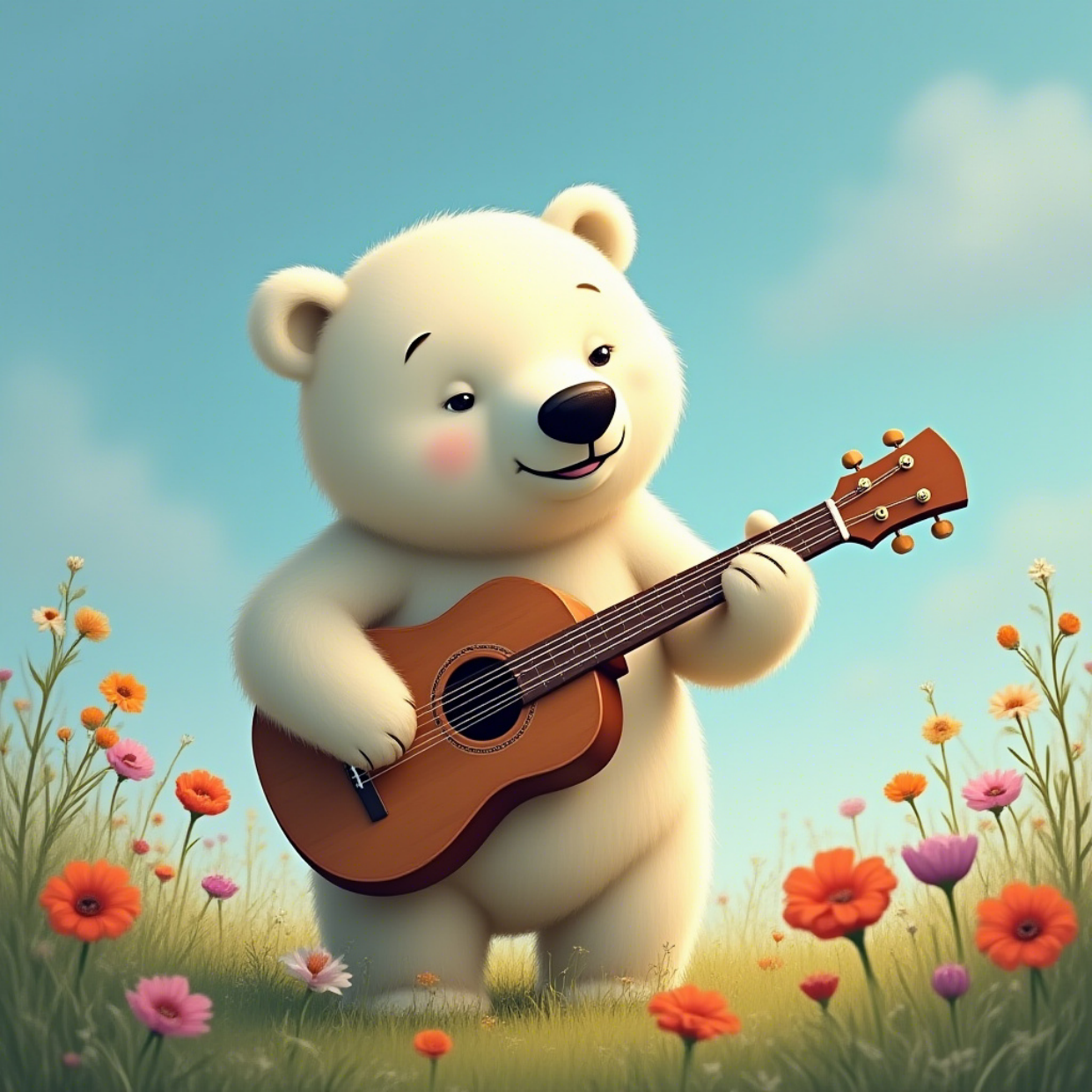}
& \includegraphics[width=0.19\textwidth]{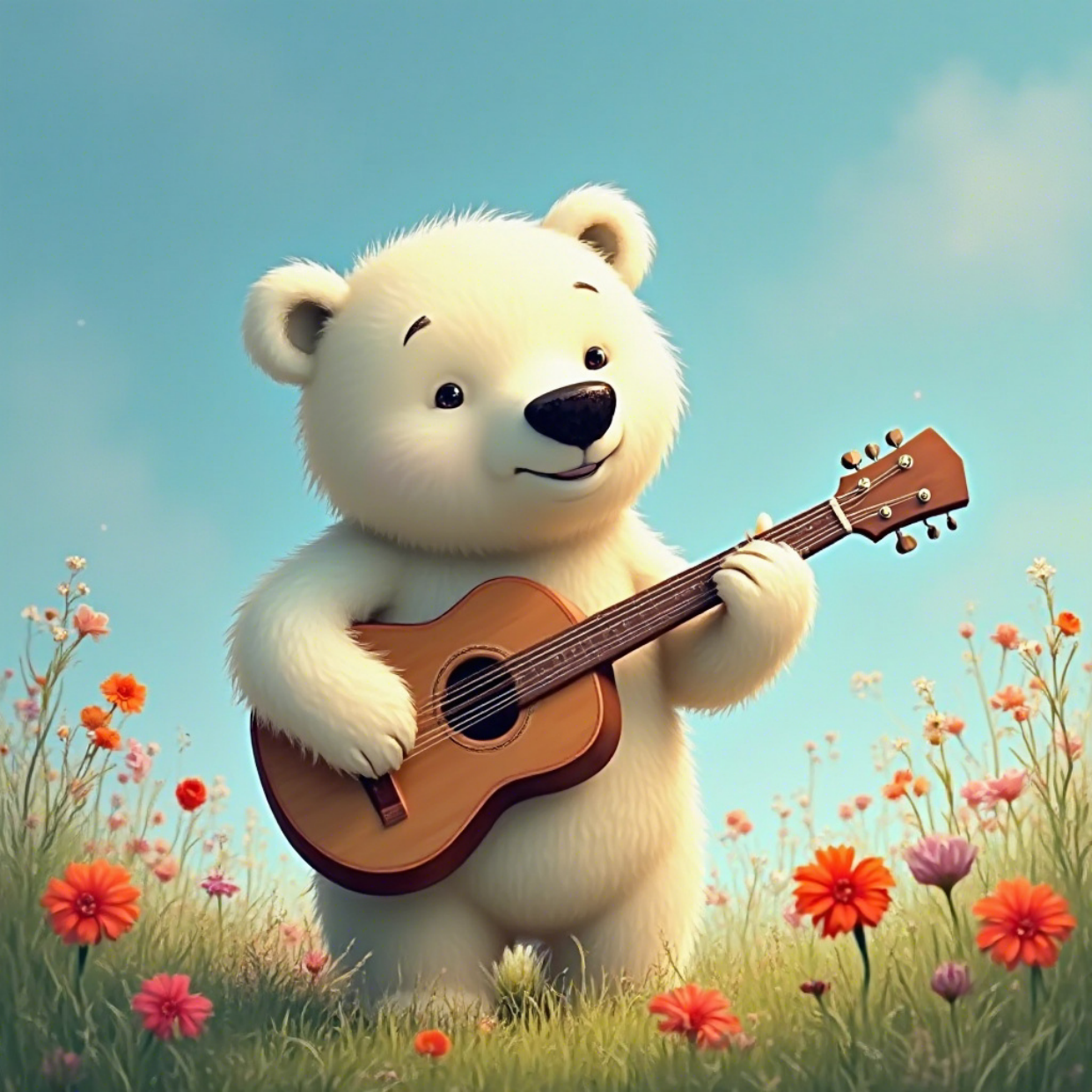}
& \includegraphics[width=0.19\textwidth]{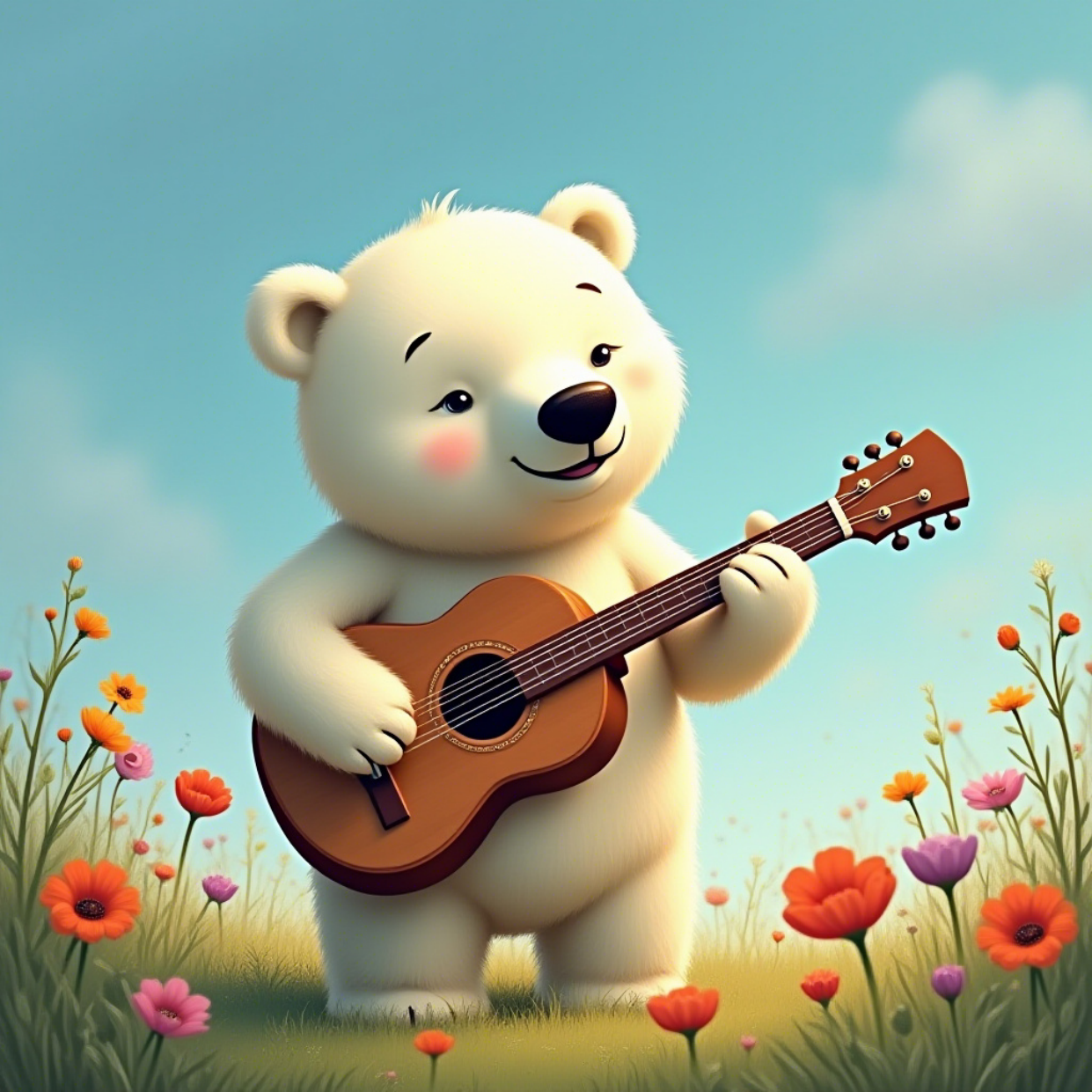}
\\[-0.25em]

\includegraphics[width=0.19\textwidth]{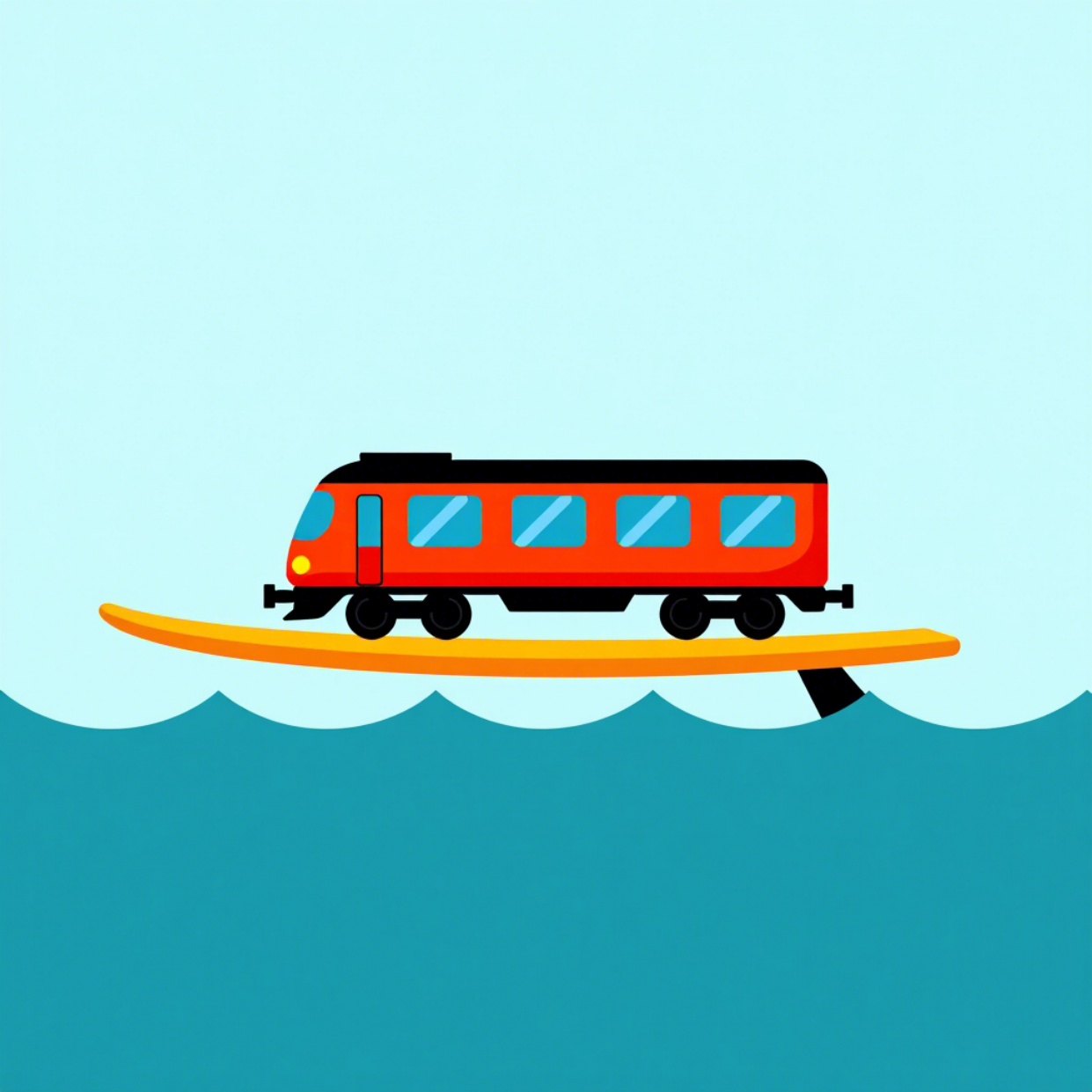}
& \includegraphics[width=0.19\textwidth]{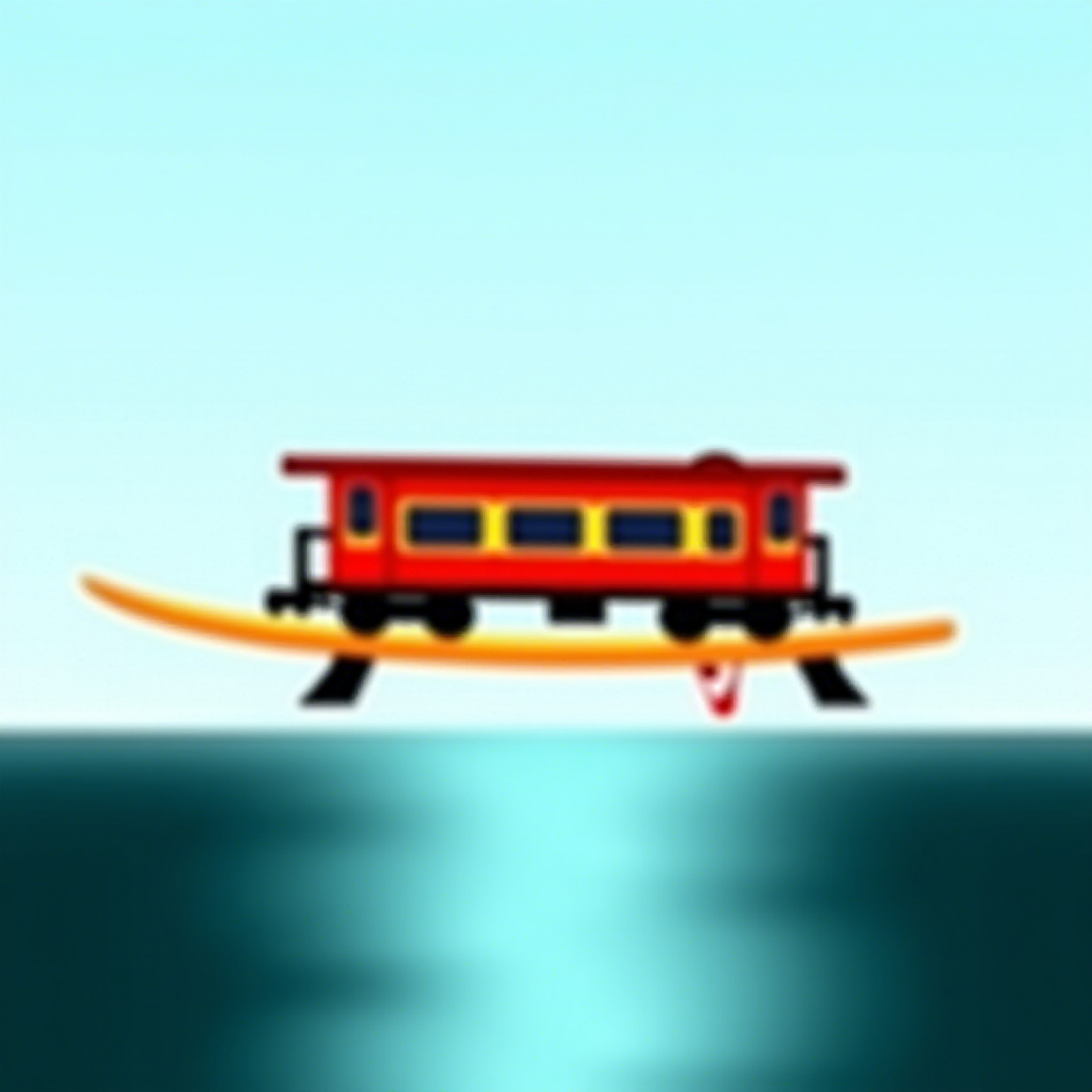}
& \includegraphics[width=0.19\textwidth]{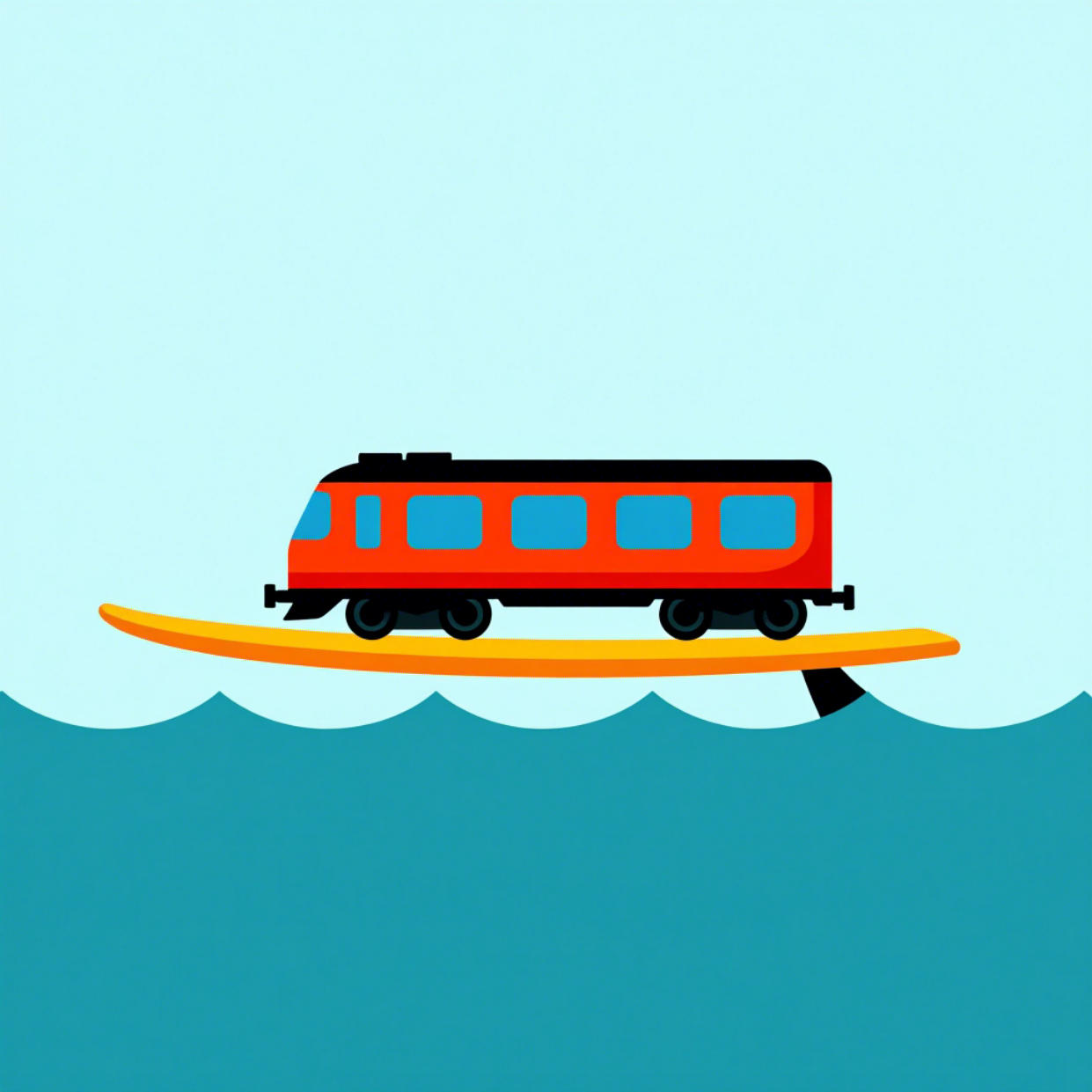}
& \includegraphics[width=0.19\textwidth]{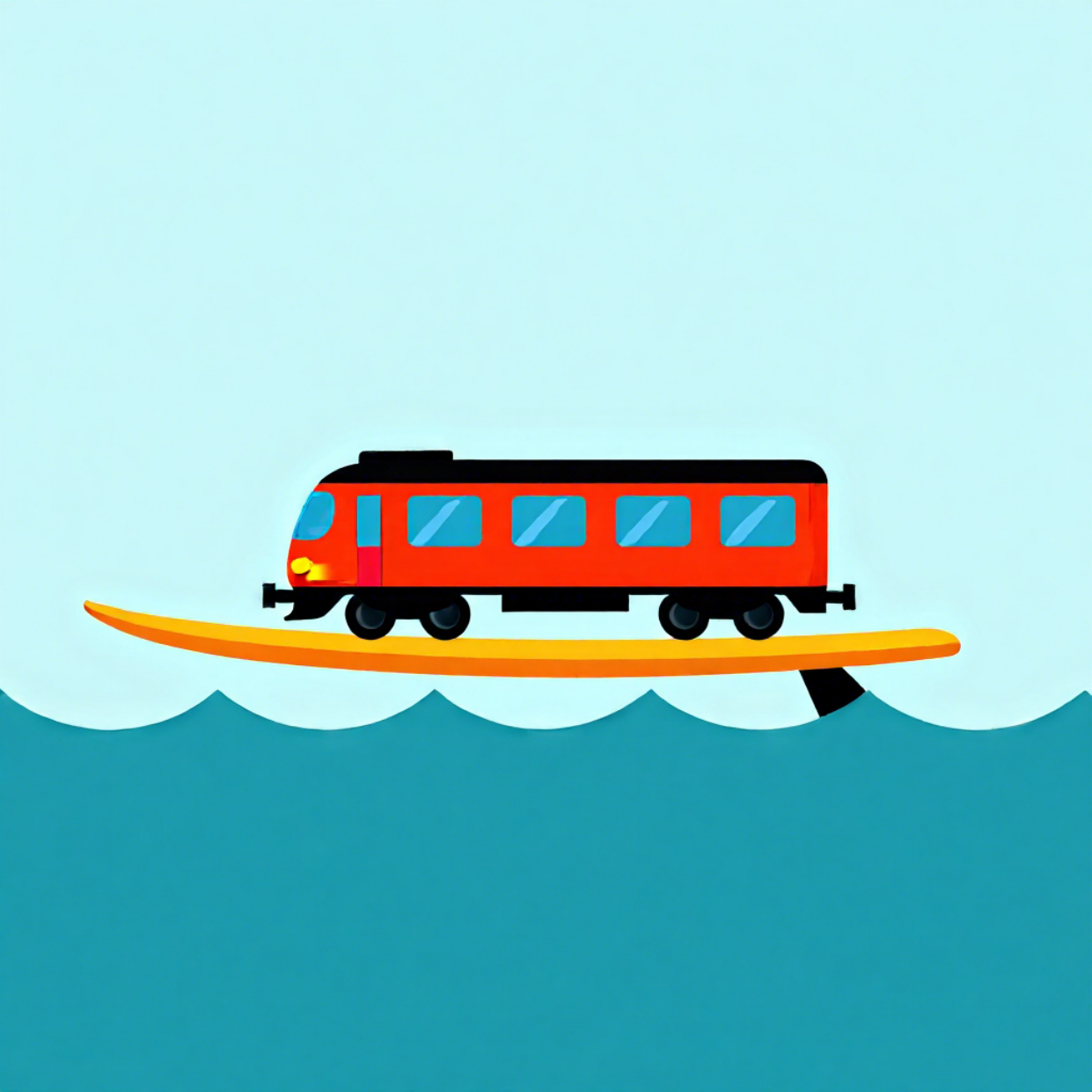}
& \includegraphics[width=0.19\textwidth]{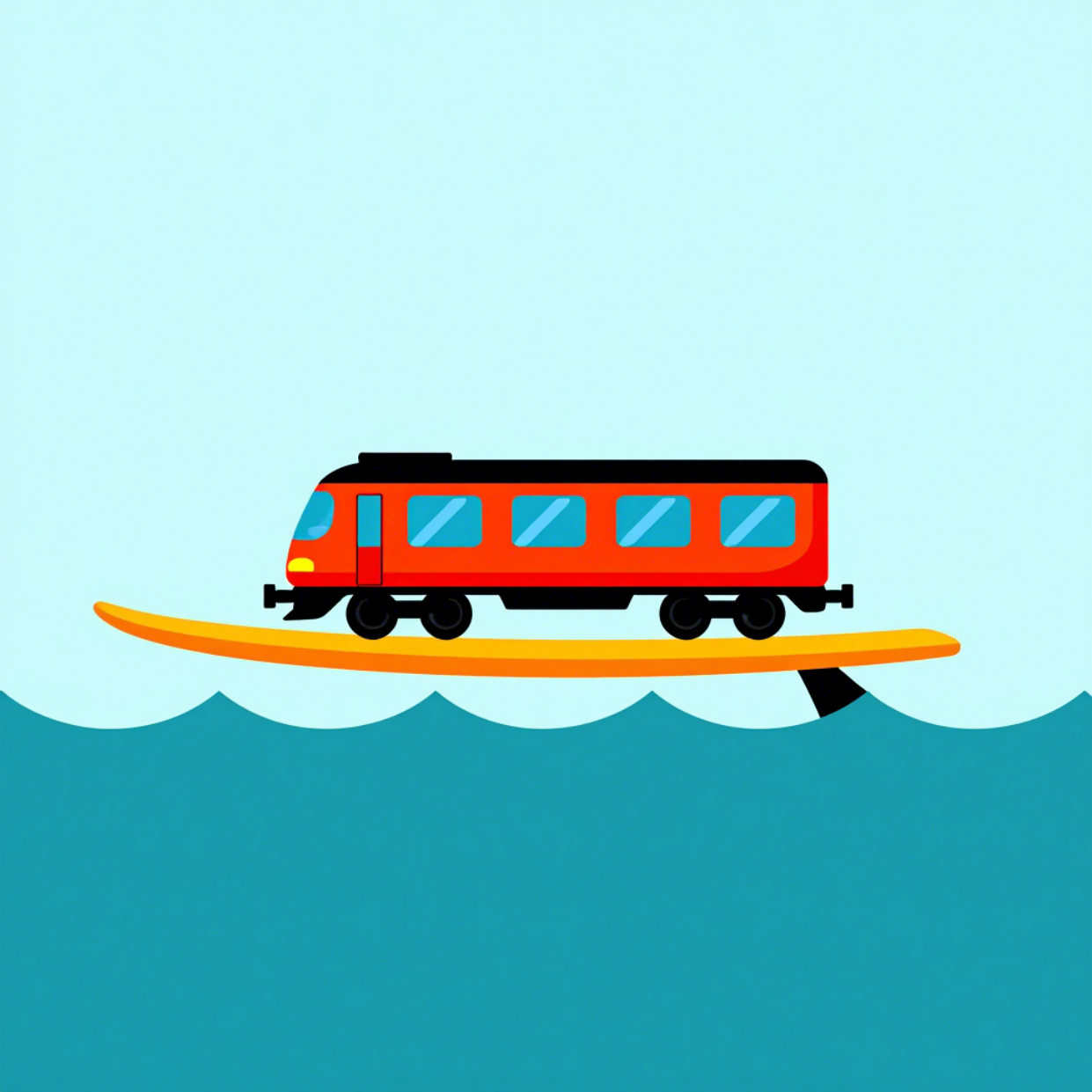}
\\[-0.25em]

\includegraphics[width=0.19\textwidth]{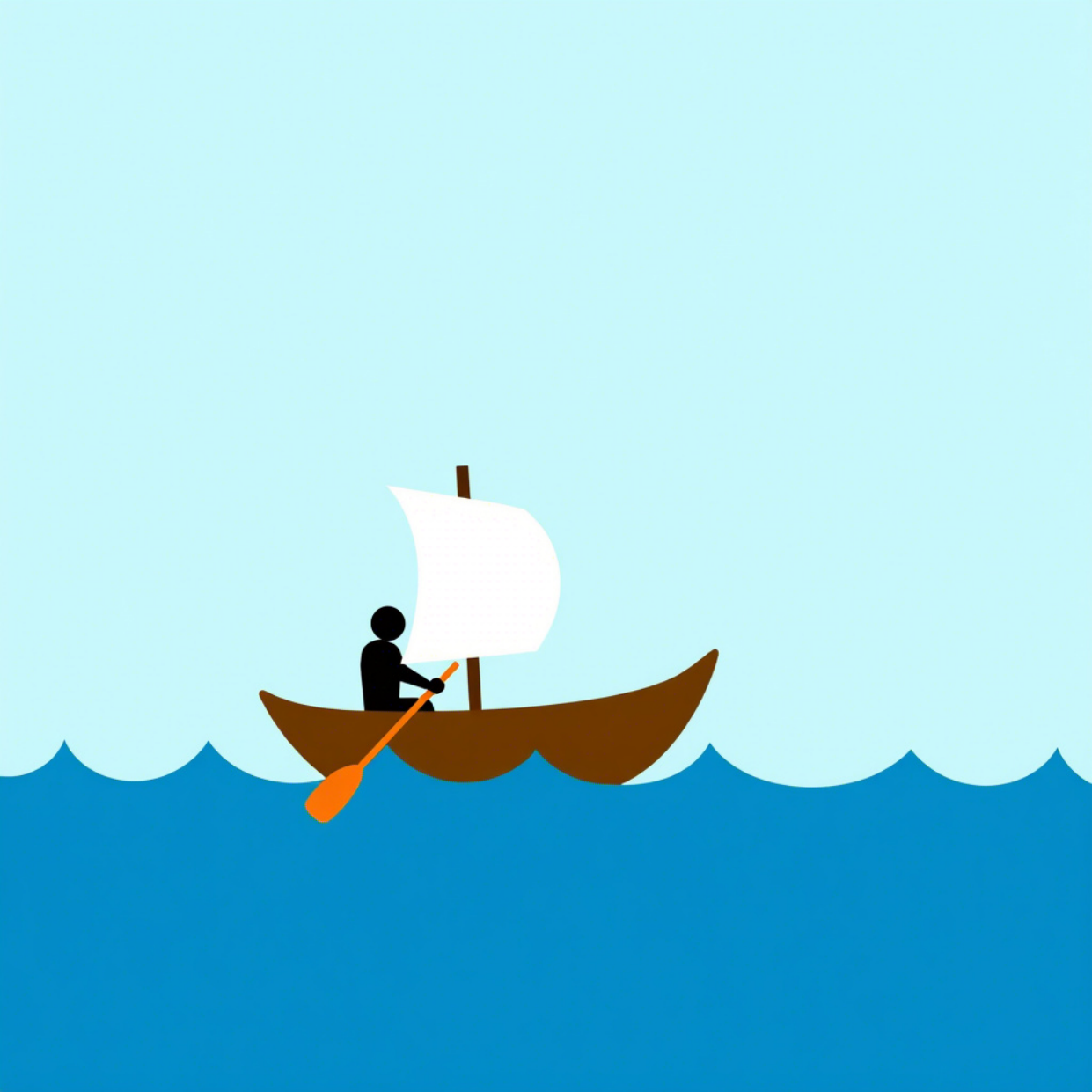}
& \includegraphics[width=0.19\textwidth]{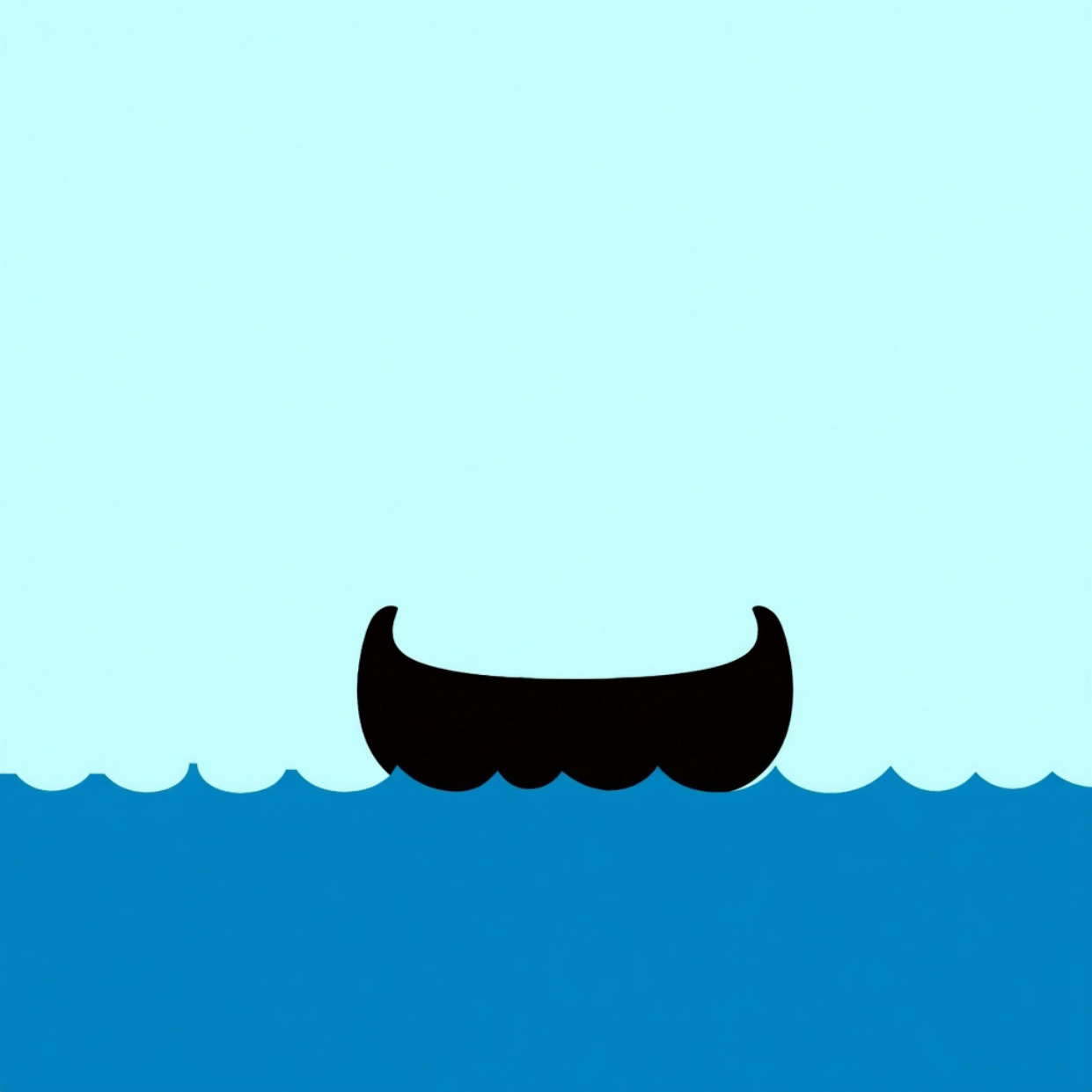}
& \includegraphics[width=0.19\textwidth]{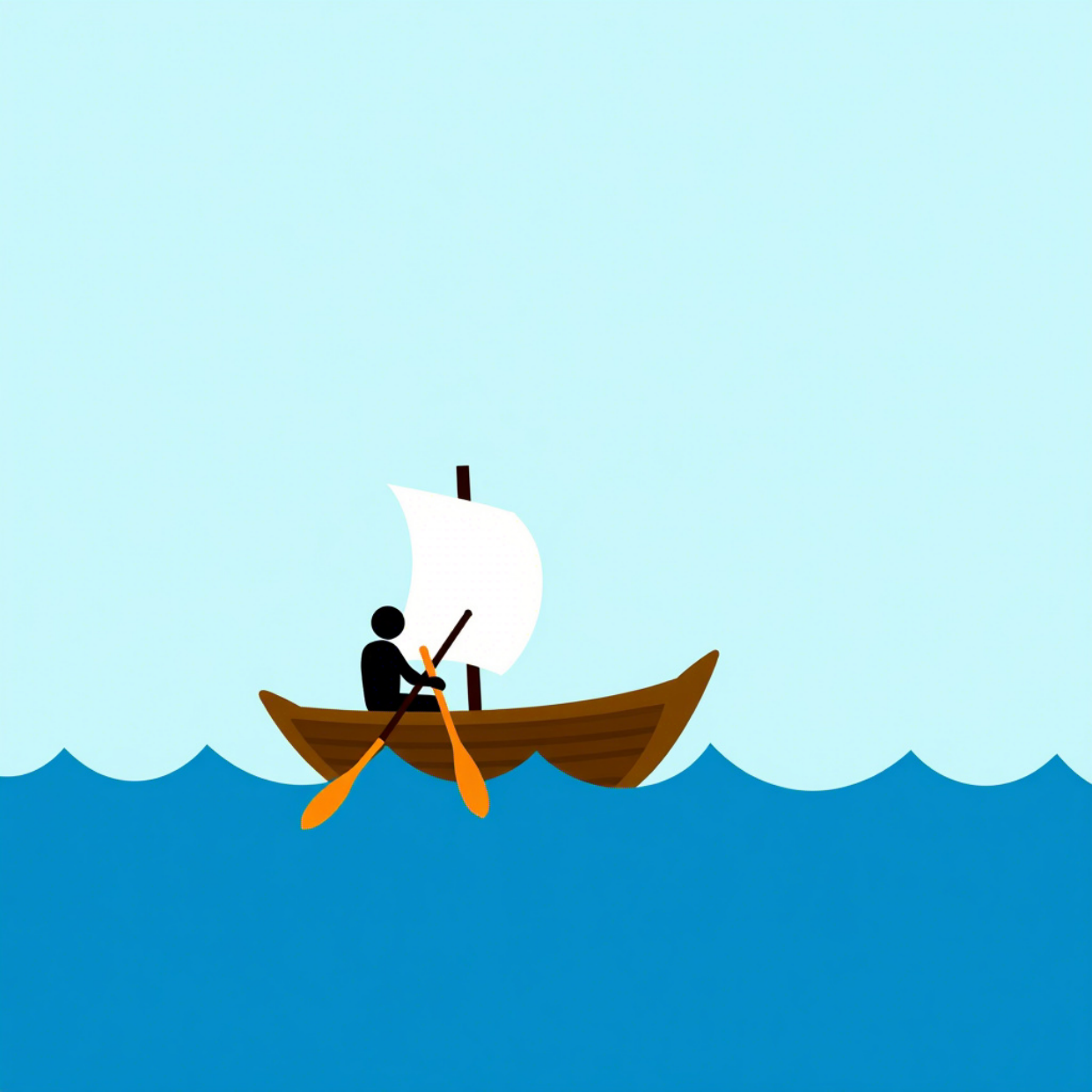}
& \includegraphics[width=0.19\textwidth]{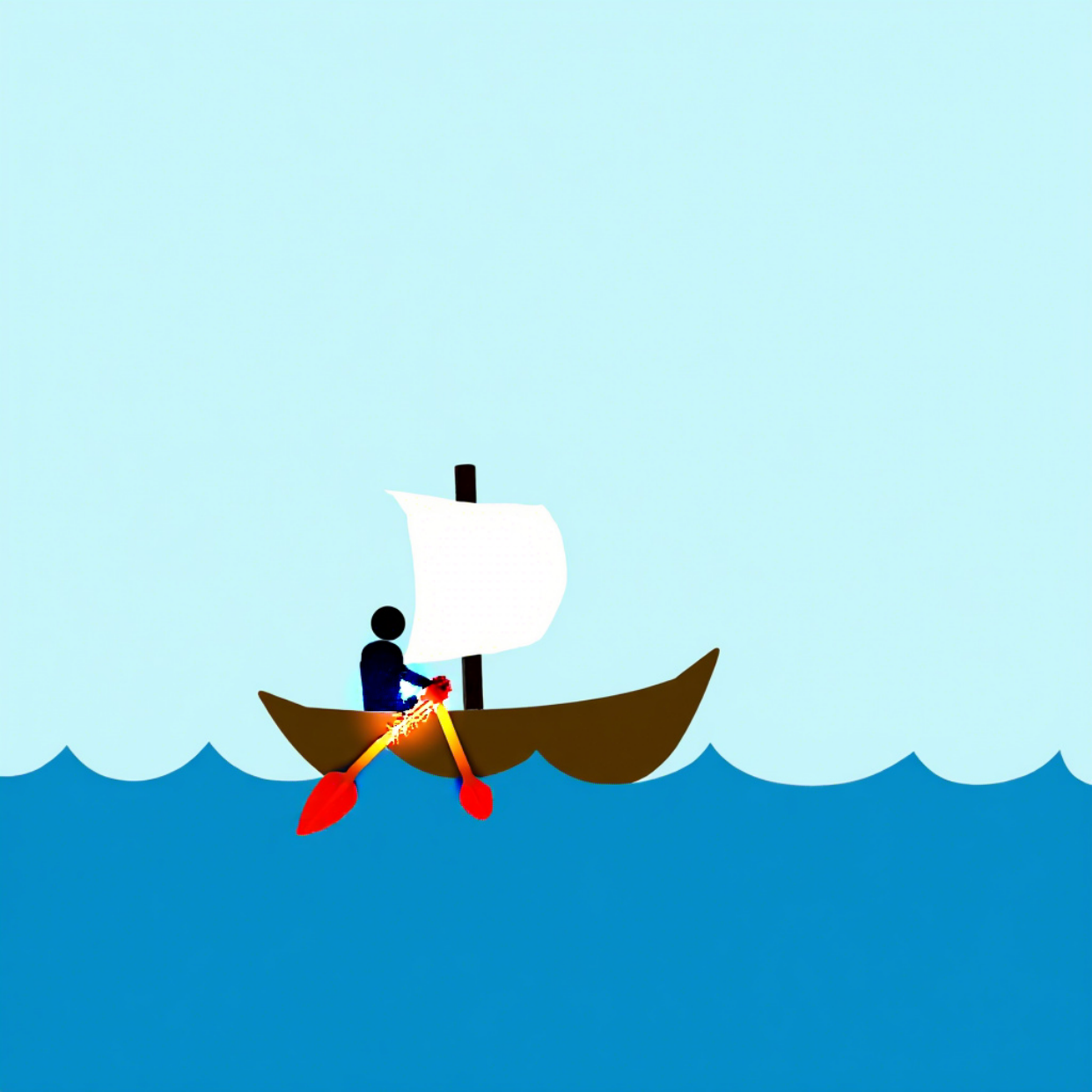}
& \includegraphics[width=0.19\textwidth]{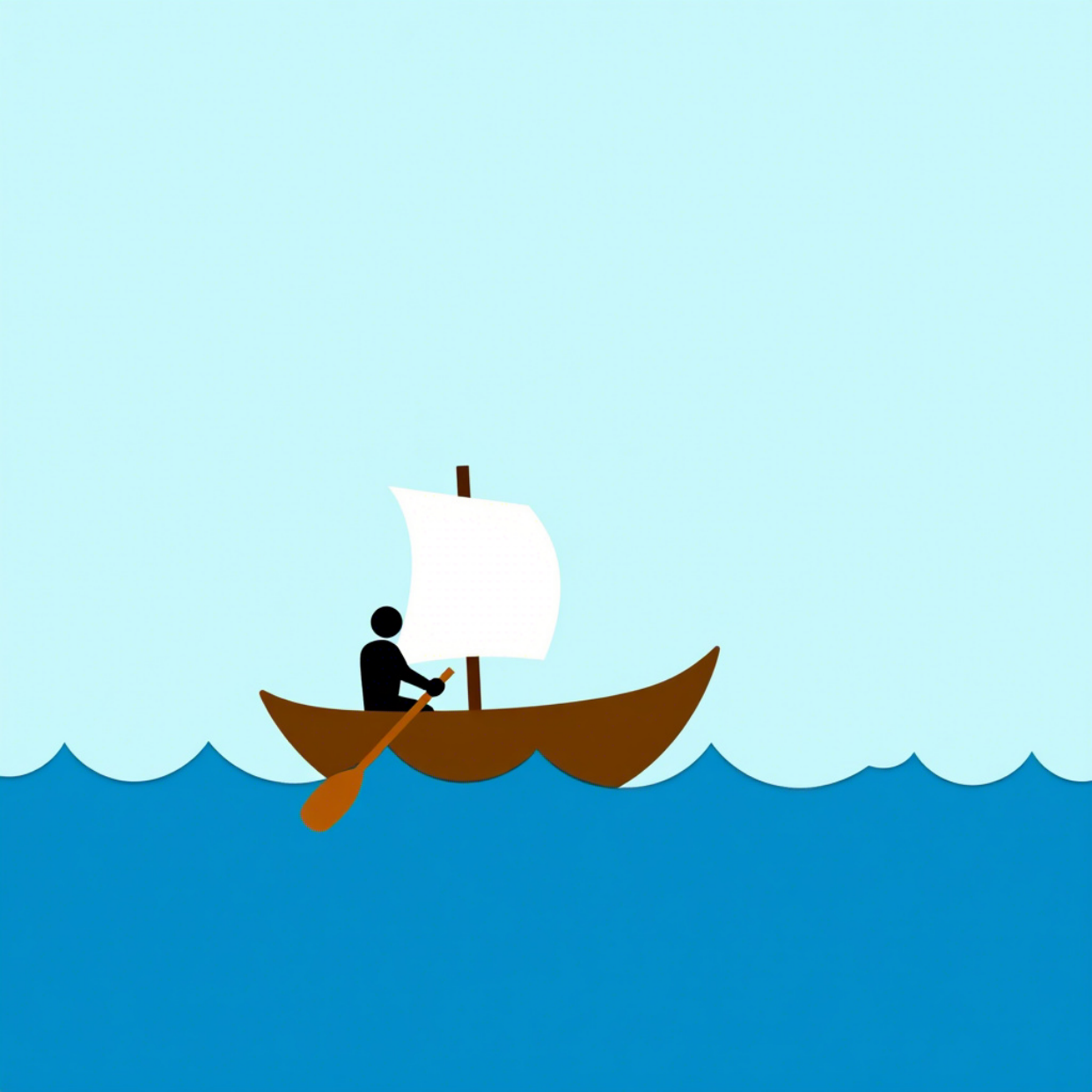}
\end{tabular}%
}
\caption{
\textbf{Qualitative comparison on FLUX.1-dev.}
}
\end{figure}
% figure-local macros
\ifdefined\cogw\else\newlength{\cogw}\fi
\ifdefined\cogh\else\newlength{\cogh}\fi
\setlength{\cogw}{0.302\textwidth}
\setlength{\cogh}{0.170\textwidth}

\providecommand{\vcell}[1]{\ensuremath{\vcenter{\hbox{#1}}}}
\providecommand{\methodcell}[1]{%
    \vcell{\rotatebox[origin=c]{90}{\makebox[\cogh][c]{\textbf{#1}}}}%
}
\providecommand{\framecell}[1]{%
    \vcell{\includegraphics[width=\cogw,height=\cogh]{#1}}%
}
\providecommand{\rowgap}{\noalign{\vskip 3pt}}  % 控制行间距，只改这里

\begin{figure*}[t]
    \centering
    \setlength{\tabcolsep}{1.2pt}
    \renewcommand{\arraystretch}{0.0}
    \scriptsize

    \begin{tabular}{@{}c@{\hspace{1.5pt}}ccc@{}}
        & \multicolumn{3}{c}{\large \textbf{A polar bear is playing guitar}}
        \\[0.25em]

        \methodcell{Original}
        & \framecell{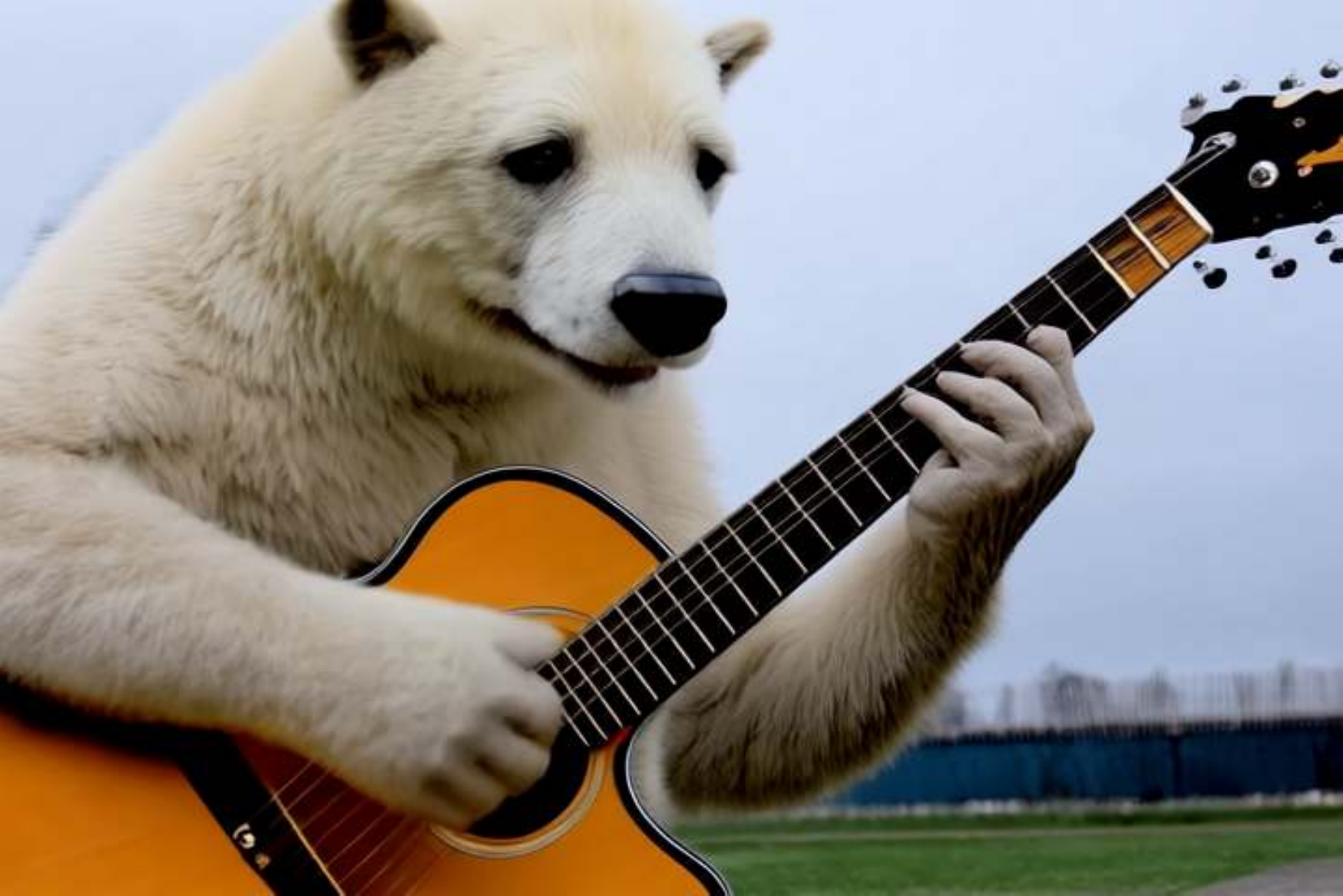}
        & \framecell{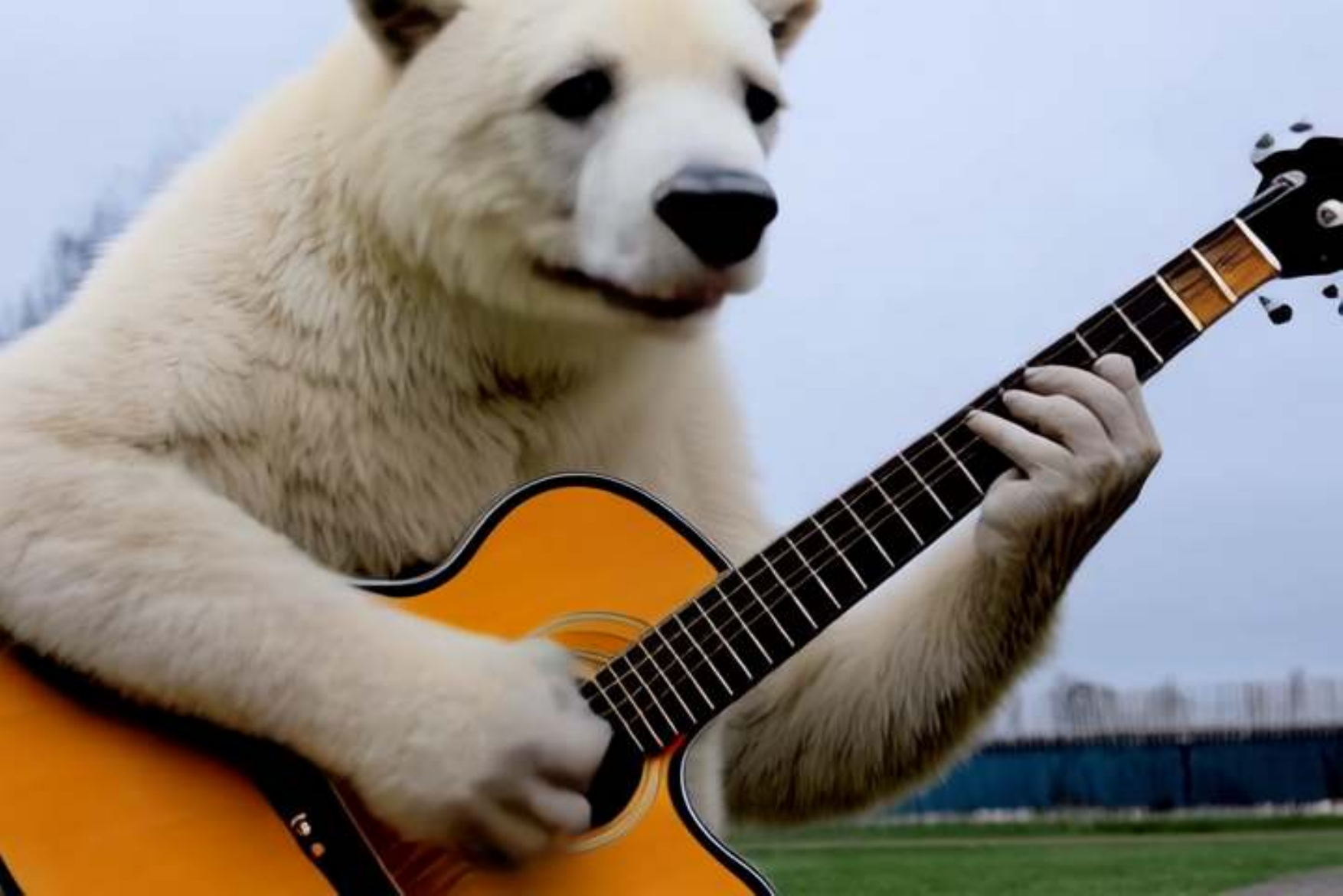}
        & \framecell{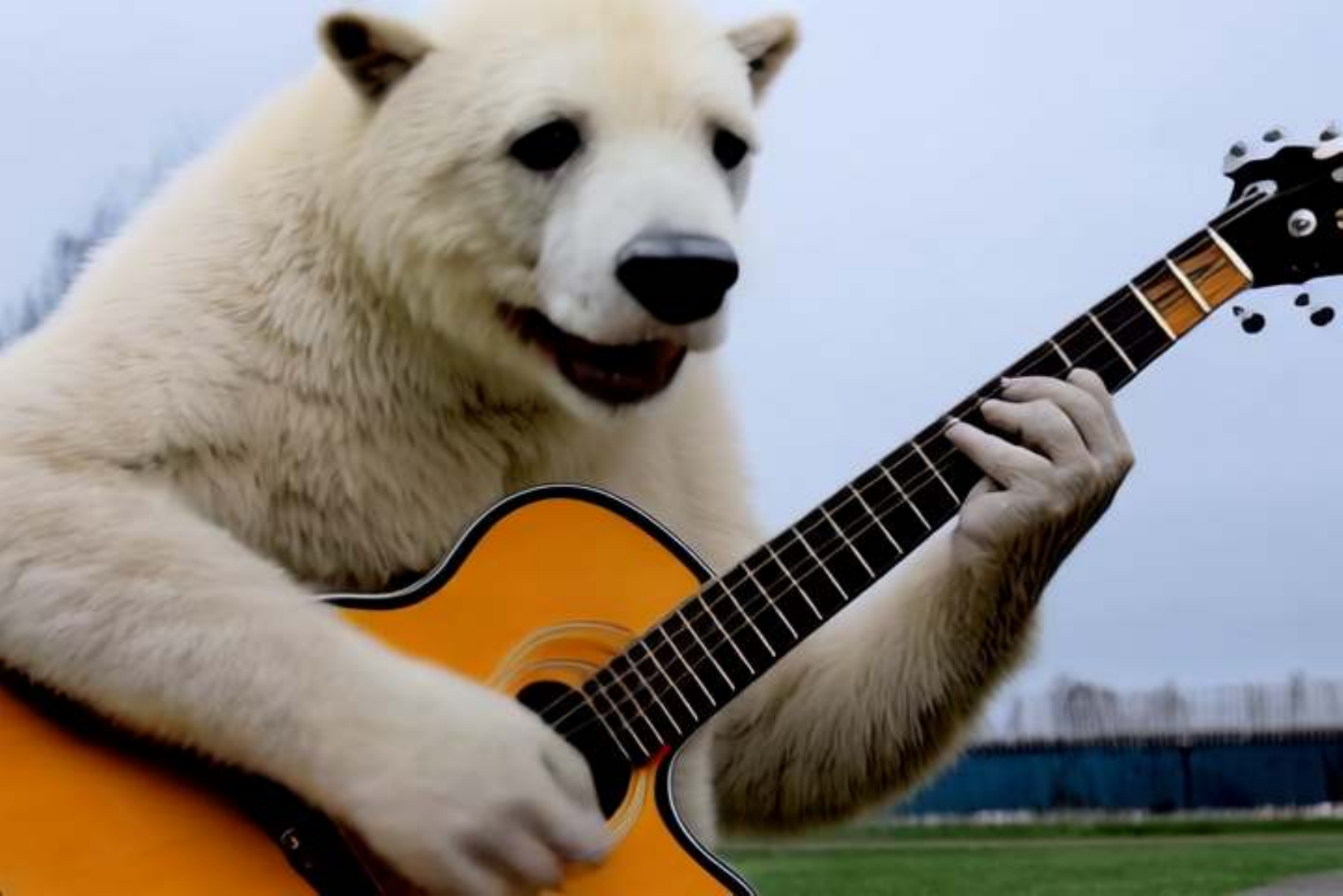}
        \\ \rowgap

        \methodcell{MagCache ($2.32\times$)}
        & \framecell{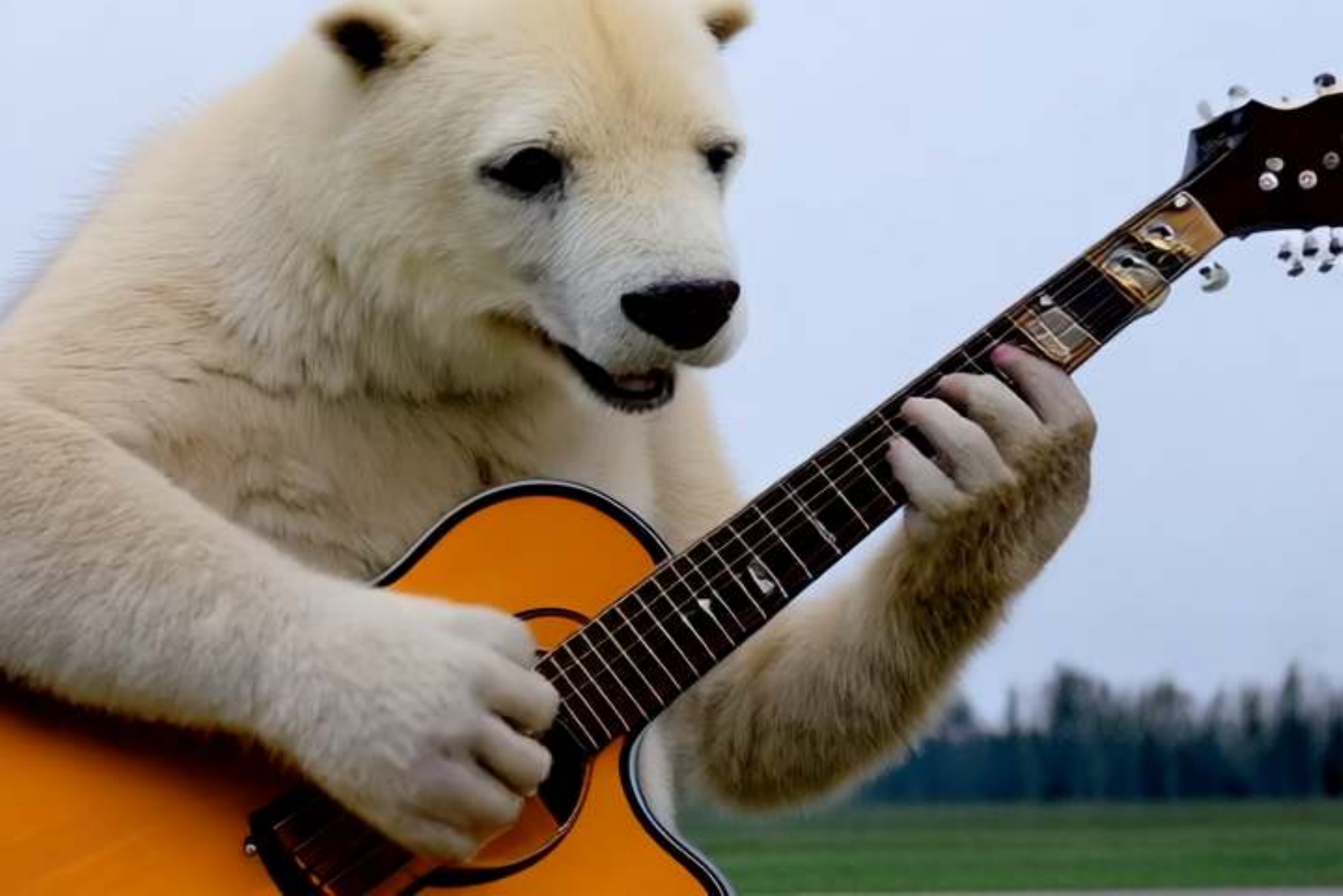}
        & \framecell{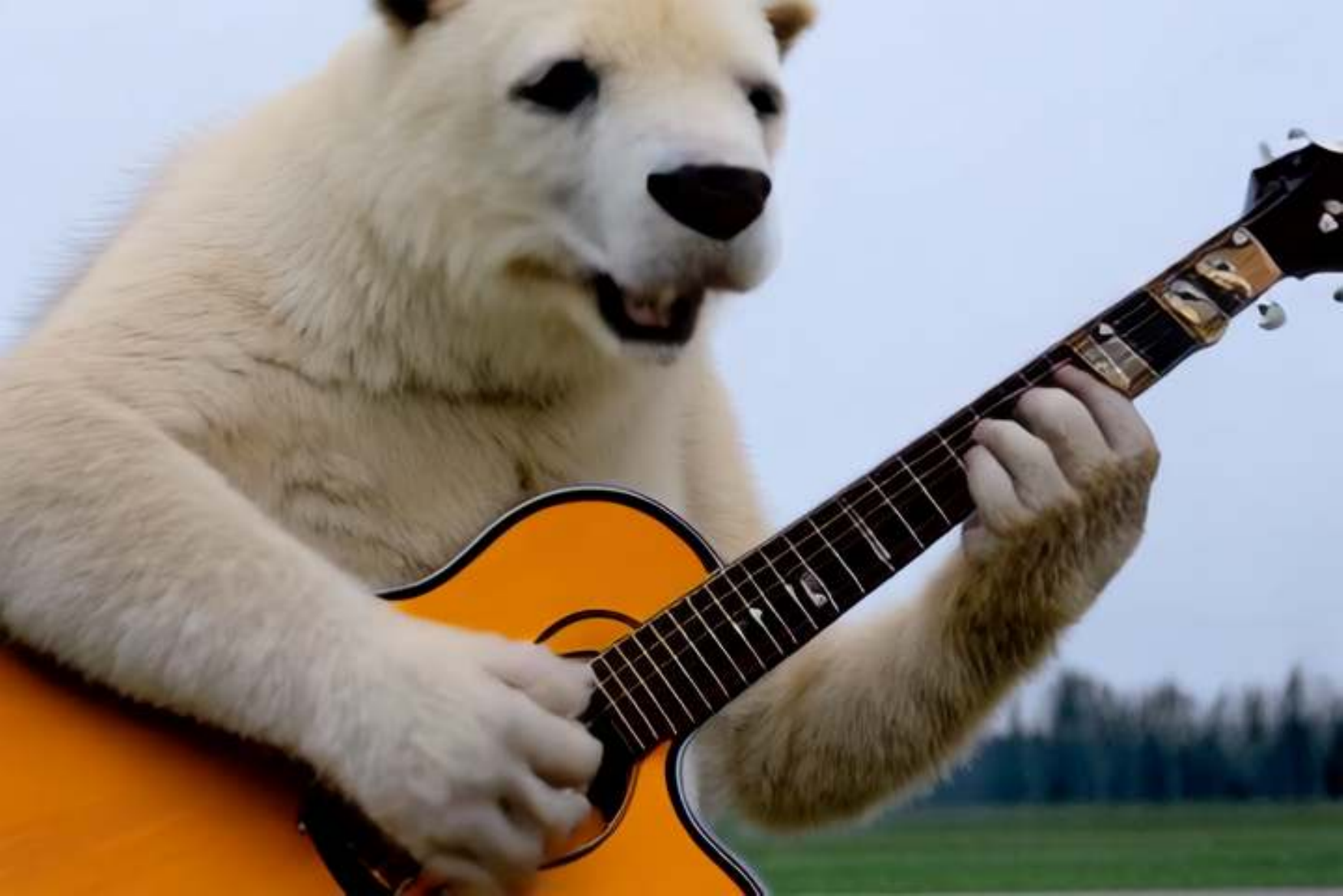}
        & \framecell{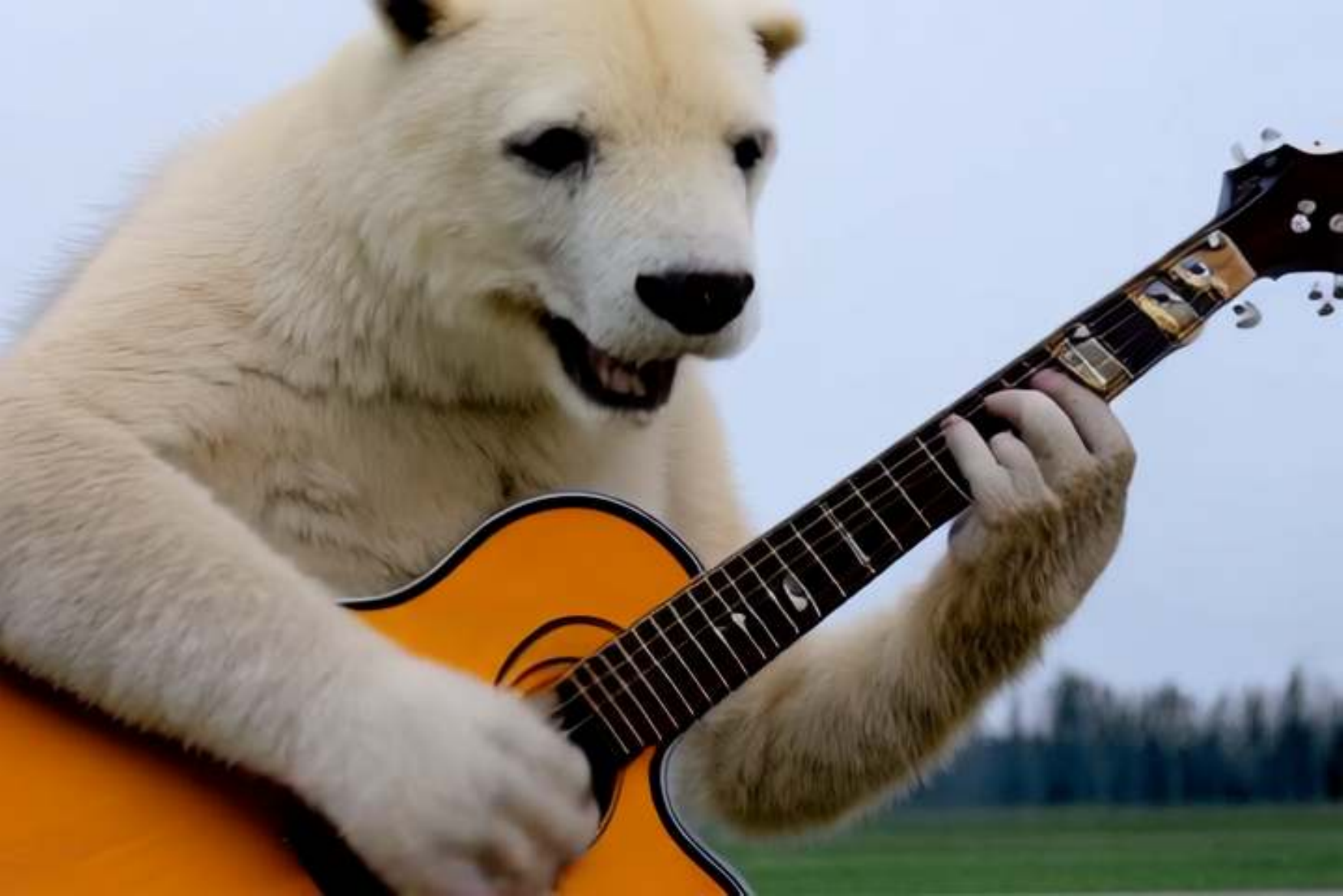}
        \\ \rowgap

        \methodcell{DiCache ($2.42\times$)}
        & \framecell{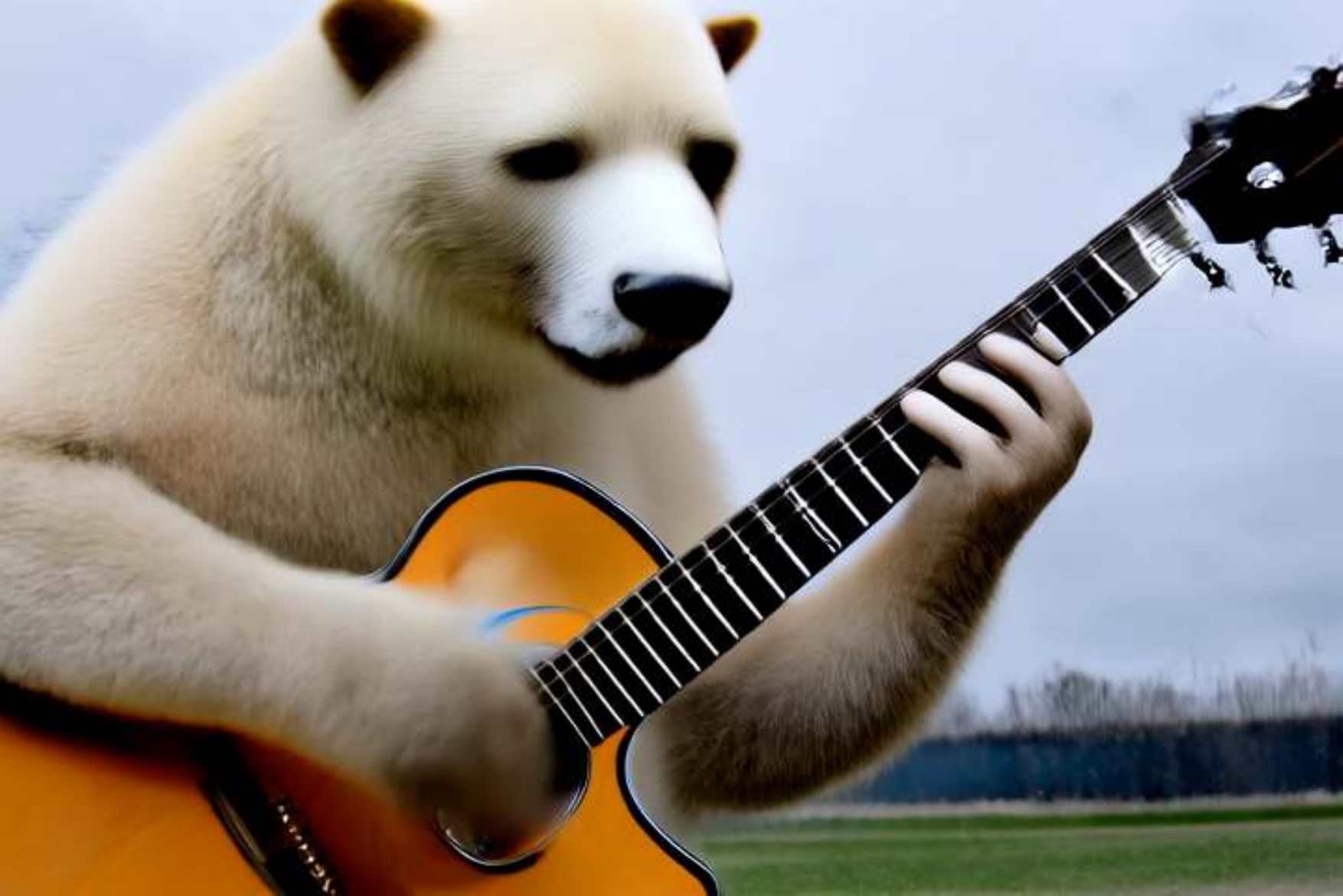}
        & \framecell{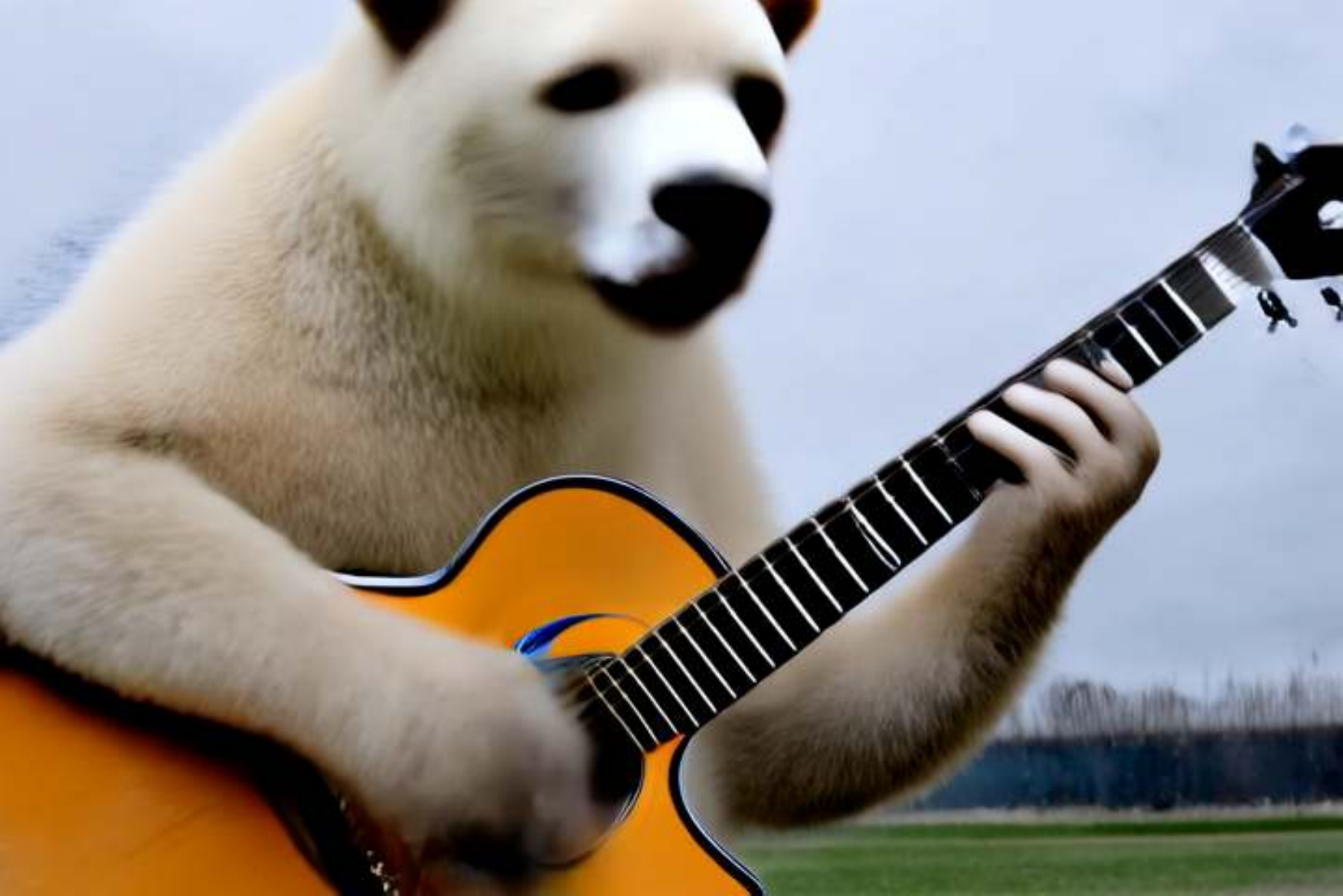}
        & \framecell{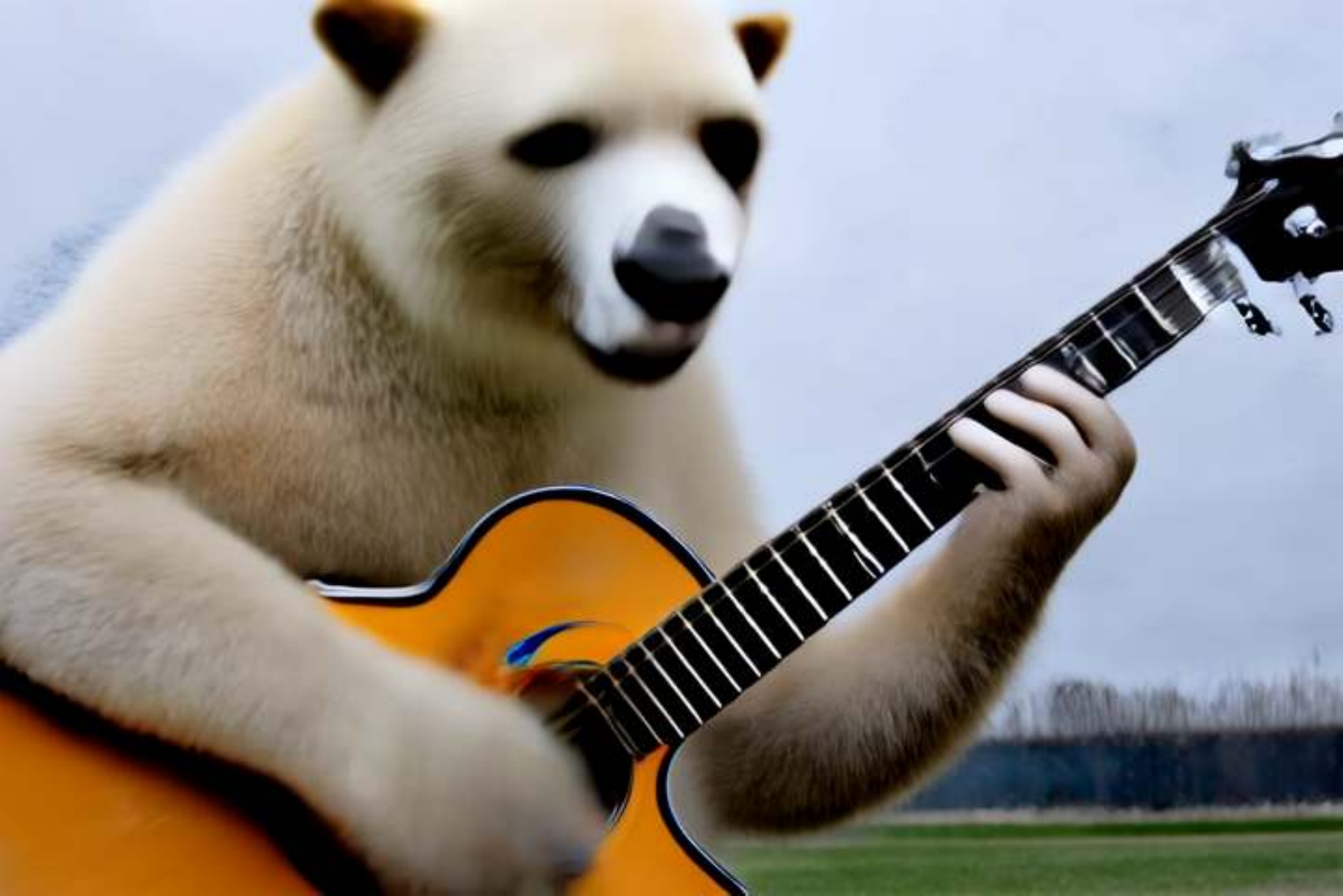}
        \\ \rowgap

        \methodcell{Ours ($2.38\times$)}
        & \framecell{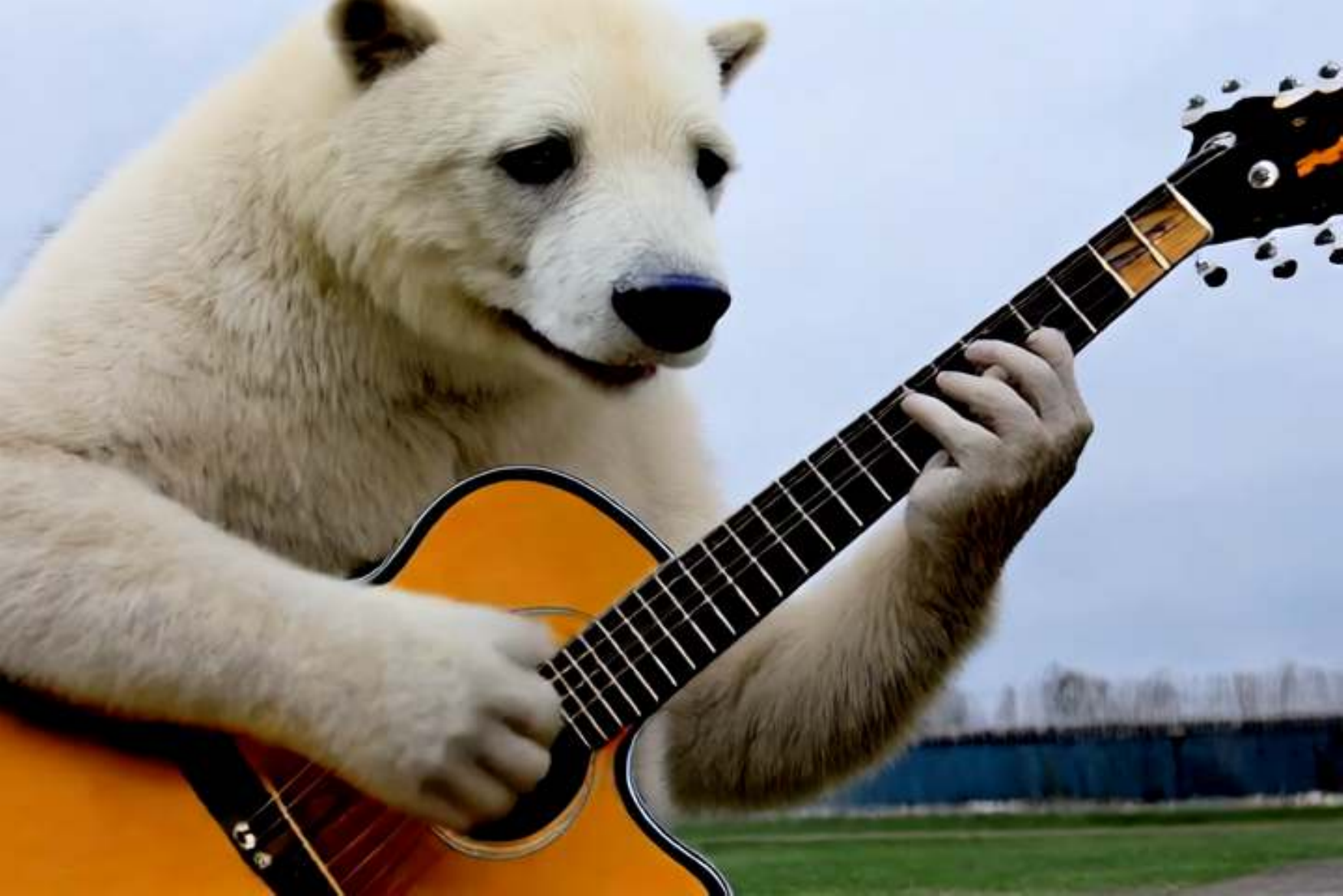}
        & \framecell{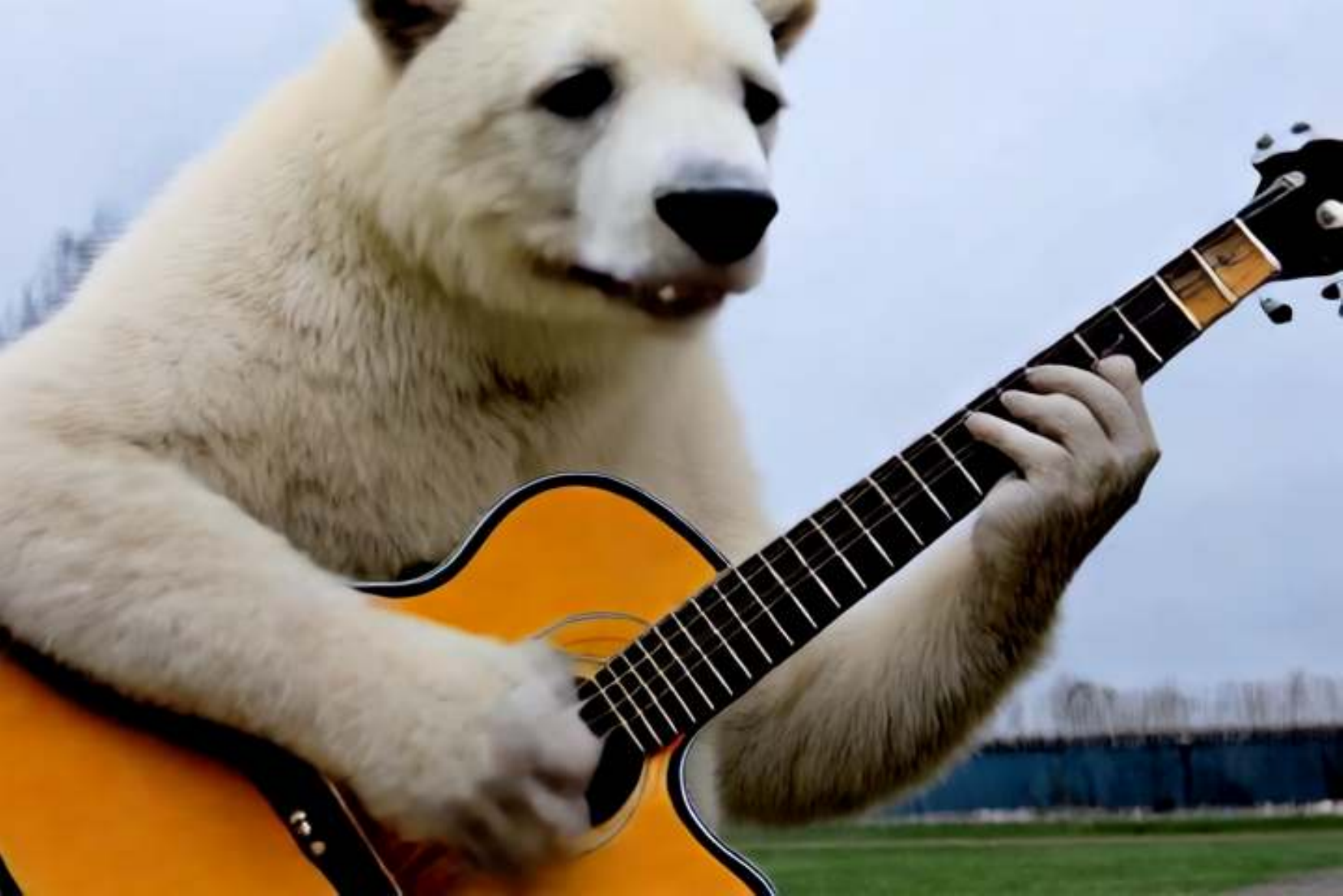}
        & \framecell{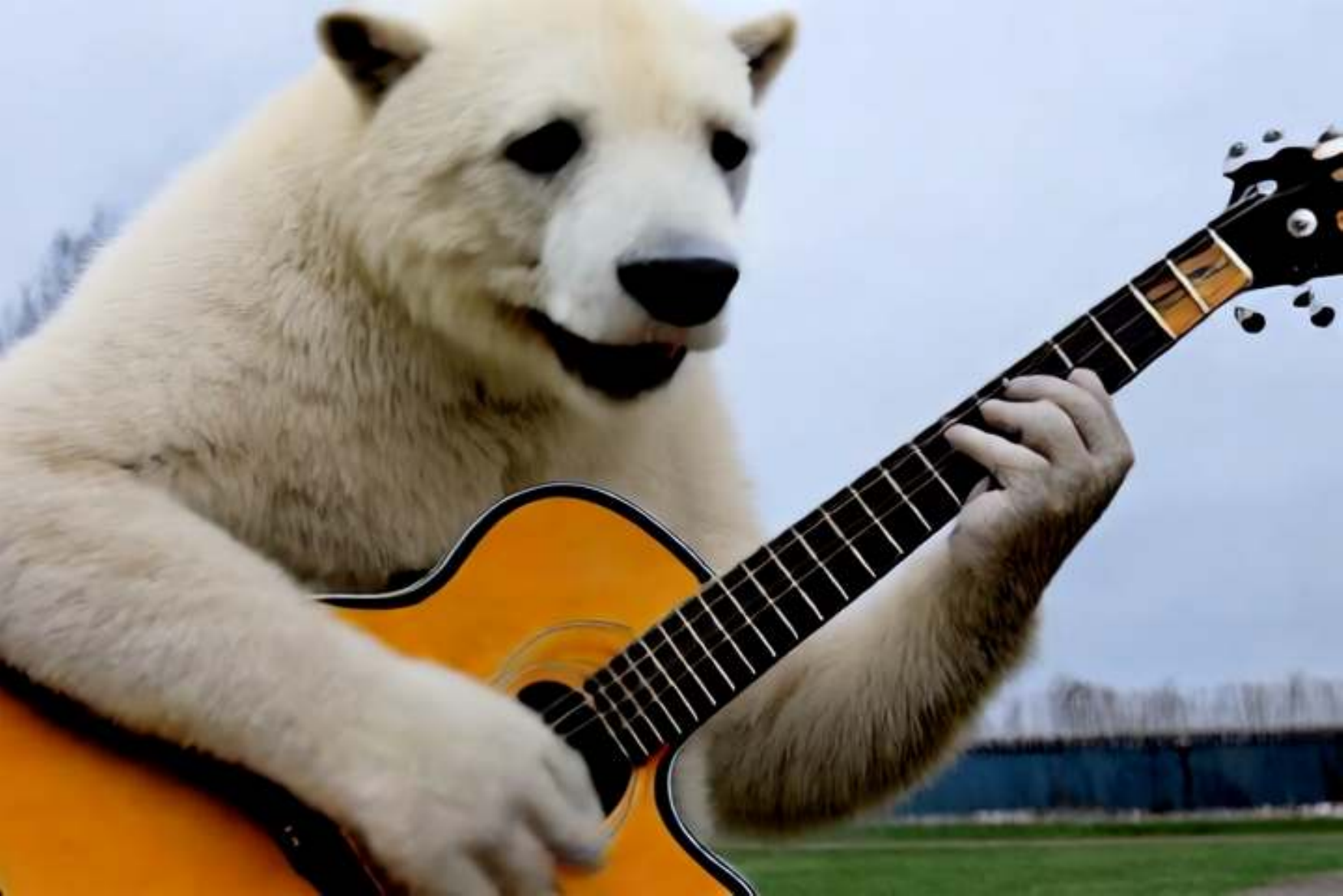}
        \\
    \end{tabular}
    \par\vspace{0.8em}
    \begin{tabular}{@{}c@{\hspace{1.5pt}}ccc@{}}
        & \multicolumn{3}{c}{\large \textbf{A bicycle leaning against a tree}}
        \\[0.25em]

        \methodcell{Original}
        & \framecell{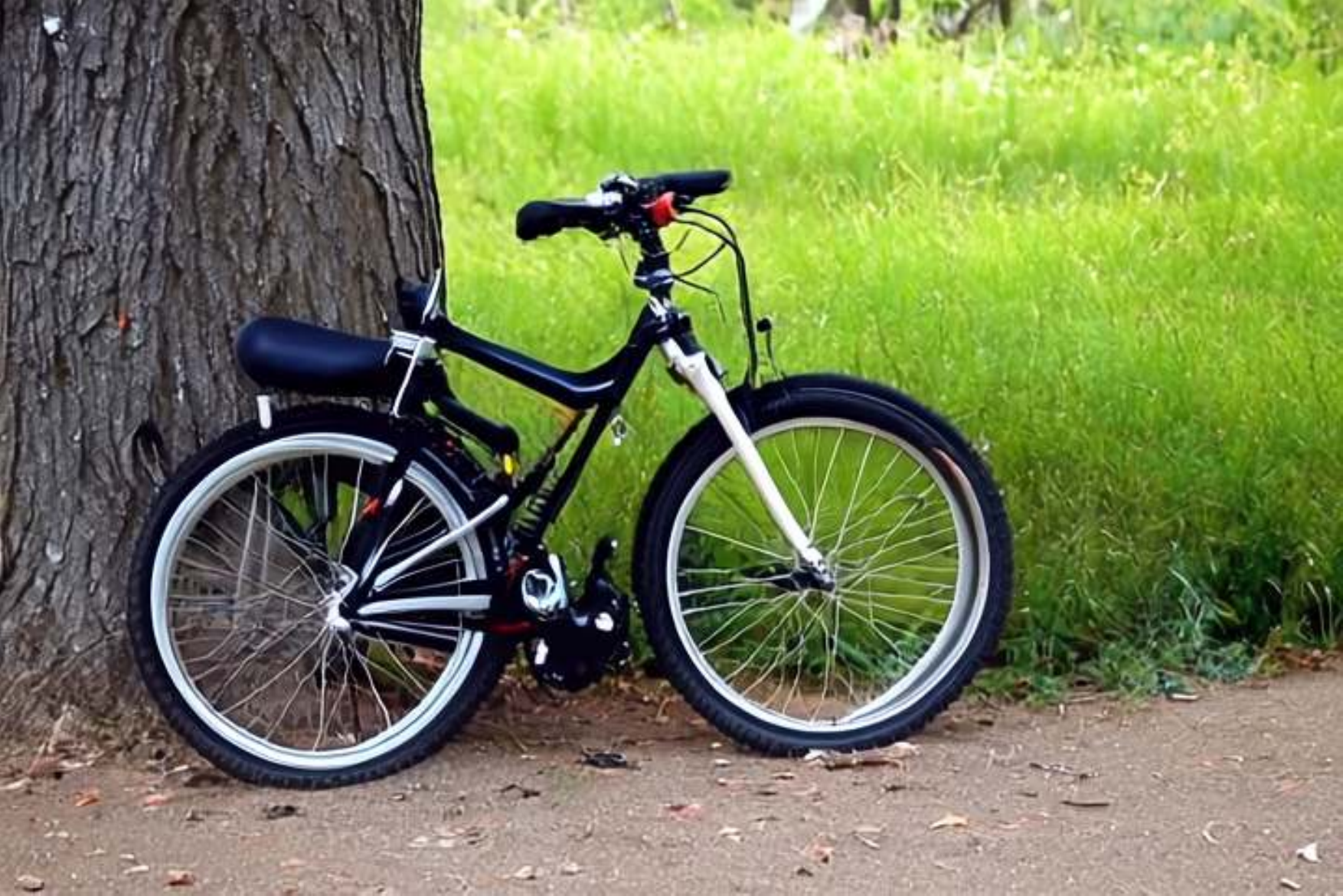}
        & \framecell{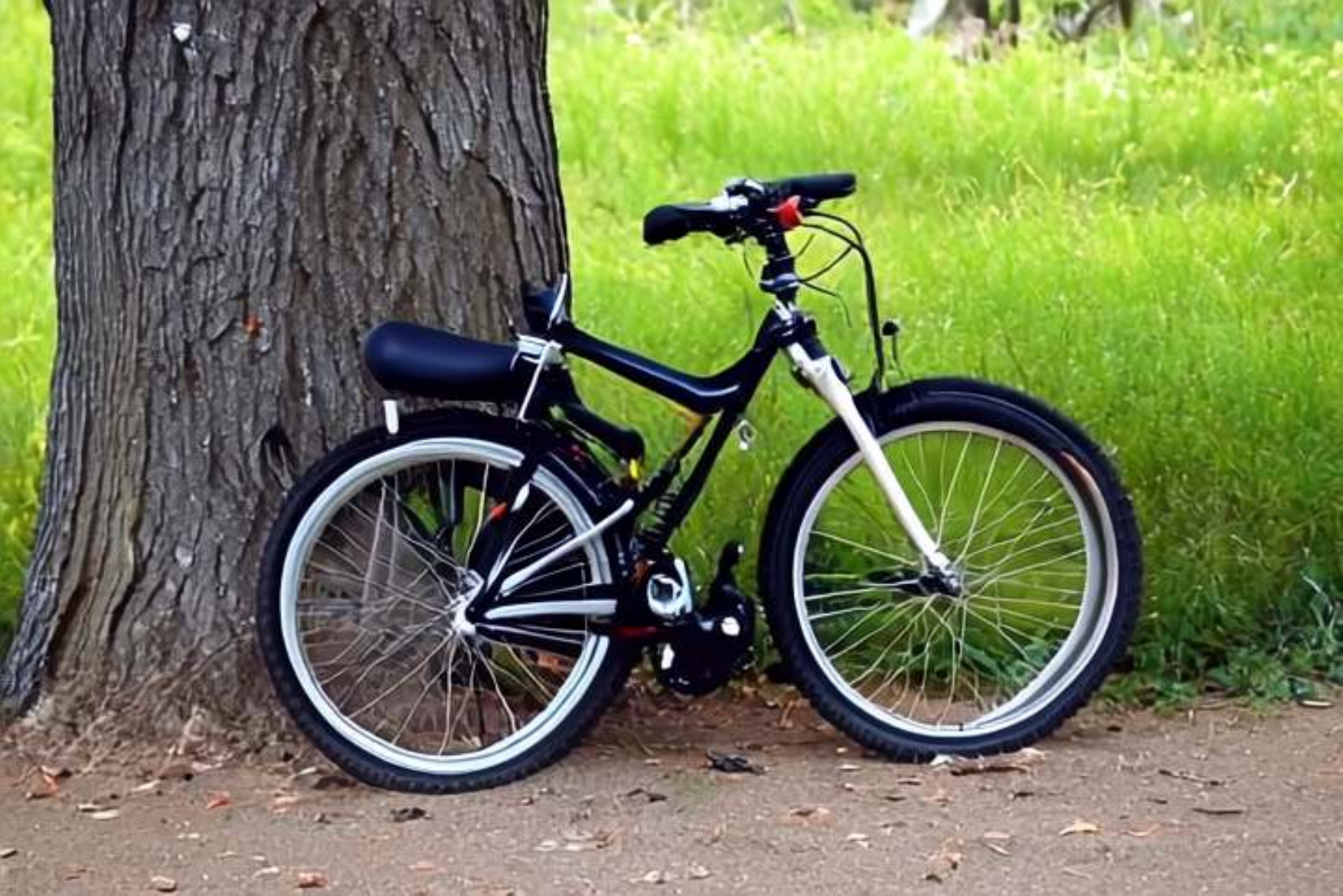}
        & \framecell{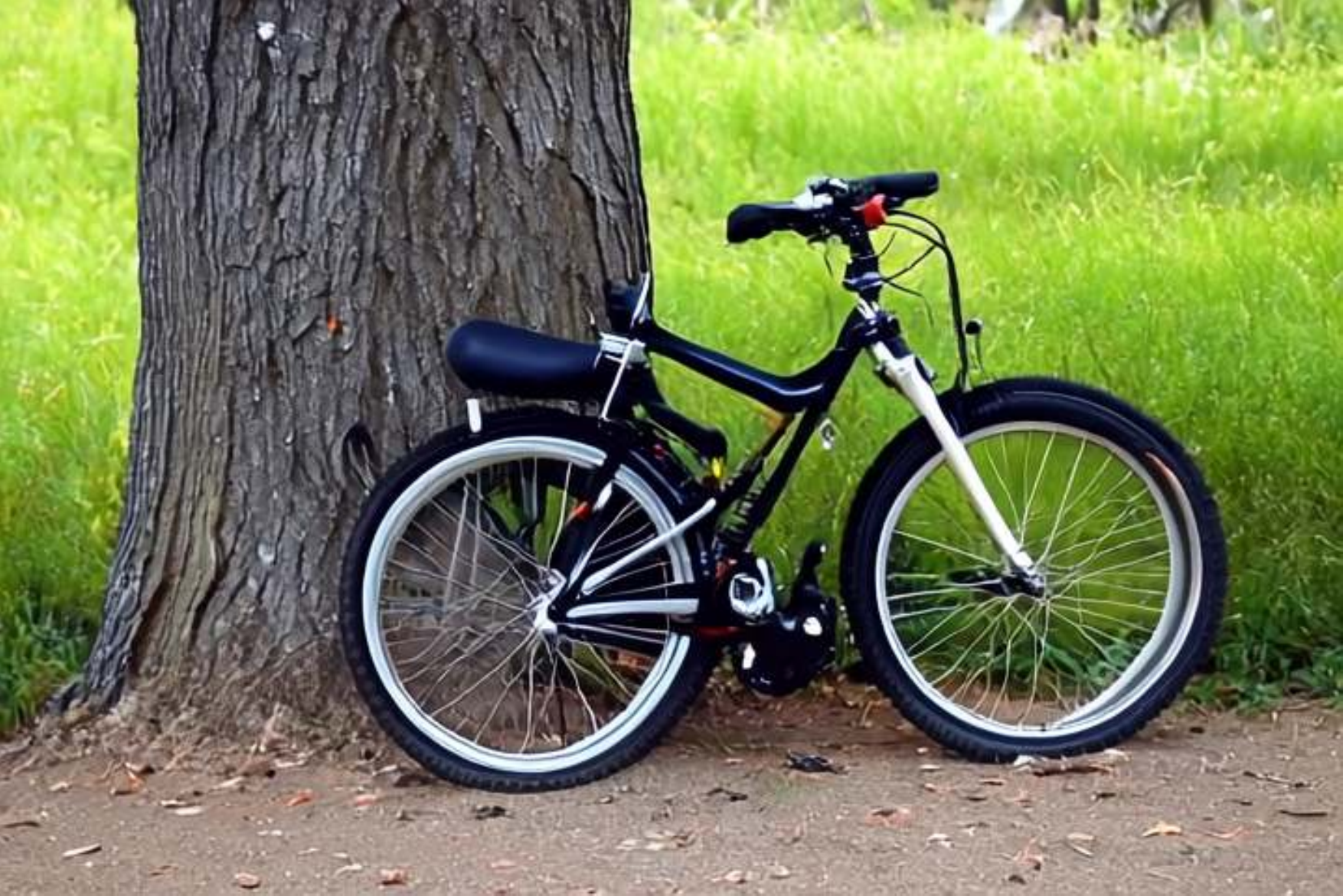}
        \\ \rowgap

        \methodcell{MagCache ($2.32\times$)}
        & \framecell{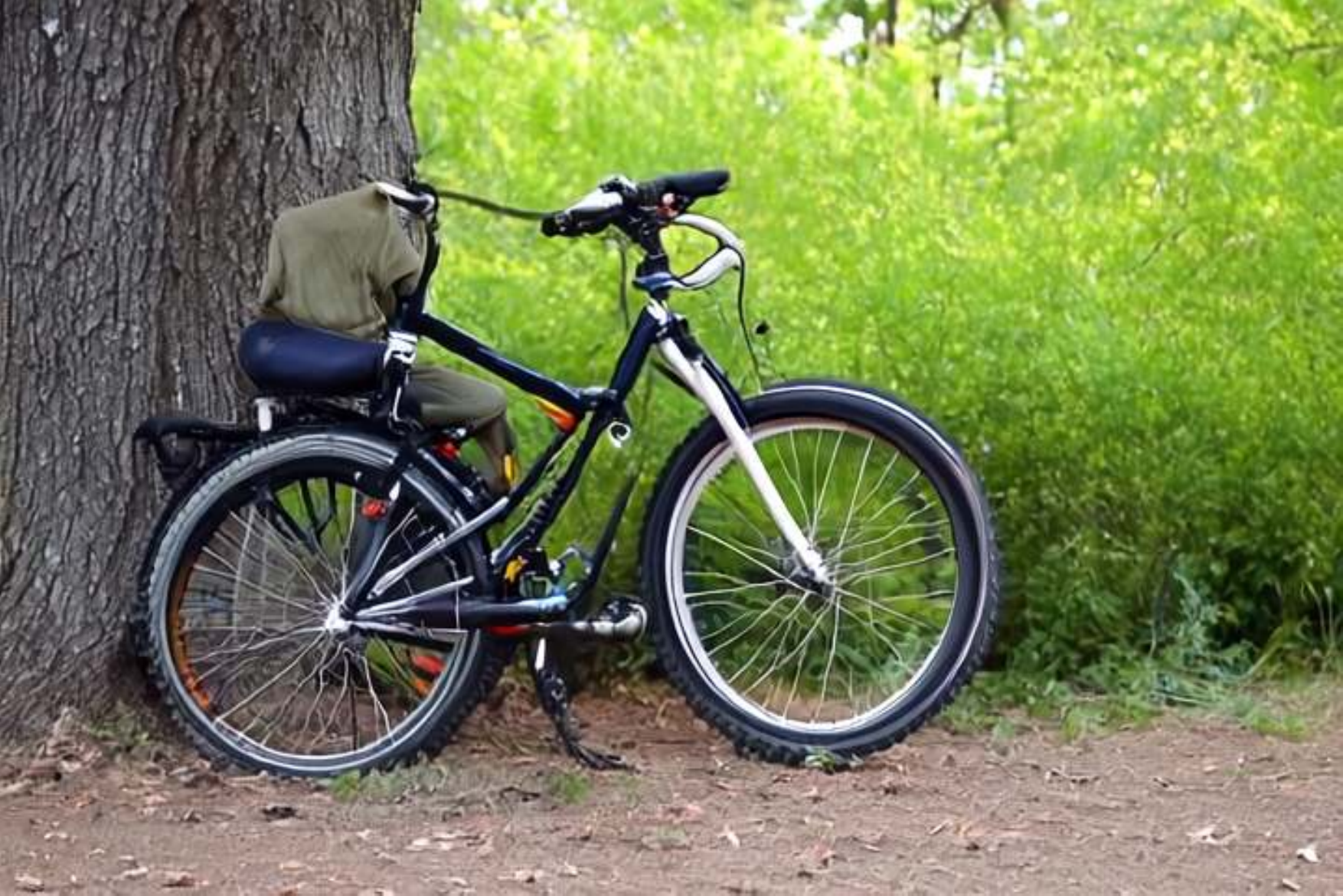}
        & \framecell{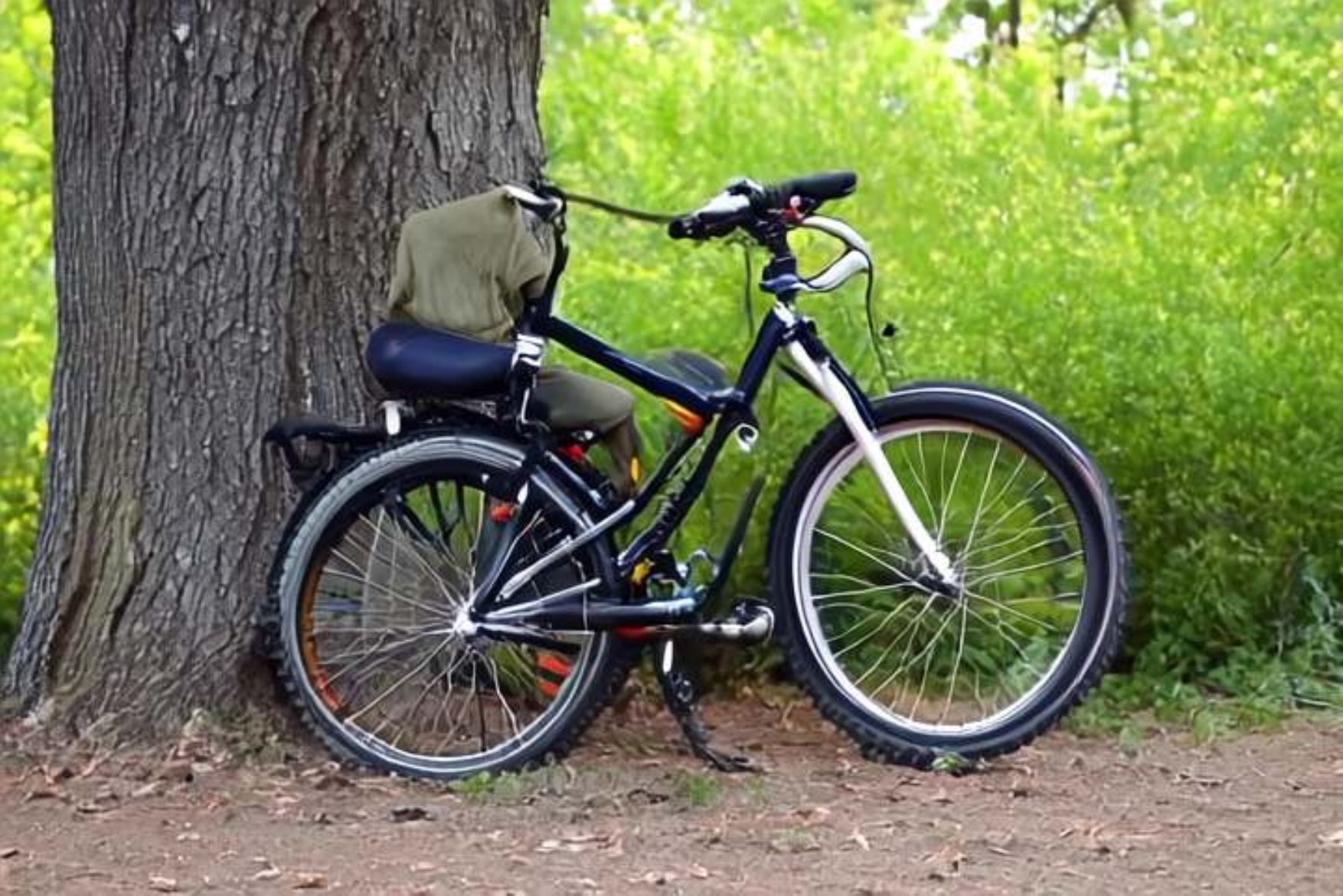}
        & \framecell{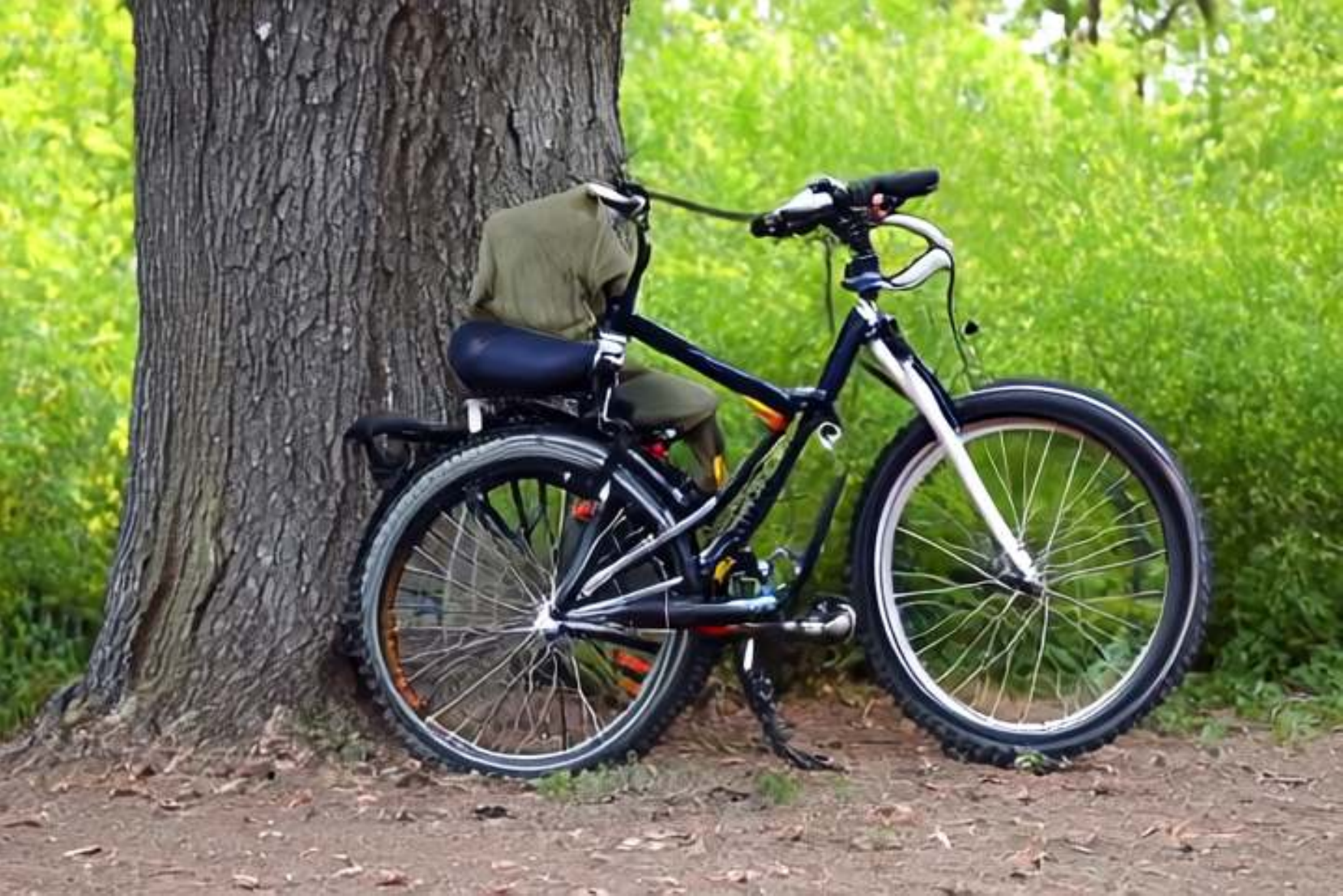}
        \\ \rowgap

        \methodcell{DiCache ($2.42\times$)}
        & \framecell{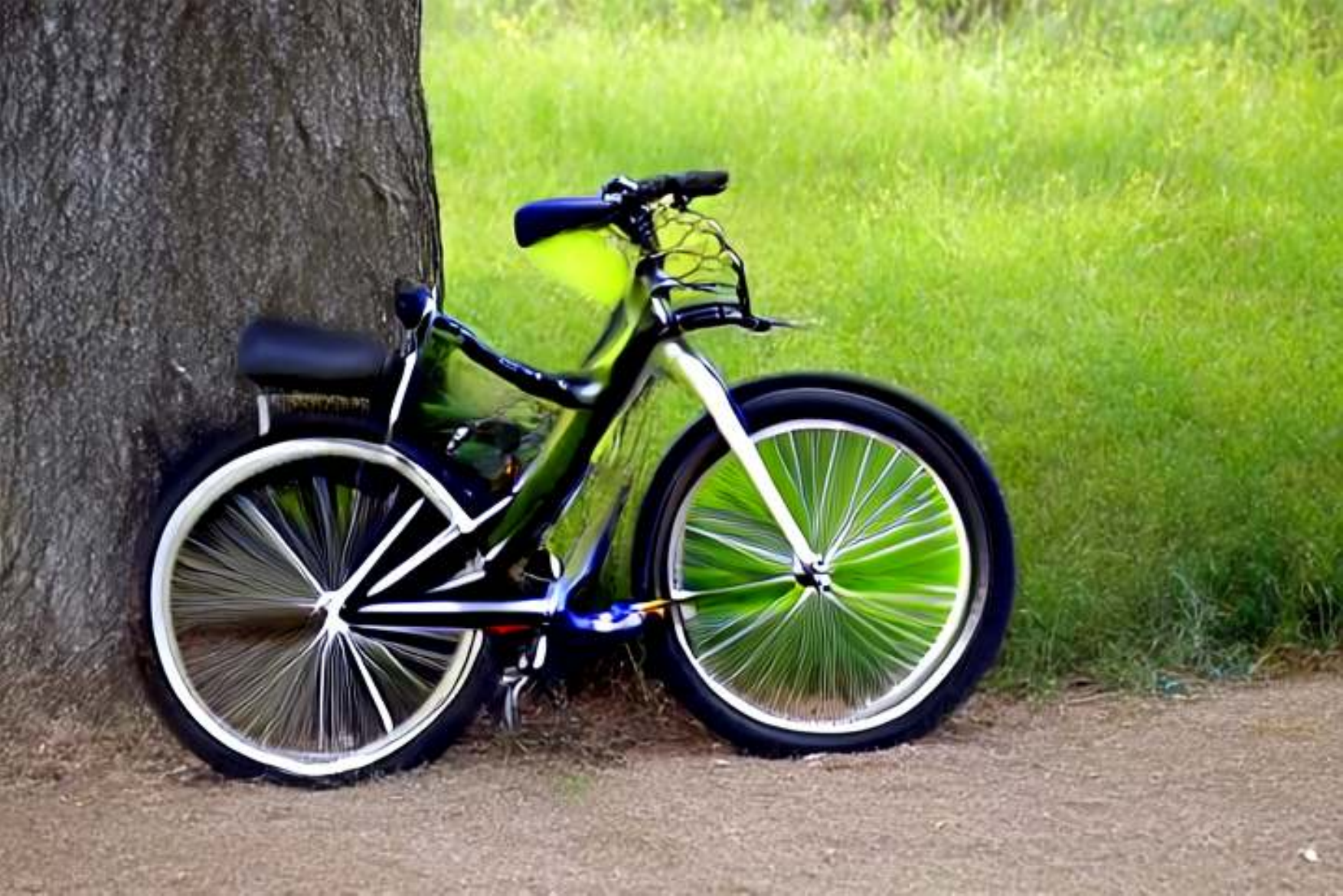}
        & \framecell{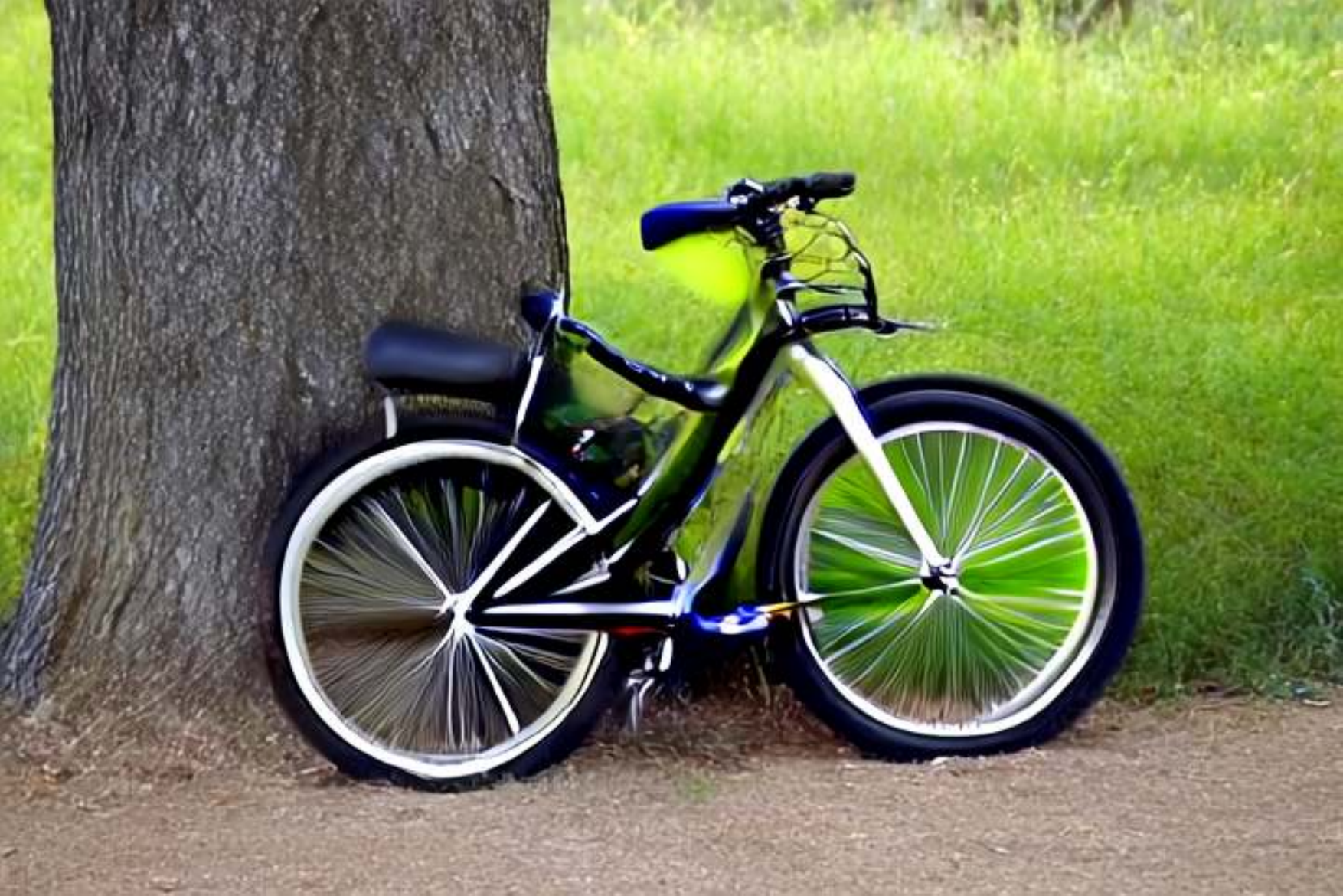}
        & \framecell{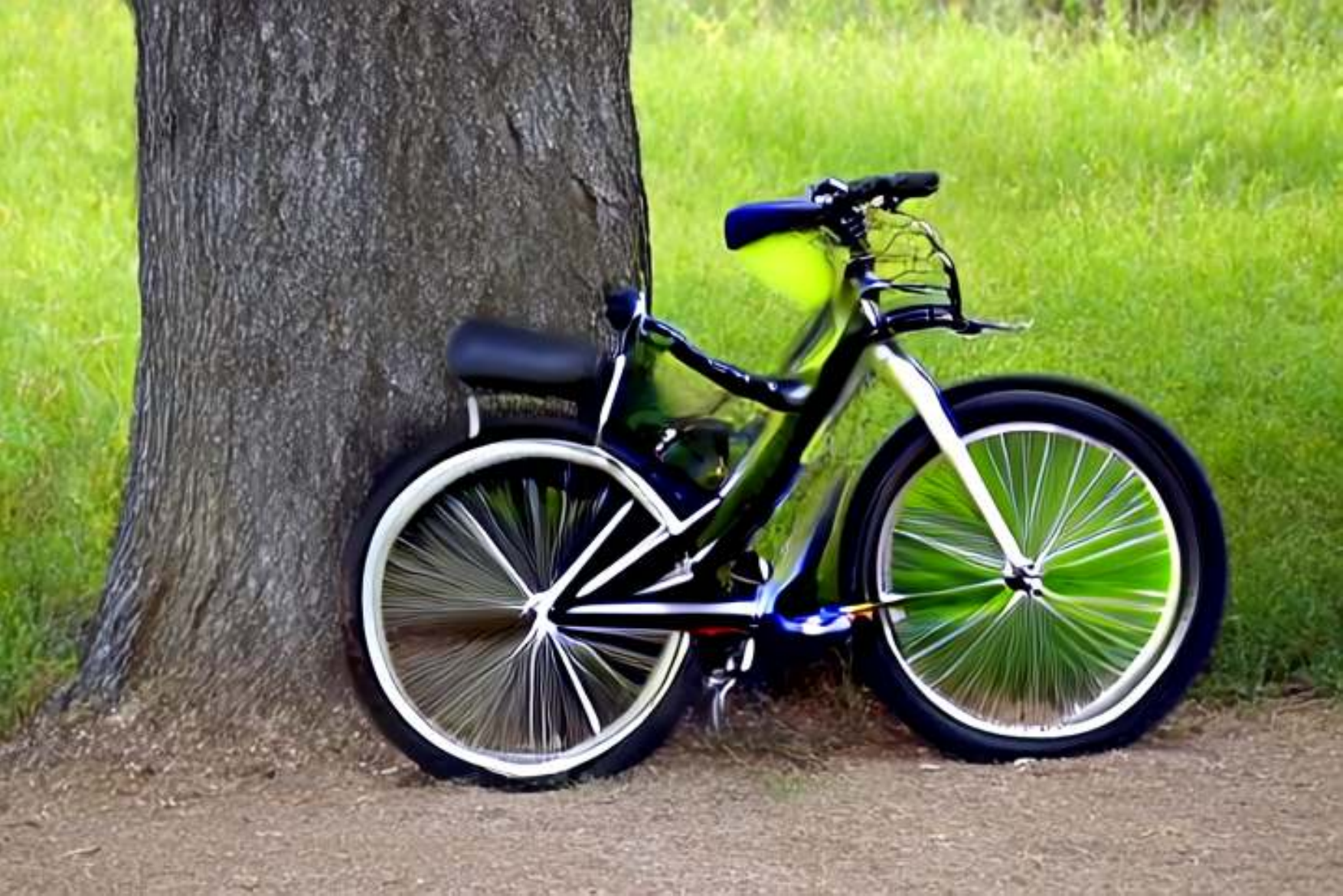}
        \\ \rowgap

        \methodcell{Ours ($2.38\times$)}
        & \framecell{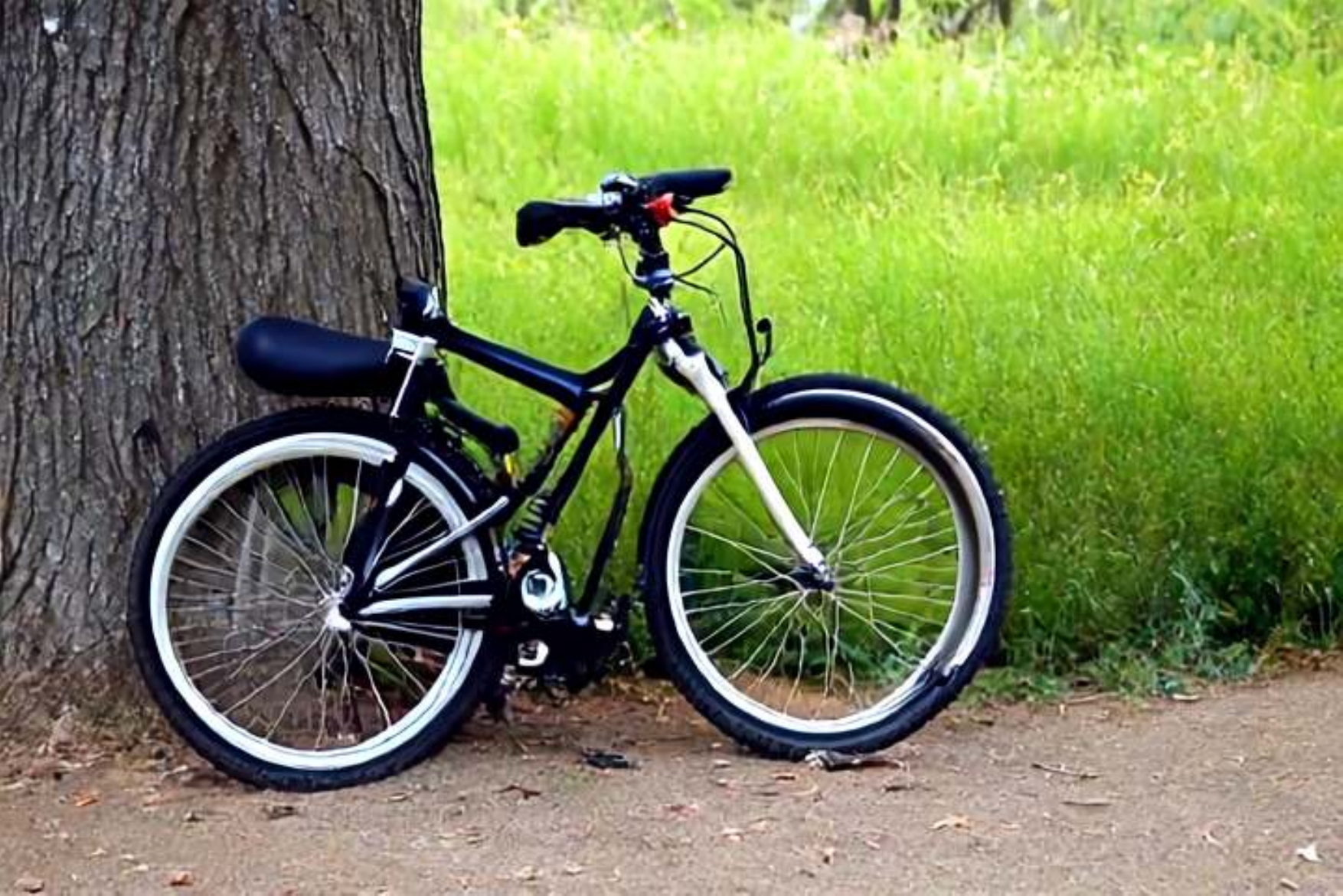}
        & \framecell{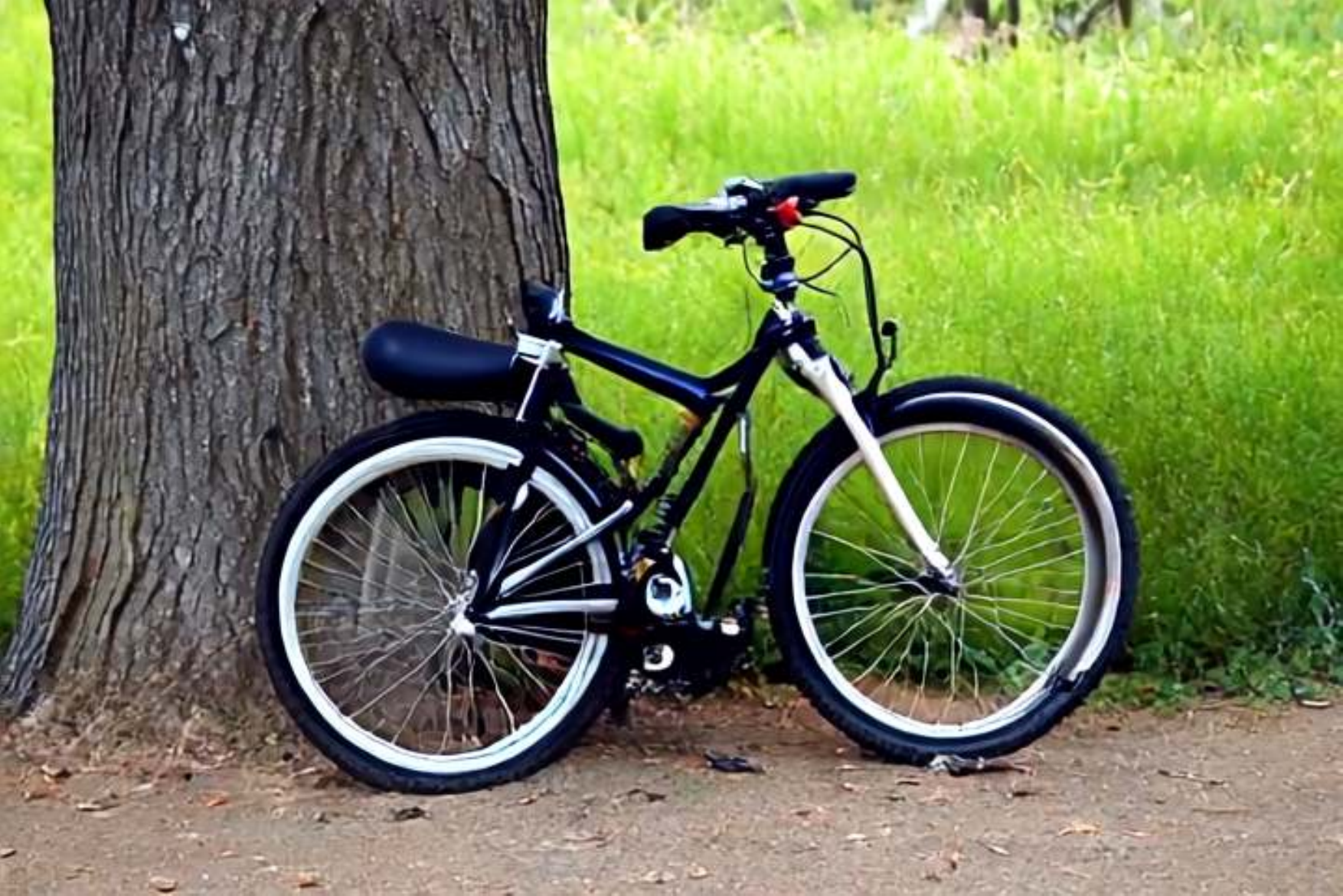}
        & \framecell{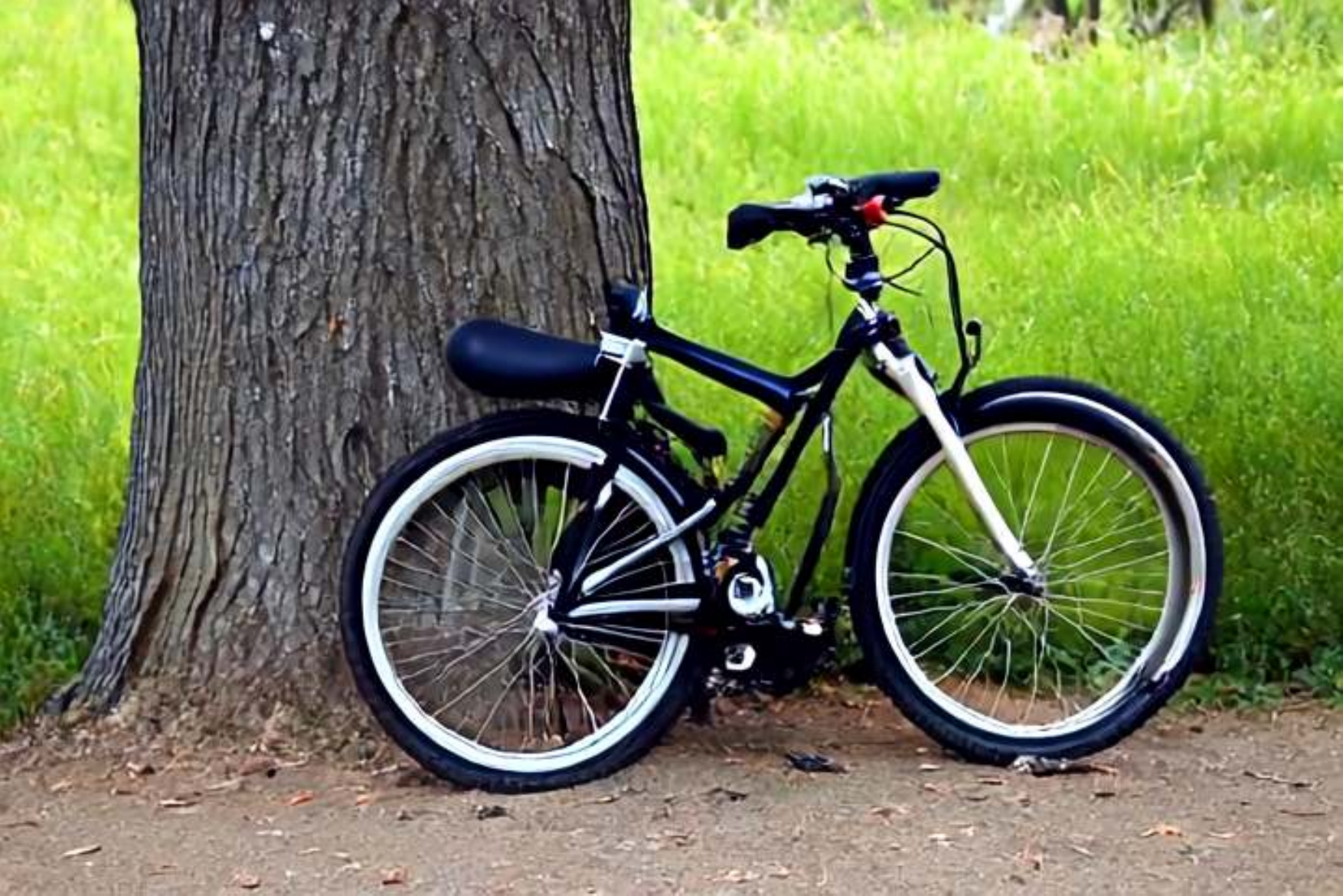}
        \\
    \end{tabular}
    \vspace{-0.3em}
    \caption{
    \textbf{Qualitative comparison on CogVideoX-2B (Part I).}
    }
    \label{fig:qualitative_cogvideox}
    \vspace{-0.8em}
\end{figure*}
% figure-local macros
\ifdefined\cogw\else\ifdefined\cogw\else\newlength{\cogw}\fi\fi
\ifdefined\cogh\else\ifdefined\cogh\else\newlength{\cogh}\fi\fi
\setlength{\cogw}{0.302\textwidth}
\setlength{\cogh}{0.170\textwidth}

\providecommand{\vcell}[1]{\ensuremath{\vcenter{\hbox{#1}}}}
\providecommand{\methodcell}[1]{%
    \vcell{\rotatebox[origin=c]{90}{\makebox[\cogh][c]{\textbf{#1}}}}%
}
\providecommand{\framecell}[1]{%
    \vcell{\includegraphics[width=\cogw,height=\cogh]{#1}}%
}
\providecommand{\rowgap}{\noalign{\vskip 3pt}}  % 控制行间距，只改这里

\begin{figure*}[t]
    \centering
    \setlength{\tabcolsep}{1.2pt}
    \renewcommand{\arraystretch}{0.0}
    \scriptsize

    \begin{tabular}{@{}c@{\hspace{1.5pt}}ccc@{}}
        & \multicolumn{3}{c}{\large \textbf{Close up of grapes on a rotating table.}}
        \\[0.25em]

        \methodcell{Original}
        & \framecell{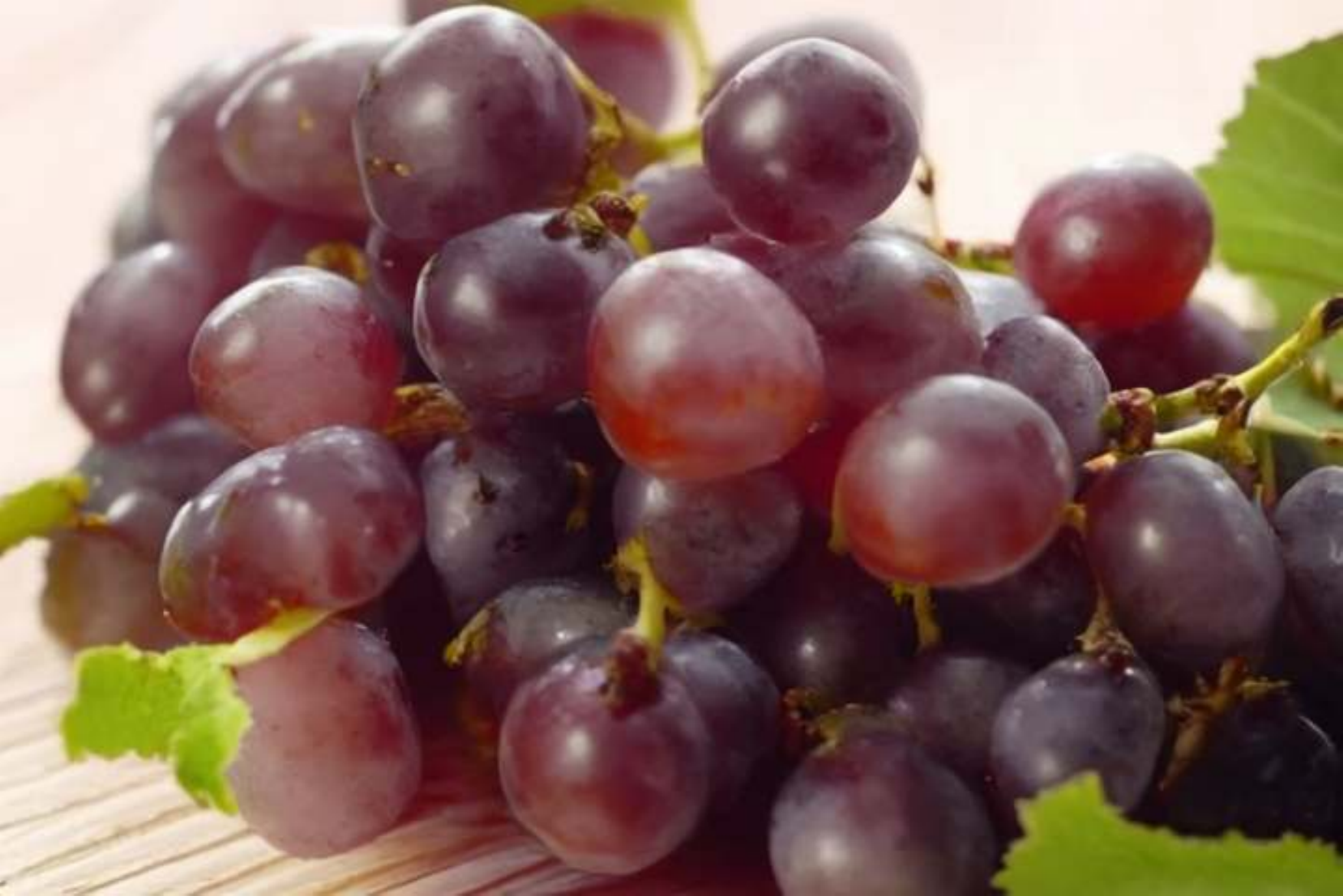}
        & \framecell{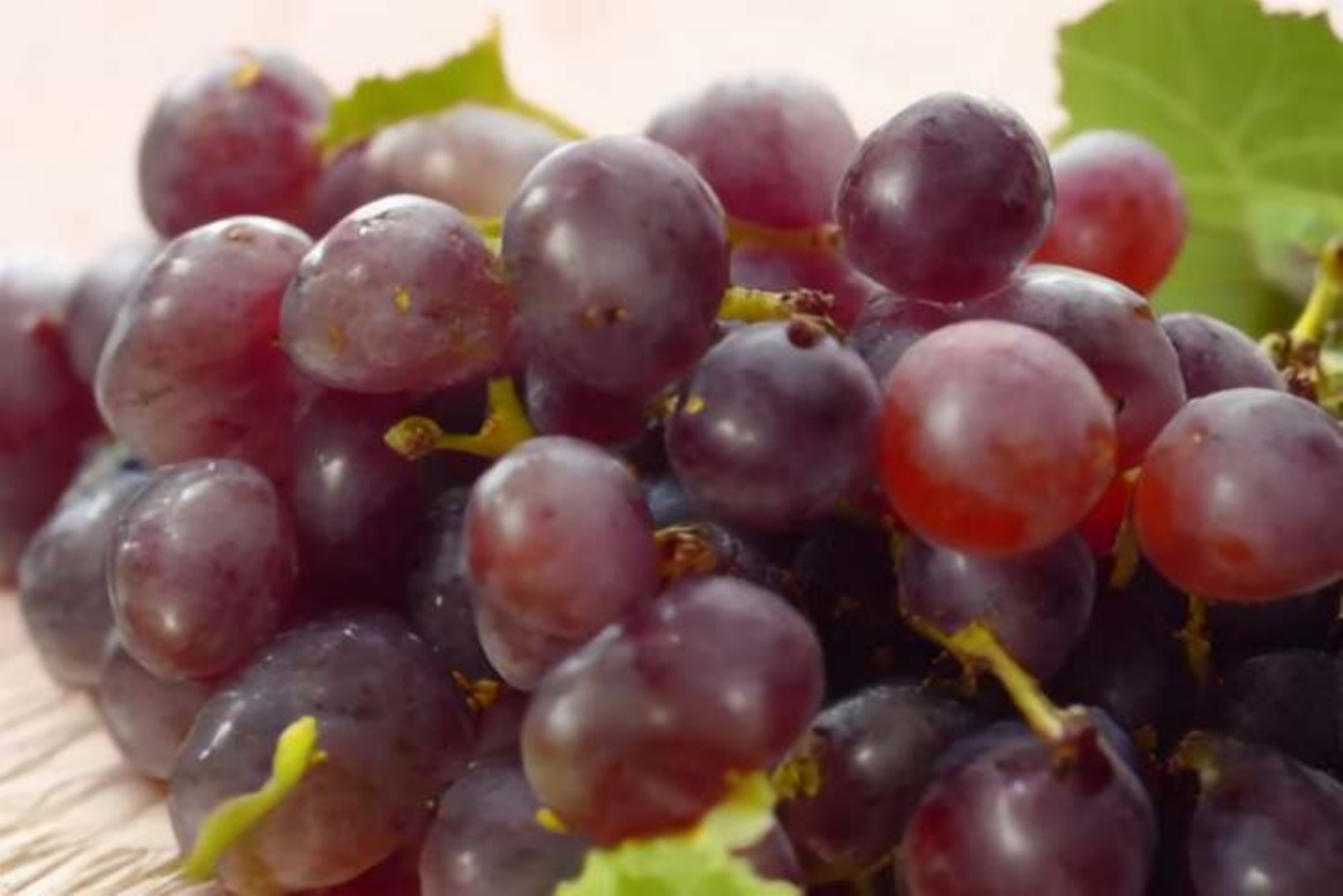}
        & \framecell{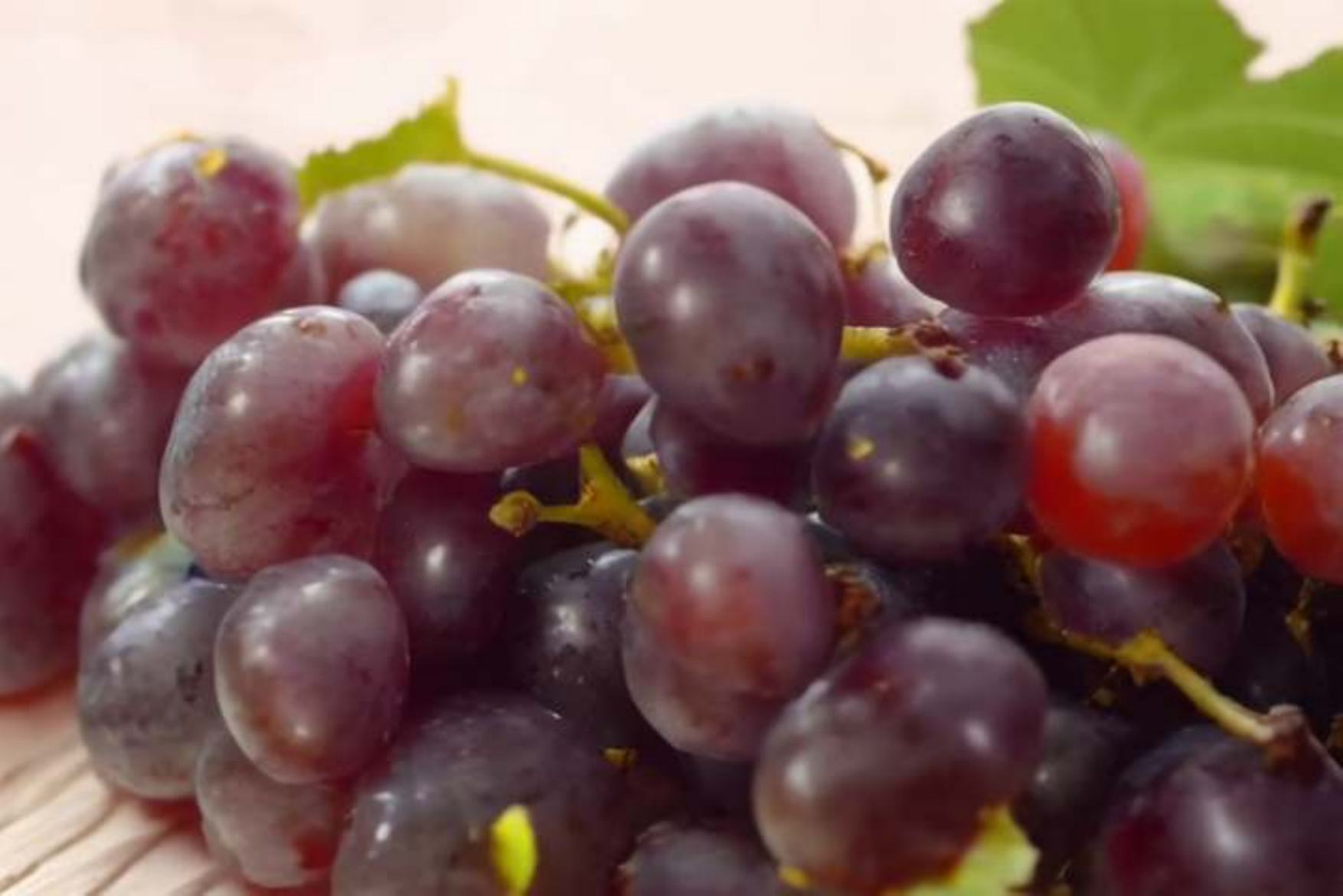}
        \\ \rowgap

        \methodcell{MagCache ($2.32\times$)}
        & \framecell{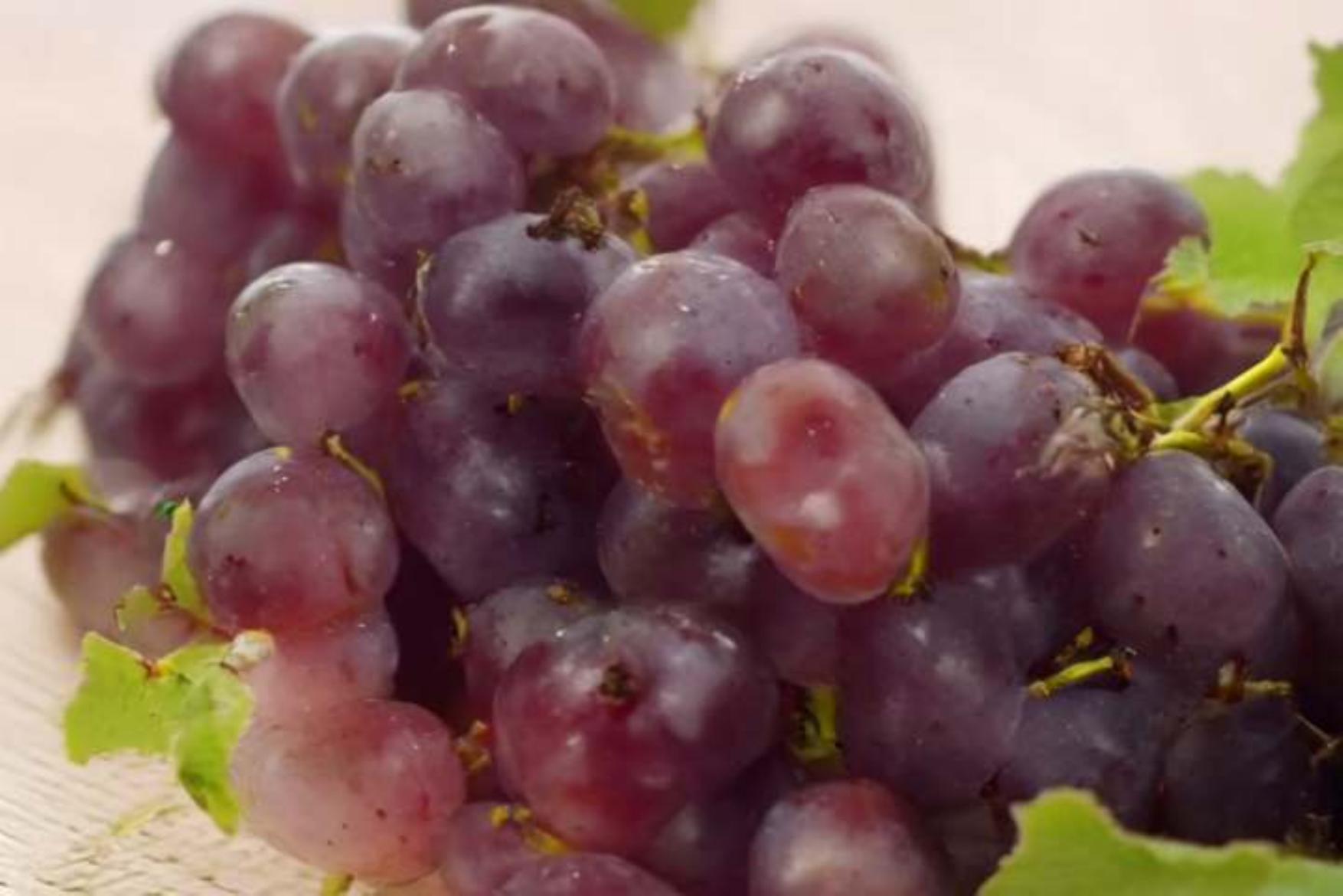}
        & \framecell{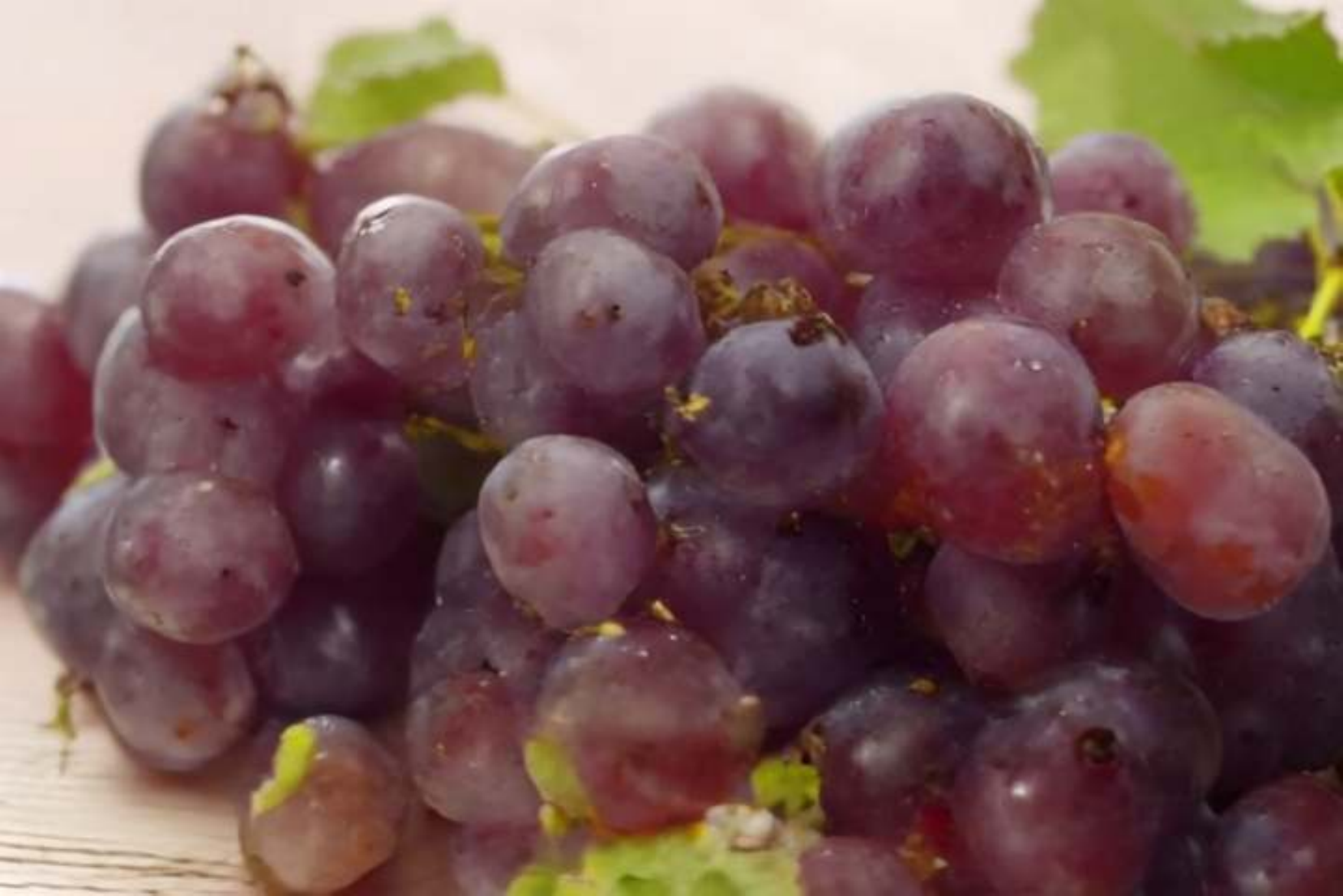}
        & \framecell{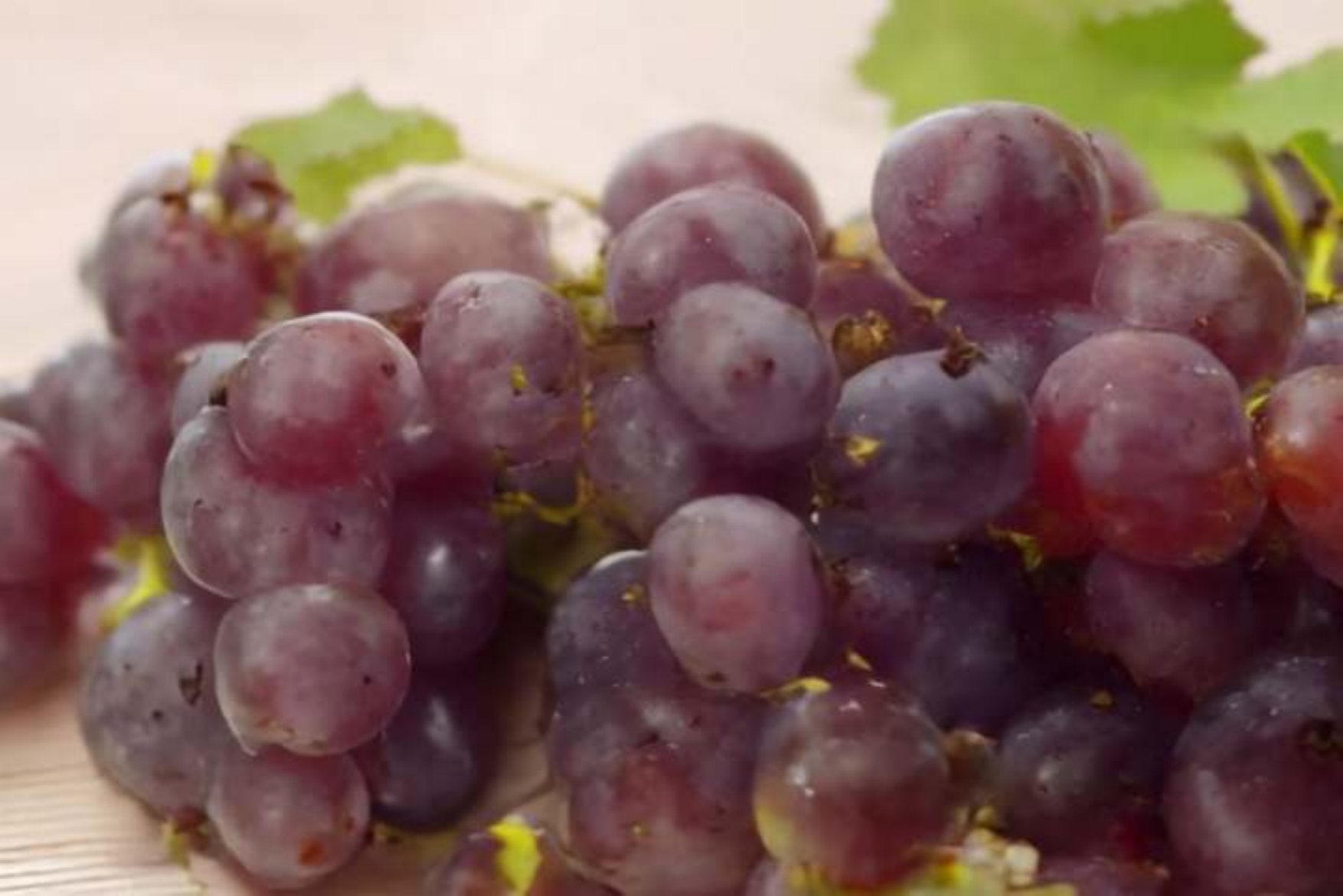}
        \\ \rowgap

        \methodcell{DiCache ($2.42\times$)}
        & \framecell{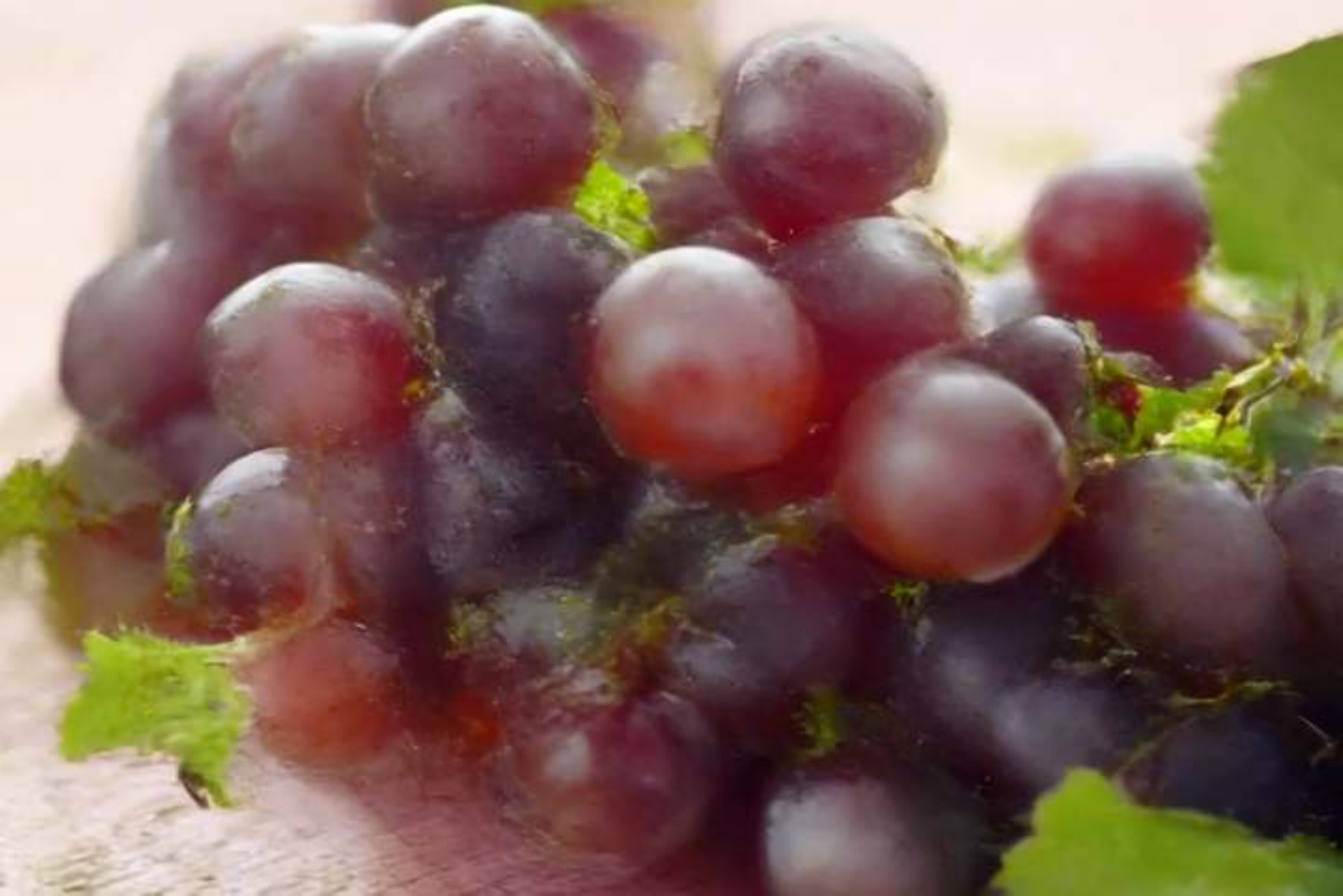}
        & \framecell{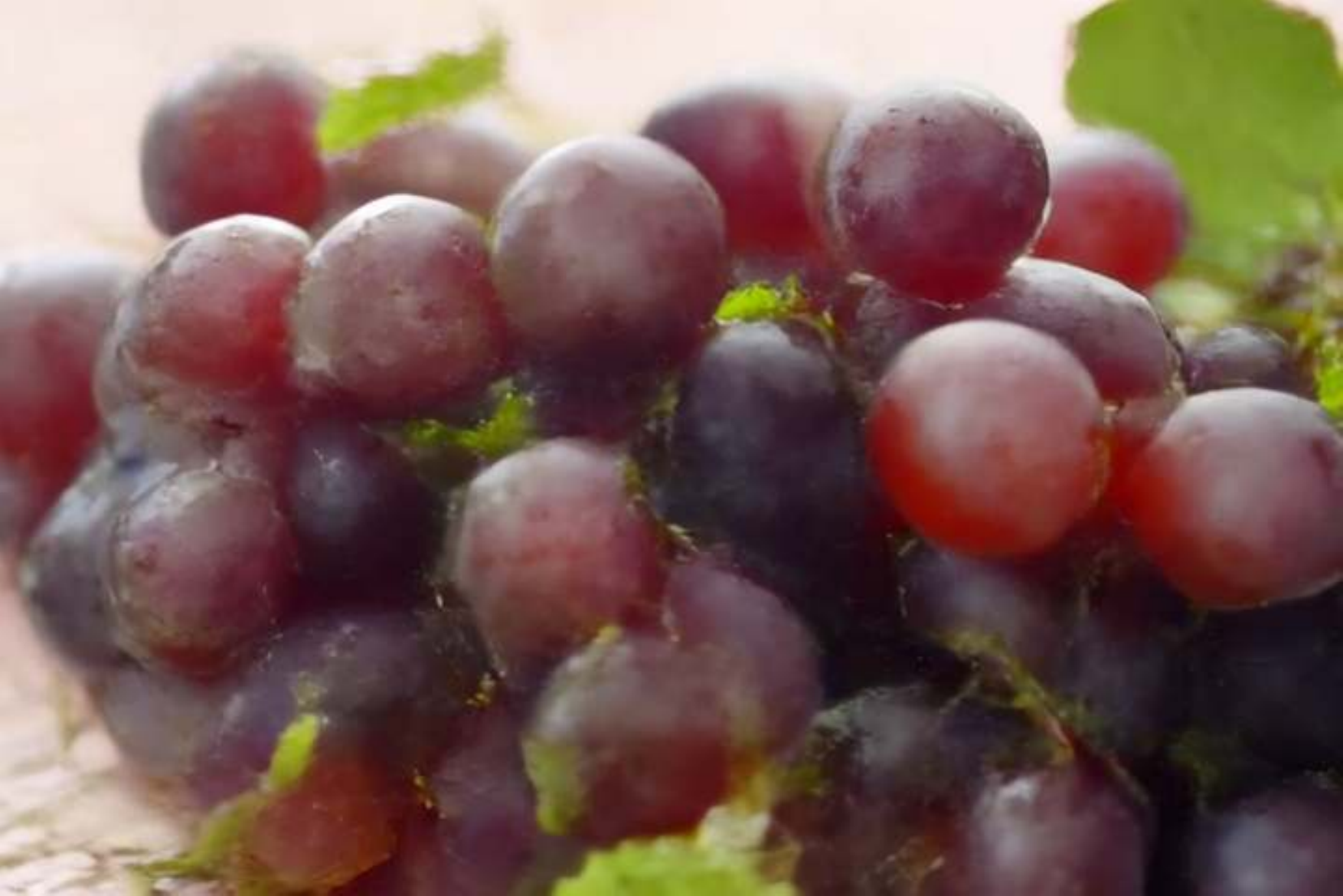}
        & \framecell{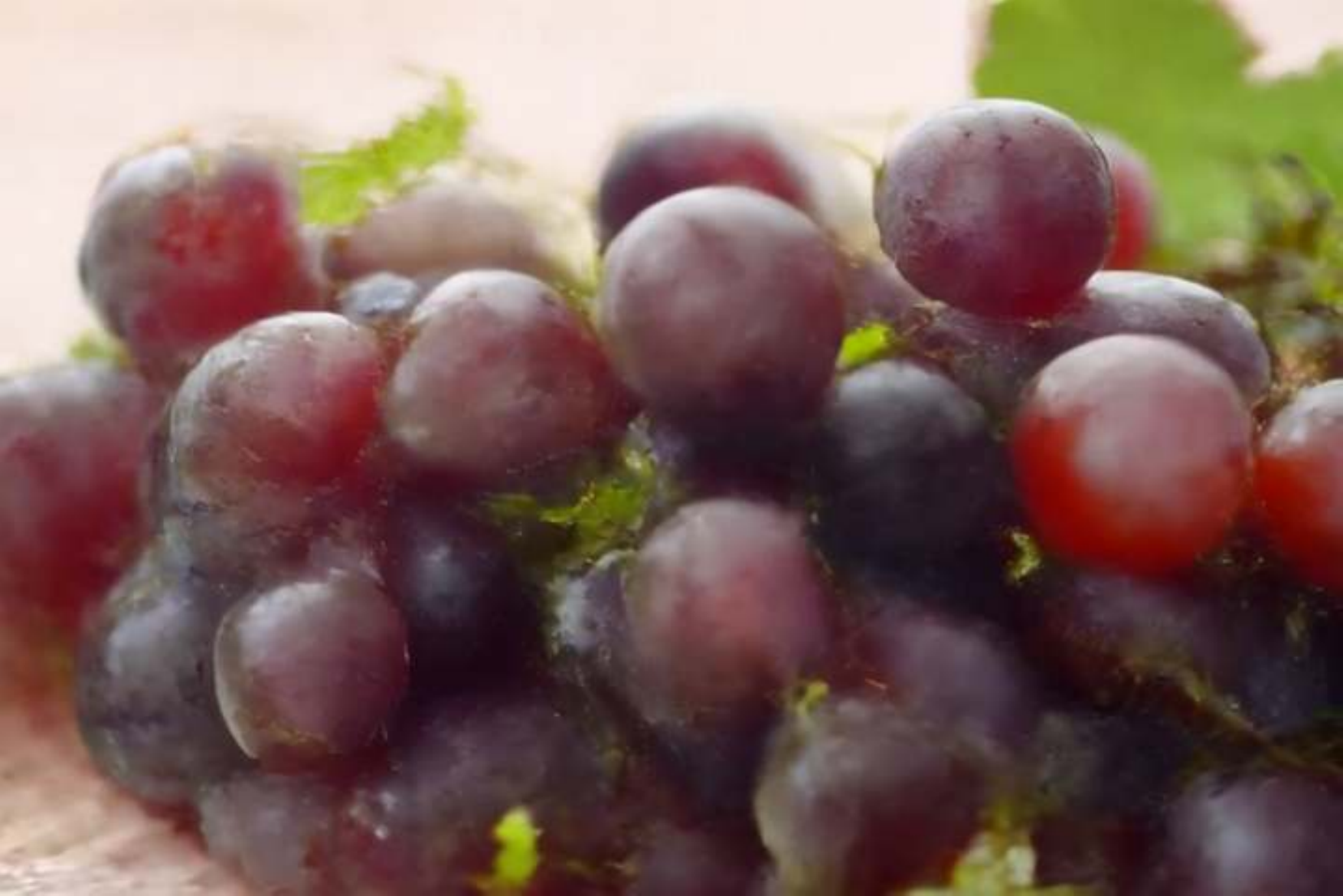}
        \\ \rowgap

        \methodcell{Ours ($2.38\times$)}
        & \framecell{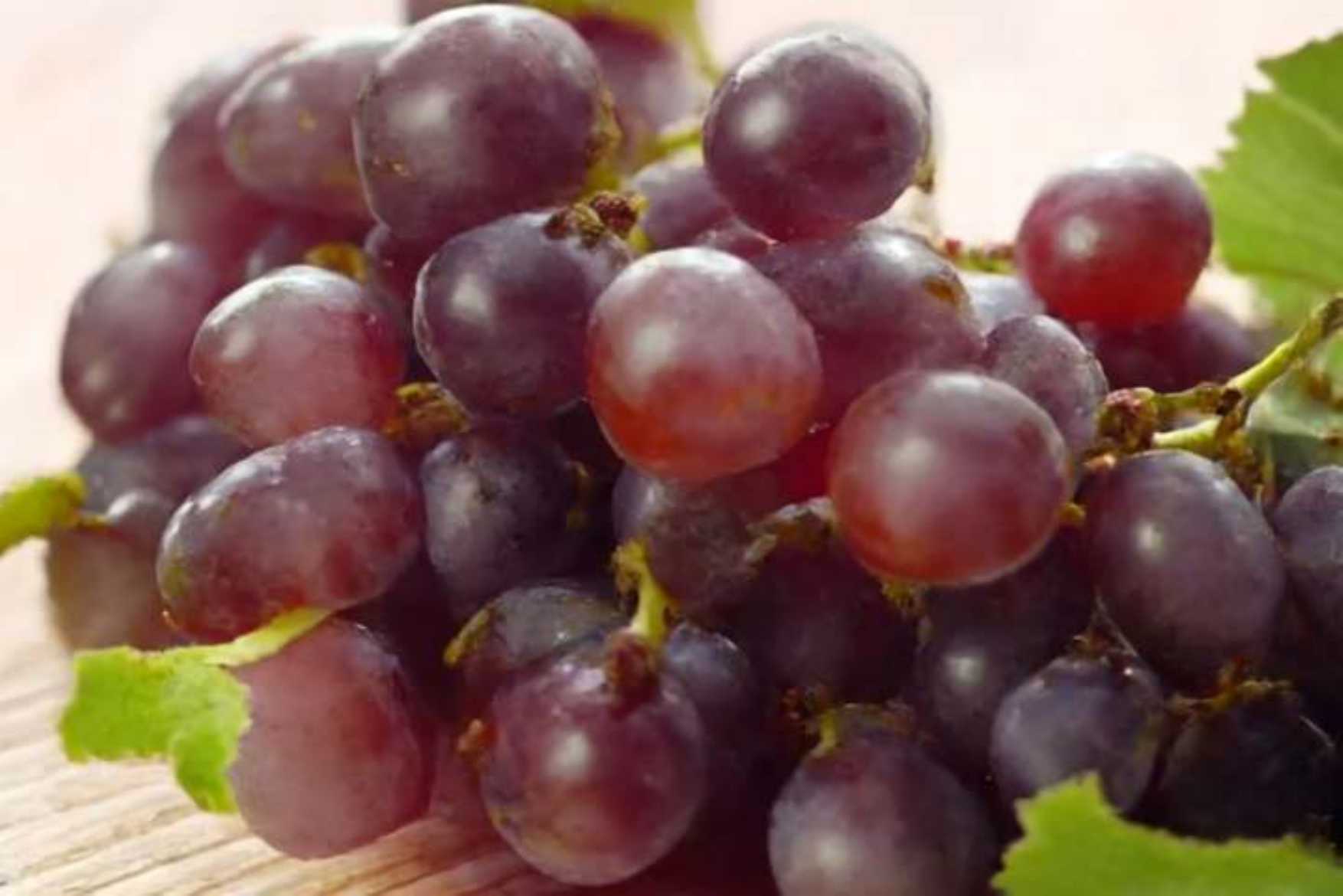}
        & \framecell{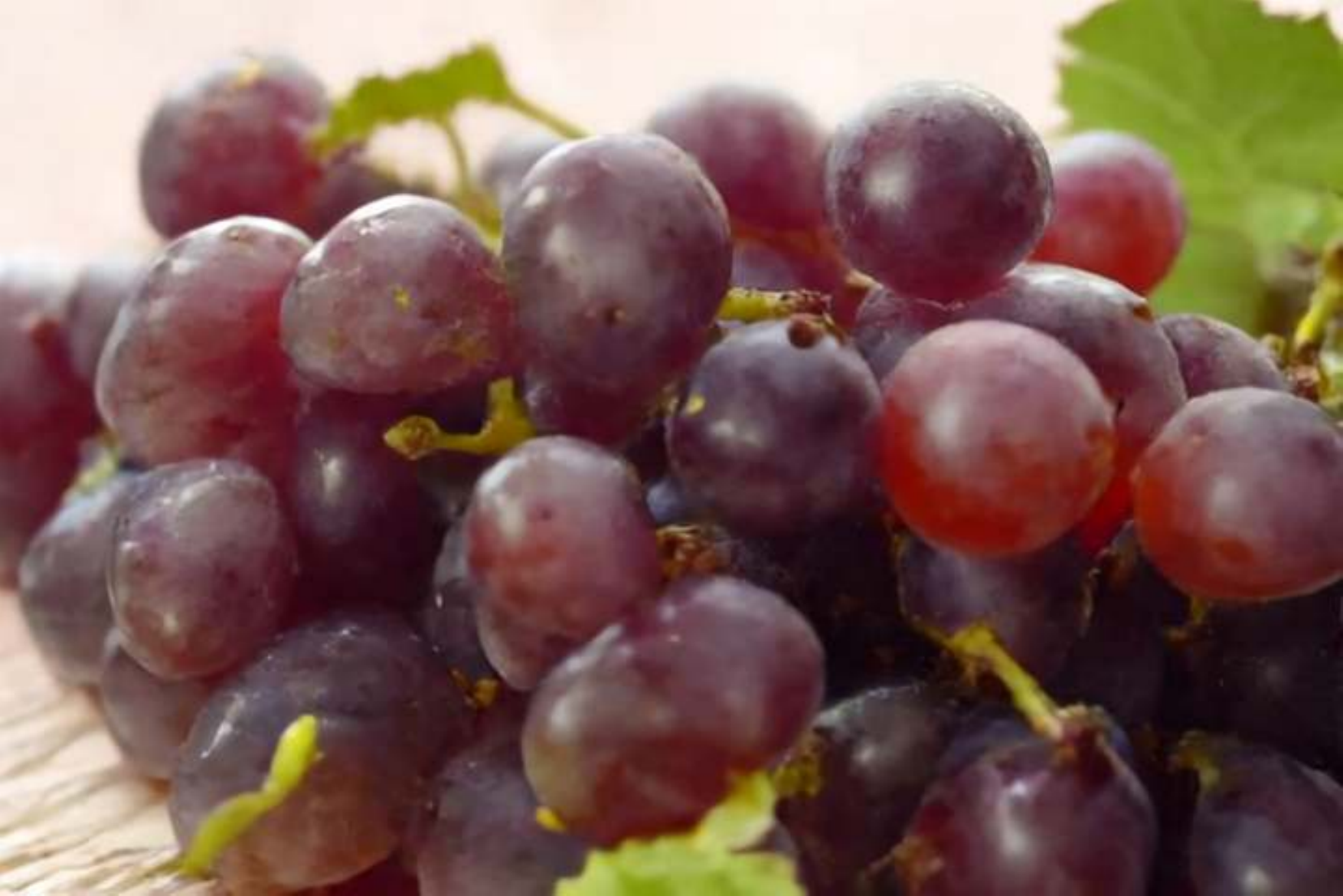}
        & \framecell{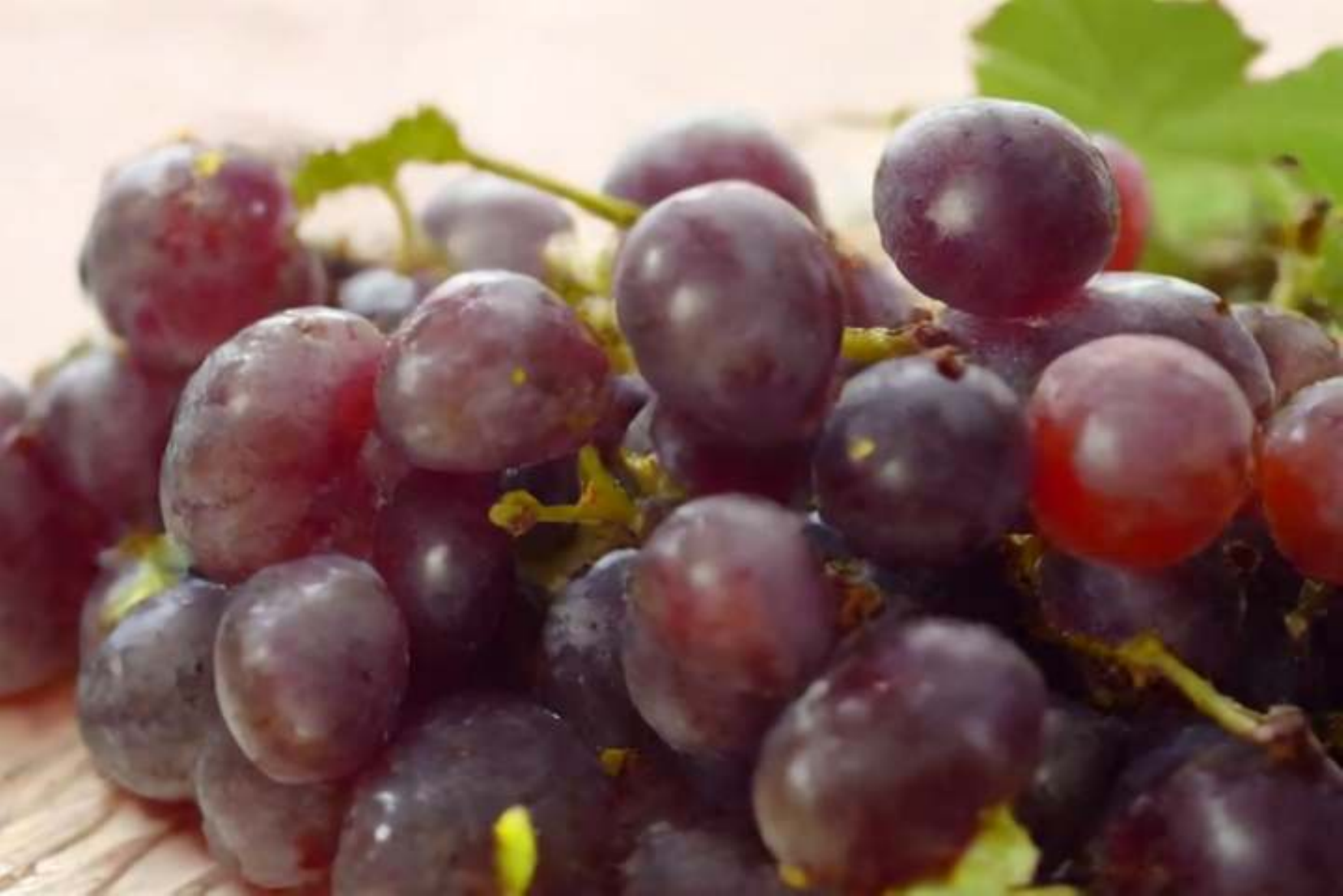}
        \\
    \end{tabular}
    \par\vspace{0.8em}
    \begin{tabular}{@{}c@{\hspace{1.5pt}}ccc@{}}
        & \multicolumn{3}{c}{\large \textbf{A tranquil tableau of bedroom}}
        \\[0.25em]

        \methodcell{Original}
        & \framecell{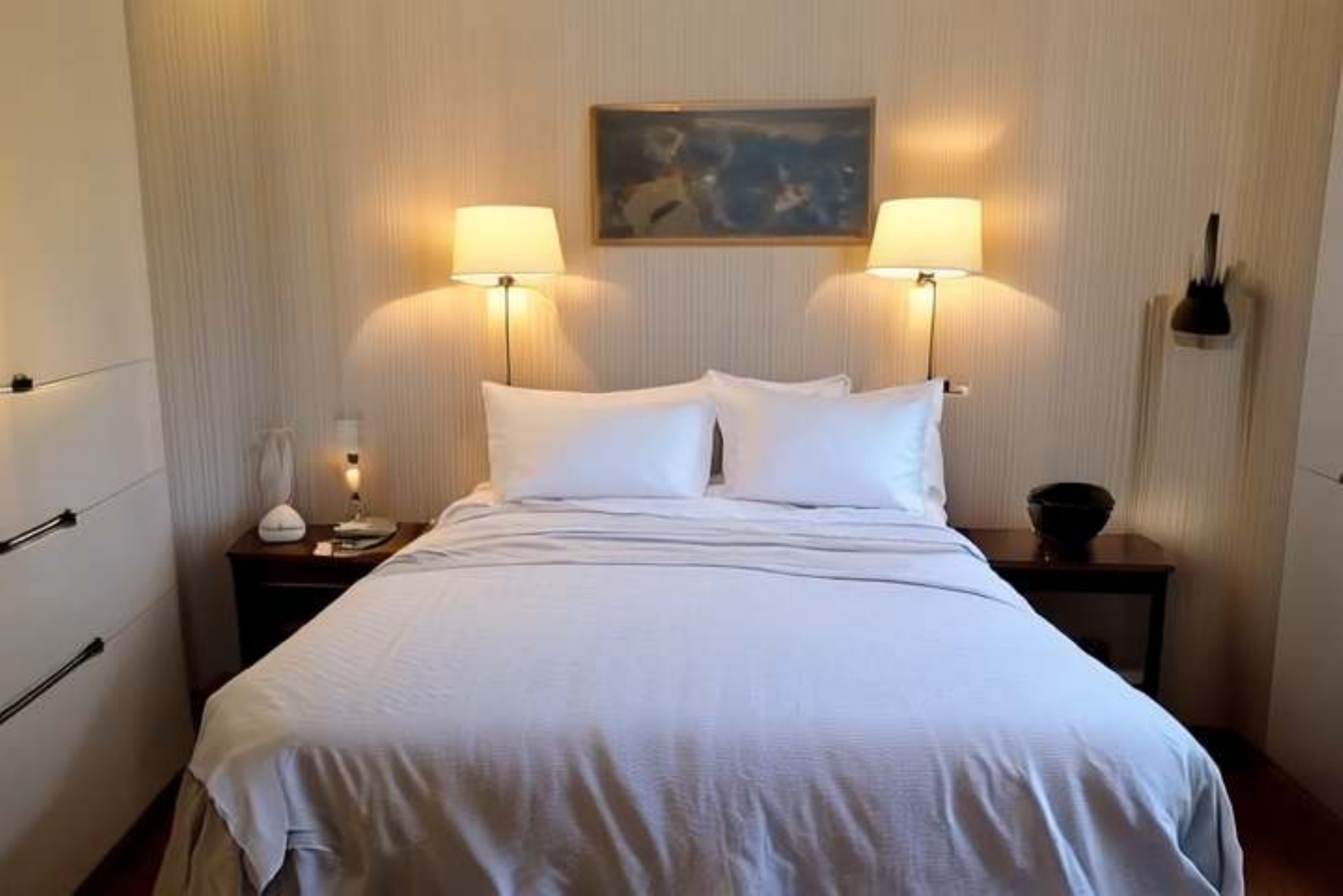}
        & \framecell{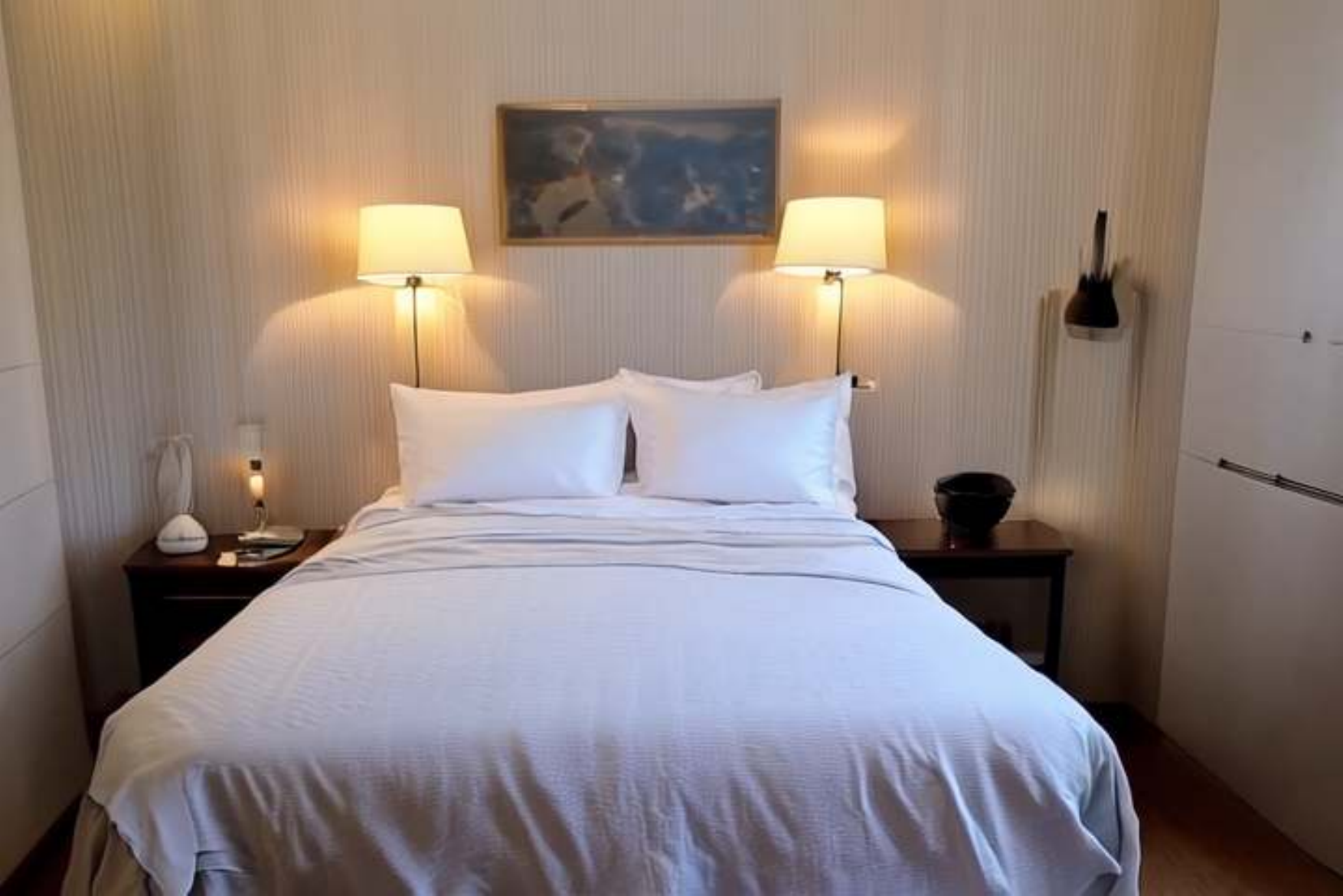}
        & \framecell{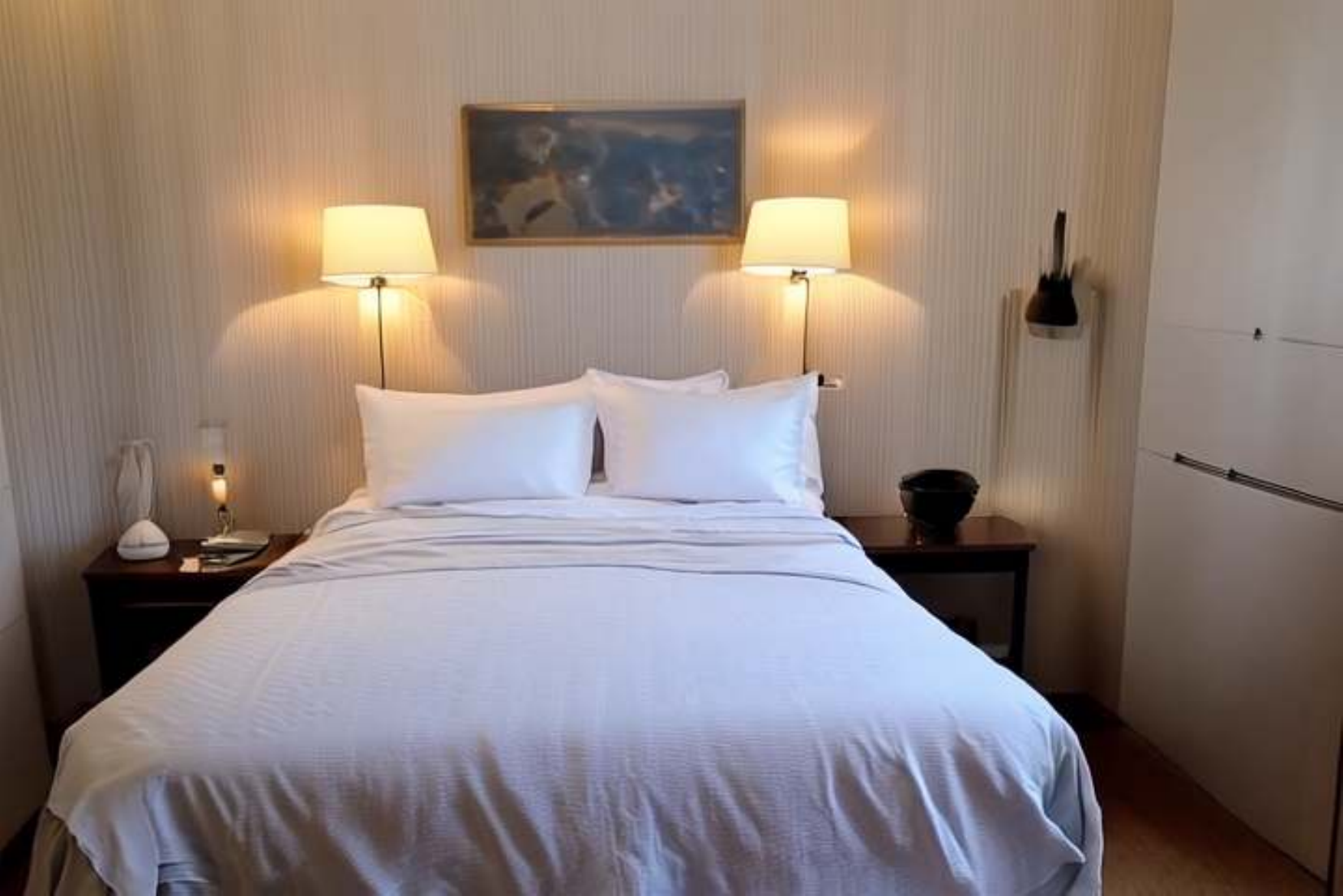}
        \\ \rowgap

        \methodcell{MagCache ($2.32\times$)}
        & \framecell{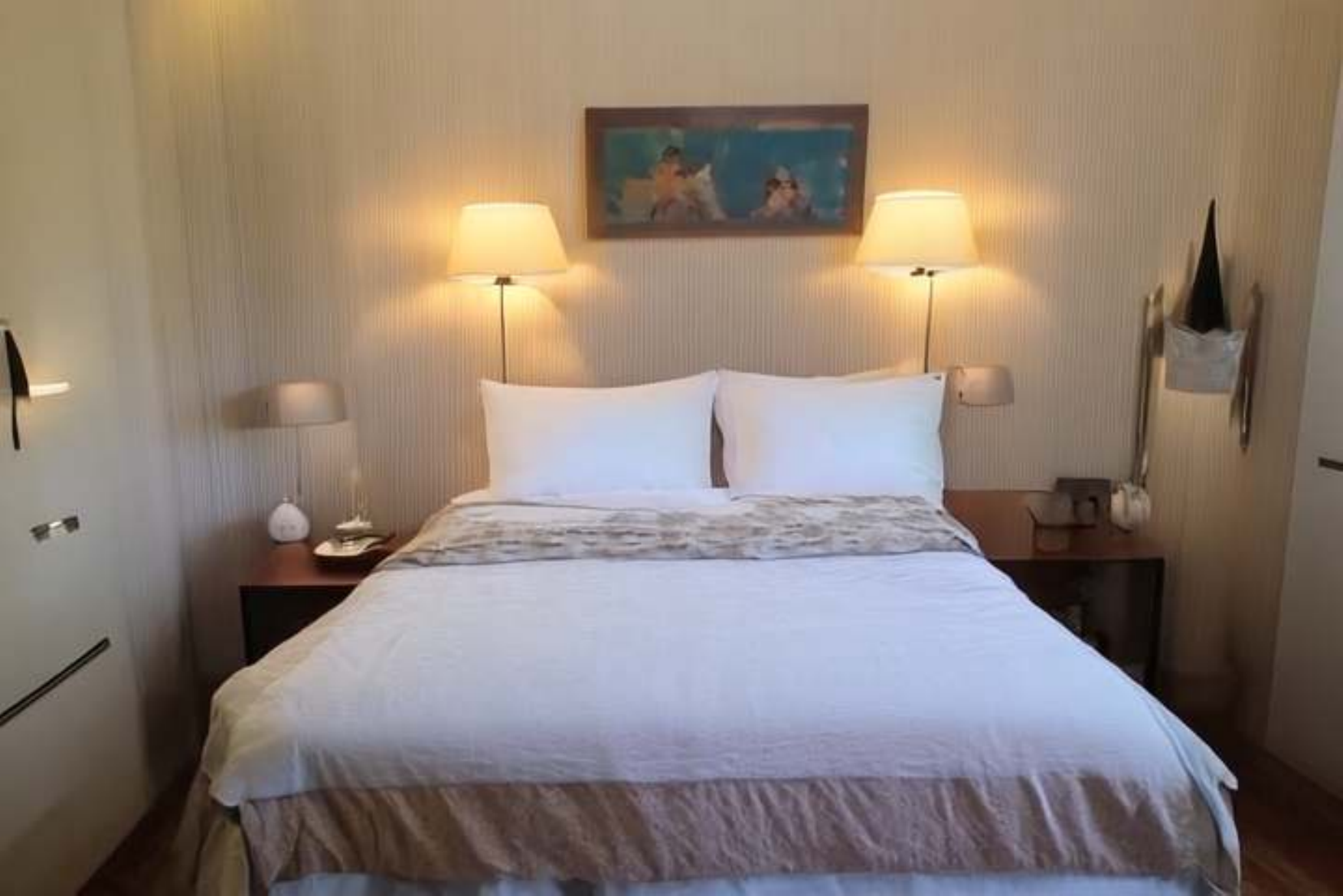}
        & \framecell{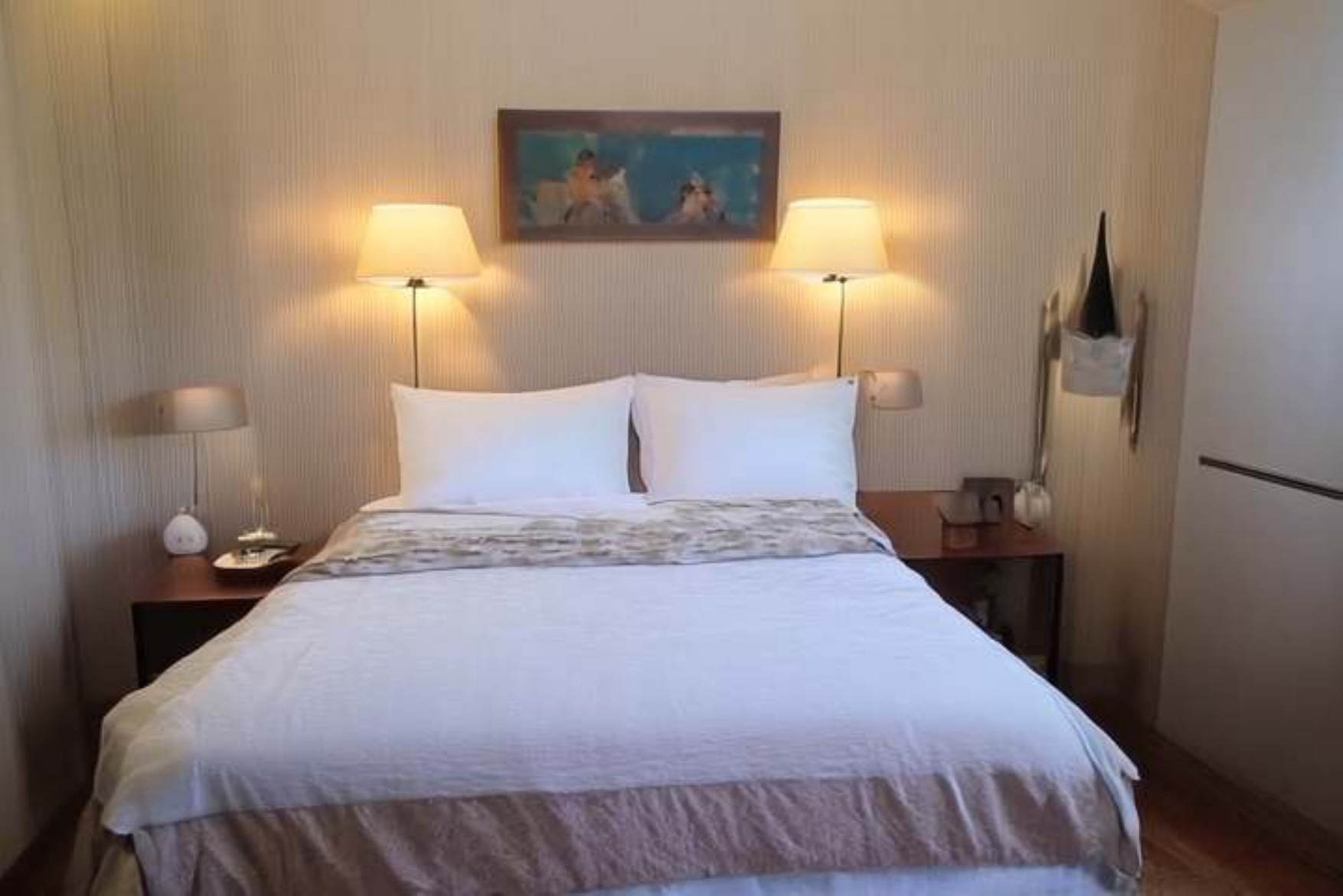}
        & \framecell{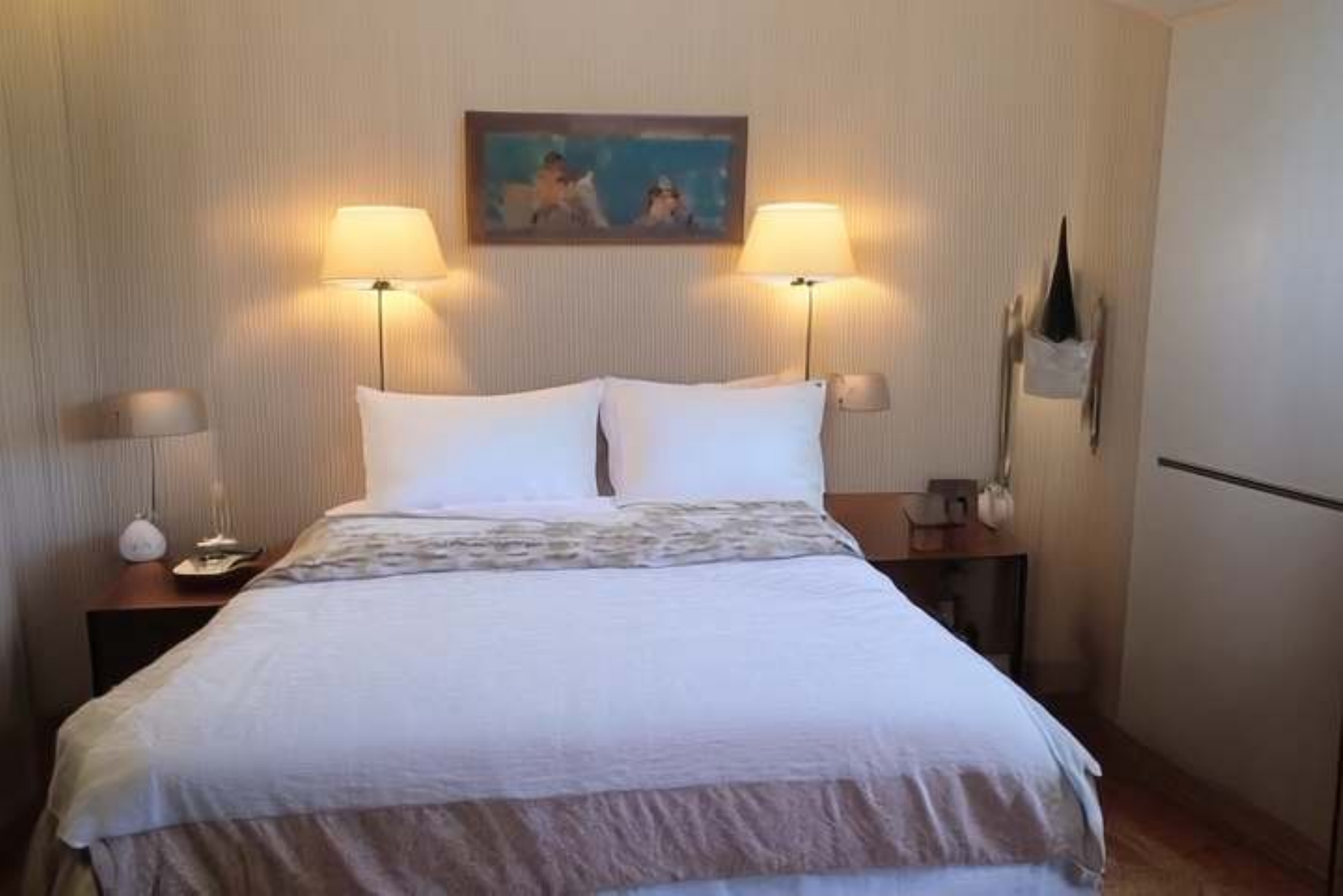}
        \\ \rowgap

        \methodcell{DiCache ($2.42\times$)}
        & \framecell{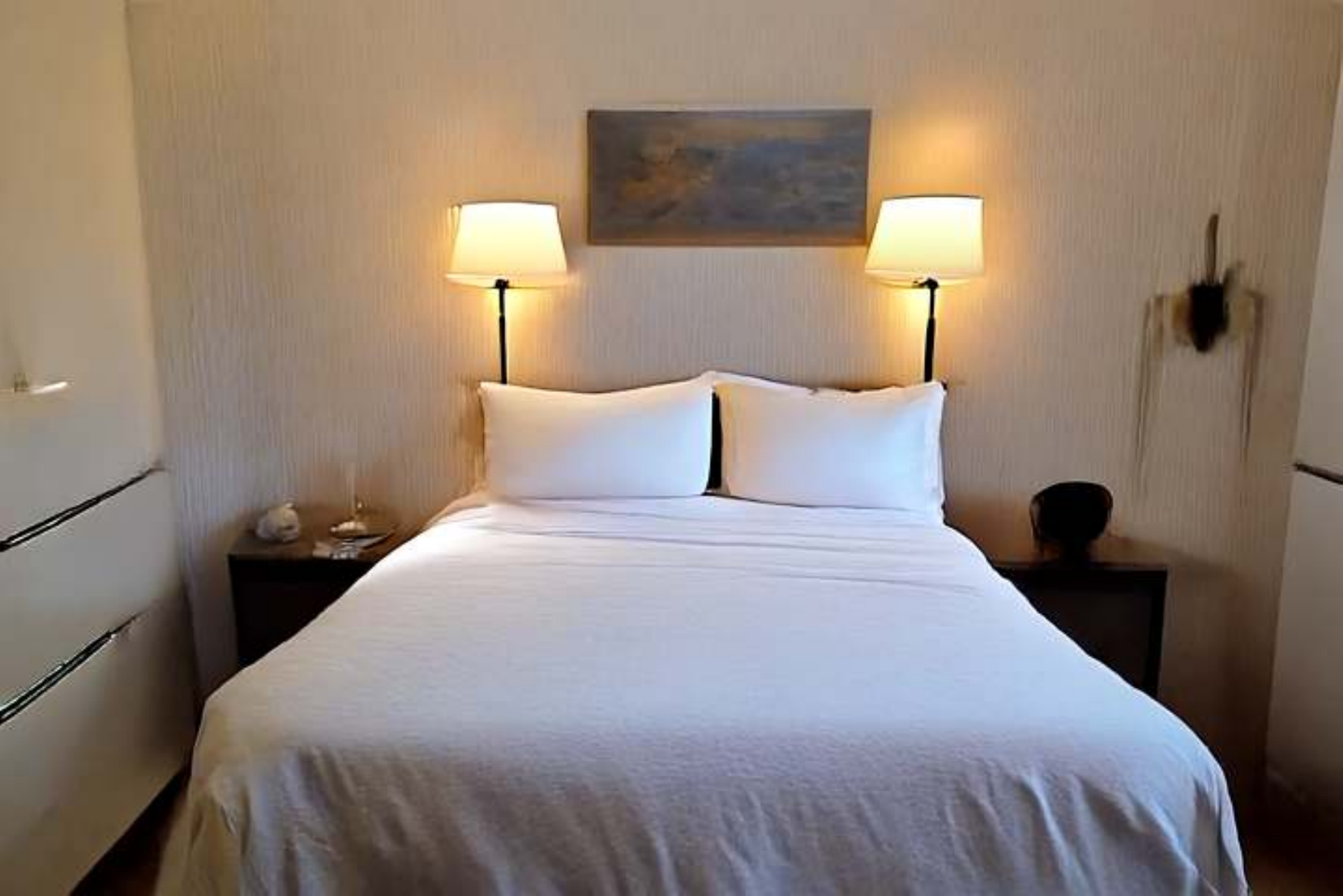}
        & \framecell{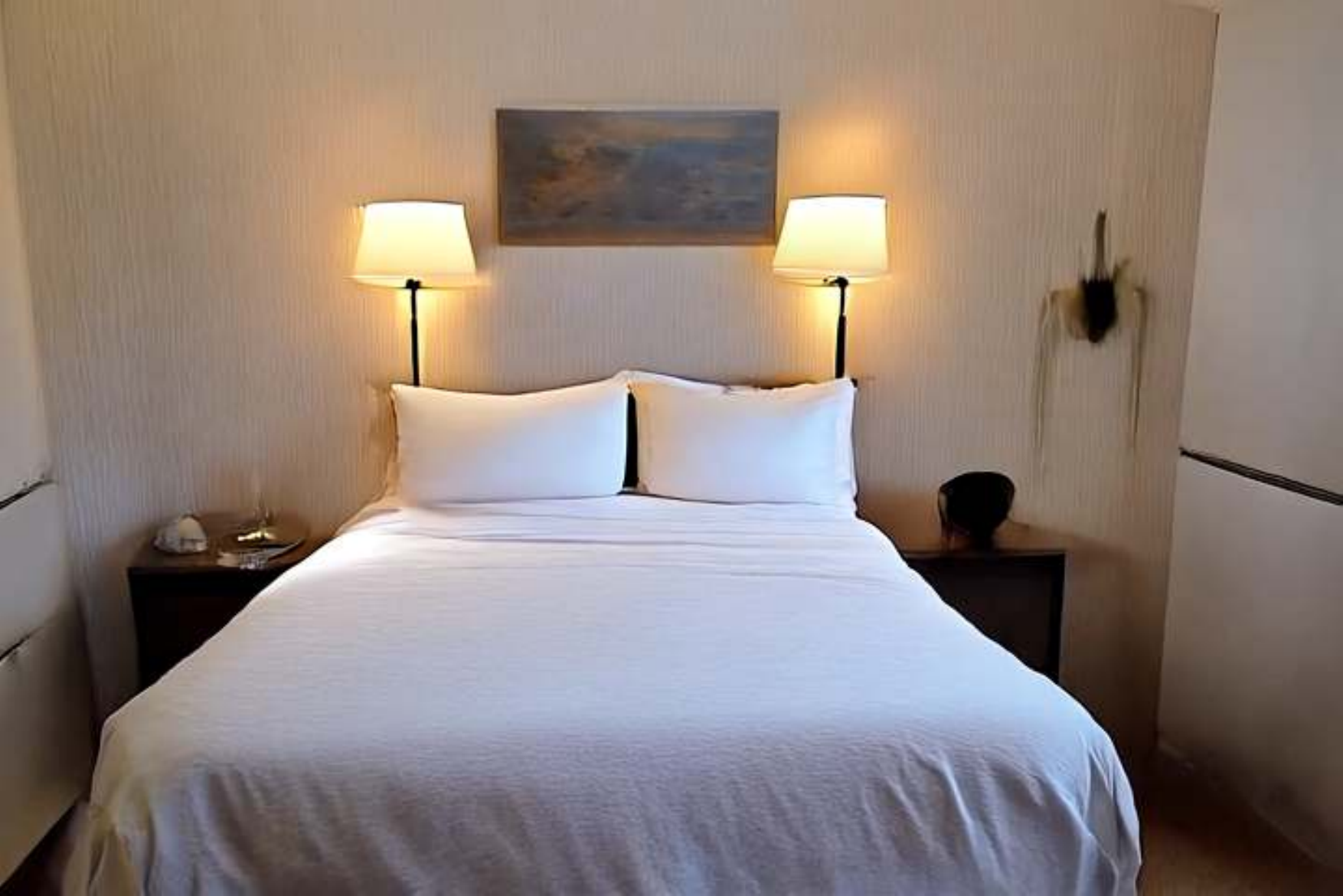}
        & \framecell{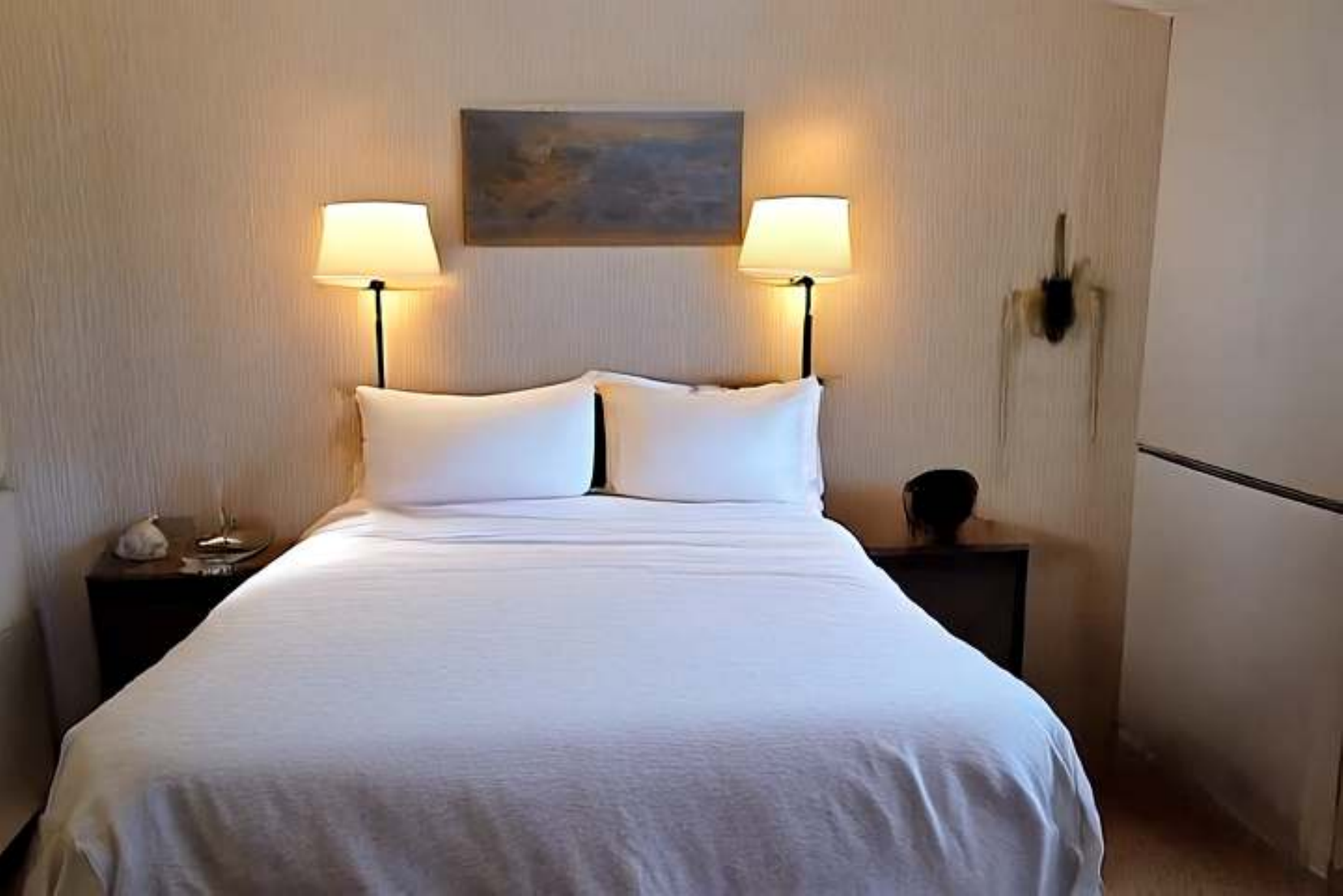}
        \\ \rowgap

        \methodcell{Ours ($2.38\times$)}
        & \framecell{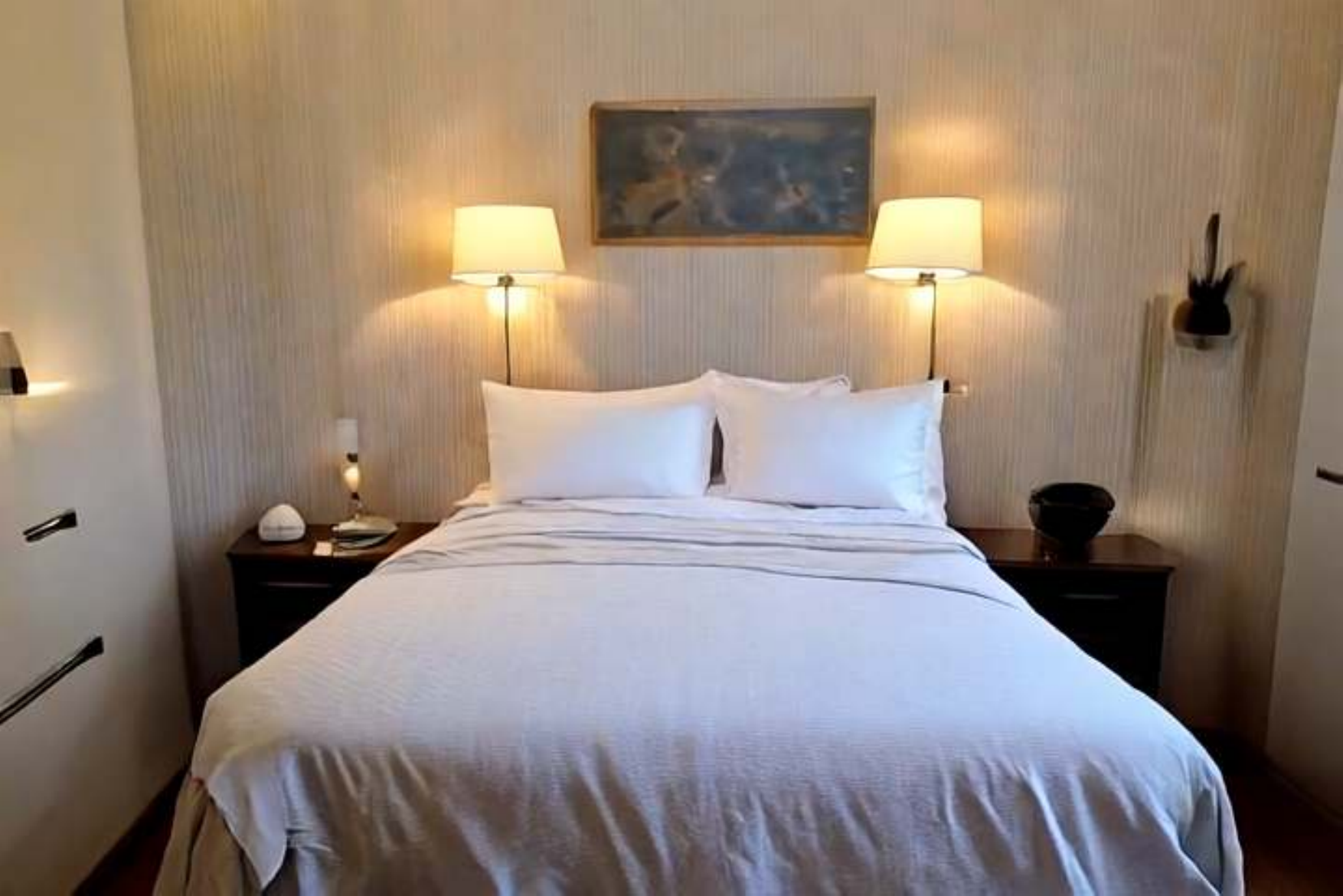}
        & \framecell{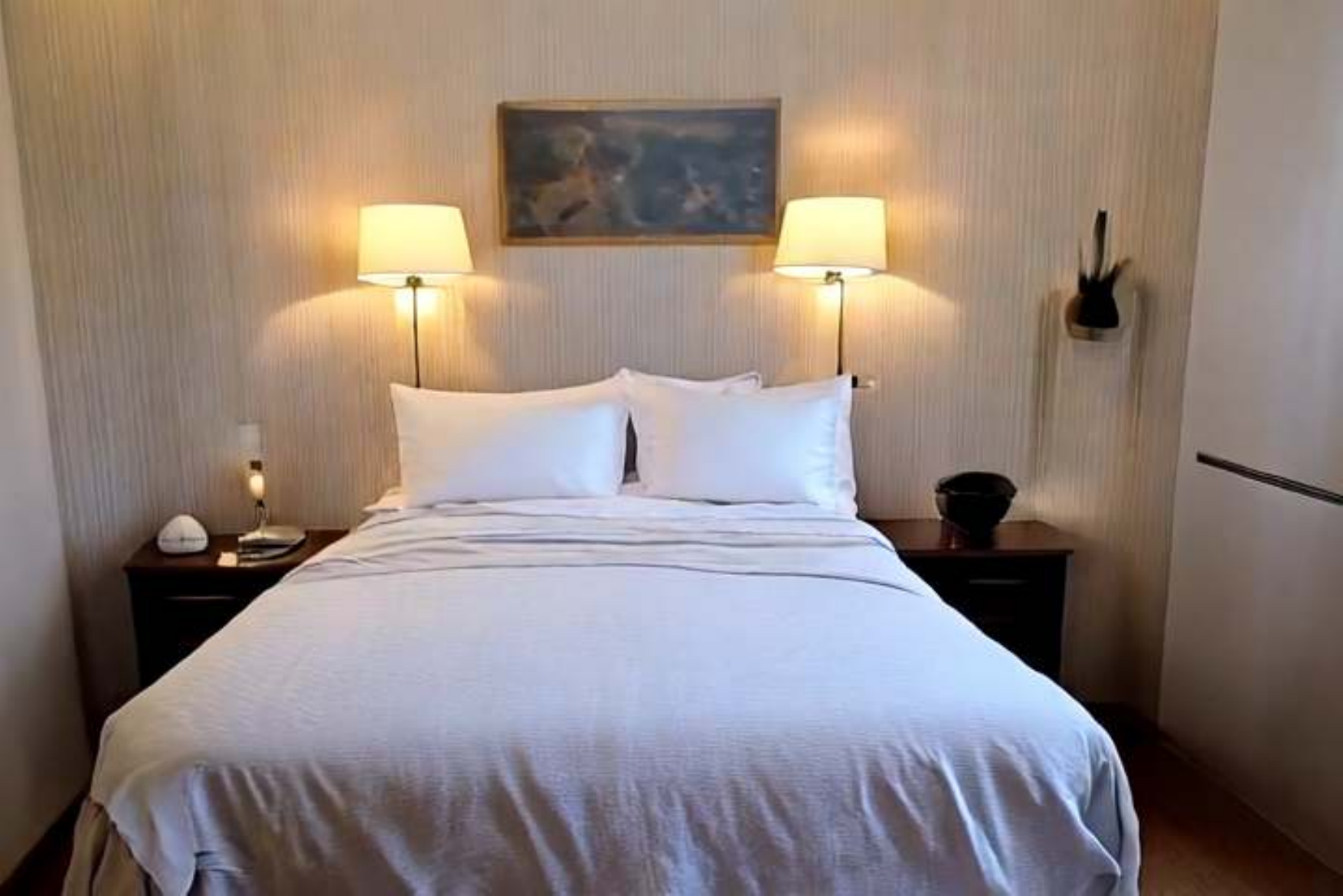}
        & \framecell{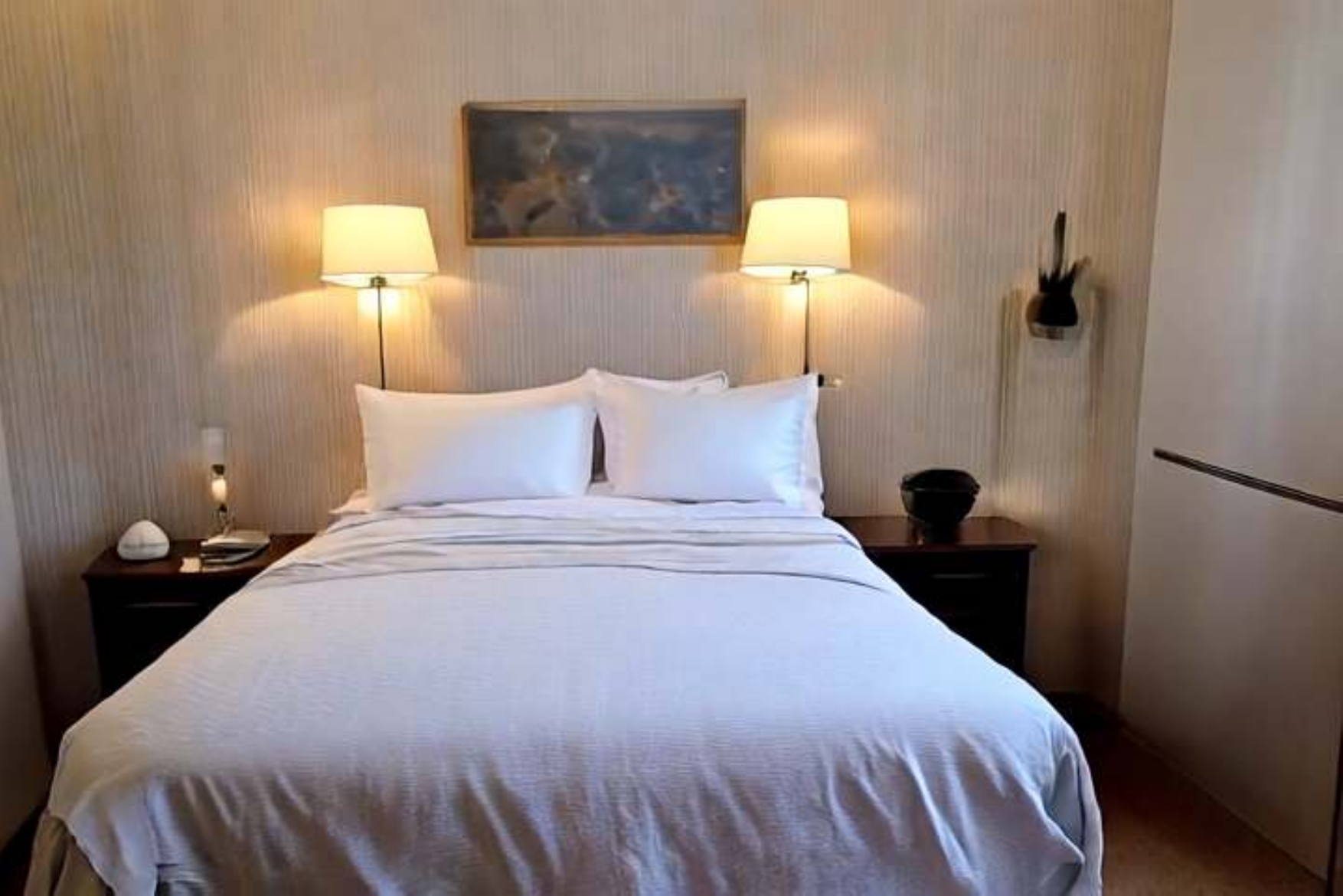}
        \\
    \end{tabular}
    \vspace{-0.3em}
    \caption{
    \textbf{Qualitative comparison on CogVideoX-2B (Part II).}
    }
    \label{fig:qualitative_cogvideox}
    \vspace{-0.8em}
\end{figure*}
% figure-local macros
\ifdefined\cogw\else\newlength{\cogw}\fi
\ifdefined\cogh\else\newlength{\cogh}\fi
\setlength{\cogw}{0.302\textwidth}
\setlength{\cogh}{0.170\textwidth}

\providecommand{\vcell}[1]{\ensuremath{\vcenter{\hbox{#1}}}}
\providecommand{\methodcell}[1]{%
    \vcell{\rotatebox[origin=c]{90}{\makebox[\cogh][c]{\textbf{#1}}}}%
}
\providecommand{\framecell}[1]{%
    \vcell{\includegraphics[width=\cogw,height=\cogh]{#1}}%
}
\providecommand{\rowgap}{\noalign{\vskip 3pt}}  % 控制行间距，只改这里

\begin{figure*}[t]
    \centering
    \setlength{\tabcolsep}{1.2pt}
    \renewcommand{\arraystretch}{0.0}
    \scriptsize

    \begin{tabular}{@{}c@{\hspace{1.5pt}}ccc@{}}
        & \multicolumn{3}{c}{\large \textbf{An astronaut flying in space.}}
        \\[0.25em]

        \methodcell{Original}
        & \framecell{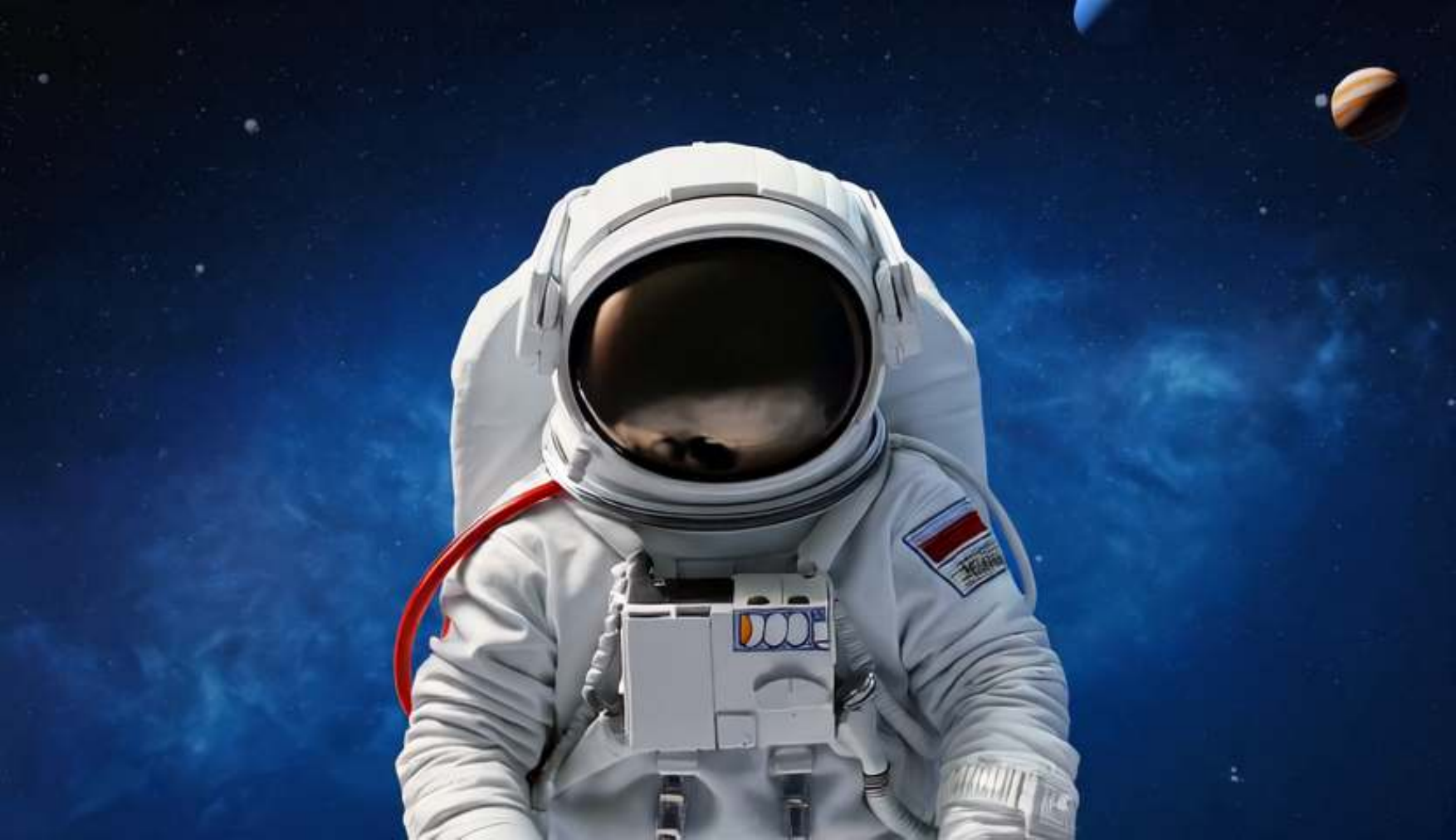}
        & \framecell{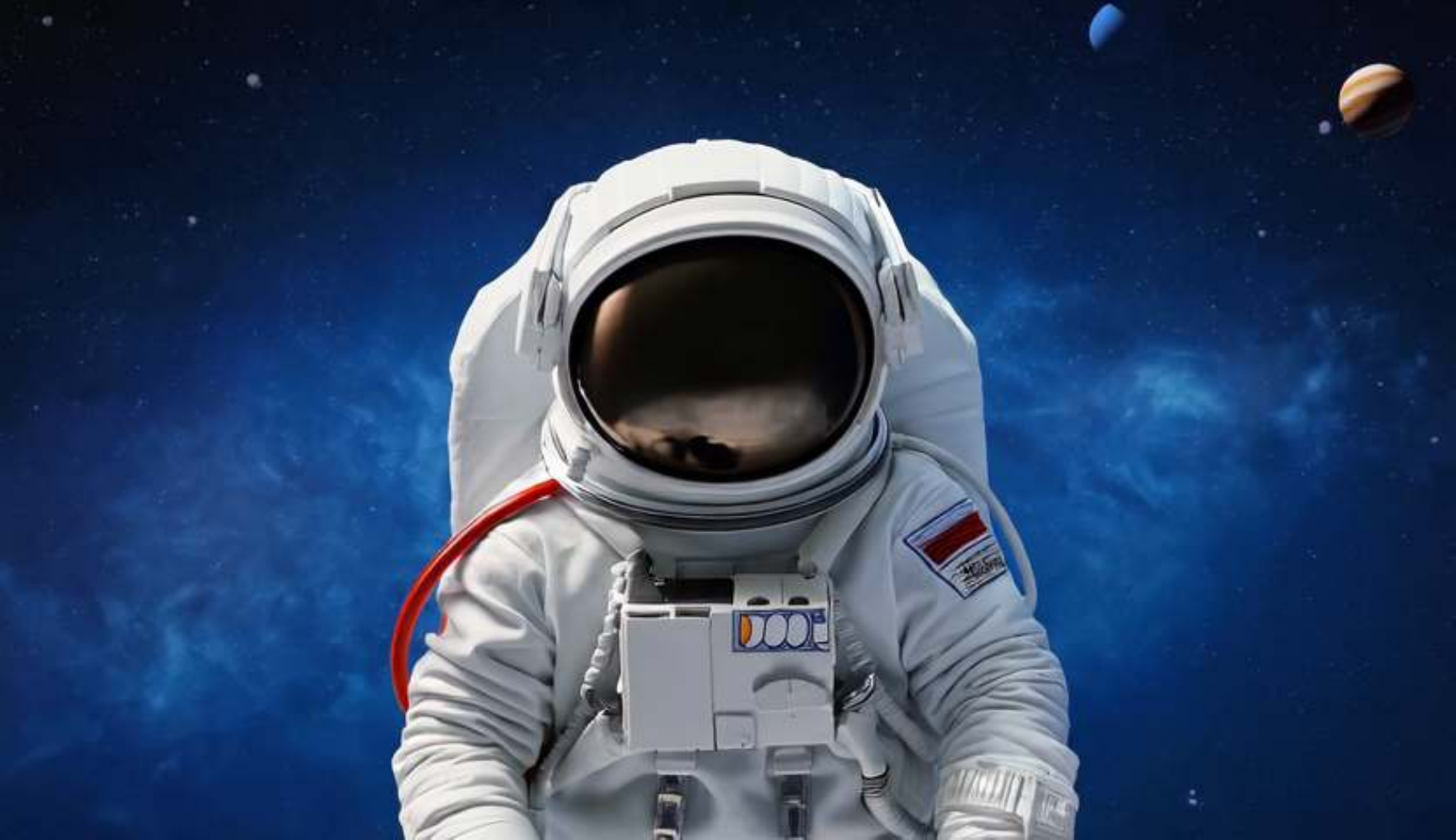}
        & \framecell{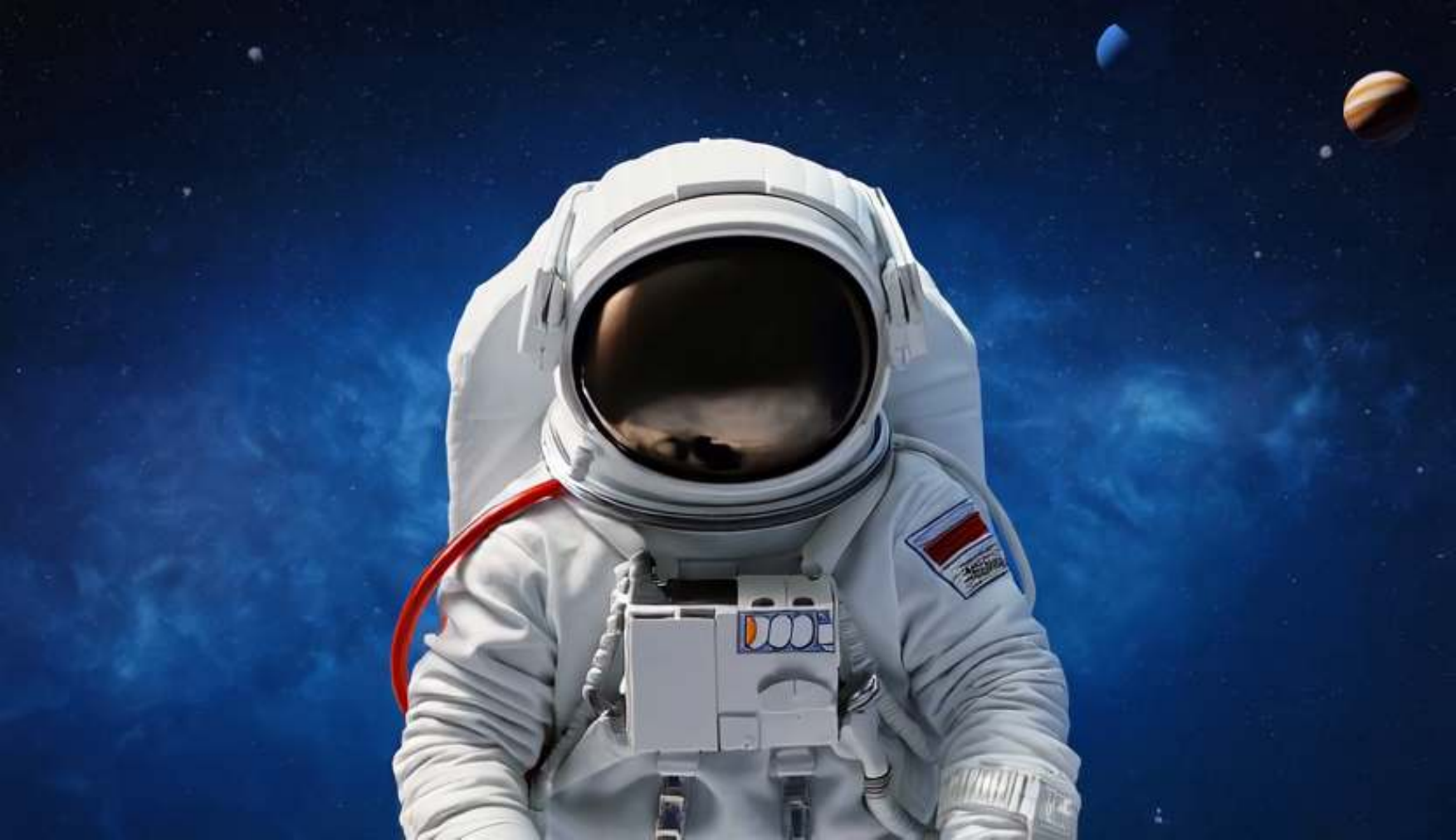}
        \\ \rowgap

        \methodcell{MagCache ($2.54\times$)}
        & \framecell{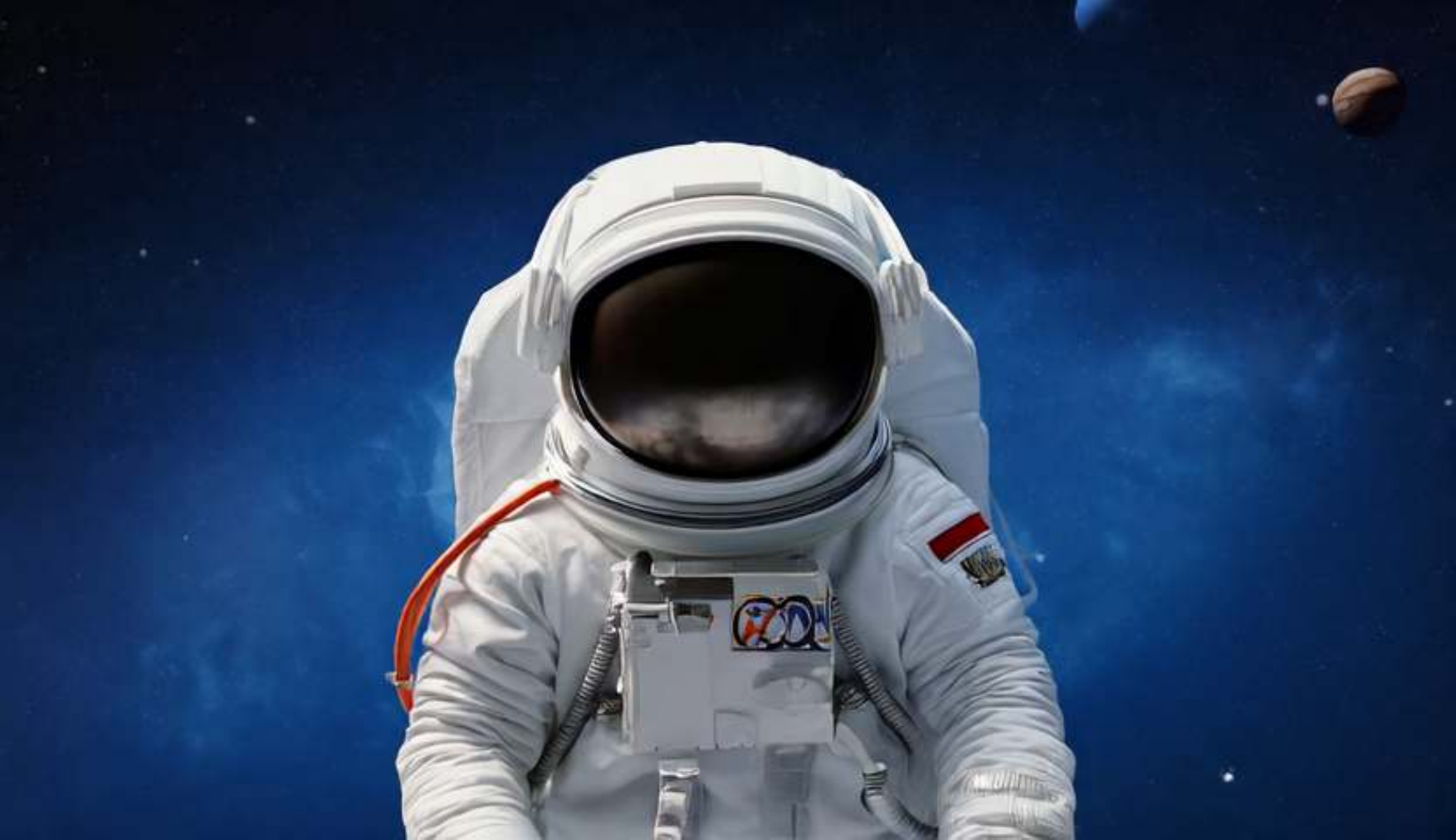}
        & \framecell{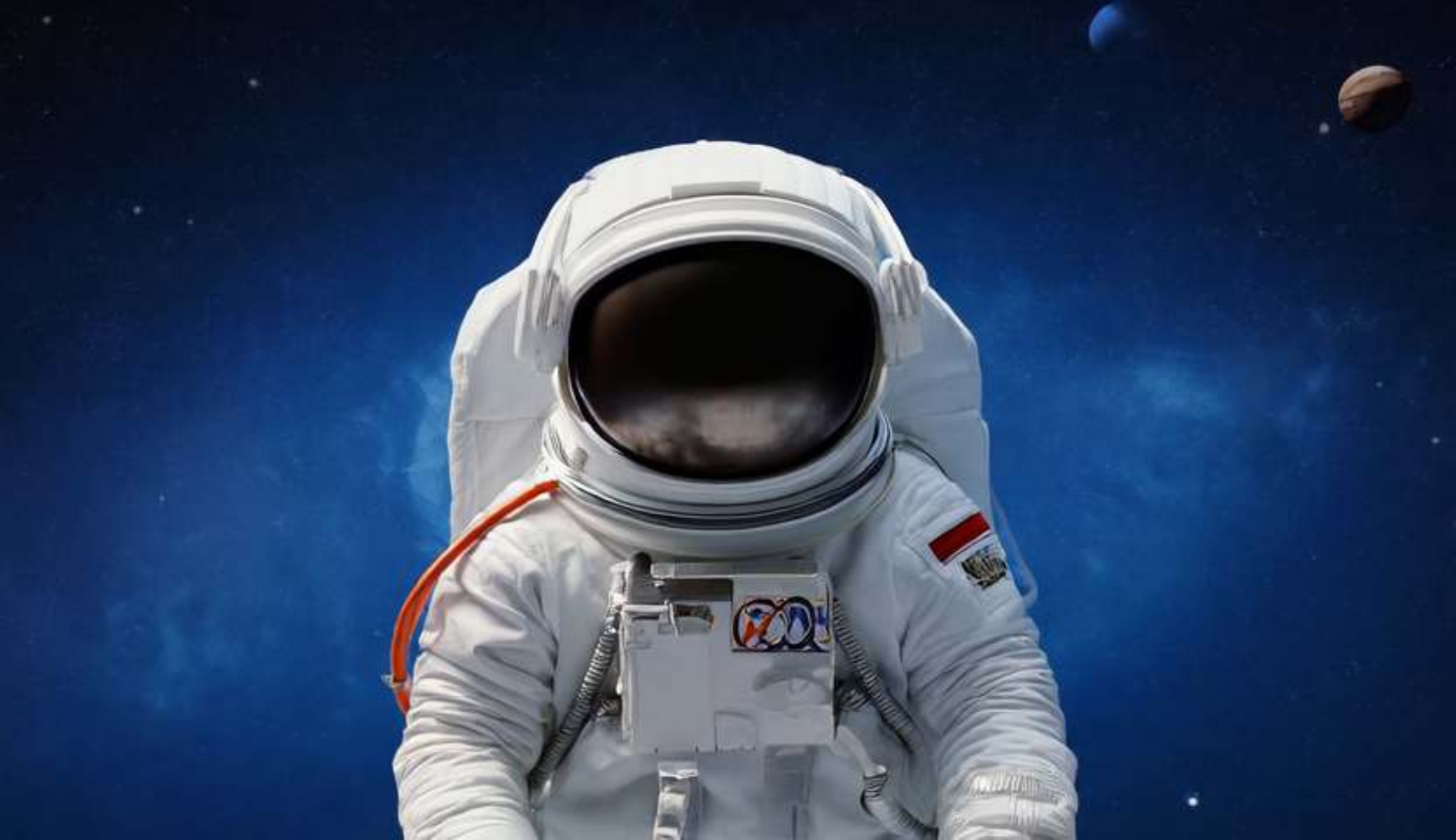}
        & \framecell{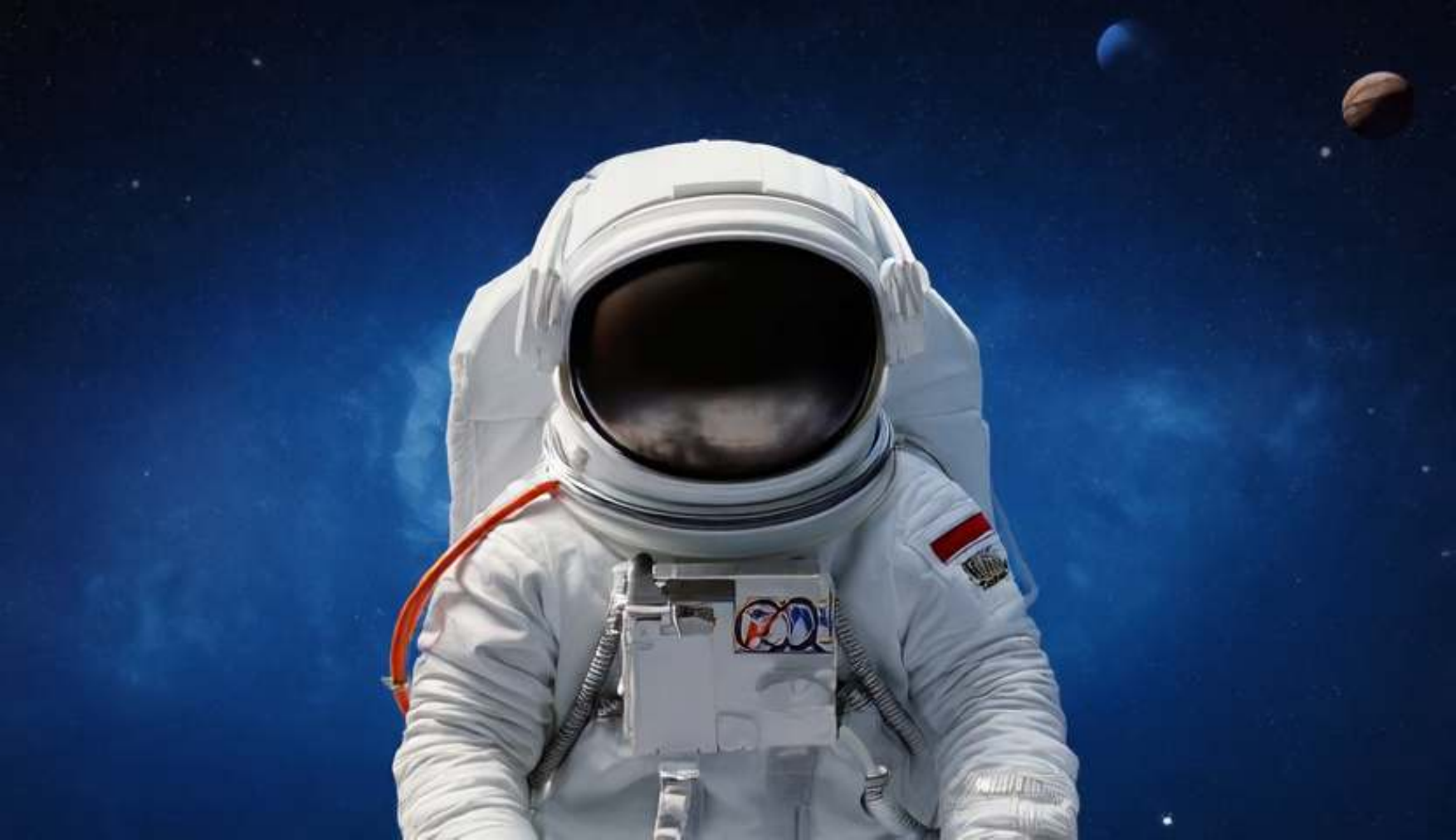}
        \\ \rowgap

        \methodcell{DiCache ($2.55\times$)}
        & \framecell{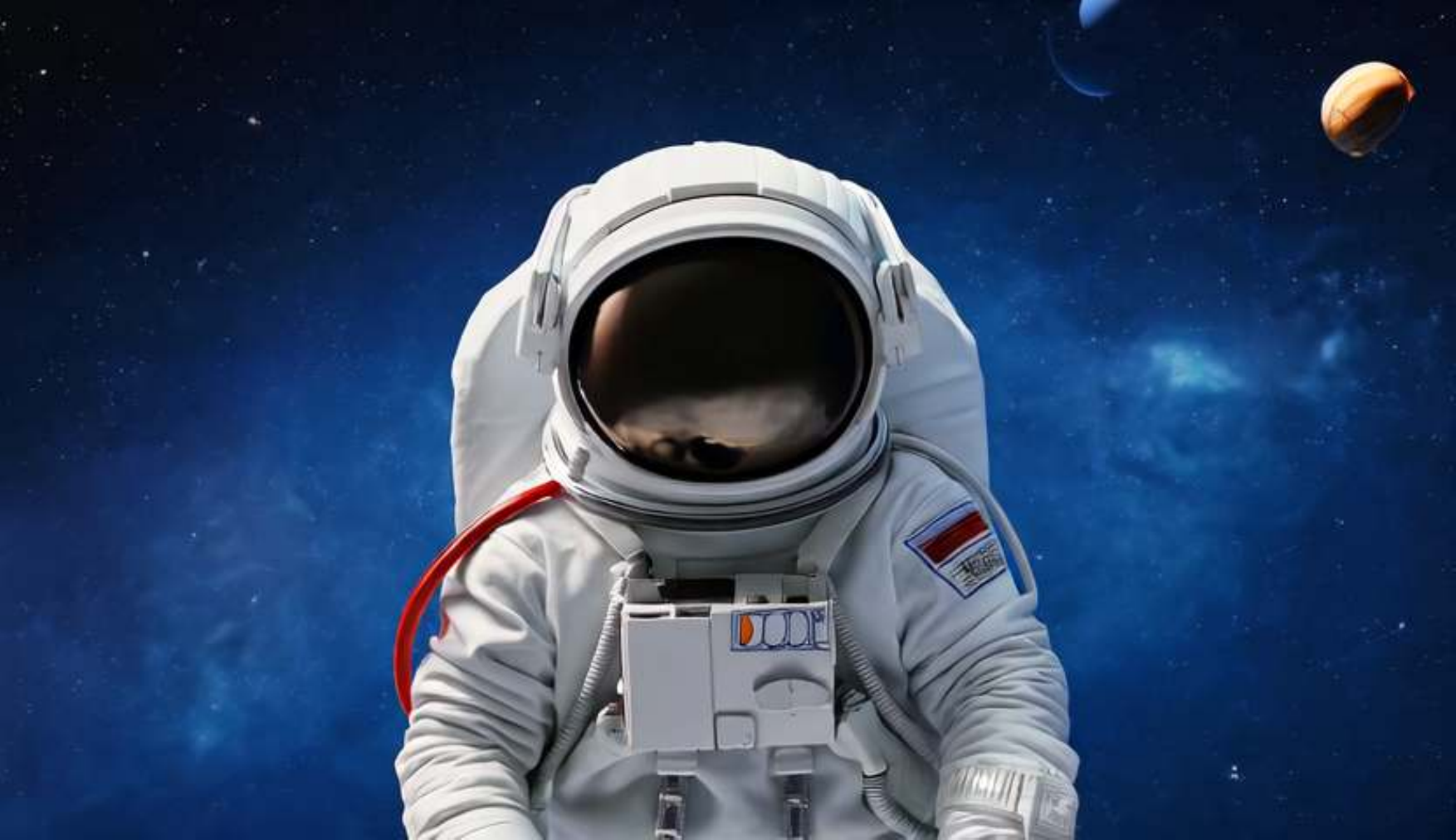}
        & \framecell{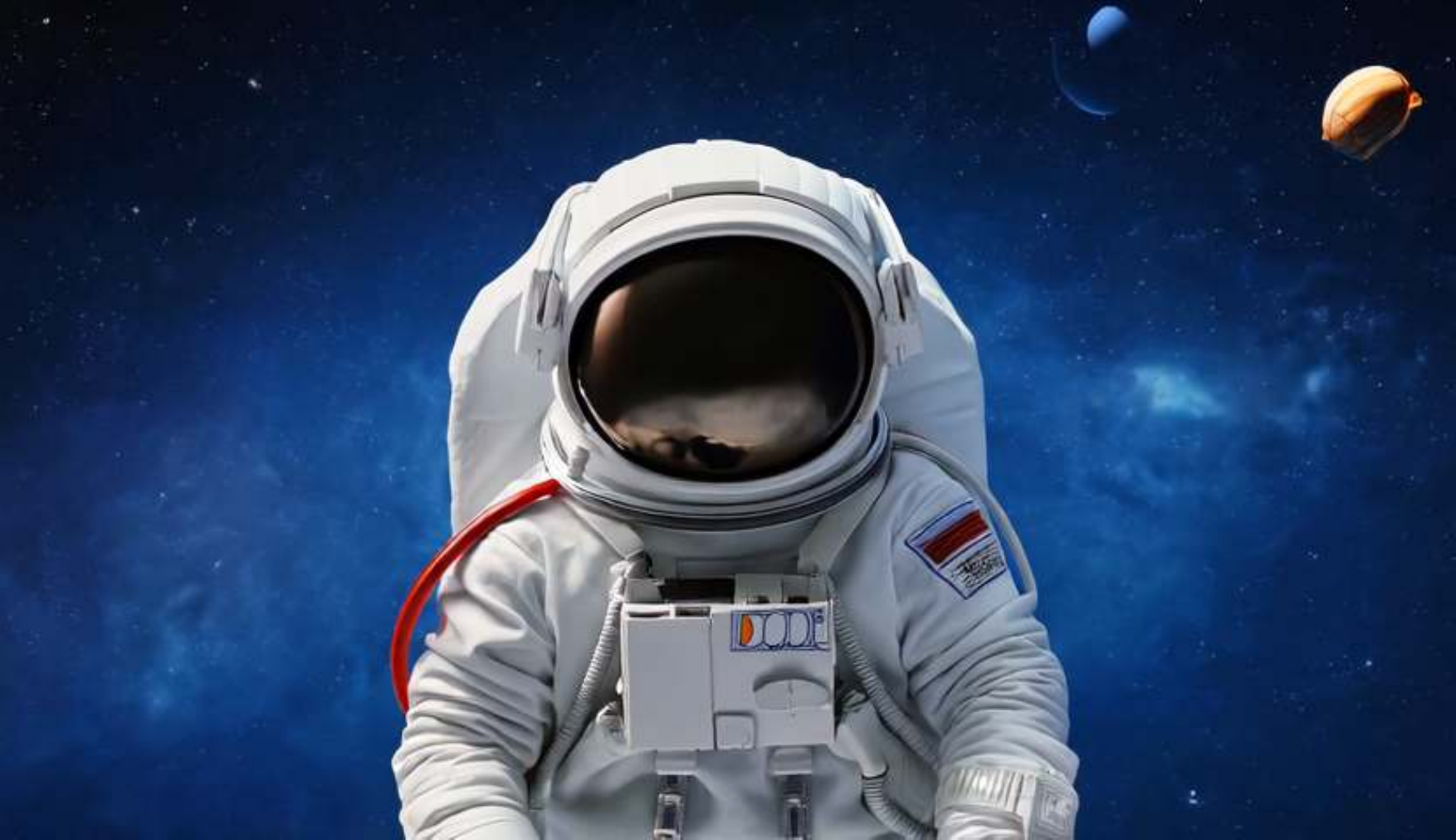}
        & \framecell{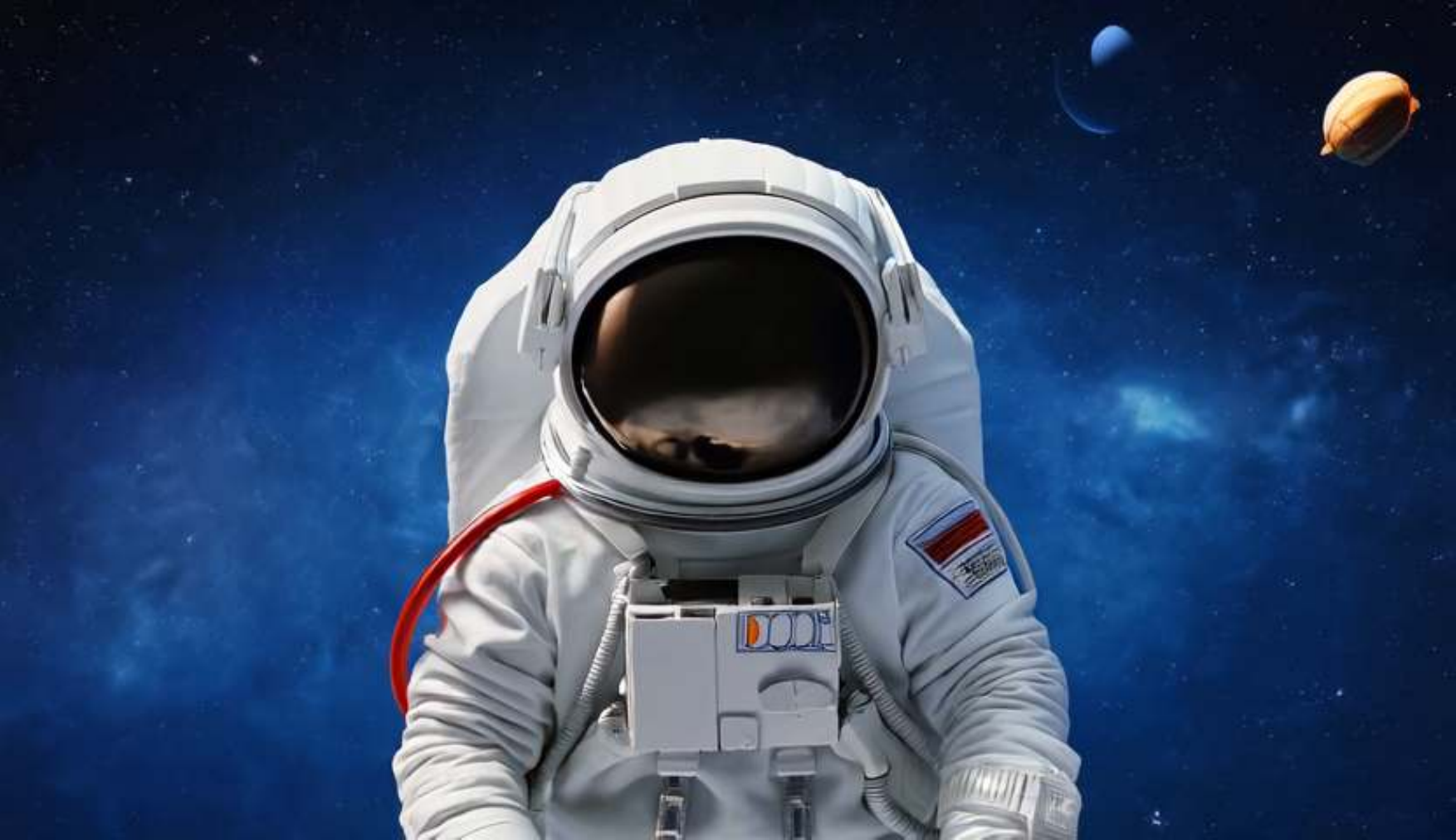}
        \\ \rowgap

        \methodcell{Ours ($2.56\times$)}
        & \framecell{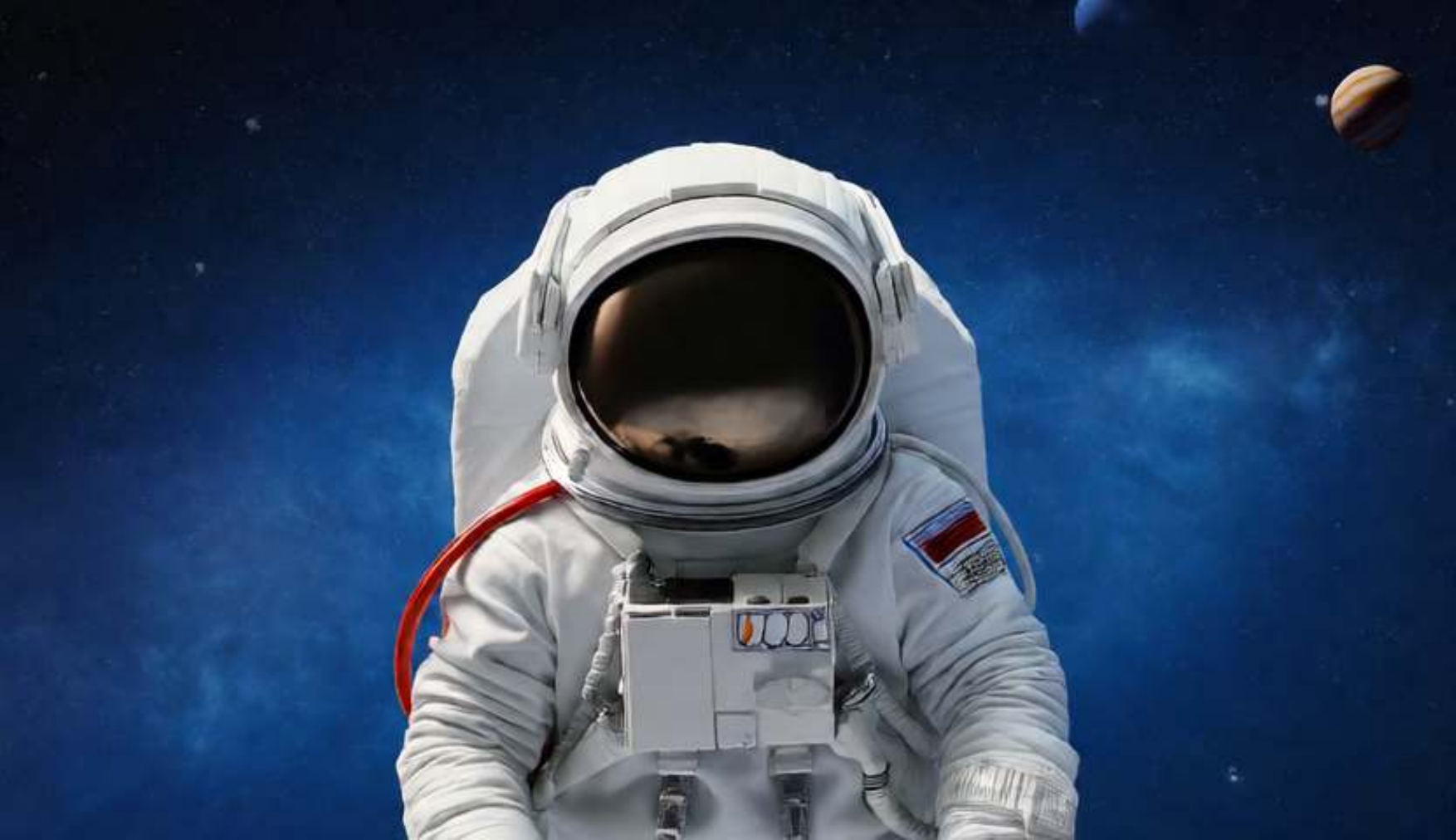}
        & \framecell{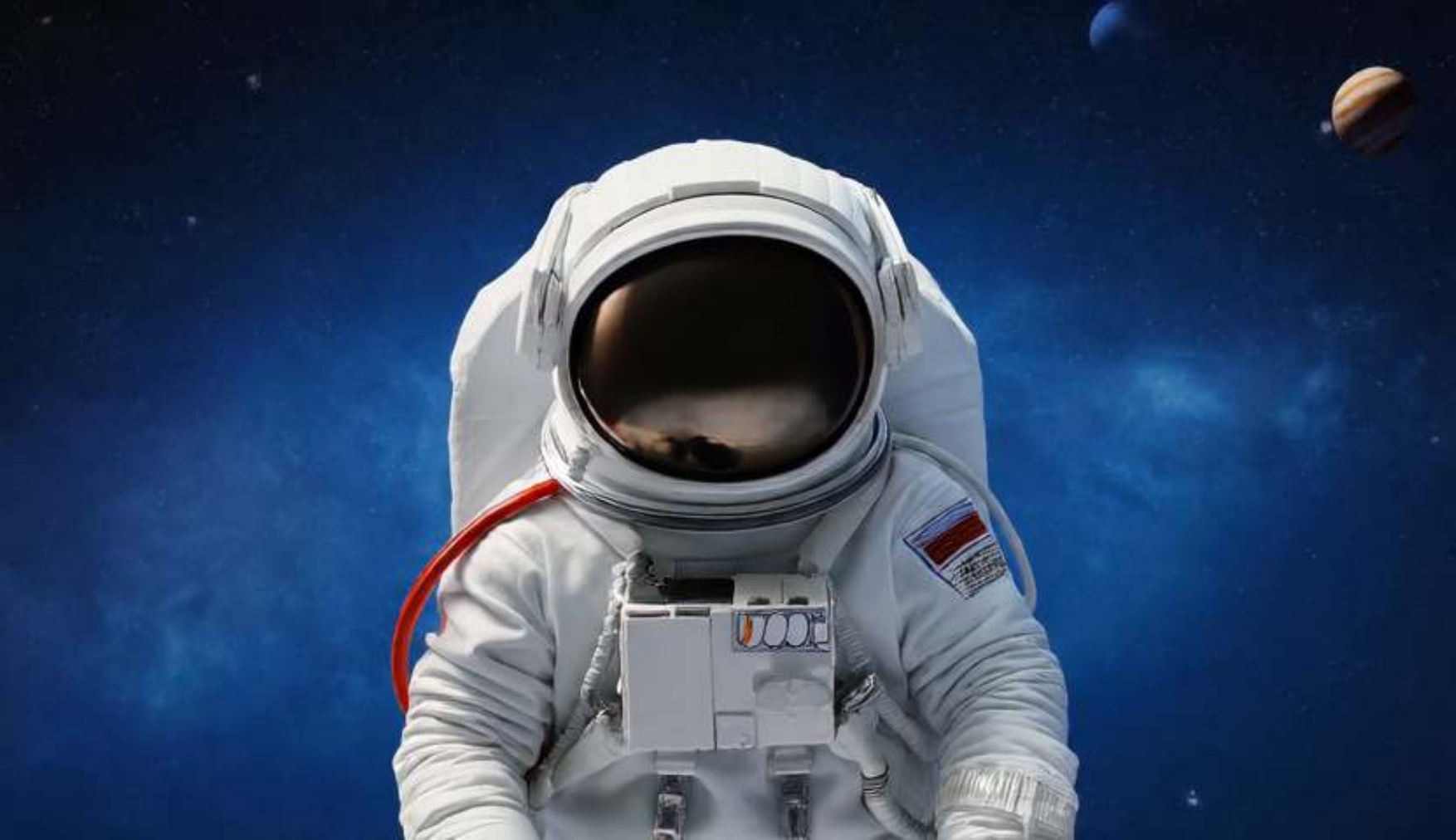}
        & \framecell{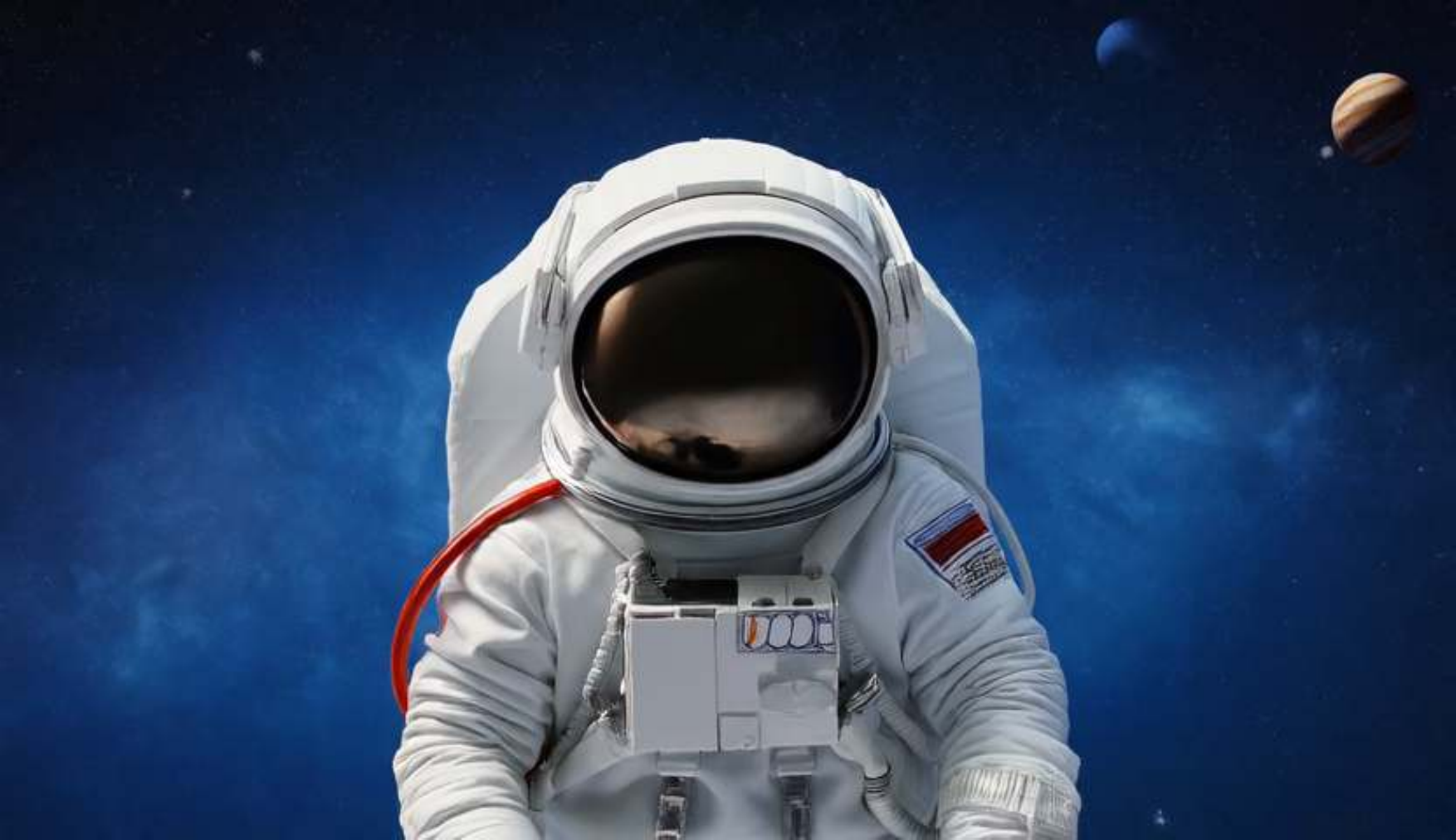}
        \\
    \end{tabular}
    \par\vspace{0.8em}
    \begin{tabular}{@{}c@{\hspace{1.5pt}}ccc@{}}
        & \multicolumn{3}{c}{\large \textbf{A motorcycle turning a corner.}}
        \\[0.25em]

        \methodcell{Original}
        & \framecell{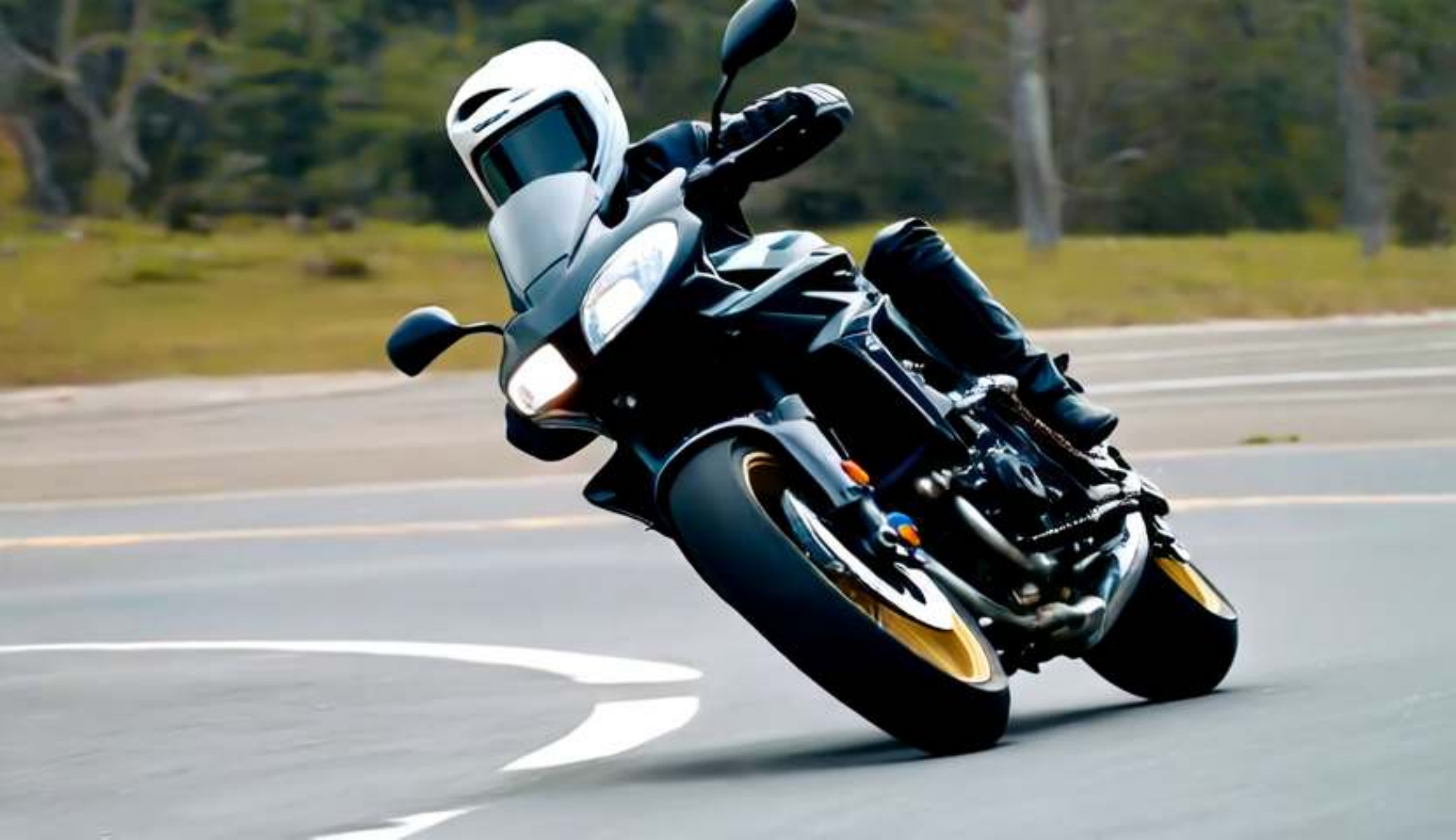}
        & \framecell{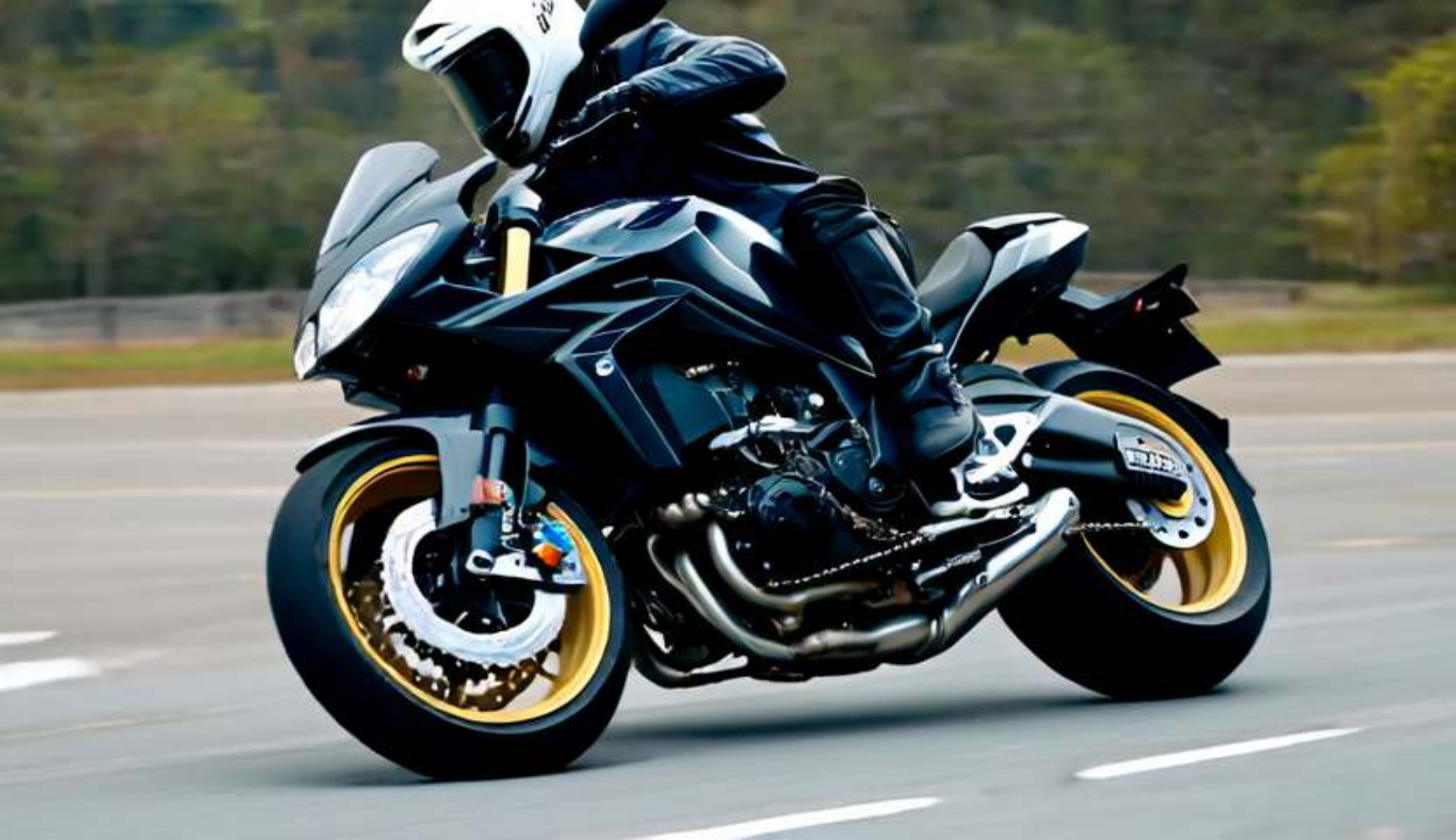}
        & \framecell{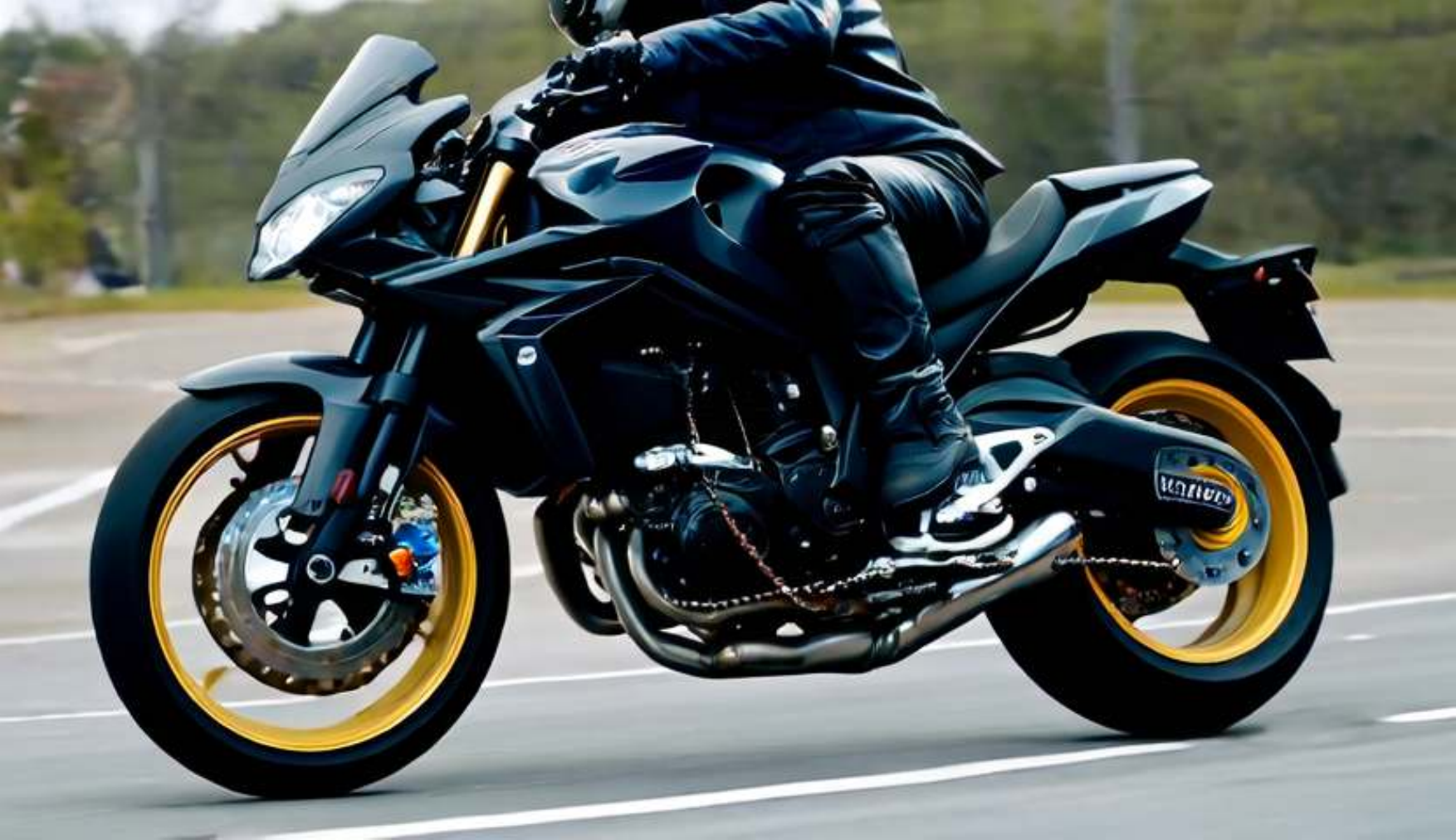}
        \\ \rowgap

        \methodcell{MagCache ($2.54\times$)}
        & \framecell{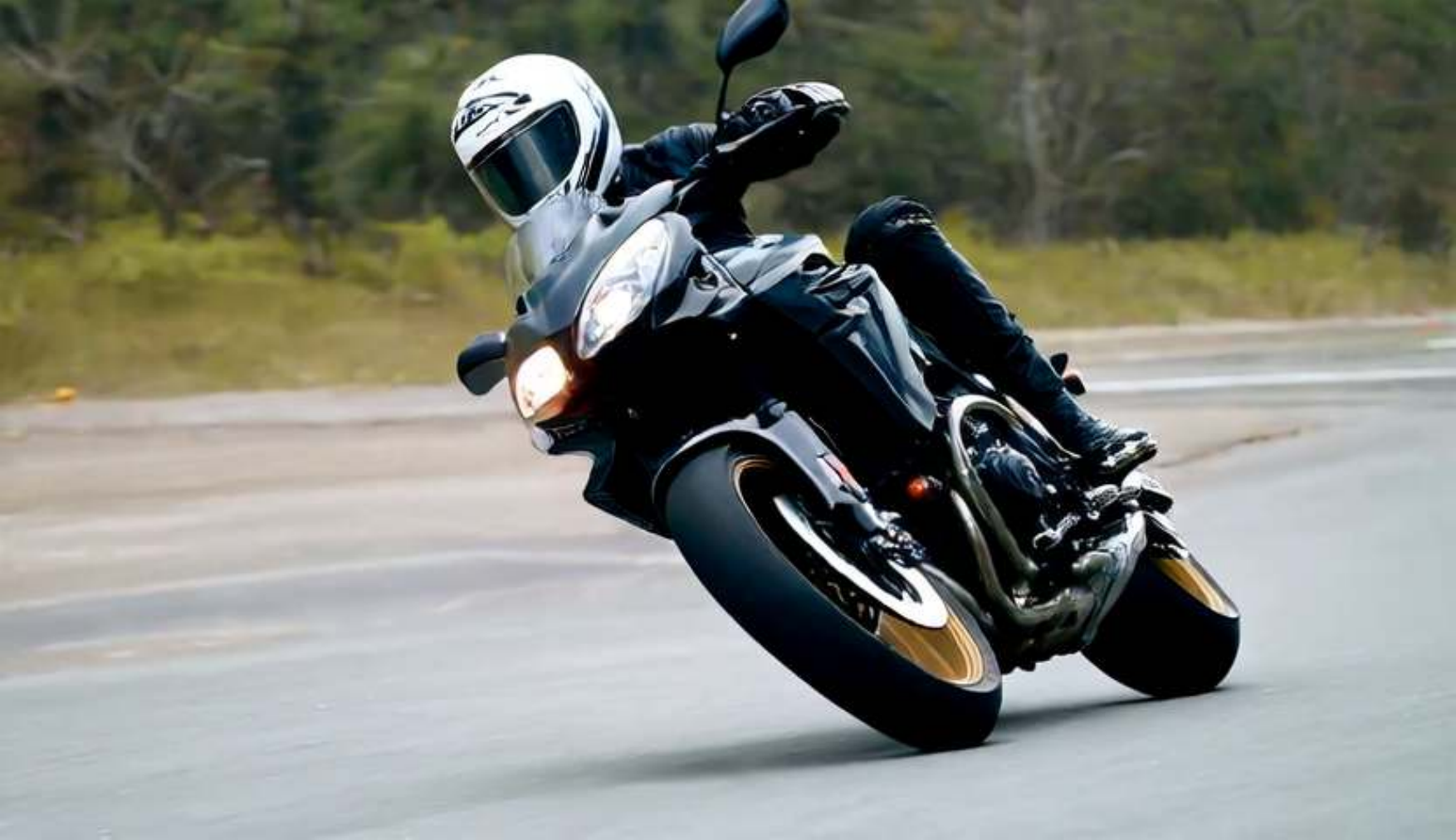}
        & \framecell{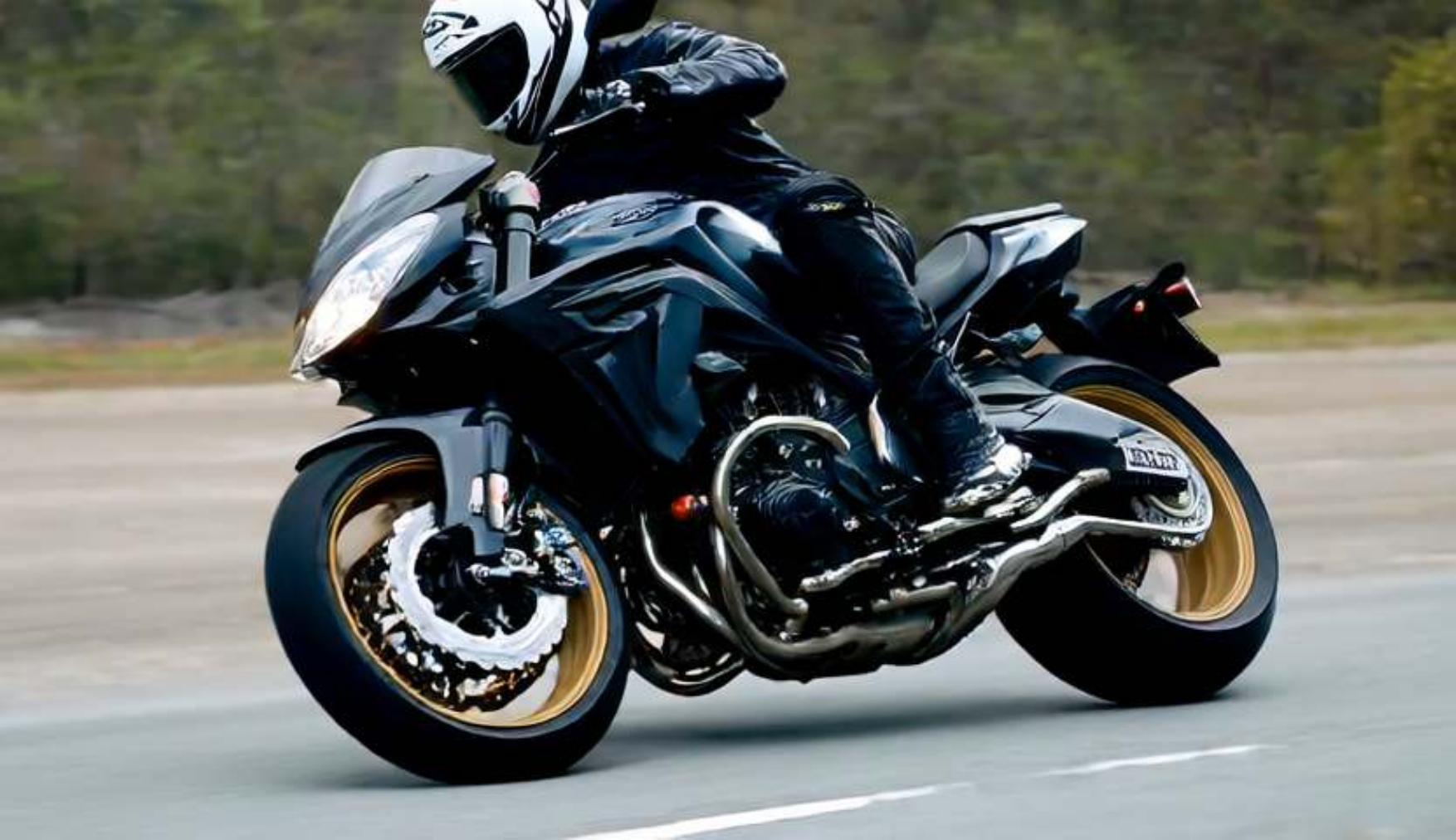}
        & \framecell{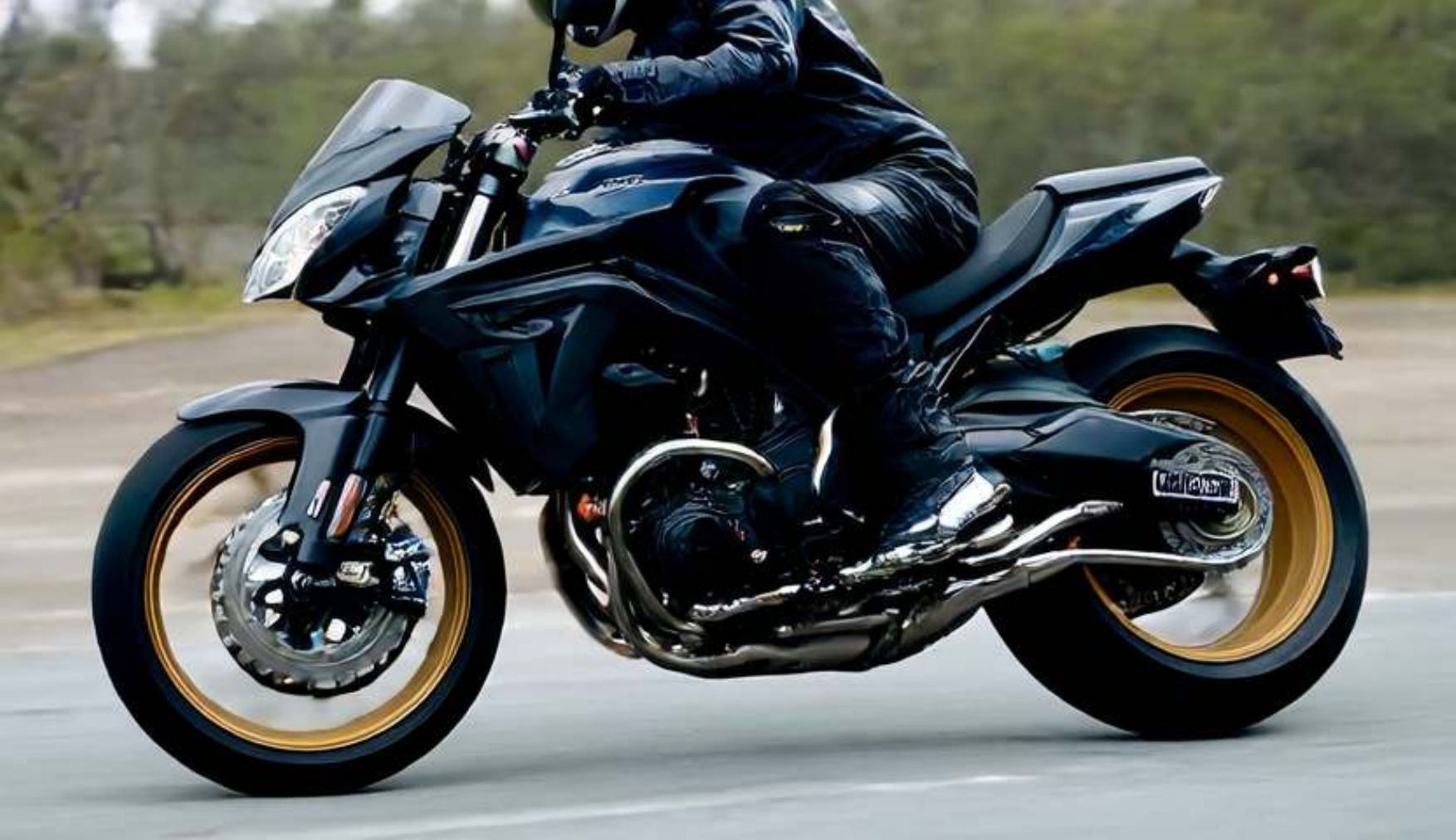}
        \\ \rowgap

        \methodcell{DiCache ($2.55\times$)}
        & \framecell{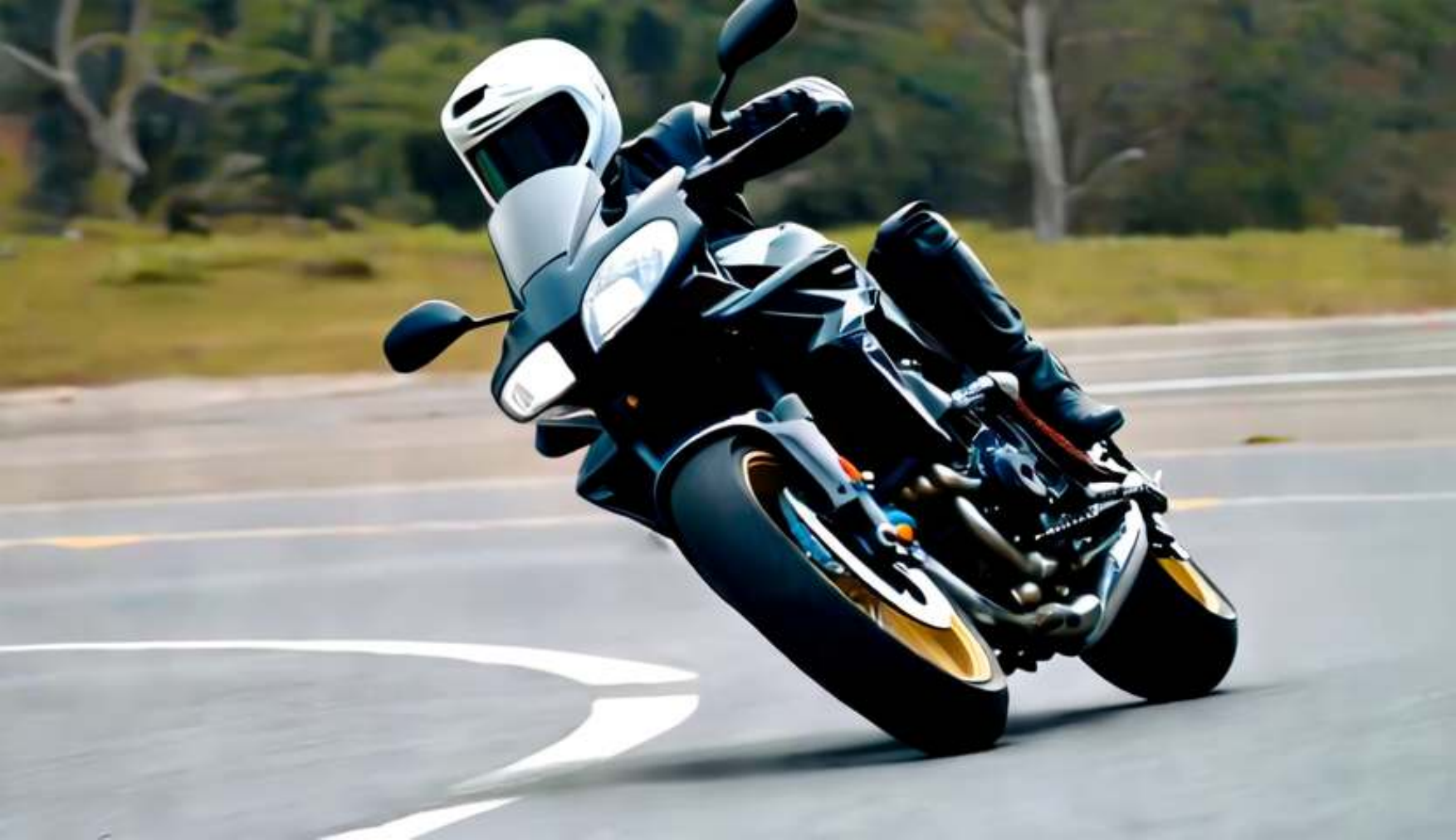}
        & \framecell{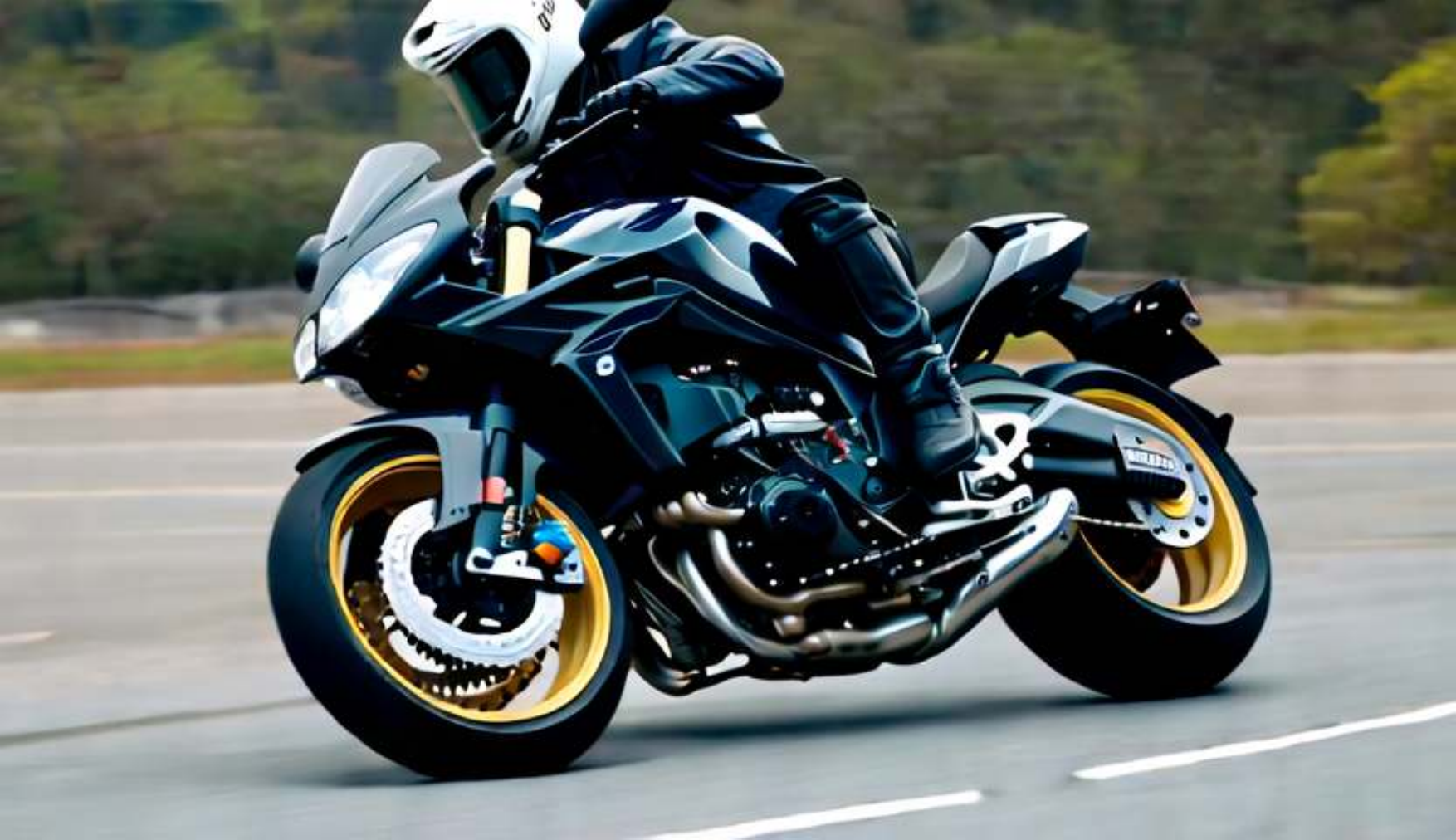}
        & \framecell{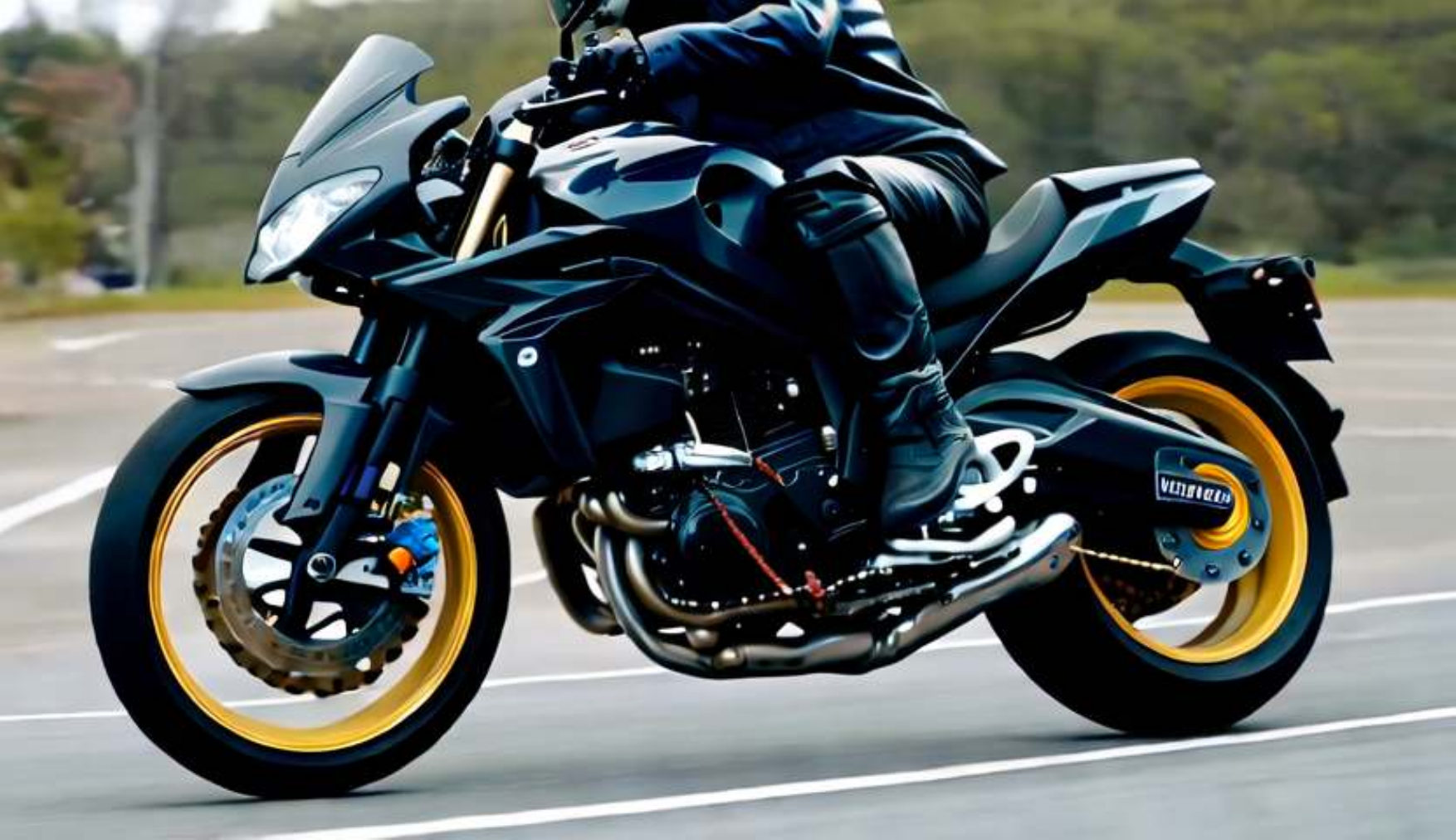}
        \\ \rowgap

        \methodcell{Ours ($2.56\times$)}
        & \framecell{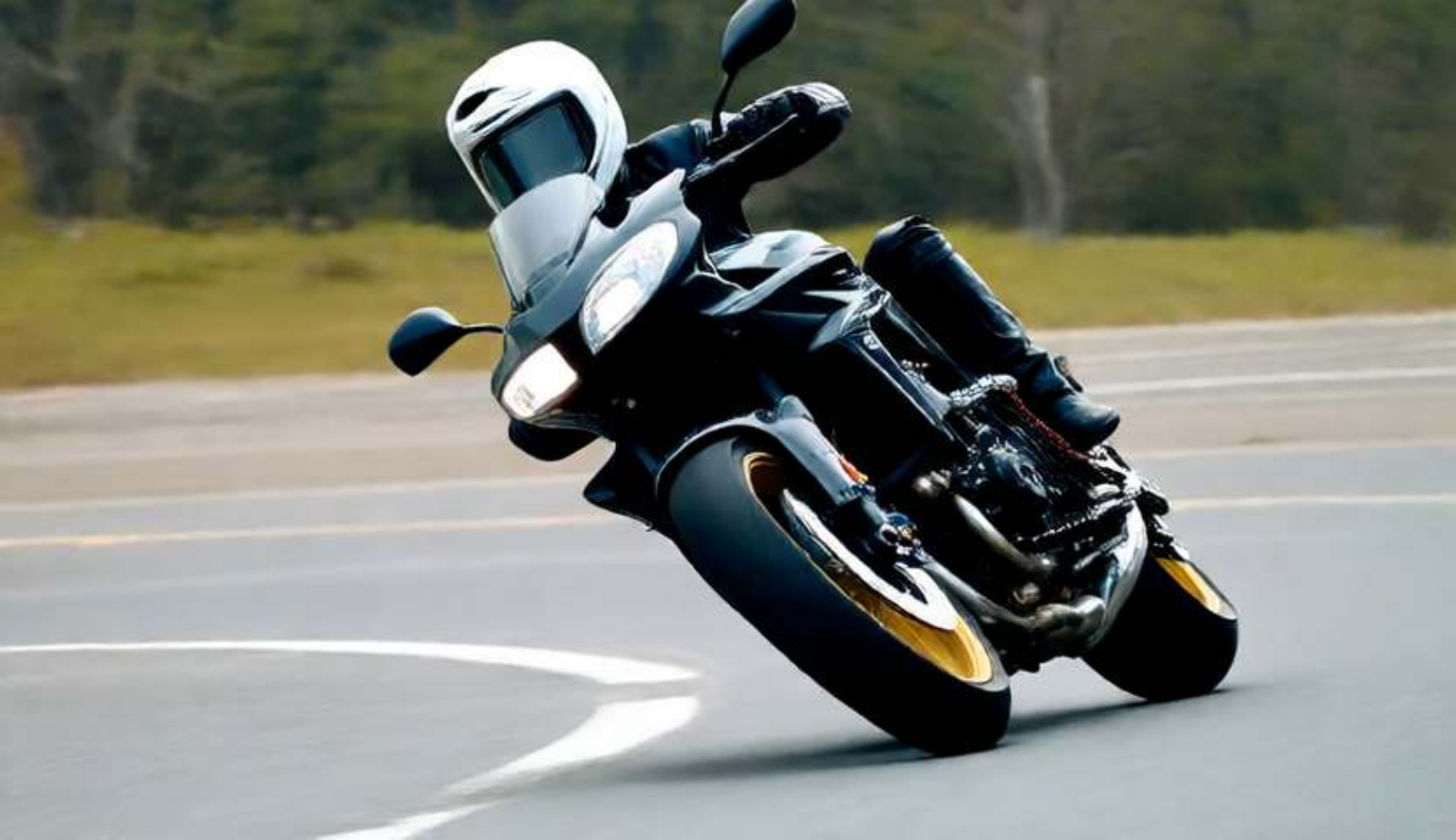}
        & \framecell{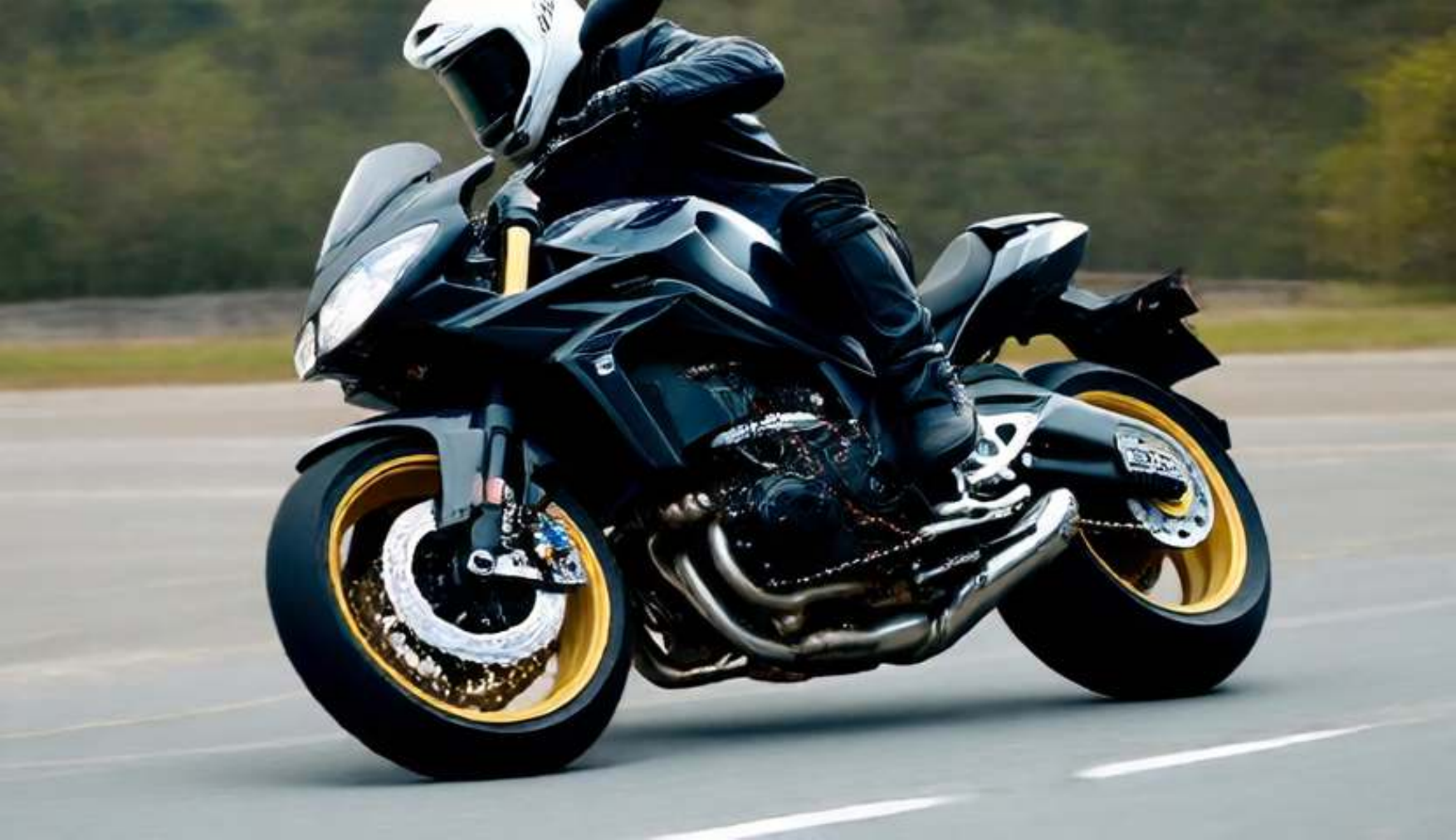}
        & \framecell{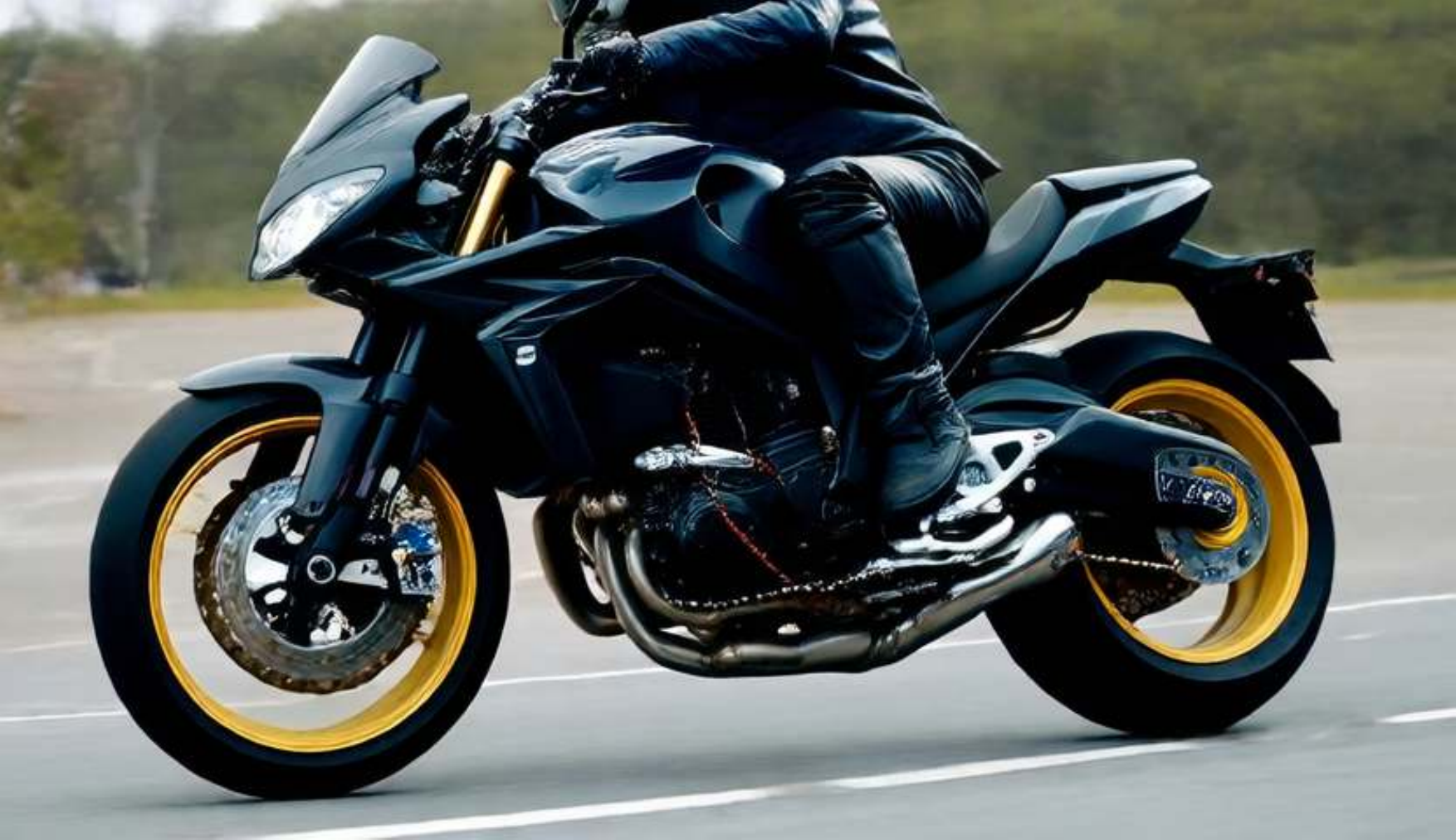}
        \\
    \end{tabular}
    \vspace{-0.3em}
    \caption{
    \textbf{Qualitative comparison on Wan2.1-T2V-1.3B (Part I).}
    }
    \label{fig:qualitative_cogvideox}
    \vspace{-0.8em}
\end{figure*}
% figure-local macros
\ifdefined\cogw\else\newlength{\cogw}\fi
\ifdefined\cogh\else\newlength{\cogh}\fi
\setlength{\cogw}{0.302\textwidth}
\setlength{\cogh}{0.170\textwidth}

\providecommand{\vcell}[1]{\ensuremath{\vcenter{\hbox{#1}}}}
\providecommand{\methodcell}[1]{%
    \vcell{\rotatebox[origin=c]{90}{\makebox[\cogh][c]{\textbf{#1}}}}%
}
\providecommand{\framecell}[1]{%
    \vcell{\includegraphics[width=\cogw,height=\cogh]{#1}}%
}
\providecommand{\rowgap}{\noalign{\vskip 3pt}}  % 控制行间距，只改这里

\begin{figure*}[t]
    \centering
    \setlength{\tabcolsep}{1.2pt}
    \renewcommand{\arraystretch}{0.0}
    \scriptsize

    \begin{tabular}{@{}c@{\hspace{1.5pt}}ccc@{}}
        & \multicolumn{3}{c}{\large \textbf{A train accelerating to gain speed.}}
        \\[0.25em]

        \methodcell{Original}
        & \framecell{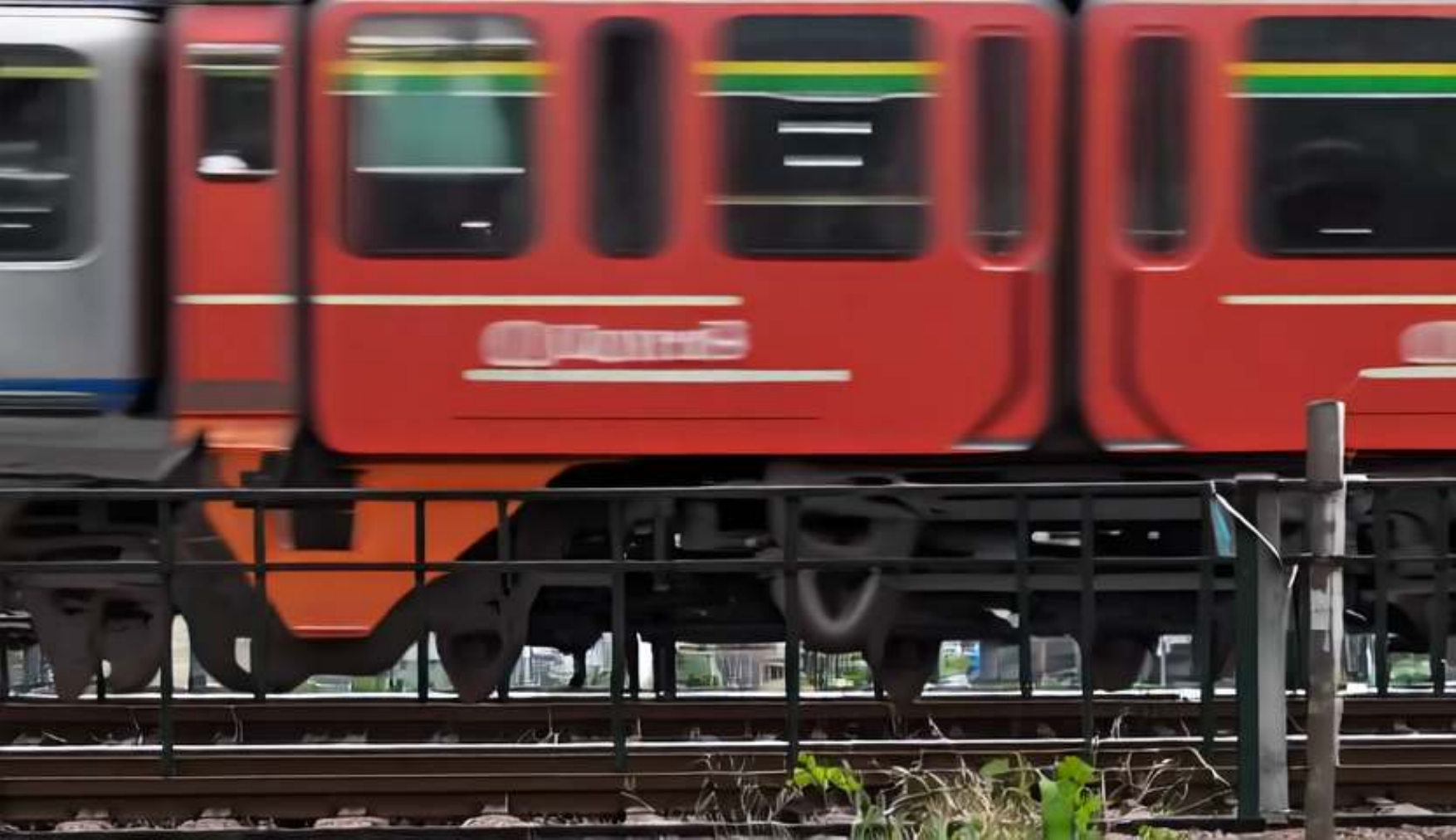}
        & \framecell{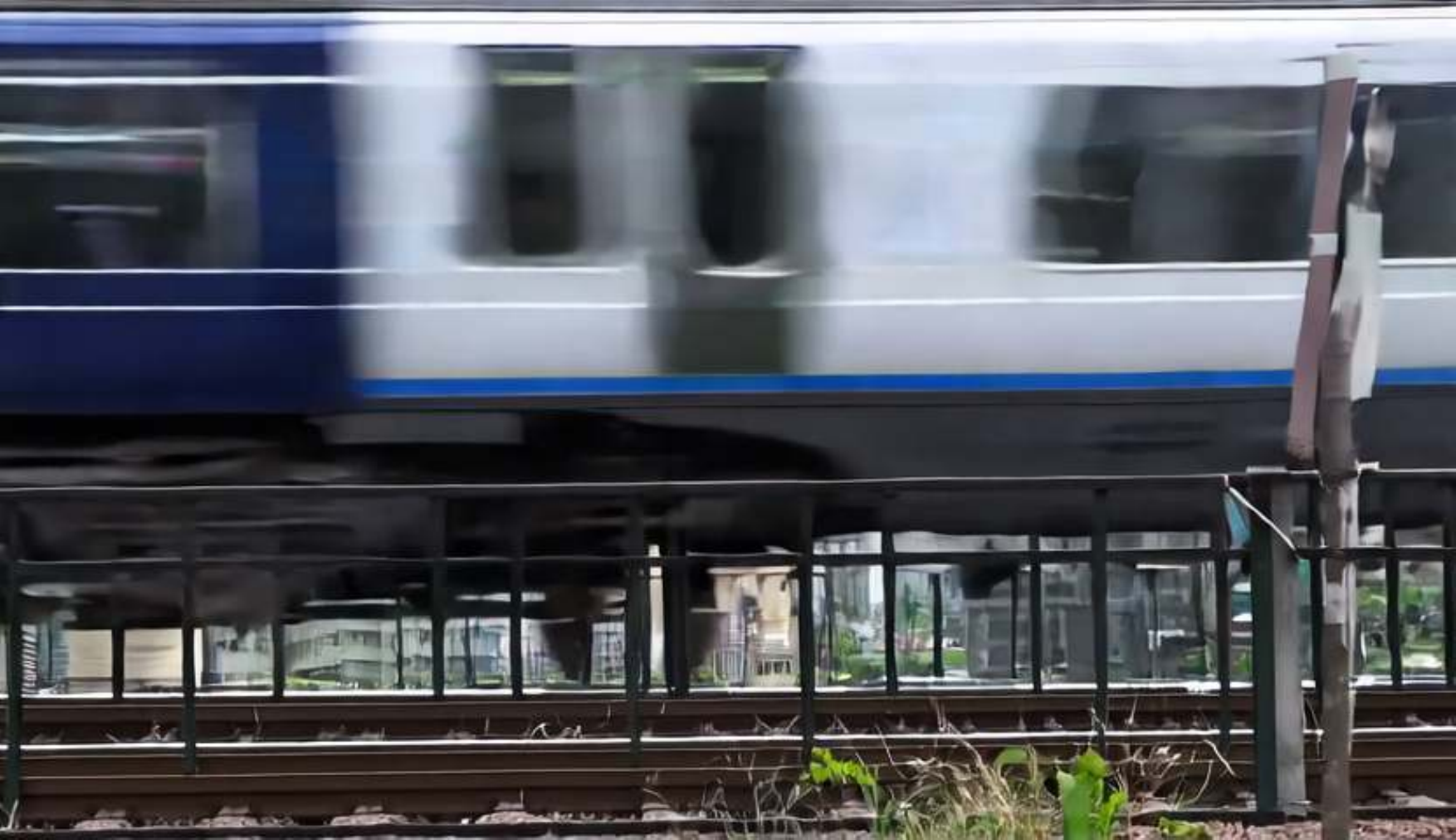}
        & \framecell{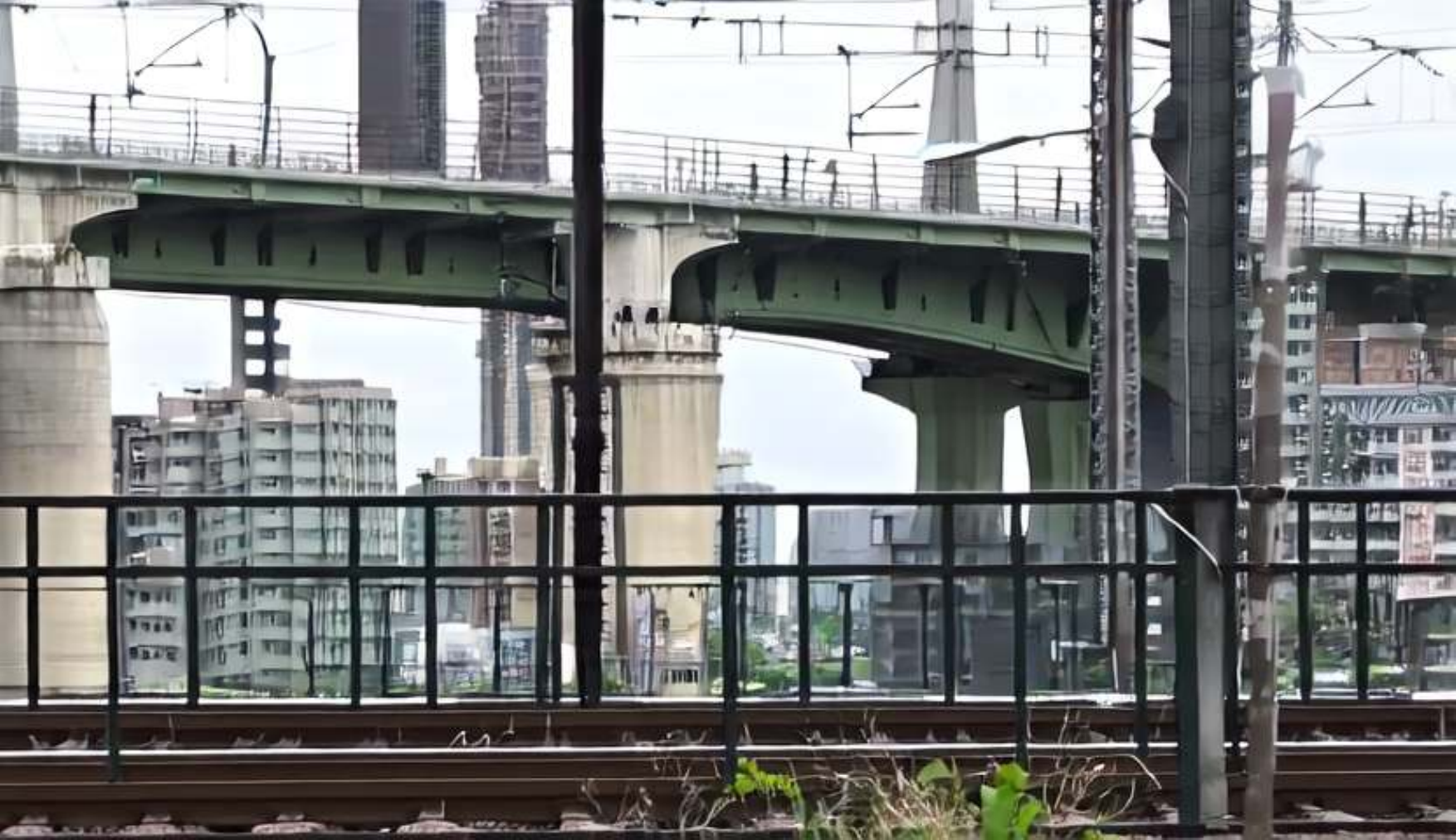}
        \\ \rowgap

        \methodcell{MagCache ($2.54\times$)}
        & \framecell{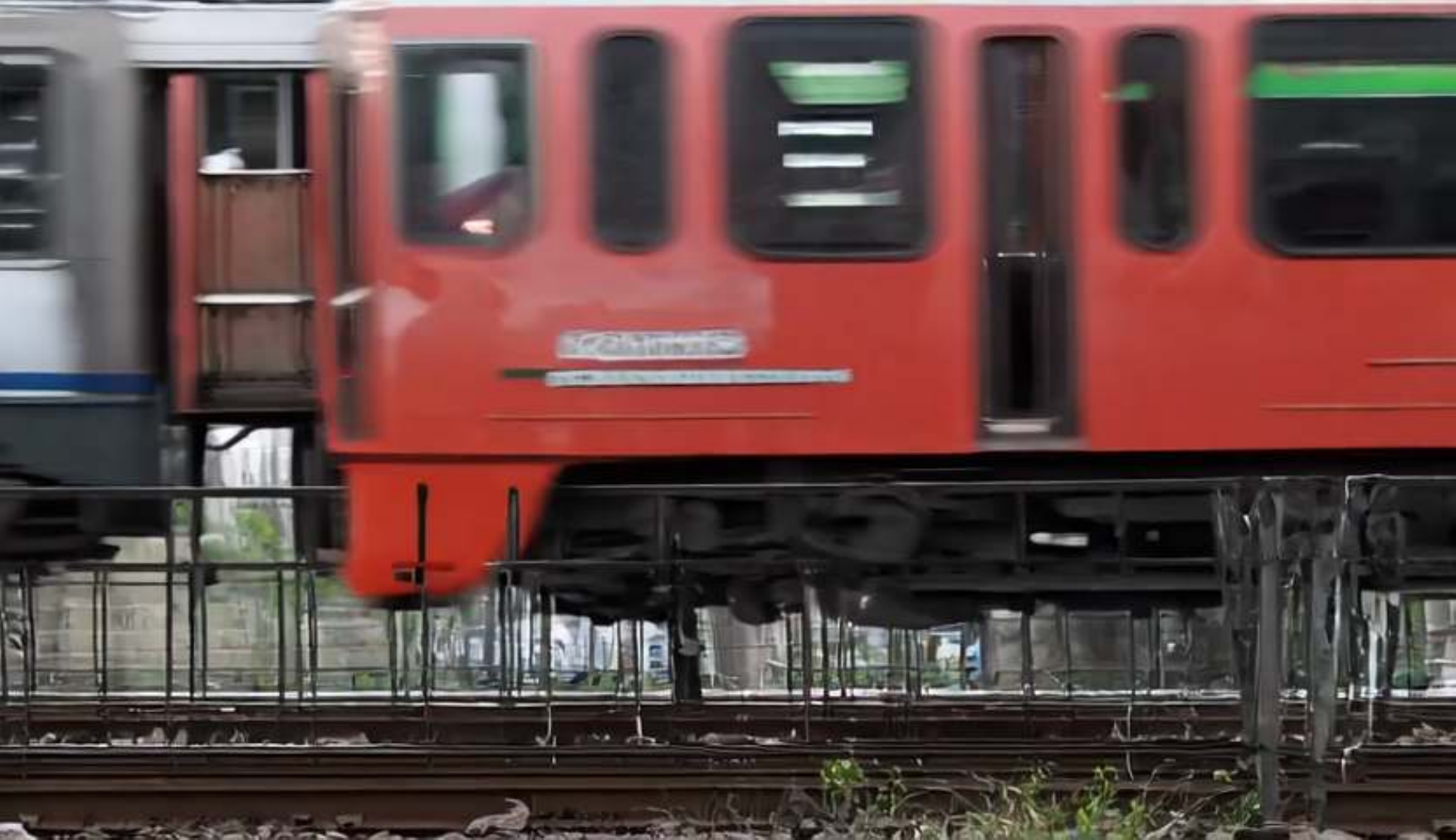}
        & \framecell{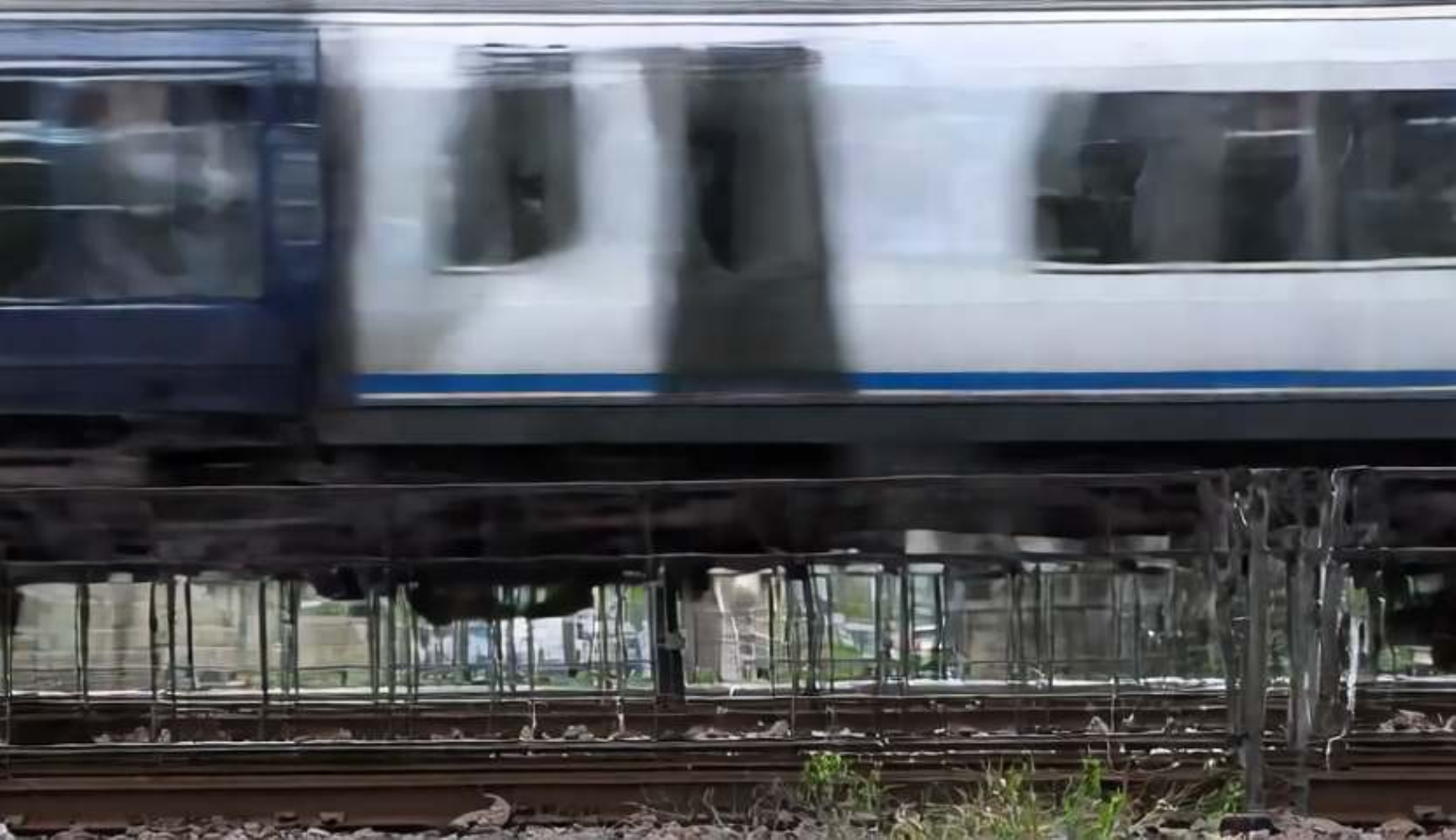}
        & \framecell{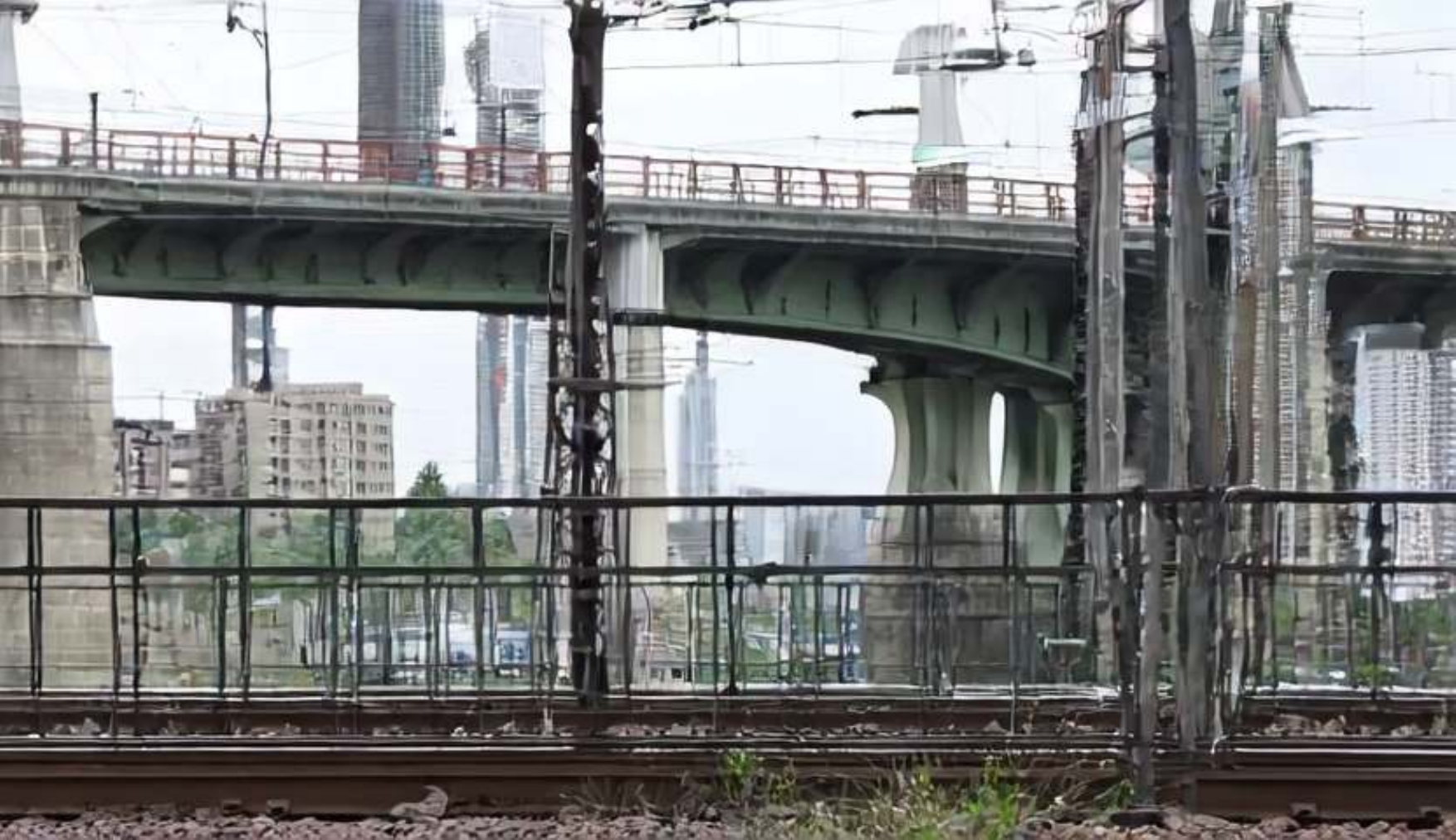}
        \\ \rowgap

        \methodcell{DiCache ($2.55\times$)}
        & \framecell{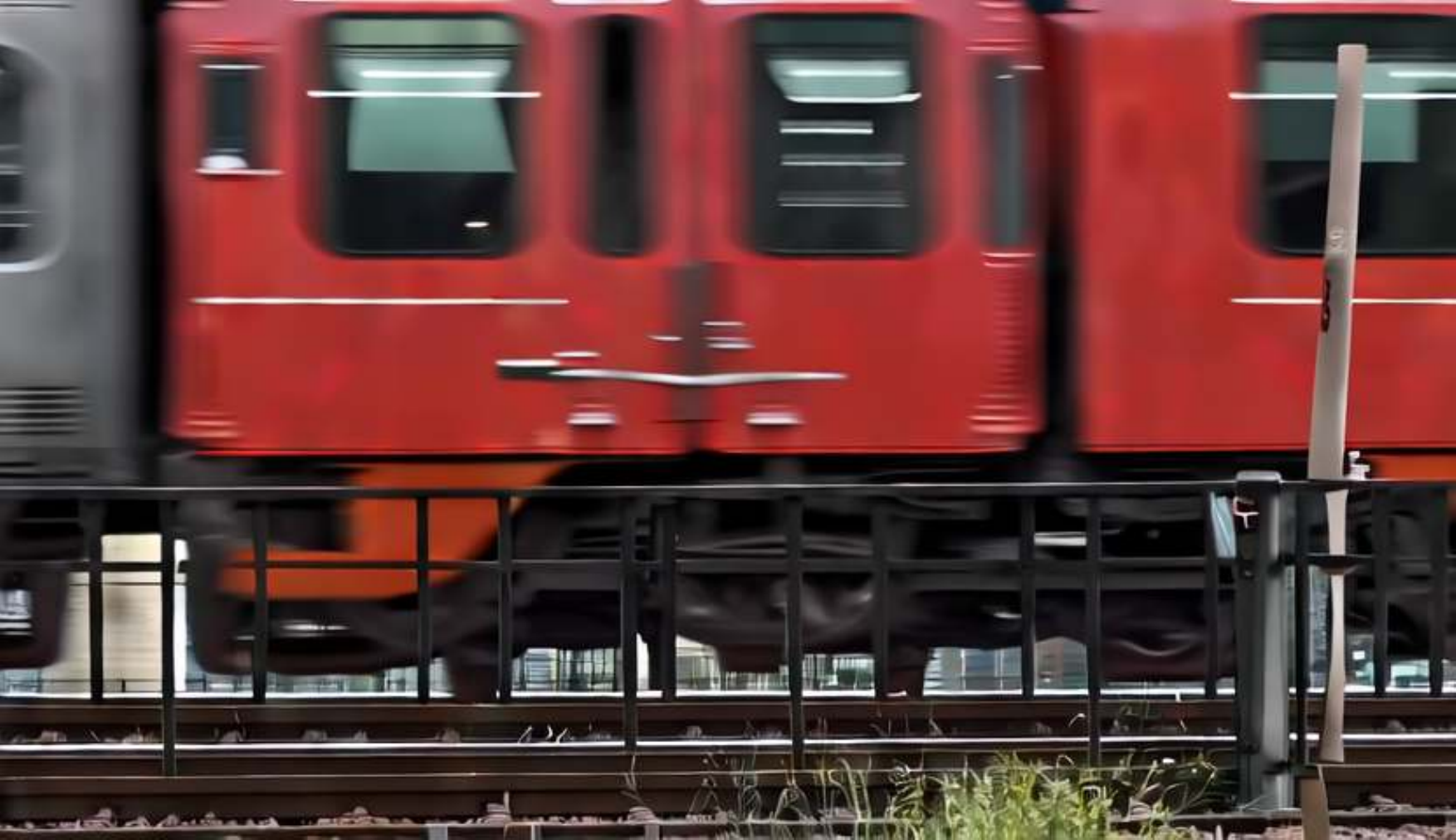}
        & \framecell{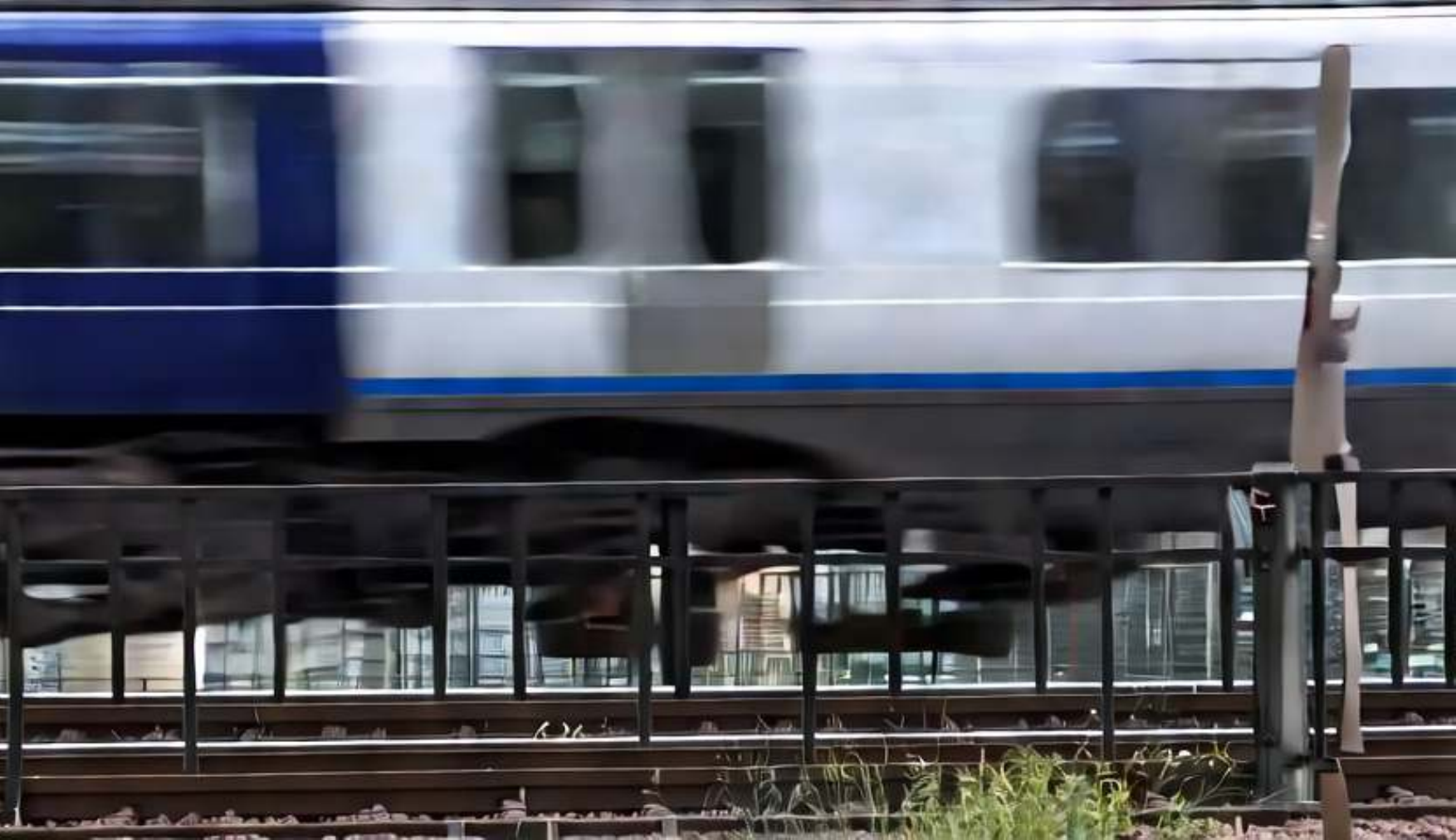}
        & \framecell{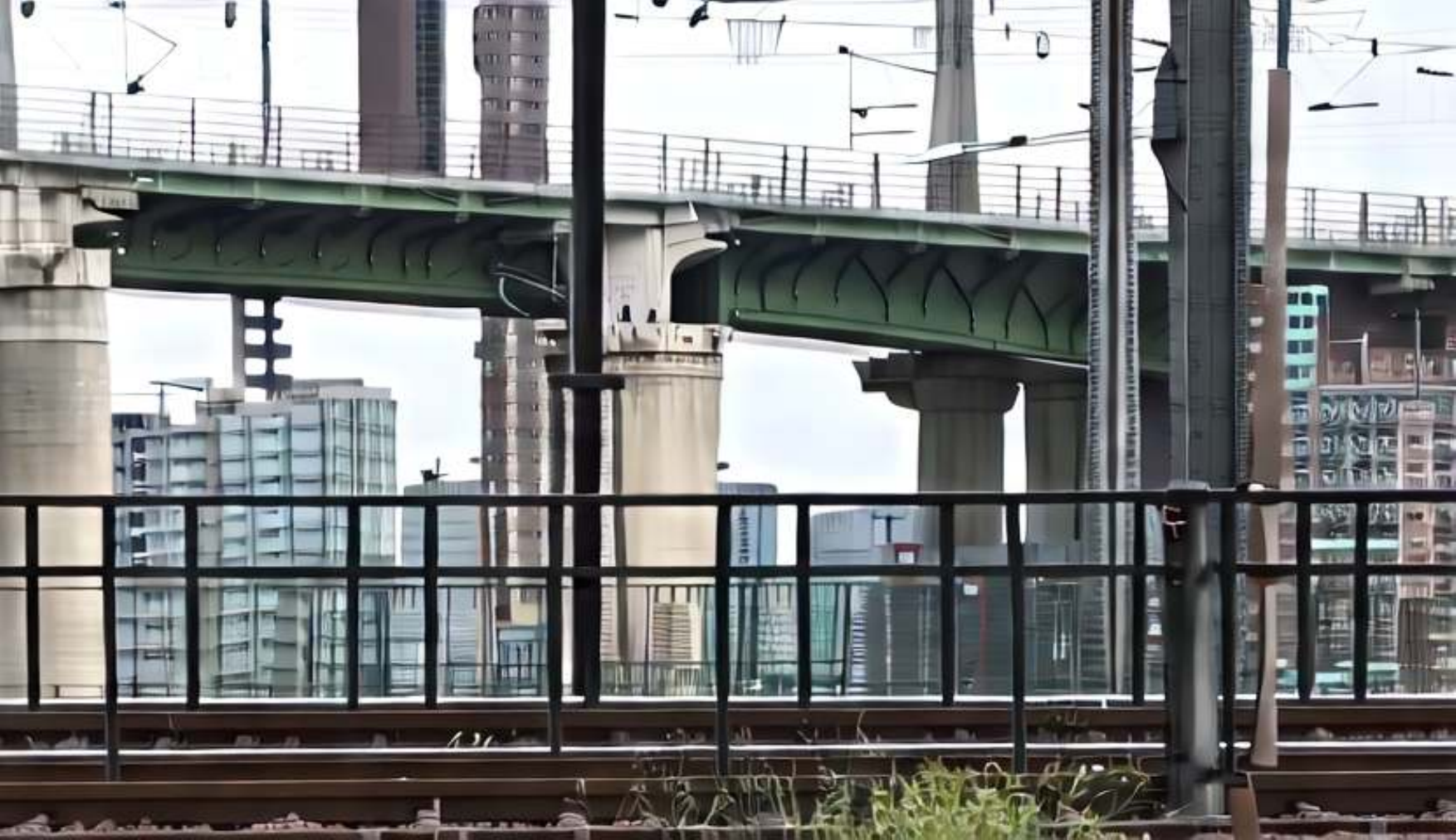}
        \\ \rowgap

        \methodcell{Ours ($2.56\times$)}
        & \framecell{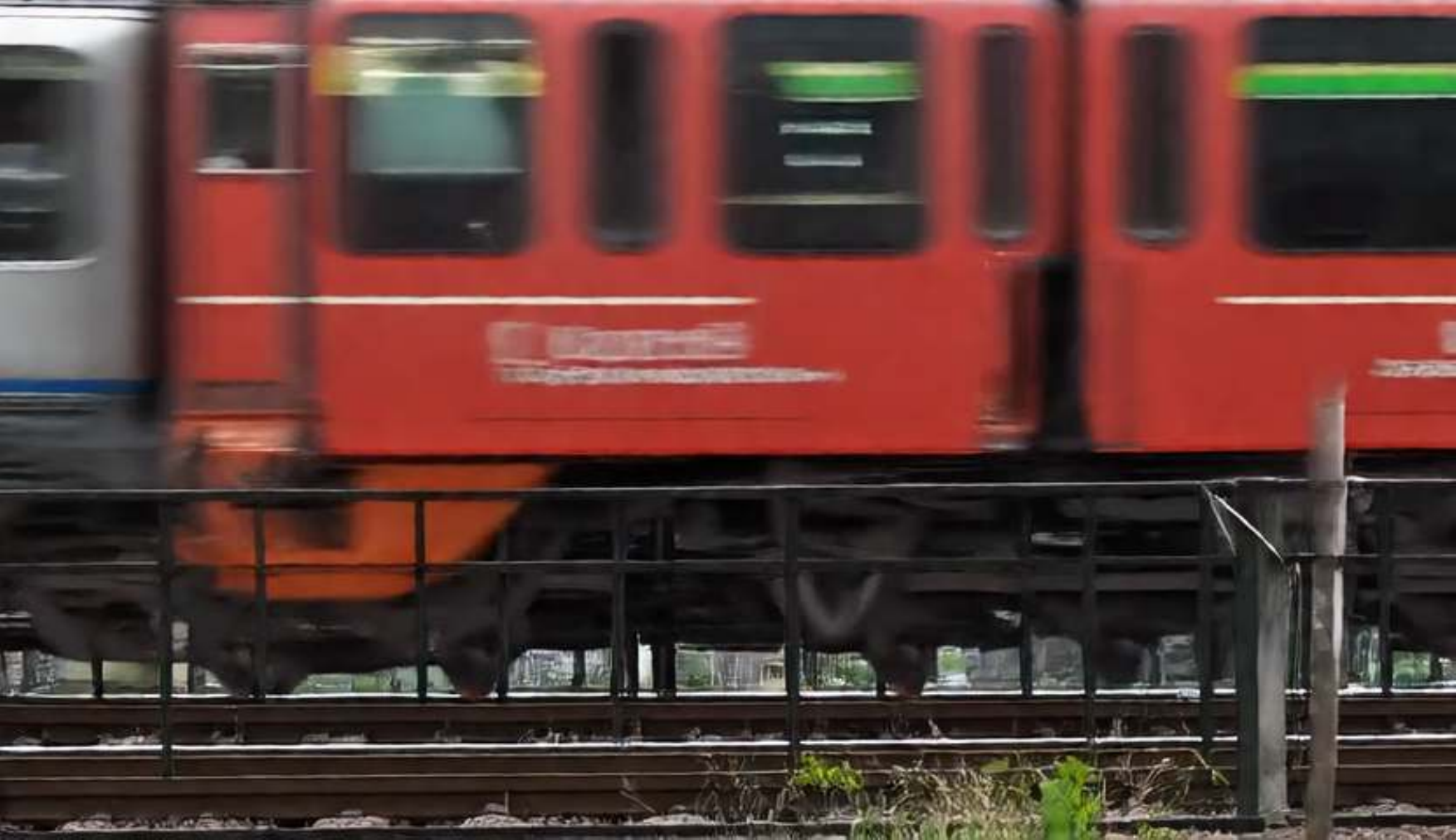}
        & \framecell{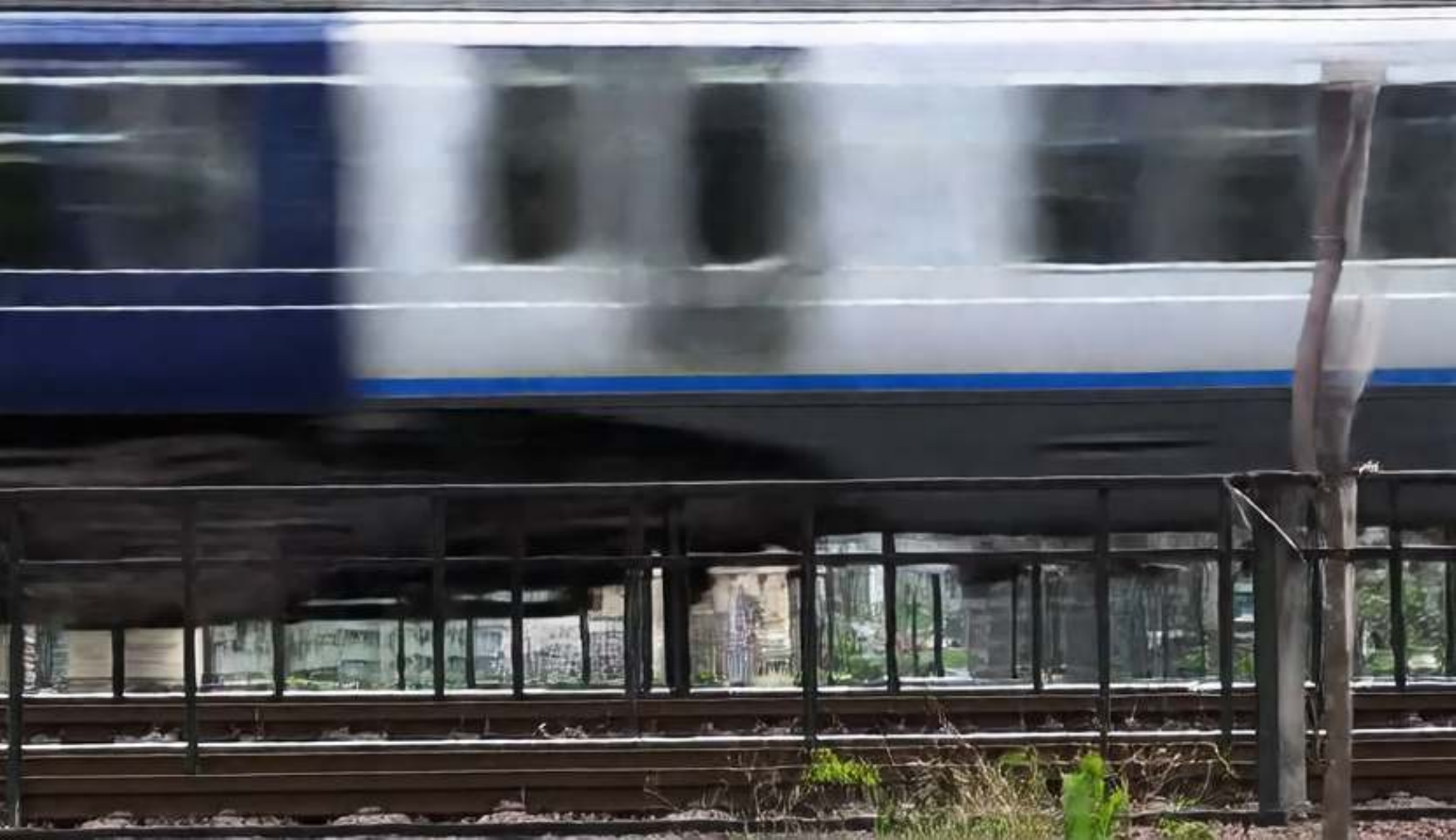}
        & \framecell{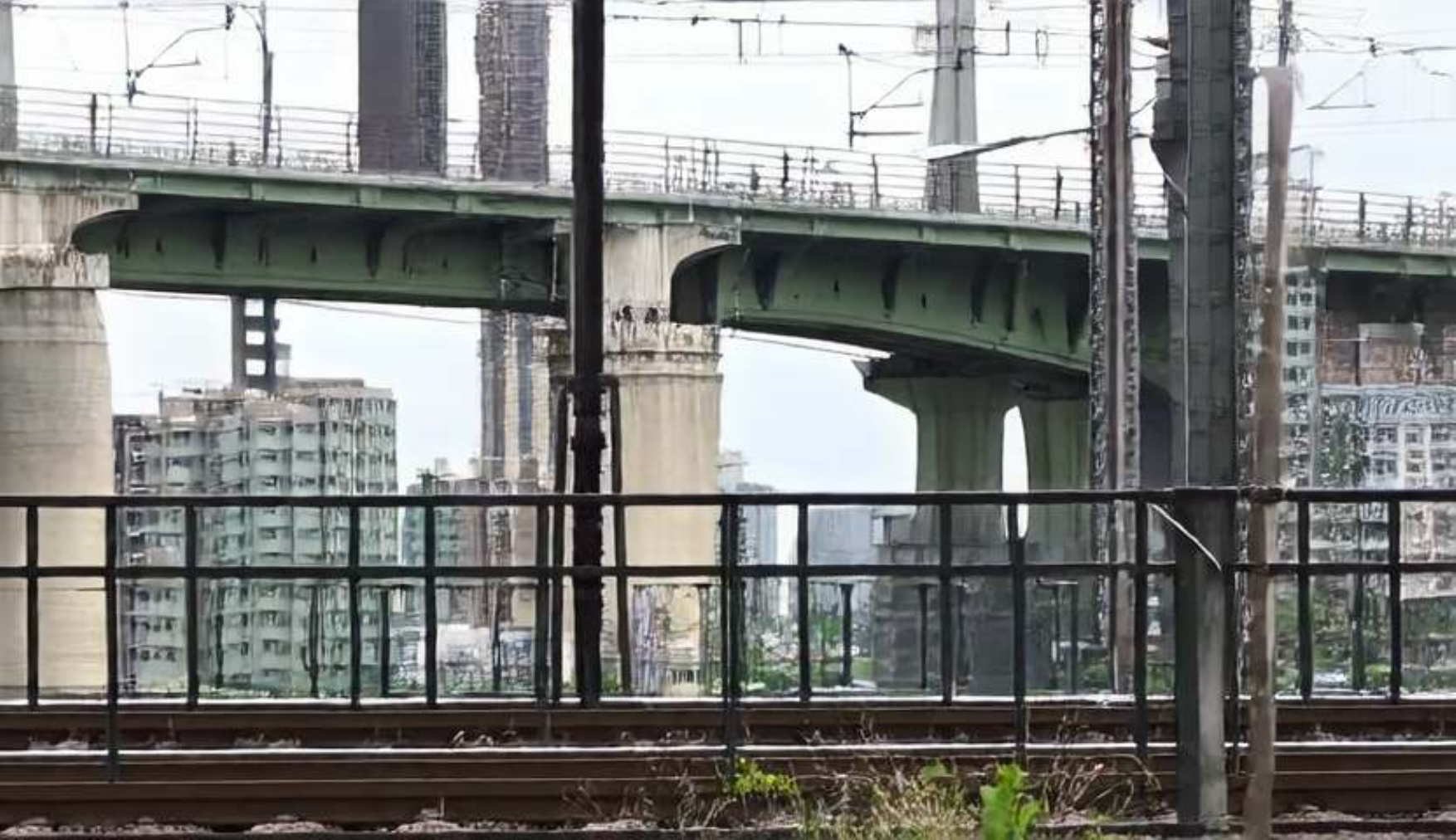}
        \\
    \end{tabular}
    \par\vspace{0.8em}
    \begin{tabular}{@{}c@{\hspace{1.5pt}}ccc@{}}
        & \multicolumn{3}{c}{\large \textbf{A person is clay pottery making.}}
        \\[0.25em]

        \methodcell{Original}
        & \framecell{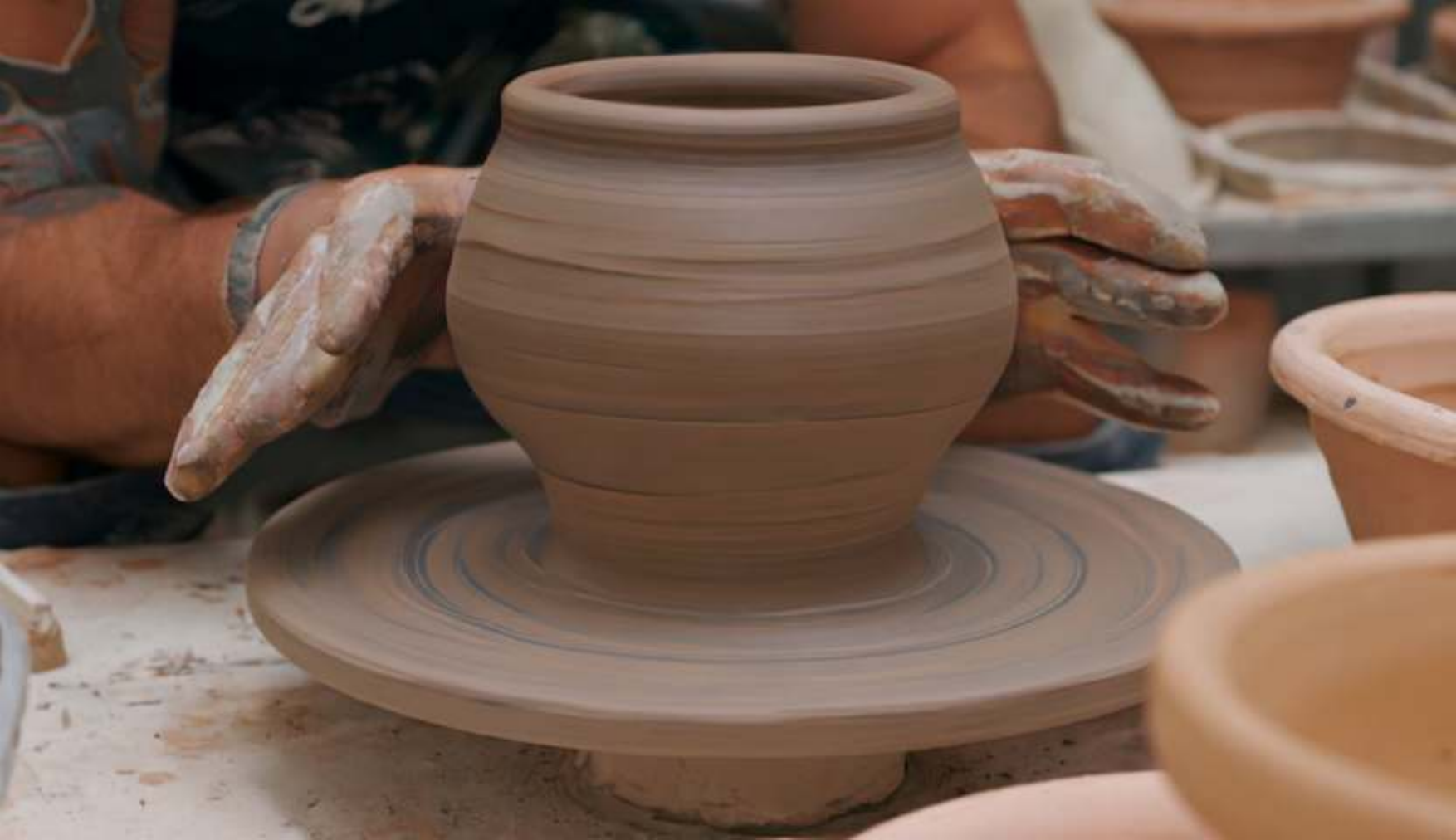}
        & \framecell{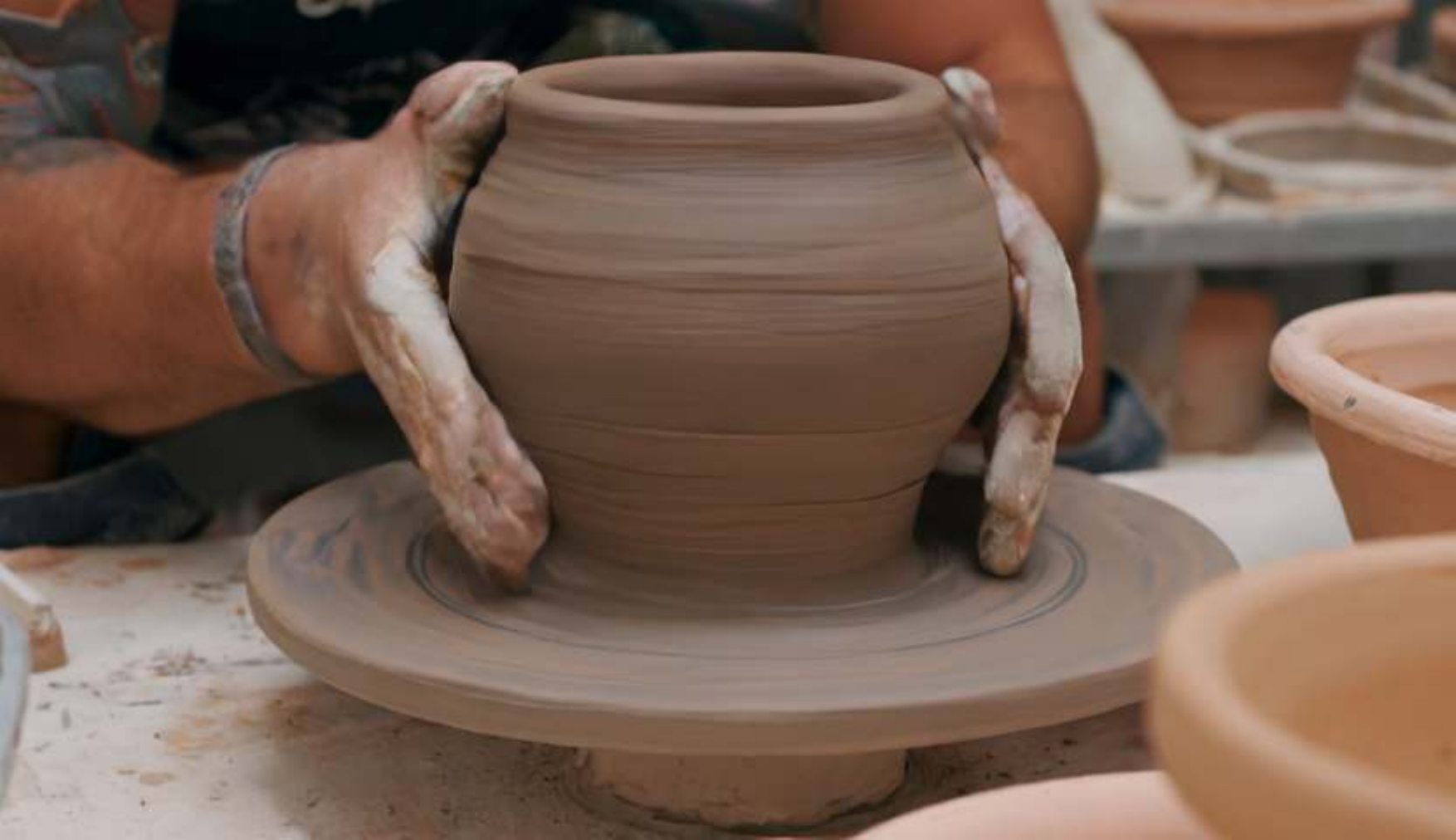}
        & \framecell{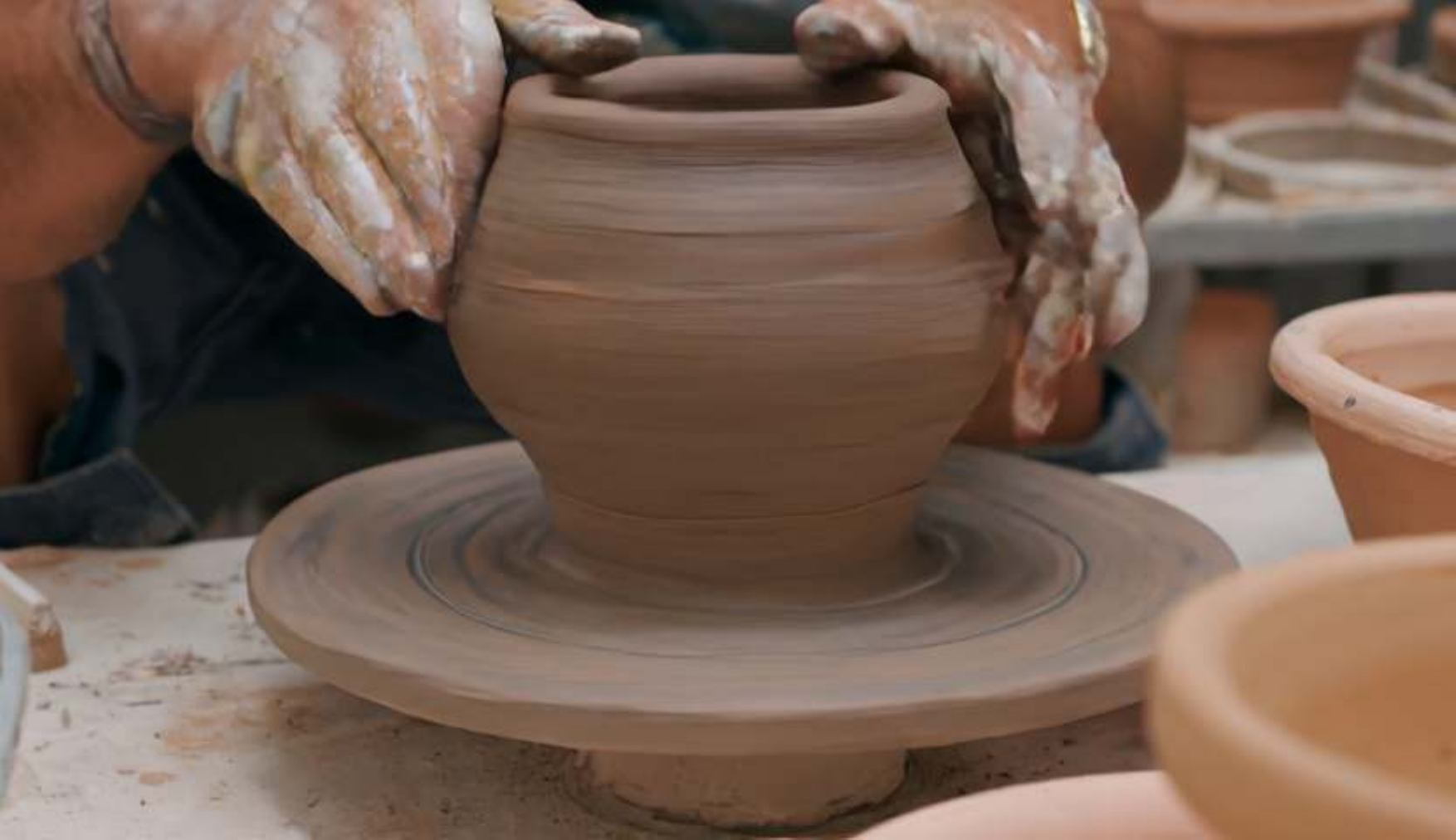}
        \\ \rowgap

        \methodcell{MagCache ($2.54\times$)}
        & \framecell{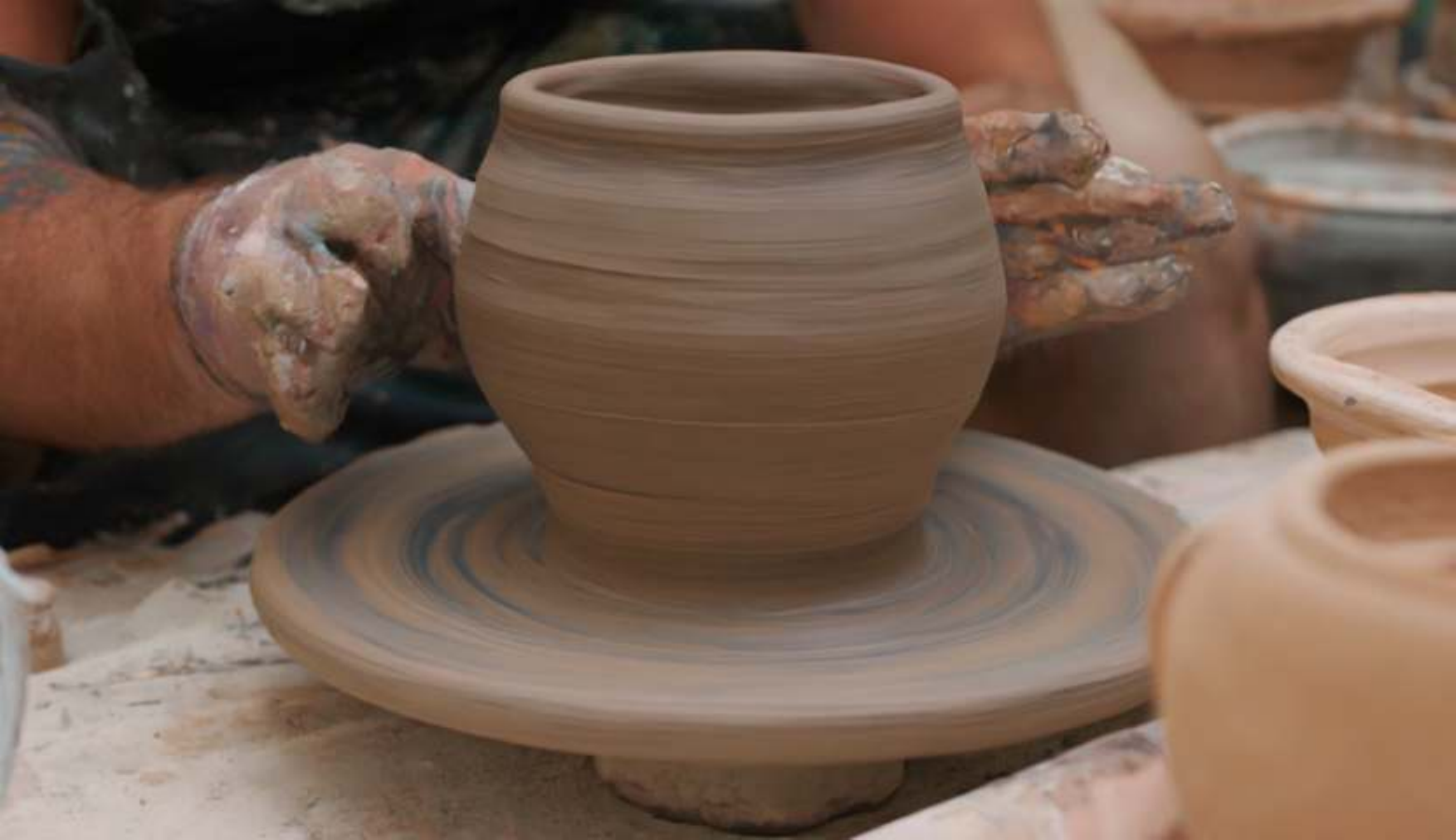}
        & \framecell{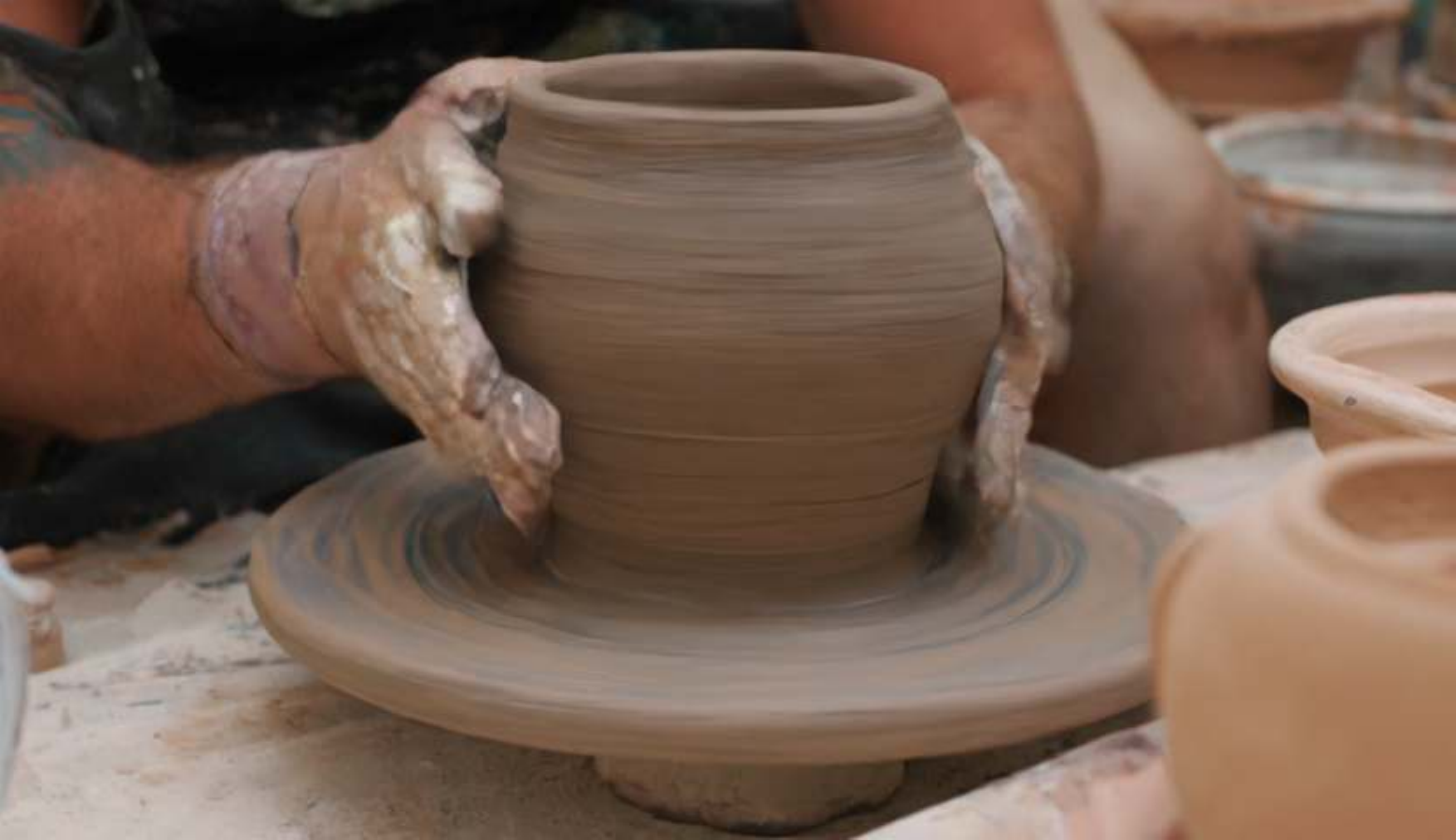}
        & \framecell{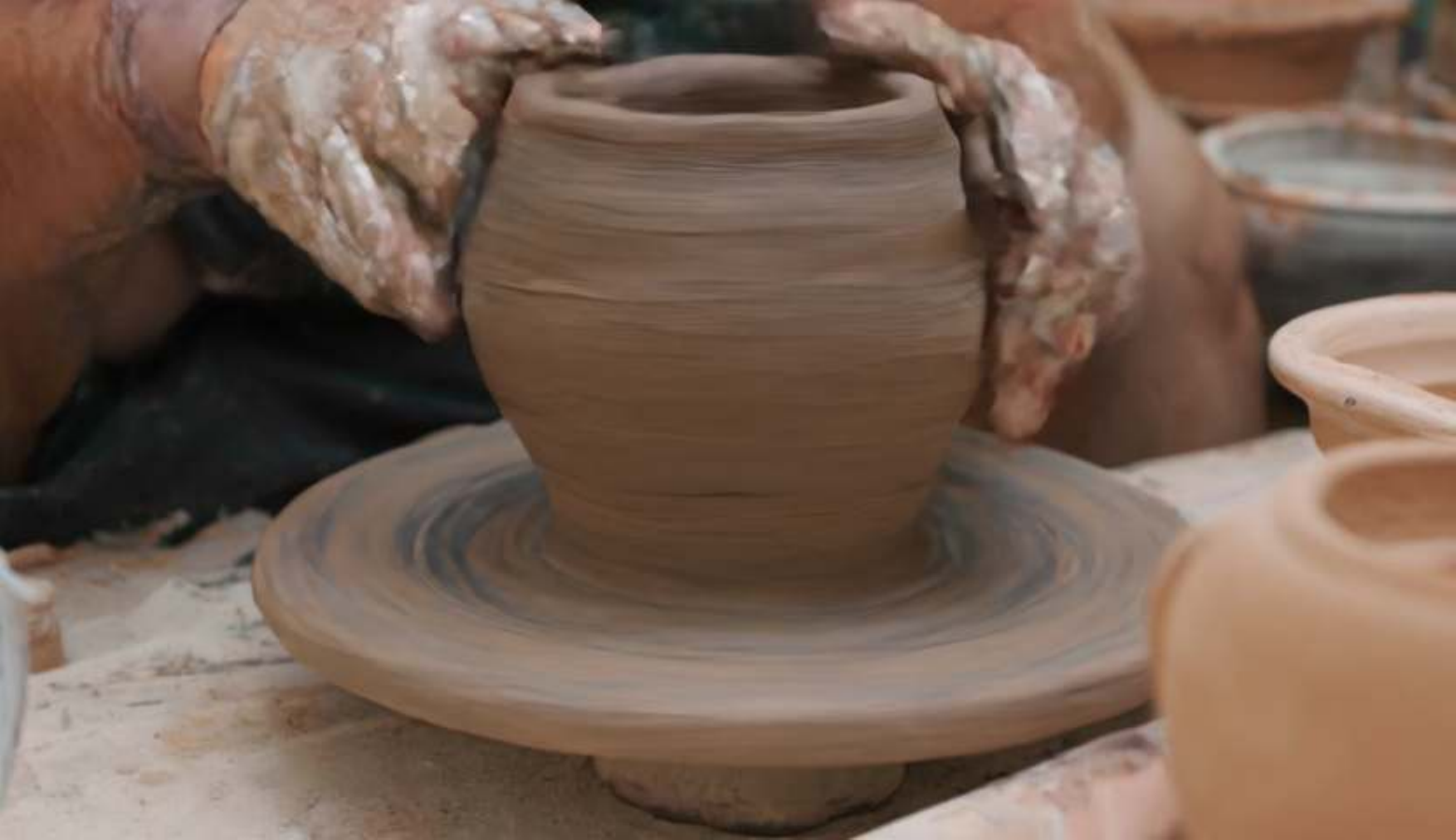}
        \\ \rowgap

        \methodcell{DiCache ($2.55\times$)}
        & \framecell{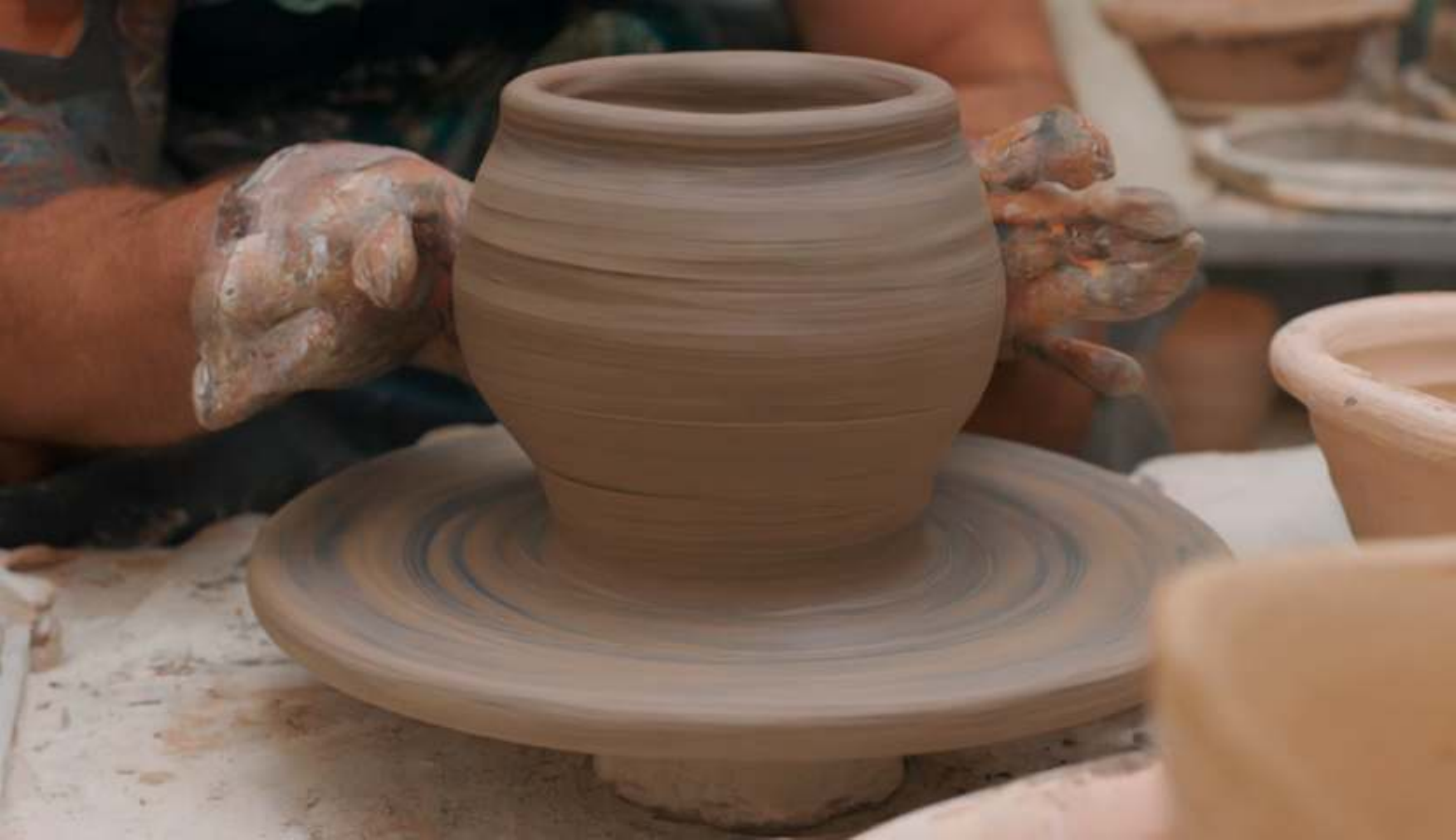}
        & \framecell{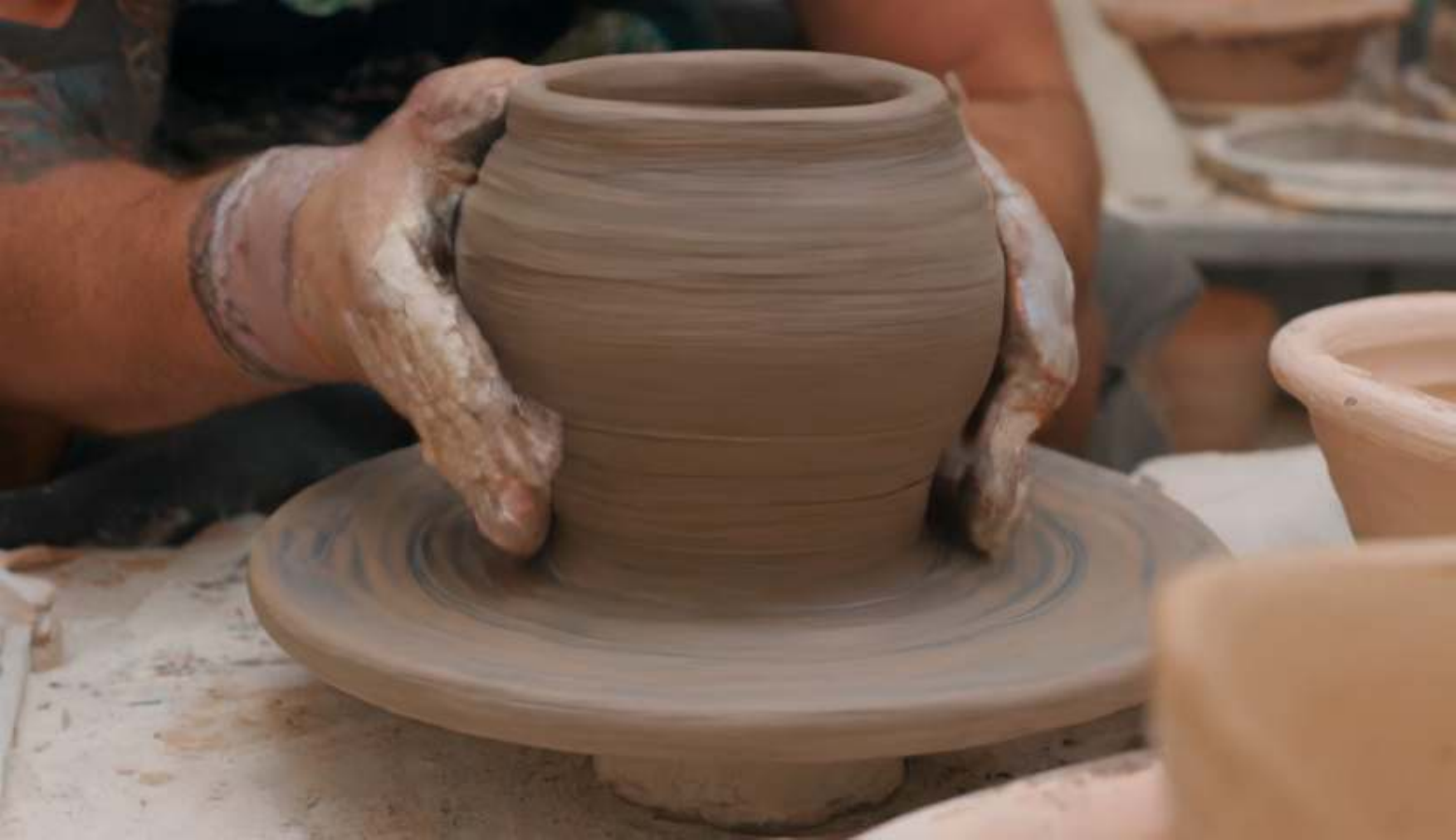}
        & \framecell{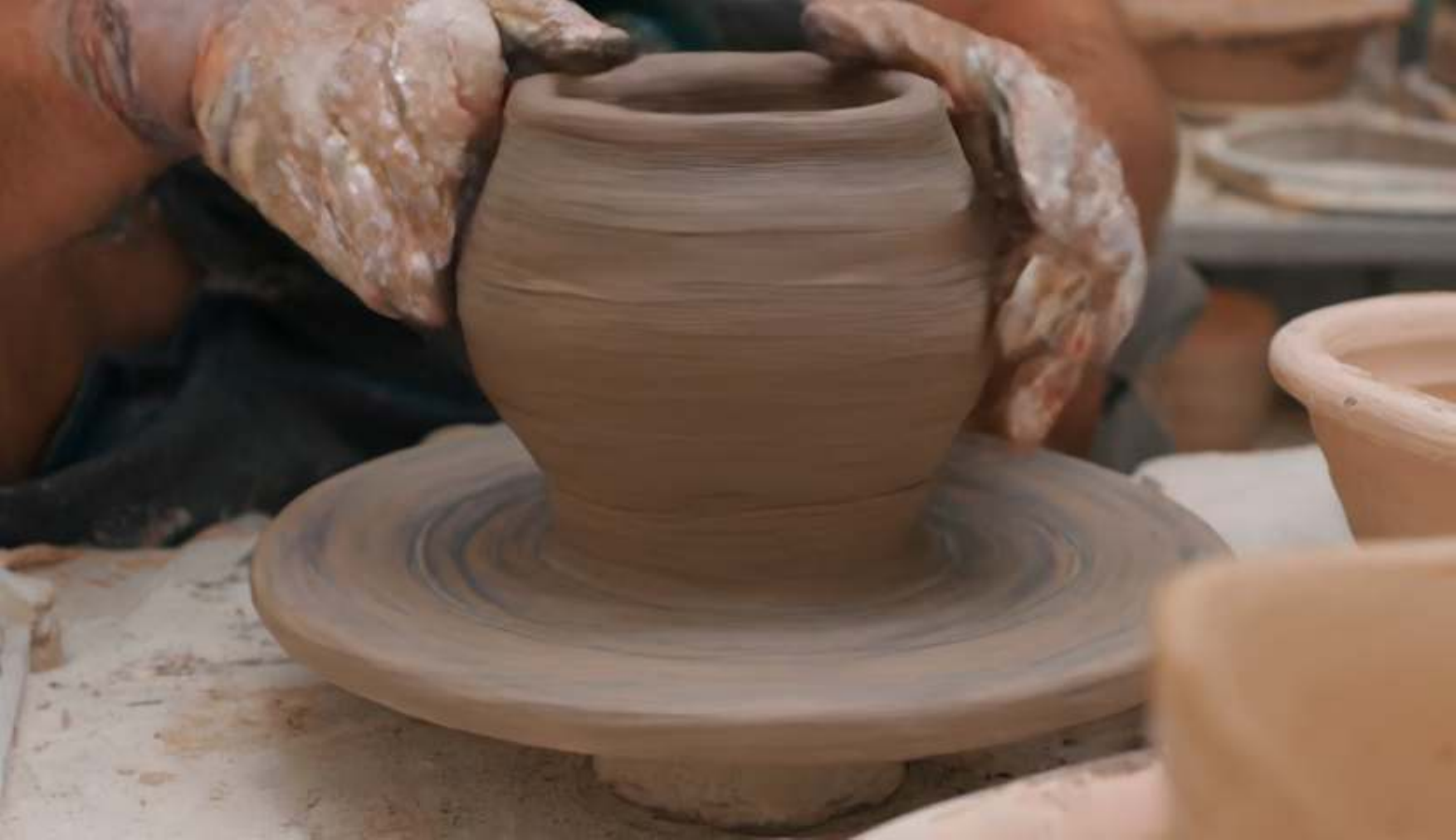}
        \\ \rowgap

        \methodcell{Ours ($2.56\times$)}
        & \framecell{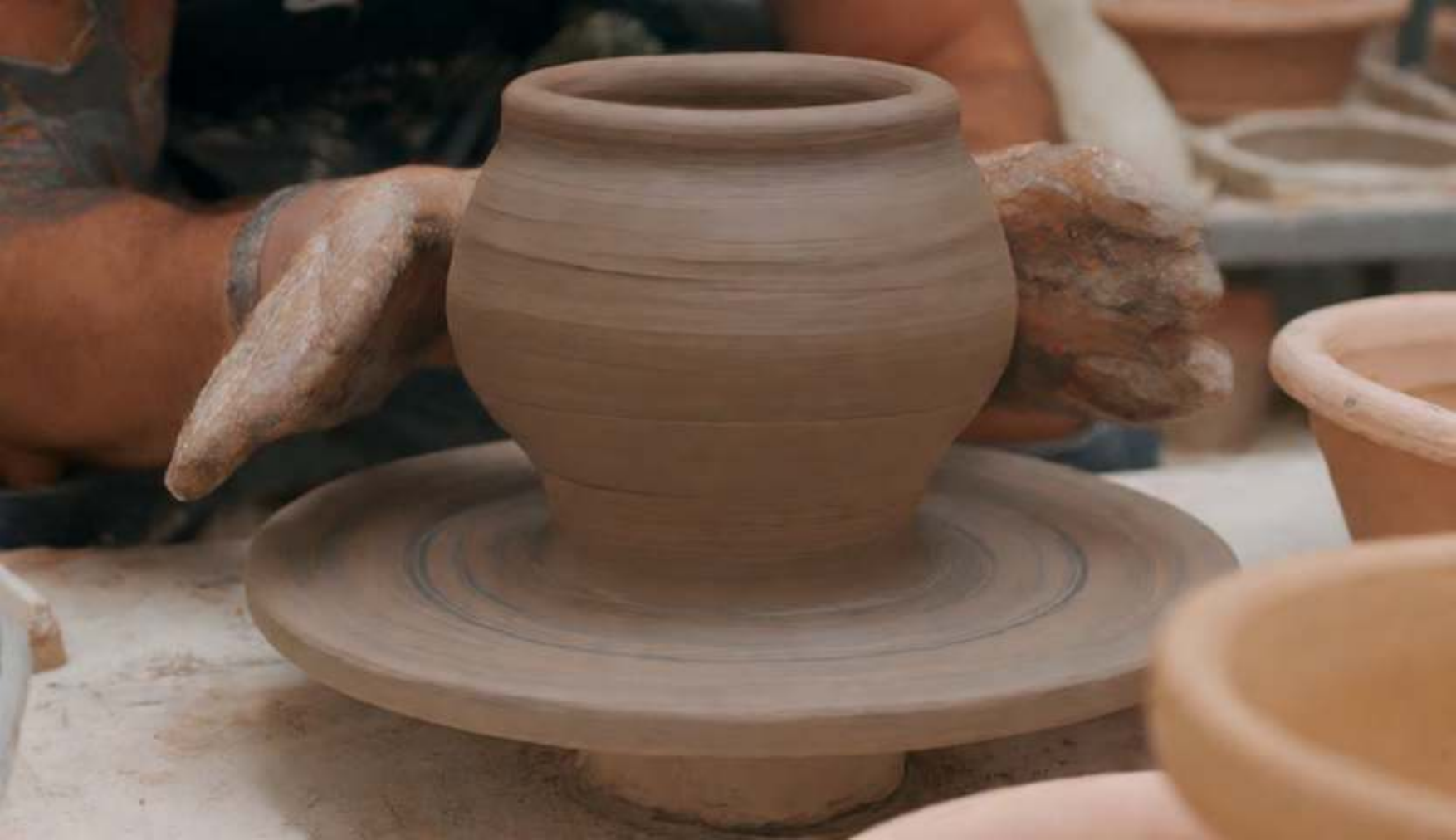}
        & \framecell{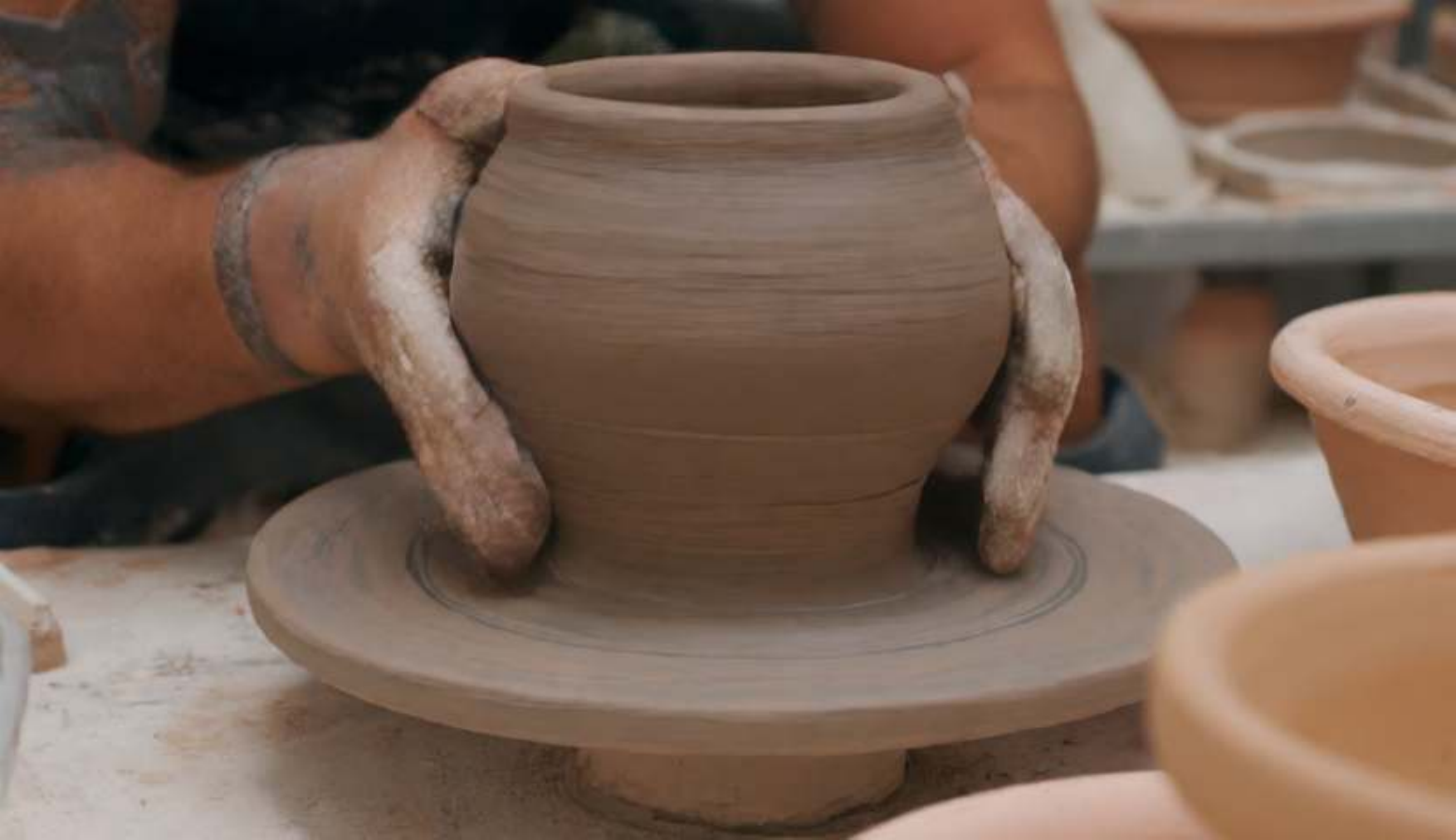}
        & \framecell{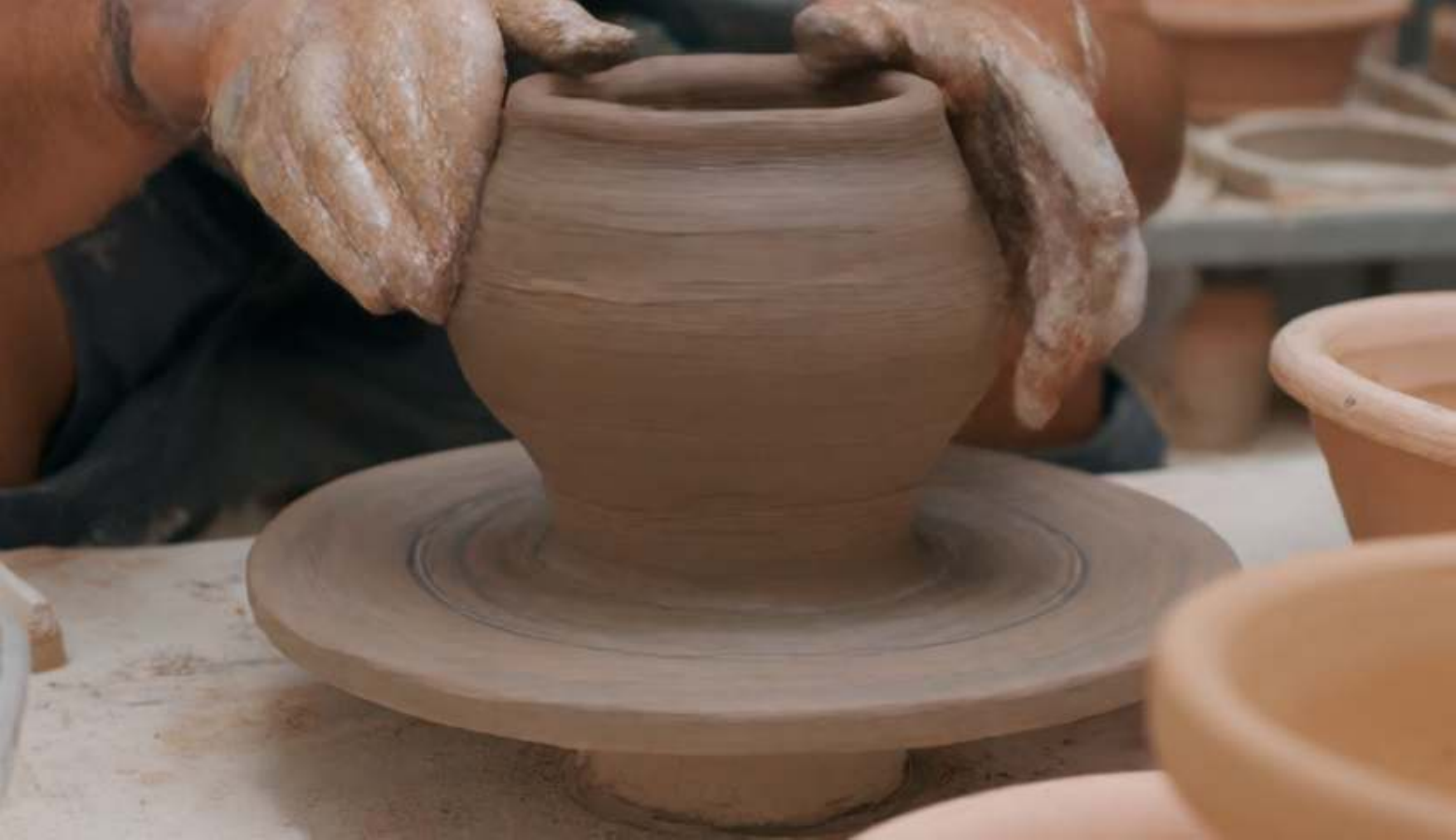}
        \\
    \end{tabular}
    \vspace{-0.3em}
    \caption{
    \textbf{Qualitative comparison on Wan2.1-T2V-1.3B (Part II).}
    }
    \label{fig:qualitative_cogvideox}
    \vspace{-0.8em}
\end{figure*}
\clearpage

%%%%%%%%%%%%%%%%%%%%%%%%%%%%%%%%%%%%%%%%%%%%%%%%%%%%%%%%%%%%

% \newpage
% \section*{NeurIPS Paper Checklist}
% \input{sections/08_checklist}

\end{document}